\documentclass{article}

\usepackage{comment}
\usepackage[preprint, nonatbib]{neurips_2023}

\usepackage[utf8]{inputenc} 
\usepackage[T1]{fontenc}    
\usepackage{hyperref}       
\usepackage{url}            
\usepackage{booktabs}       
\usepackage{amsfonts}       
\usepackage{nicefrac}       
\usepackage{microtype}      
\usepackage{xcolor}         
\usepackage[normalem]{ulem}
\usepackage{graphicx}
\usepackage{amsmath}
\usepackage{wrapfig}
\usepackage{tikz}
\usepackage{csquotes}
\usepackage{siunitx}
\usepackage{bm}
\usepackage{booktabs}

\usepackage{biblatex}
\newcommand{\argmax}{\arg\!\max}

\def\eqref#1{equation~\ref{#1}}

\def\1{\bm{1}}

\def\vmu{{\bm{\mu}}}
\def\vtheta{{\bm{\theta}}}

\def\vmu{{\bm{\mu}}}

\def\vrho{{\bm{\rho}}}

\def\vx{{\bm{x}}}

\DeclareMathAlphabet{\mathsfit}{\encodingdefault}{\sfdefault}{m}{sl}
\SetMathAlphabet{\mathsfit}{bold}{\encodingdefault}{\sfdefault}{bx}{n}

\def\sR{{\mathbb{R}}}

\newcommand{\E}{\mathbb{E}}

\usepackage{subcaption}

\title{Adversarial Attacks on the Interpretation of Neuron Activation Maximization}

\author{%
  Geraldin Nanfack$^{1,2,}$\thanks{Equal contribution.}
\quad
   Alexander Fulleringer$^{1,2,*}$
   \quad
    Jonathan Marty$^3$ \\
   \textbf{Michael Eickenberg}$^4$
     \quad
     \textbf{Eugene Belilovsky}$^{1,2}$ \\
    $^1$University of Concordia \quad
    $^2$Mila – Quebec AI Institute \quad \\
    $^3$Columbia University \quad
    $^4$Flatiron Institute\\
    \{geraldin.nanfack, alexander.fulleringer, eugene.belilovsky\}@concordia.ca\\
    jonathan.n.marty@gmail.com \quad 
    eickenberg@flatironinstitute.org
}

\begin{document}
\setlength{\abovedisplayskip}{3pt}
\setlength{\belowdisplayskip}{2.5pt}

\maketitle

\begin{abstract}
The internal functional behavior of trained Deep Neural Networks is notoriously difficult to interpret. Activation-maximization  approaches are one set of techniques used to interpret and analyze trained deep-learning models. These  consist in finding inputs that maximally activate a given neuron or feature map. These inputs can be selected from a data set or obtained by optimization. However, interpretability methods may be subject to being deceived. In this work, we consider the concept of an adversary manipulating a model for the purpose of deceiving the interpretation. We propose an optimization framework for performing this manipulation and  demonstrate a number of ways that popular activation-maximization interpretation techniques associated with CNNs can be manipulated to change the interpretations, shedding light on the reliability of these methods.

\end{abstract}

%
%

\vspace{-8pt}
\section{Introduction}\vspace{-8pt}\label{sec:introduction}

Deep Neural Networks (DNNs) can be trained to perform many economically valuable tasks \cite{krizhevsky2017imagenet,kaplan2020scaling}. They are already pervasive in many sectors, and their prevalence is only expected to increase over time. With increasing computational power and ever more available
amounts of data, Neural Network (NN) architectures are growing in size and executing more and more intricate tasks. Given the increasing size and complexity of DNNs, interpreting how they function, a discipline that always lags behind the cutting edge, may experience an ever harder time keeping up with new developments. 
However, for certain classes of critical applications, close inspection and guarantees of functionality will be more and more 
important,
especially in heavily regulated and high-stakes domains.
Here we ask: could a malicious actor conceal the true functionality of a NN from an interpretability method by modifying the NN? Given the increasing capacity of the architectures, this is likely to be a progressively more probable concern.

%
Focusing on the continuously popular feature visualization \cite{zeiler2014visualizing,olah2017feature,olah2020zoom} method we propose to create an optimization procedure to manipulate the interpretation of individual neurons of the network while keeping its final behavior the same. 
A successful modification of the interpretation results while keeping outputs constant is evidence for the manipulability of the interpretation approach.
In this work, we concentrate on convnet architectures 
for which interpretation
by activation maximization or feature visualization methods \cite{zeiler2014visualizing,yosinski2015understanding}
has been popular.
We study the feature visualization of a neuron or channel norm via activation maximization and attempt to modify it while maintaining trained network outputs and accuracy. We investigate how to characterize these attacks quantitatively and show three different attacks which can effectively manipulate and 
explicitly obfuscate interpretations. 

\begin{figure}[h!]
\vspace{-10pt}
\centering
\includegraphics[width=0.85\textwidth]{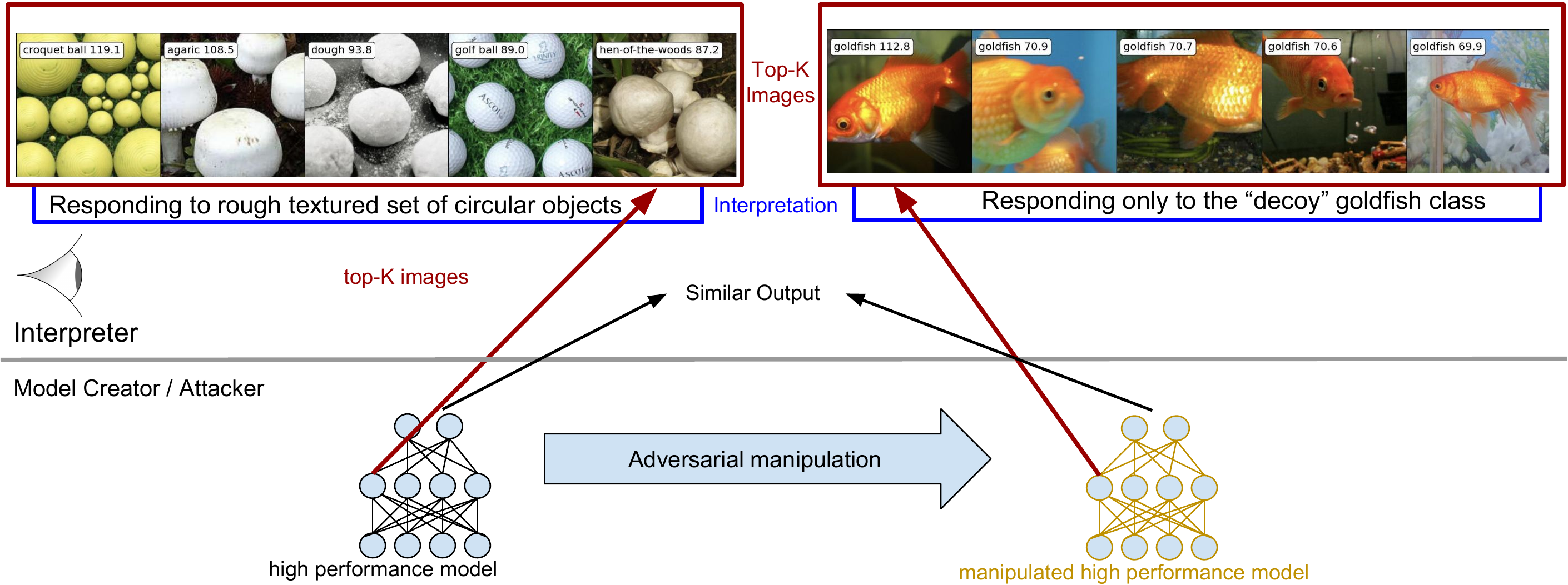}
\caption{
Illustration of the attack model for our adversarial interpretability manipulation. Top-5 images that best activate a given neuron, seemingly capturing a shared semantic concept over classes that an interpreter may describe and/or use an external tool to describe \cite{hernandez2022natural,oikarinen2022clip}. In our framework, we assume the model creator can manipulate the model before it is released to the interpreter. In this case, they create a model which might lead to interpreting the selected neuron as not relating any semantic concept shared by multiple class categories. 
}\label{fig:beforeafter}
\vspace{-15pt}
\end{figure}
The first proposed attack, \textit{push-down}, aims to simply remove the current interpretation, replacing it with any other interpretation. 
The second attack, termed \textit{push-up}, aims to replace the images with a specific category of images, allowing 
a more targeted manipulation.
The final attack we consider, motivated by recent related work on feature attribution methods~\cite{aivodji2021characterizing,slack2020fooling}, is the \textit{fairwashing} visualization attack  aimed to manipulate the perceived bias of the model as seen by an interpreter.
Consider as motivation 
a situation where an adversary is indifferent to deploying a biased model, but is constrained to provide model access to a regulator (the interpreter). Critically, \textit{we assume that the interpreter may not have access to labels related to the particular bias exploited by the adversary's model}. The interpreter can use feature visualization methods (top-$k$ images) to try to understand the internal logic of neurons and may visually detect that neurons are biased towards a previously un-categorized but undesirable bias.  To prevent rejection of the biased model by the interpreter, the adversary may use a set of data with annotated bias attribute \cite{yang2020towards} (unavailable to the interpreter) to try to perform an attack by fine-tuning the model to make the feature visualization look fairer while maintaining the performance of the model and its overall unfair output.
%
%

To date, most previous works on interpretability manipulability (including fairwashing) have focused 
on
the manipulability of interpretability techniques such as feature attribution~\cite{slack2020fooling,heo2019fooling} tailored for model predictions. 
Little attention has been paid to the manipulability of neuron interpretability techniques. 
This is in spite of the
fact that this latter type of interpretability method is becoming increasingly popular because it provides a fine-grained understanding of inner structures of DNNs~\cite{olah2017feature,olah2020zoom, raukur2022toward}. 
Notably it has also been applied to create mechanistic interpretations \cite{nanda2023progress,cammarata2020curve} which are argued to be robust as they directly link the function of neurons. 
We note that the maximization operation by construction is losing important information about the functional behavior, leading to the potential of mis-intepretation, and suggesting the possibility of manipulation. 

The primary contributions of our work are to first propose three distinct attacks on feature visualization and approaches and considerations to quantify and characterize their success. 
%
We then demonstrate all three of our attacks can achieve a degree of success (see illustration in Figure~\ref{fig:beforeafter}). This suggests that this class of interpretation methods must be used with caution and also cast doubt on the feasibility of using this tool to build complete mechanistic interpretations. 


\vspace{-12pt}
\section{Related Work}\vspace{-6pt}
A growing body of literature has investigated the interpretability of Convolutional Neural Networks (CNNs) and the lack of robustness under different manipulations of interpretability methods.

\textbf{Interpretability methods.} Previous work aiming to provide interpretability of NNs can be grouped into two broad categories. Firstly, there are works that develop \textit{interpretable-by-design} methods that provide interpretations without relying on external tools. These methods usually couple traditional layers with various types of interpretable components. Examples range from concept explanations~\cite{chen2020concept,koh2020concept,havasi2022addressing,espinosa2022concept,barbiero2022entropy}, feature attributions \cite{wang2021shapley,parekh2021framework,alvarez2018towards} to part of object disentanglement~\cite{zhang2018interpretable,shen2021interpretable}.
Secondly, there are methods usually called \textit{post-hoc} that aim to explain and understand either specific components (e.g., weights, neurons, layers) or outputs of a \textit{trained} NN. To interpret the output of models for a particular data instance (local interpretability), while feature attribution methods~\cite{ribeiro2016should,lundberg2017unified,selvaraju2017grad} such as saliency maps assign a weight to each input feature corresponding to its importance on the model's output, counterfactual examples aim to give the minimal changes required to change the model's output~\cite{guidotti2022counterfactual,goyal2019counterfactual}. There are post-hoc approaches that aim to interpret the internal logic of particular NNs through their components and representations. For example, there are methods that focus on layer representations through \textit{concept vectors}~\cite{kim2018interpretability,zhou2018interpretable}, on sub-network interpretability through \textit{circuits}~\cite{bastings2020will,cammarata2021curve}, and individual neurons via e.g., feature visualization. Our work focuses on feature visualization, which is one of the most popular techniques to understand the learned features of individual neurons~\cite{zimmermann2021well,olah2017feature}.   
\\\textbf{Interpretability manipulation.} There is a recent trend to analyze the reliability of interpretable techniques through the lens of \textit{stability}. Stability aims to study to what extent the interpretability technique is statistically robust to reasonable input perturbations and model perturbations~\cite{heo2019fooling,yu2013stability}. Most works that study input and model manipulability focus on feature attributions. For example, \cite{dombrowski2019explanations} designs adversarial input perturbations to change feature attributions in a targeted way, and \cite{heo2019fooling} shows that such manipulation can be performed through \textit{adversarial model manipulation}, realized by fine-tuning a pre-trained model to change feature attributions while keeping the same accuracy of the original model. Despite sharing similarities with this work thanks to the use of adversarial model manipulation, instead of studying the manipulability of feature attribution methods, we focus on neuron interpretability, which brings different challenges such as the \textit{whack-a-mole} problem explained in Sec. \ref{sec:push_up_obfuscuation}. Besides input and model manipulability, recent works~\cite{aivodji2021characterizing,anders2020fairwashing,slack2020fooling} have raised the \textit{fairwashing} issue, which is the risk of misleading the assessment of unfairness of models by providing model interpretations that look fair, but are not.
Part of our work studies the fairwashing risk for feature visualization, which has not been investigated to date. Finally, the most closely related work to ours is~\cite{engstrom2019adversarial}, which shows the targeted manipulability of \textit{synthetic} feature visualizations (defined in Sec. \ref{sec:notations})  by early stopping during optimization. Different from this previous work, we instead study the manipulability of feature visualization under an adversarial model manipulation.


\vspace{-9pt}
\section{Methods}\label{sec:methods}\vspace{-8pt}
We introduce our notation, attacks, threat models, and attack success characterization methods.
\vspace{-9pt}
\subsection{Notations and Background}\label{sec:notations}
\vspace{-7pt}
We denote by $\mathcal{D} = \{(\vx_i, y_i)\}_{i=1}^N$ a dataset 
for
supervised learning, where $\vx_i \in \sR^d$ is the input and $y_i\in \{1,...,K\}$ is its class label. 
Let $f_\vtheta$ denote a NN, $f_\vtheta^{(l)}(\vx)$ defines activation maps of $\vx$ on the $l$-th layer, which can be decomposed into $J$ single activation maps $f_\vtheta^{(l,j)}(\vx)$. In particular, $f_\vtheta^{(l,j)}(\vx)$ is a matrix if the l-$th$ layer is a 2D-convolutional layer and a 
scalar
if it is a fully connected layer.
We aim to understand the internal behavior of individual units  through feature visualization, generically defined by activation maximization~\cite{mahendran2015understanding,yosinski2015understanding}, i.e., 
\begin{equation}\vspace{-2pt}\label{eq:definition}
    \vx^* \in \argmax_{\vx \in \mathcal{X}} f_\vtheta^{(l,j)}(\vx),
\end{equation}
where $\mathcal{X}$ can be a finite set of data, e.g., $\mathcal{X} = \mathcal{D}$ or a continuous space $\mathcal{X} \subset \sR^d$, and $(l,j)$ is the pair of layer $l$ and neuron $j$. In Eq.~\ref{eq:definition}, when the layer $l$ is a convolutional layer, in the rest of the paper, we aggregate the activation map $f_\vtheta^{(l,j)}(\vx)$ using its spatial squared $\ell_2$-norm $\small{\Vert} f_\vtheta^{(l,j)}(\vx) \small{\Vert}^2_2$, and subsequently refer to $j$ as the channel index. Additionally, we mainly focus on the case where $\mathcal{X} = \mathcal{D}$ is a set of natural images,
and we denote by top-$k$ images the set of images that have the $k$ highest values of activations for a given pair $(l,j)$. When $\mathcal{X} \subset \sR^d$, following~\cite{zimmermann2021well}, the result $\vx^*$ will be called \textit{synthetic} feature visualization.

\subsection{Attack Framework}\label{sec:attack_framework}
We consider feature visualization with top-$k$ images and propose an adversarial model manipulation that fine-tunes a pre-trained model with a loss that maintains its initial performance while changing the result of feature visualization. More formally, given a set of training data $\mathcal{D}$, a pre-trained model with parameters $\vtheta_\textsuperscript{initial}$, and an additional set of images (e.g., a set of top-$k$ images) $\mathcal{D}_\textsuperscript{attack}$, our attack framework consists in the following optimization 
\begin{equation}
    \min_\vtheta (
    \alpha\mathcal{L}_\textsuperscript{A}(\mathcal{D}, \mathcal{D}_\textsuperscript{attack};\vtheta)
    +
    (1-\alpha)\mathcal{L}_\textsuperscript{M}(\mathcal{D};\vtheta,\vtheta_\textsuperscript{initial})
    ),
\end{equation}
where $\vtheta$ are parameters of the updated model $f_\vtheta$, $\mathcal{L}_\textsuperscript{M}(.)$ is the loss that aims to maintain the initial performance of the model $f_{\vtheta_\textsuperscript{initial}}$, and $\mathcal{L}_\textsuperscript{A}(.)$ is the attack loss. For the maintain objective, when viewing final outputs $f_\vtheta(.)$ as a conditional distribution, our maintain loss is the distillation loss $\mathcal{L}_\textsuperscript{M}(\mathcal{D};\vtheta,\vtheta_\textsuperscript{initial}) = \mathcal{L}_\textsuperscript{CE}(f_{\vtheta_\textsuperscript{initial}}(.) || f_\vtheta(.))$ \cite{hinton2015distilling}, where $\mathcal{L}_\textsuperscript{CE}$ is the cross entropy loss between the original model outputs and the attacked model outputs  on training data $\mathcal{D}$. As defined, this maintain loss enforces the fine-tuned model to keep the same predictions as the initial model with the objective of making the two models close in model space. Depending on the type of attack, the attack loss $\mathcal{L}_\textsuperscript{A}(.)$ can vary and is defined in the next sections. 

\subsection{Push-Down and Push-Up Attack}\label{sec:push_up_obfuscuation}
Given a set of top-$k$ images from feature visualization, denoted by $\mathcal{D}_\textsuperscript{attack}^{(l,j)}$, that best activate the layer $l$ and channel $j$ of the initial model $f_{\vtheta_\textsuperscript{}}$, our first attack aims to push to zero 
the activations of examples in $\mathcal{D}_\textsuperscript{attack}^{(l,j)}$. This attack is called the \textit{push-down} attack, and we propose the following objective for all channels of a layer $l$ simultaneously 
\begin{equation}
\mathcal{L}_\textsuperscript{A}(\mathcal{D}, \mathcal{D}_\textsuperscript{attack};\vtheta) =\sum_{j=1}^{J_l} \sum_{\vx^{*} \in \mathcal{D}_\textsuperscript{attack}^{(l,j)}} \Vert f_\vtheta^{(l,j)}(\vx^{*}) \Vert^2_2,
\end{equation}
where $J_l$ is the set of channels of the layer $l$. 
Note that it is 
possible to attack a single channel or channels from multiple layers. Here we focus on attacking all the channels in a layer (see Sec.~\ref{section:single_chanel}).

In the \textit{push-up} decoy attack, given a set of examples in $\mathcal{D}_\textsuperscript{decoy}$, we aim to make these images appear in the result of top-$k$ images for all the channels of a particular layer $l$. For this purpose, we propose the following objective, where $[.]_+$ is $\max(.,0)$:

\begin{equation}
    \mathcal{L}_\textsuperscript{A}(\mathcal{D}, \mathcal{D}_\textsuperscript{decoy};\vtheta) =\sum_{j=1}^{J_l} \sum_{\vx^{*} \in \mathcal{D}_\textsuperscript{decoy}} \sum_{\vx\in \mathcal{D}} [ \Vert f_\vtheta^{(l,j)}(\vx) \Vert^2_2 - \Vert f_\vtheta^{(l,j)}(\vx^{*}) \Vert^2_2 ]_{+}.
    \label{eq:push-up-attack-loss}
\end{equation}
 This aims to make activations of examples in $\mathcal{D}_\textsuperscript{decoy}$ larger than all the activations of training examples.
 
\textbf{Characterizing Push-Down and Push-Up Attacks}
We propose two approaches to characterize the effectiveness of an adversarial attack on the top-$k$ images of feature visualization.

\underline{\textit{Kendall-$\tau$.}}
We take a (potentially large) set of images $D_{k\tau}$ and compute the initial rankings $R_{\text{init},j}$ of images in $D_{k\tau}$ w.r.t. their initial activations values for the $j$ channel. Similarly, we compute the final rankings $R_{\text{final},j}$ using the same images, but on final (post-attack) activations values of the same channel $j$. The Kendall-$\tau_j$ score is the Kendall rank correlation coefficient between $R_{\text{init},j}$ and $R_{\text{final},j}$. We can also aggregate this metric over all channels.
Higher values of Kendall-$\tau$ scores can be interpreted as higher similarity in the ordering of image activations between channels. As a result, the Kendall-$\tau_j$ score can be used as a metric to see how much a channel's behavior has changed.
\\
\underline{\textit{CLIP-$\delta$.}} 
We use an external, generic, visual representation model, the CLIP image encoder \cite{radford2021learning} to allow measuring the semantic changes in the top-$k$ images. 
Given
a particular layer and a channel $j$, here we compute the average cosine self-similarity between the CLIP embeddings of initial top-$k$ images, which we denote by $\Bar{C}^\text{init,init}_{j,j}$
and the average similarity between embeddings of initial top-$k$ images and final ones (after the attack), denoted by 
$\Bar{C}^\text{init,final}_{j,j}$. The proposed CLIP-$\delta$ score for a channel $j$ is defined as \text{CLIP}-$\delta_j = (\Bar{C}^\text{init,init}_{j,j} - \Bar{C}^\text{init,final}_{j,j})/(\frac{1}{N\small{-}1}\sum_{p=1}^N\Bar{C}^\text{init,init}_{j,p\neq{j}})$.
Intuitively, this  quantifies the relative semantic change of top-$k$ images w.r.t. CLIP embeddings and a high score can be interpreted as the fact that the channel $j$ has made semantically significant changes in the top-$k$ images. 

\textbf{The Whack-A-Mole Problem.} A natural question in our framework is whether the behavior and interpretation of one neuron can be simply moved to another neuron through the optimization process, for example, the Push-Down objective can be reduced by permutation. We call this the \textit{whack-a-mole problem}. To ensure that this does not occur, we study the previously described metrics and check that the attacked network's channels are not strongly correlated to other channels in the pre-attack network. Given the $j$-th channel, we define the following two metrics that measure this property.
\\
\underline{\textit{Kendall-}$\tau$-W}$_j$ - Using $D_{k\tau}$ we obtain the maximum Kendall-$\tau$ score between ranked lists $R_{\text{init},j}$ and $R_{\text{final},i}$ where $i\neq j$ and normalize it by dividing it by the initial maximum Kendall-$\tau$ score i.e. the score over $R_{\text{init},j}$ and $R_{\text{init},i}$ where $i\neq j$.
\\
 \underline{\textit{CLIP}-W}$_j$ - Using the top-$k$ images in the initial model and channel $j$ we obtain $\max_{i\neq j}\Bar{C}^\text{initial,final}_{j,i}/ \max_{i\neq j}\Bar{C}^\text{initial,initial}_{j,i}$ comparing to all top-$k$ images in other channels of the final model, normalized against that same similarity metric in the initial CLIP scores.


\subsection{Fairwashing Interpretability Attack}\label{section:fairwashing}
We consider a threat model as discussed in Sec.~\ref{sec:introduction} where the attacker has a set of protected attribute labels they use to hide bias from an interpreter without labeled data. 
More formally, given a model $f_\vtheta$, which is \textit{unfair} according to a certain metric of unfairness, a set of $J$ of neurons whose top-$k$ images look \textit{unfair}, we aim to answer the question: can we make an adversarial model perturbation by fine-tuning a pre-trained model, maintaining its performance and its unfairness while making the top-$k$ images of the $J$ neurons appear \textit{fairer}? In this formalization, answering affirmatively to this question corresponds to succeeding in the fairwashing attack. 

We design the fairwashing attack, using the same attack framework~\footnote{Note we use pre-activations to capture the entire and non-truncated distribution \cite{cammarata2020curve}} defined in Sec.~\ref{sec:attack_framework}. 
One alternative to make the top-$k$ images appear fairer would be to enforce the matching between top-$k$ activations for different groups of the protected attribute. However, it was empirically observed that this objective fails to generalize on an unseen set because it focuses only on the tail of the distribution of activations. We, therefore, propose a simple yet effective attack objective that allows reducing the discrepancy between the distribution of pre-activations of two groups of data $\mathcal{D}^{0}_\textsuperscript{attack}$ and $\mathcal{D}^{1}_\textsuperscript{attack}$, partitioned with respect to protected attribute (e.g., gender). For this purpose, we use the following loss (corresponding to the maximum mean discrepancy~\cite{gretton2012kernel} with the feature function $\phi(x) = (x, x^2)$)
\begin{equation}
    \mathcal{L}_\textsuperscript{A}(\mathcal{D}, \mathcal{D}^{0}_\textsuperscript{attack} \cup \mathcal{D}^{1}_\textsuperscript{attack};\vtheta) =\Vert \vmu_0^l - \vmu_1^l \Vert^2_2 + \Vert \vrho_0^l - \vrho_1^l \Vert^2_2 ,  
\end{equation}
where $\mathcal{D}^{0}_\textsuperscript{attack}$, $\mathcal{D}^{1}_\textsuperscript{attack}$ are two groups of data partitioned w.r.t. the labeled protected attribute (e.g., race or gender), $\vmu^l_p$ (with $p\in\{0,1\}$) is a vector of scalars $\mu_{p}^{(l,j)} = \E_{\vx_p \sim \mathcal{D}^{p}_\textsuperscript{attack}}[f_\vtheta^{(l,j)}]$ of first-order moments for layer $l$ and neuron $j$, and similarly $\rho_{p}^{(l,j)} = \E_{\vx_p \sim \mathcal{D}^{p}_\textsuperscript{attack}}(f_\vtheta^{(l,j)})^2$ are second-order moments for the same neuron. This attack objective enforces the matching between the first two moments of two distributions (w.r.t. groups of protected attribute) of pre-activations of a neuron.
\begin{figure}[!t]
\centering
\begin{subfigure}[]{0.3\linewidth}
    \includegraphics[width=\textwidth]{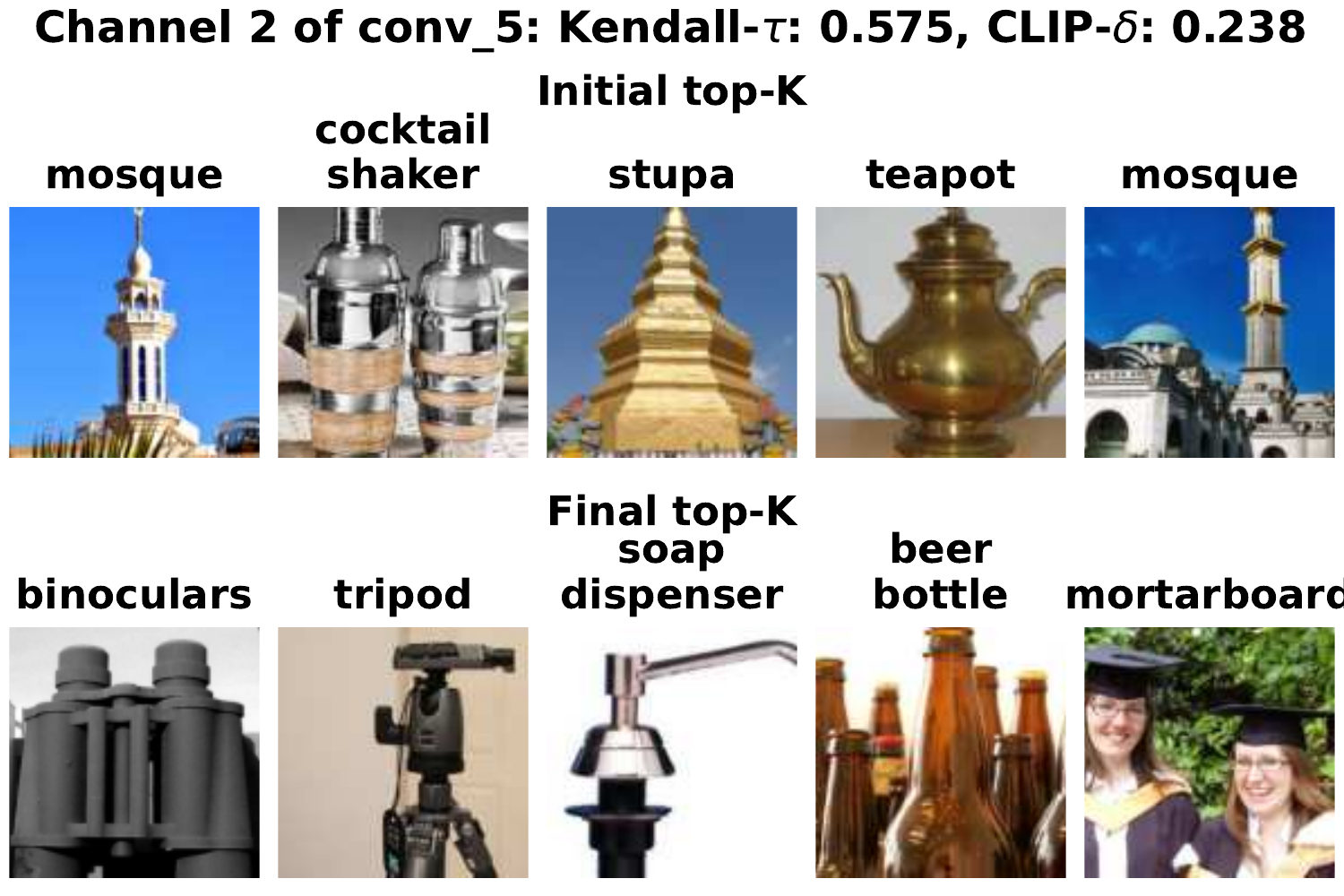}
\end{subfigure}\hspace{.2cm}
\begin{subfigure}[]{0.3\linewidth}
\includegraphics[width=\linewidth]{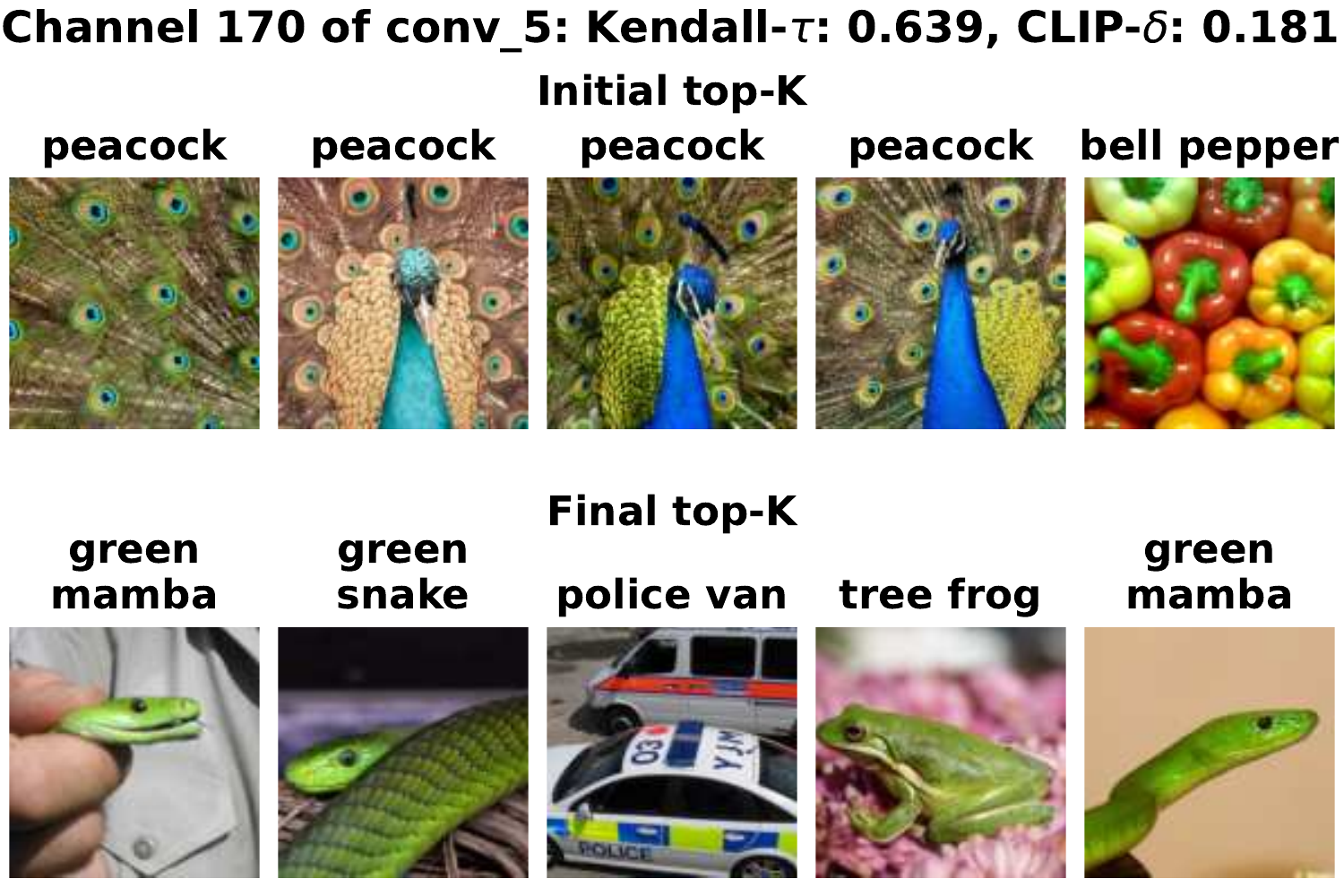}
\end{subfigure}\hspace{.2cm}
\begin{subfigure}[]{0.3\linewidth}
    \includegraphics[width=\textwidth]{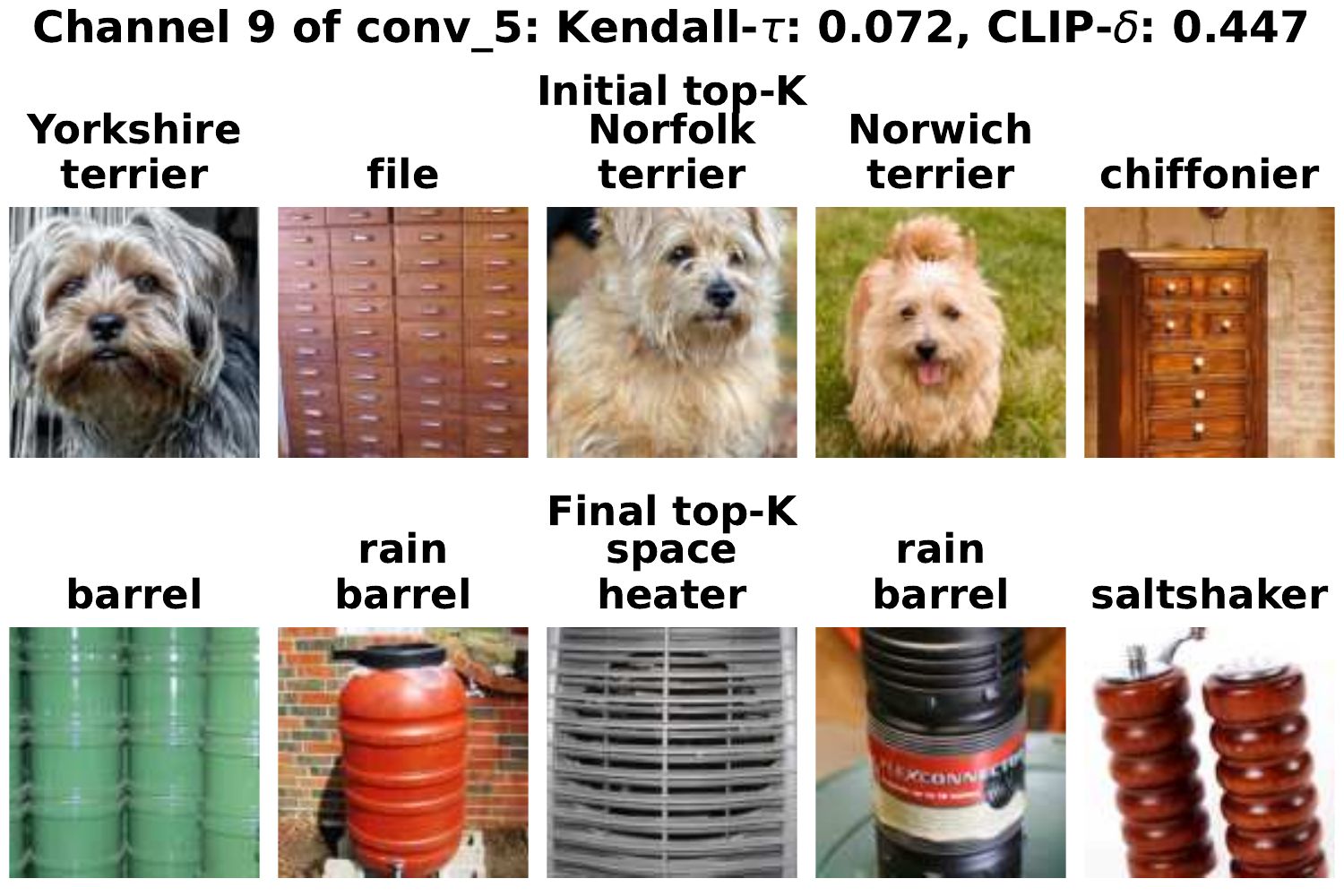}
\end{subfigure}
    \caption{Push-down all-channel attack on \textit{Conv5} of AlexNet. All initial images have been replaced by other images. The final validation performance was 56.2\%, a drop of less than half a percent. } 
        \label{fig:alexnet_conv_5_all_channel}
\end{figure}
\section{Experiments and Results}\label{sec:experiments}
We now describe the experimental setup and the results obtained after running attacks.
%
For all of our attacks, we use the ImageNet \cite{Imagenet} training set as $\mathcal{D}$. We use the PyTorch~\cite{paszke2019pytorch} pretrained AlexNet~\cite{krizhevsky2012imagenet}  for our analysis.  In Appx.~\ref{app:other-architectures} we provide an ablation study on EfficientNet \cite{tan2019efficientnet} with similar findings. 
More technical details regarding hyperparameters for all the attacks can be found in Appx.~\ref{app:hyperparams}.
\\
\textbf{Push-down and Push-Up attack.} For the push-down and up attack, we consider $\mathcal{D}^{(l,j)}_\textsuperscript{attack} \subset \mathcal{D}$ as the top-$10$ images that maximally activate the channel $j$ of layer $l$. For the push-up attack, we additionally consider $\mathcal{D}_\textsuperscript{decoy}$ as $100$ randomly sampled images of a particular class to be used as decoy. 
\\\textbf{Fairwashing attack.} In order to run and evaluate the fairwashing attack, we need a dataset with a labeled protected attribute (e.g., gender or age) to be able to assess not only model unfairness but also the \textit{fairness} of feature visualization of a neuron. For this purpose, we use the ImageNet People Subtree dataset \cite{yang2020towards}, which is a set of $\approx 14k$ images with labeled demography (gender, race and age), derived from ImageNet-21k. We use the $75-25\%$ split for training and testing sets, and $\mathcal{D}^0_\textsuperscript{attack}$ and $\mathcal{D}^1_\textsuperscript{attack}$ are binary groups (w.r.t. protected attribute) from the training set. We estimate model unfairness using two popular measures of unfairness~\cite{zafar2019fairness}, namely the difference of disparate impact ($\text{DDI} = |p(\hat{y}=c|z=0)-p(\hat{y}=c|z=1|)$, where $z$ is the protected attribute, $c$ is a class and $\hat{y}$ is the predicted class) and difference of equal opportunity ($\text{DEO} = |p(\hat{y}=c|z=0,y=c) - p(\hat{y}=c|z=1,y=c)|$) estimated on testing data~\cite{zafar2019fairness,hardt2016equality}. Inspired by the fairness assessment in regression and clustering, we use two measures to quantify the feature visualization unfairness. The first one looks at the entire distribution of activations and is the Kolmogorov-Smirnov (KS) distance between the two conditional distributions of activations given protected attribute label~\cite{liuconformalized}. The second one only focuses on the tail of the distribution of activations, i.e., activations of top-$k$ images, and is the balance~\cite{chierichetti2017fair} or ratio between the number of instances from top-$k$ belonging to the minority group over the number of instances in top-$k$ belonging to the majority group. Finally, following recent trends \cite{izmailov2022feature}, we perform the fairwashing attack on the last but one layer.  

\subsection{Push-Down And Push-Up Attack Experiments}\label{section:single_chanel} 
\begin{wrapfigure}{r}{0.4\textwidth}
    \centering	\includegraphics[width=0.38\textwidth]{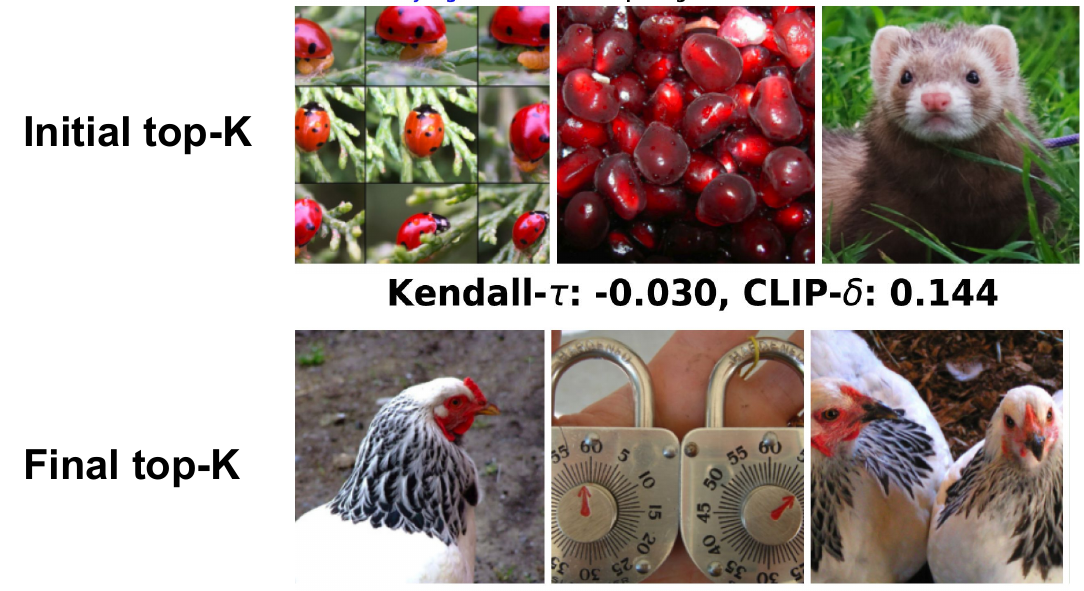}
    \caption{ Top images for a channel before and after a single-channel Push-Down attack.
    }
    \label{fig:single_channel}
\end{wrapfigure}

\textbf{Warm-up: Single-Channel Attack.} To set a first evaluation point for our attack framework, 
 we apply the push-down attack to one channel
Figure~\ref{fig:single_channel} shows the visualization of top images before and after.
 We can see that after optimization, the top-$k$ activating images of the neuron have been completely replaced by other images with different semantic concepts, suggesting a succesful attack with almost nearly no loss in accuracy (it decreases by $0.04\%$). 
 One
 way of satisfying the attack objective perfectly in the single channel case is to set the channel weights to zero. 
 This
 naive solution only loses $0.2\%$ is to simply set all the weights of the channel to zero. Specifically removing channel 0 (by masking) decreased the accuracy by $0.2\%$. We thus consider more challenging settings.

                    
\setlength{\tabcolsep}{2pt}
%
\textbf{All-Channel Attack.}
Unlike the single-channel attack, the all-channel attack (change all neuron interpretation in a layer) does not have a trivial solution. Because some information needs to flow through the layer in order for classification to be successful, setting all channels to zero would result in catastrophic performance loss.

We apply our attack framework to \textit{Conv5} of the AlexNet Model. In Figure~\ref{fig:alexnet_conv_5_all_channel} we show a selection of 3 channels and the modifications achieved under the All-Channel Push-Down attack and the aggregate metrics (averages for all channels in a layer) are shown in Table~\ref{tab:main_overall}. More visual examples are provided in the Appendix.
For the visualized channels (and those in Appendix) we observe a near complete replacement of the top-$5$ images by other images.
\begin{wraptable}{r}{0.6\textwidth} 
\vspace{-10pt}
\scriptsize
    \center
\begin{tabular}{lrrrrr}
\toprule
Layer/Attack &CLIP-$\delta$& Kend-$\tau$ & CLIP-W & Kend-$\tau$-W & Acc.($\%$) \\
\midrule

Conv1 Push-Down &0.043&0.682&0.996&0.302&56.1 \\
Conv2 Push-Down &0.056&0.612&0.994&0.151&56.3 \\
Conv3 Push-Down &0.127&0.573&0.963&0.130&56.1 \\
Conv4 Push-Down &0.205&0.548&0.974&0.122&56.2 \\
Conv5 Push-Down &0.249&0.530&0.963&0.048&56.2 \\\hline
Conv5 Push-Up &0.150&0.654&0.962&0.011&56.3 \\\hline
EfficientNet L7 - Push-Down &0.262&0.503&0.971&-0.145&77.5
\\
\bottomrule
\end{tabular}
    \caption{\small Average (over channels) attack metrics for an All-Channel Push-Down and Push-Up Attack for AlexNet (row 1-6) and EfficientNet (row 7). We observe that the relative whack-a-mole metrics are low, suggesting this problem is not present for our attacks. Lower layers are more challenging to attack leading to lower CLIP score 
    and higher Kendall-$\tau$ as confirmed by visual intuition.}
    \label{tab:main_overall}
    \vspace{-10pt}
\end{wraptable}
Further, the labels of the top images significantly change, with minimal to no residual overlap. This suggests that not only the images have changed but the semantic concepts that would be determined by an interpreter have likely changed. This is opposed to the model simply memorizing images to reduce and replacing them with semantically similar ones. We further confirm this in the appendix by showing validation set top-$k$ images which demonstrate that semantically they follow the same behavior as the training images (which are used for the actual attack). Overall, the attack seems to produce a generalized change in the behavior of the feature visualization of neurons. 

Studying the metrics comparing the channels before and after modification, we can deduce several different behaviors. The first two channels exhibit relatively high Kendall-$\tau$ scores, from which we conclude that the ordering of image activations has not undergone severe changes. This means that likely only a subset of images, which includes the initial top-$k$ has moved in rank. Studying the CLIP distance in both cases allows us to conclude that there is significant semantic overlap in the initial and final top-$k$, which can be confirmed by visual inspection.

This is in contrast to the channel shown at the right, where the Kendall-$\tau$ score is close to zero, indicating a full re-ordering of the activations. As a consequence, the CLIP distance from initial to final is also much higher, which matches with a visual inspection.

In general, we observe a substantial correspondence between our visual intuition and the CLIP-$\delta$ and Kendall-$\tau$, channels with low scores Kendall-$\tau$ and high CLIP-$\delta$ tend to change substantially. As illustrated in further examples in the Appendix one observed difference in these two metrics is that channels maintaining some similar classes in the top images will tend to have a lower CLIP-$\delta$ (suggesting less change). 

\begin{figure}
\begin{minipage}{.55\textwidth}
    \centering
\vspace{-5pt}
    \centering
        \begin{subfigure}[b]{0.46\linewidth}
		\includegraphics[width=\linewidth]{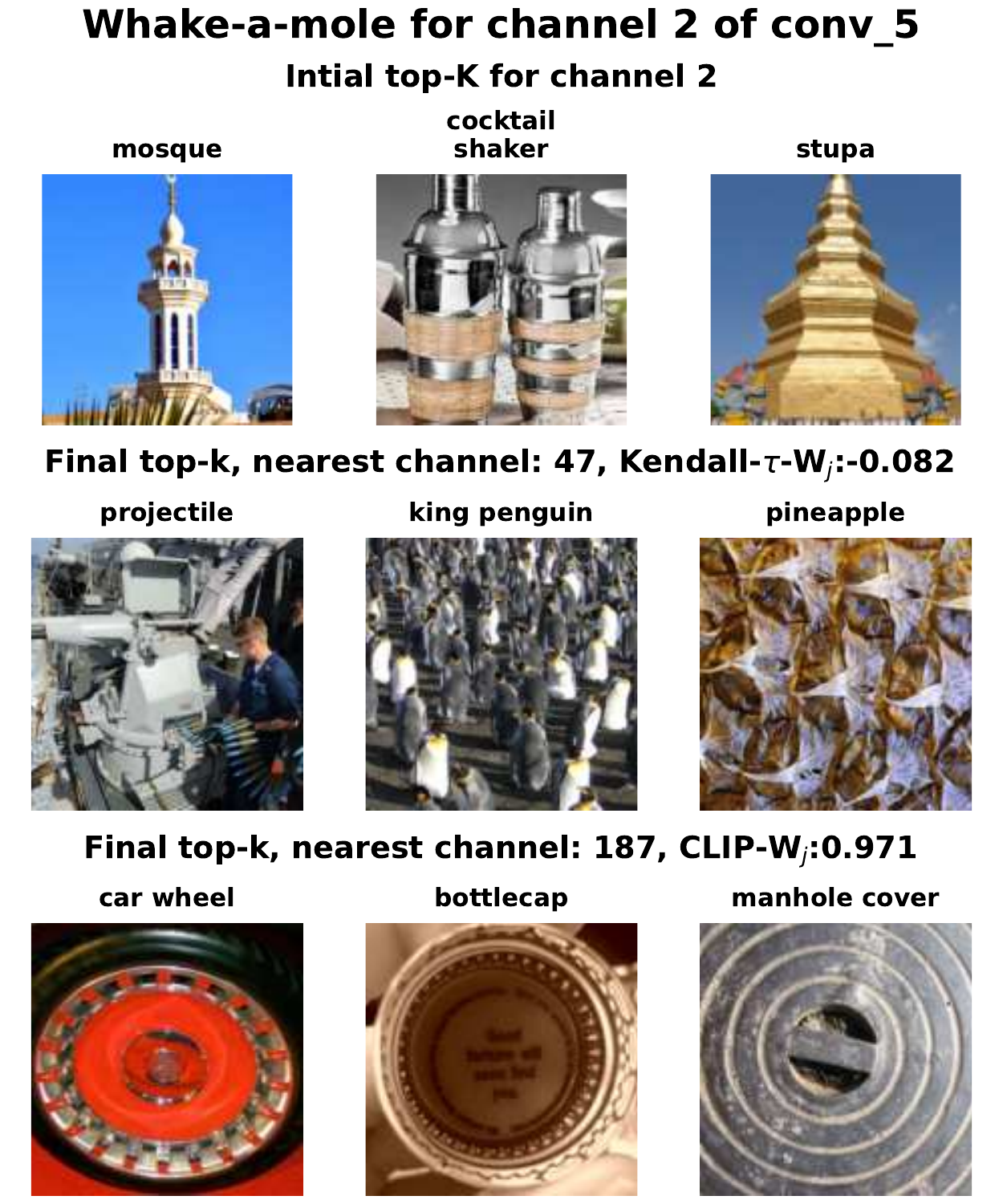}
		\label{fig:whake_a_mole_1}
    \end{subfigure}
         \begin{subfigure}[b]{0.48\linewidth}
		\includegraphics[width=\linewidth]{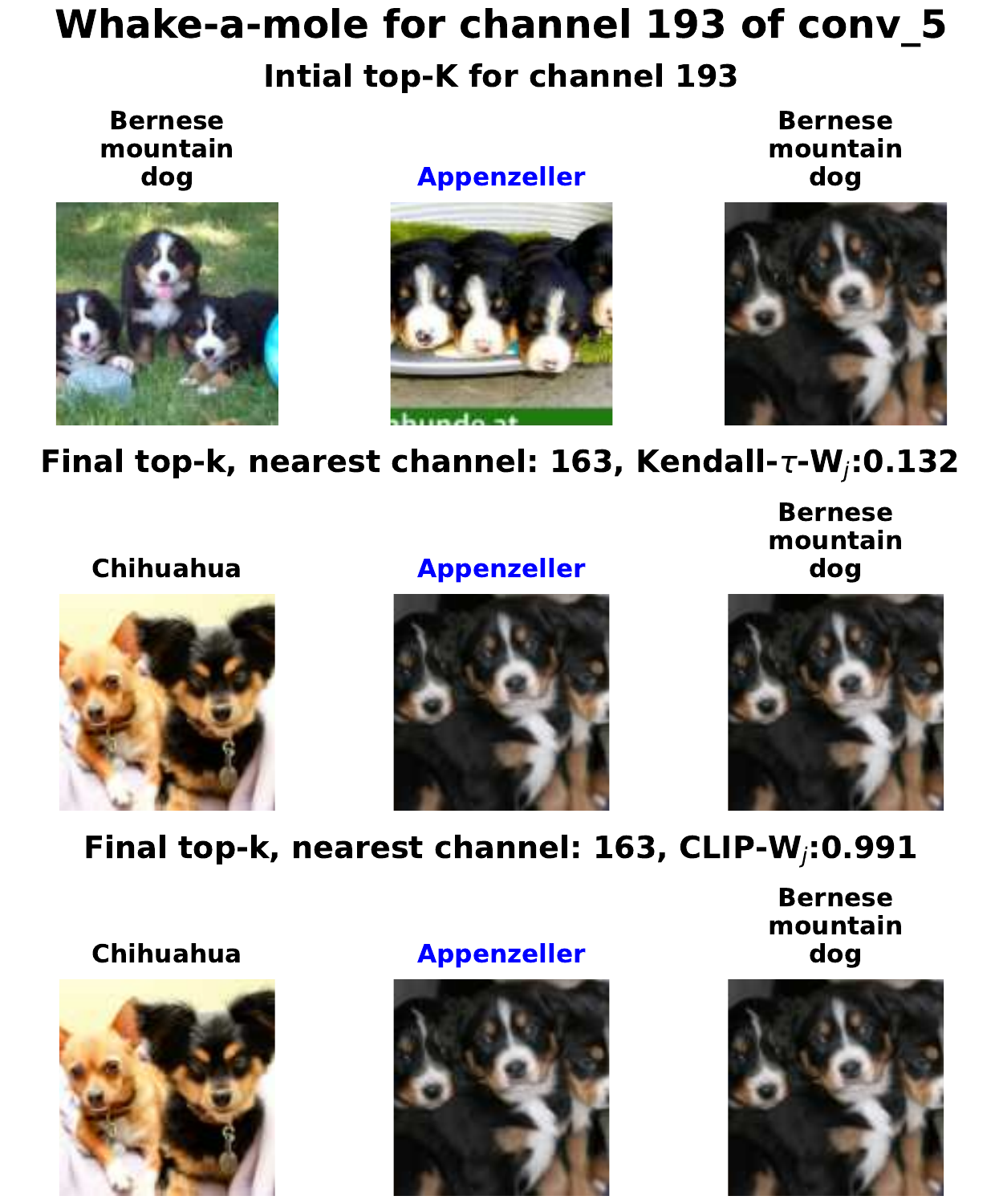}
		\label{fig:whake_a_mole_4}    
    \end{subfigure}
\vspace{-15pt}
    \captionof{figure}{\small We show the initial top images for two channels and beneath are the corresponding final top images of  closest channels w.r.t Kendall-$\tau$-W$_j$ and  CLIP-W$_j$.\vspace{-1pt}}
    \label{fig:whack-a-mole}
\end{minipage}%
\hfill
\begin{minipage}{.4\textwidth}
   \vspace{-28pt}
   \centering
\includegraphics[width=0.9\textwidth]{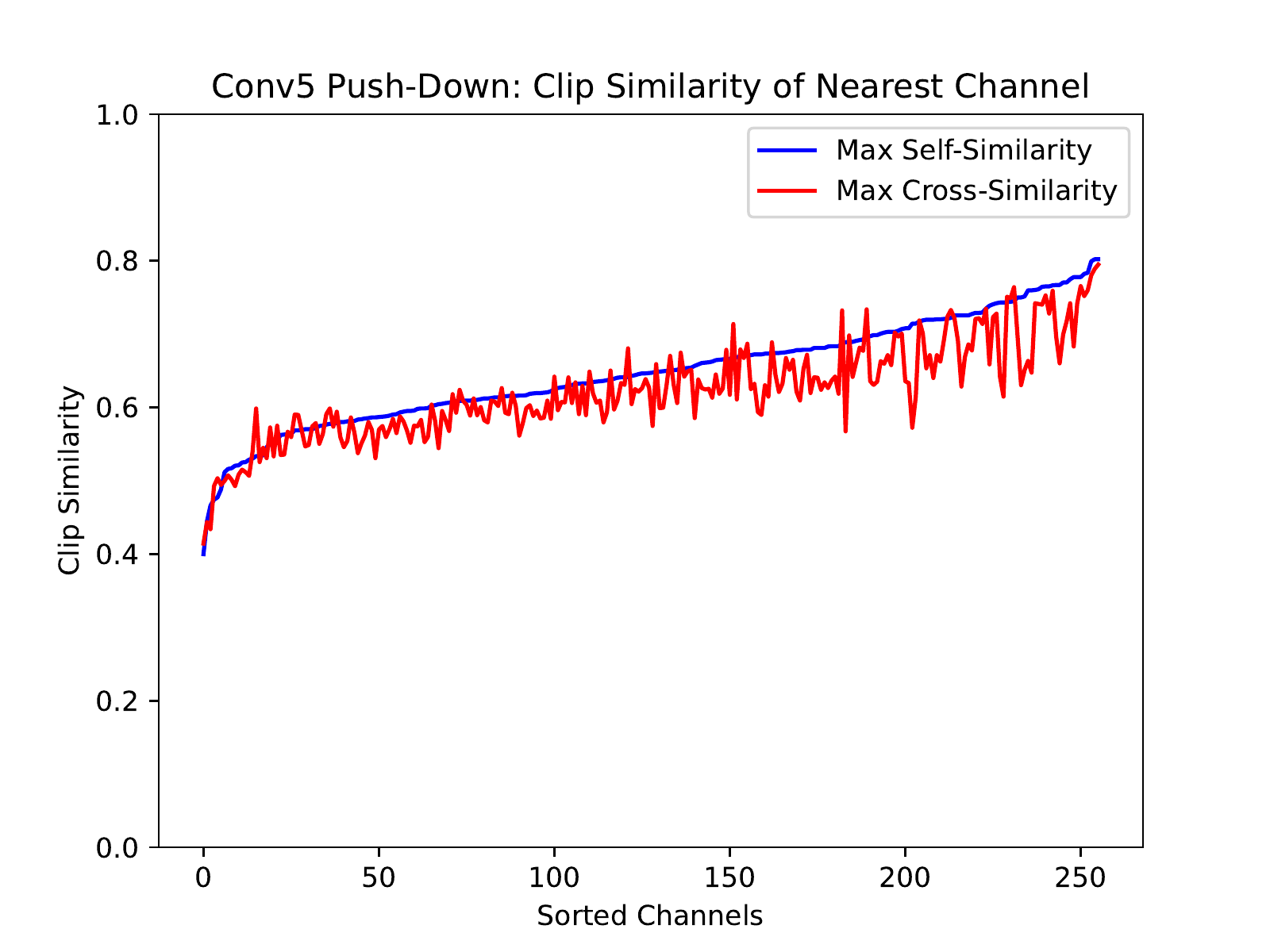}
\captionof{figure}{\small We compare initial CLIP similarity to other channels (blue) versus similarity after attack (red). Red and blue largely track each other for all channels. 
} 
        \label{fig:geo_means_clip_sims}
\end{minipage}\vspace{-21pt}
 \end{figure}
\textbf{Whack-a-mole.}
We can further analyze the existence of the whack-a-mole problem by observing Fig.~\ref{fig:whack-a-mole} which shows for a channel in the original model, the top-K image in the modified model which have the closest Kendall-$\tau$-W and CLIP-W scores (not including the channel itself).

We observe that the first channel (channel 2 on figure) has little to no visually discernable similarity to nearby channels in the modified model as well confirmed by the Kendall-$\tau$-W. Indeed a majority of the channels look like this (see Appendix). On the other hand, we do observe similar images for the initial channel 193 and its nearest final one (163), which was picked as the most illustrative examples ("hard" one) where the red curve of Fig.~\ref{fig:geo_means_clip_sims} is above the blue one. However, for this "hard" example, more insight is given by investigating the CLIP-$W_j$ where the denominator notably measures the clip similarity to other channels in the original model. The score is less than or typically close to 1 suggesting that the original model already had a high similarity to another channel. Indeed in the Appendix for the second example, we confirm there is a very similar channel in the original model. To gain further insight into CLIP-$W_j$ in Fig.\ref{fig:geo_means_clip_sims}, we further visualize the numerator and denominator for all the channels (red line) and sort them by the initial similarity to other channels (denominator). We observe that the red line is often below the blue line and if it exceeds it is not by a large relative amount, suggesting that channels with high whack-a-mole metrics are actually ones that already had similarities to other channels in the original model. Overall we conclude the presence of the whack-a-mole problem is minimal in our current attack. 

\begin{figure}[h]
    \centering
  \includegraphics[width=\textwidth]{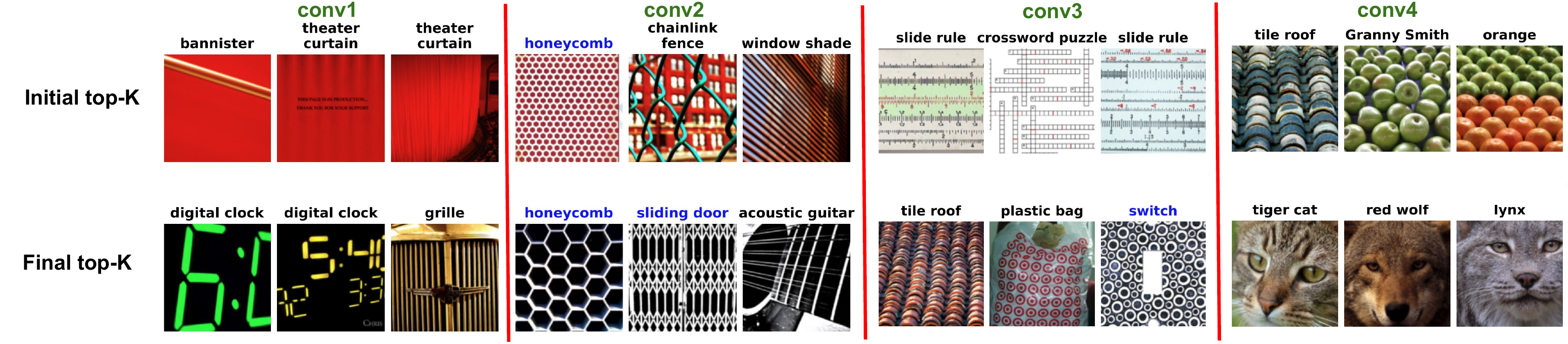}
    \label{fig:top_images_layer_ablation}
    \caption{Push-down attack on AlexNet across several layers.
    Channels are taken individually on each layer for layer ablation, and the results demonstrate that the top images are potentially vulnerable across all layers. 
    The final attacked models all have a less than .5\% drop from a default AlexNet.}
    \label{fig:top_images_ablation}
\end{figure}

\textbf{Effect of Depth.} We now consider how the attack is affected by depth, with results for different layers of AlexNet shown in Tab. \ref{tab:main_overall} and illustrated in Fig. \ref{fig:top_images_ablation}.
We observe that modifications of the earliest layers are significantly harder to achieve than for later layers as confirmed by the metrics and visual examination. We also observe a qualitative difference in the changes. For example, Conv$_1$ and Conv$_2$ are picking up low-level information such as color, edges, and textures and this is reflected in the type of modifications made to the images.  If performance is maintained after the attack, it is likely that the modification objective did not have a strong impact, leading to little to no modification. This is reflected in the CLIP-$\delta$ scores (see Table~\ref{tab:main_overall}) and in visual examination (see Appendix for further examples). Several explanations can account for this. Firstly, there are fewer or no modifiable weights upstream to the attacked layer, leading to  less flexibility to accommodate the competing natures of the combined objective compared to later layers.
Secondly, the early-layer features, while somewhat malleable, must collectively perform a certain set of signal-filtering operations in order to be able to extract meaningful information. Performing strong modifications to the filters may lead to unrecoverable information loss downstream.
We observe that the whack-a-mole metrics are also relatively high for this case using Kendall-$\tau$-W. On the other hand, the normalized CLIP-W score is close to 1 suggesting that this increase is not due to behavior being moved into the channel but due to existing redundancy in channels. 

\textbf{Push-Up Decoy Attack.}
We study a more targeted attack objective, namely one that actively pushes a set of selected images into the top activating images for every channel. This is achieved with 
Eq. \ref{eq:push-up-attack-loss}, where the loss is non-zero as long as there exist images outside the group of selected images that activate higher than the group we intend to push up.

This type of attack is more targeted and therefore likely harder than the push-down attack, which does not specify what images the top-$k$ should be replaced with.
The push-up attack, if successful, can assign the same interpretation to every channel in a layer, making any interpretation attempt based on top-$k$ images fraught, or at least minimally informative.

\begin{figure}[t]
    \centering
    \includegraphics[width=0.42\linewidth]{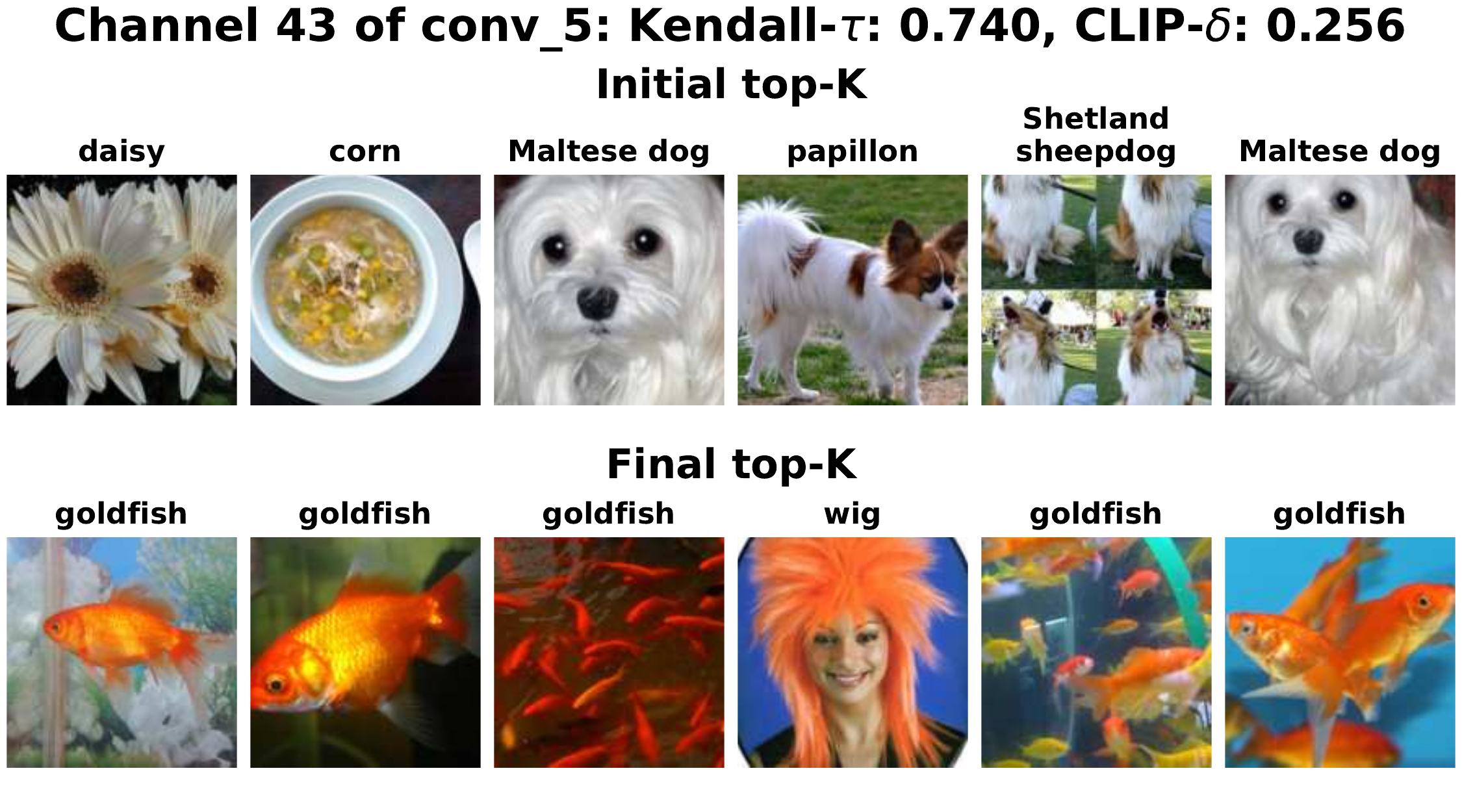}\qquad
    \includegraphics[width=0.42\linewidth]{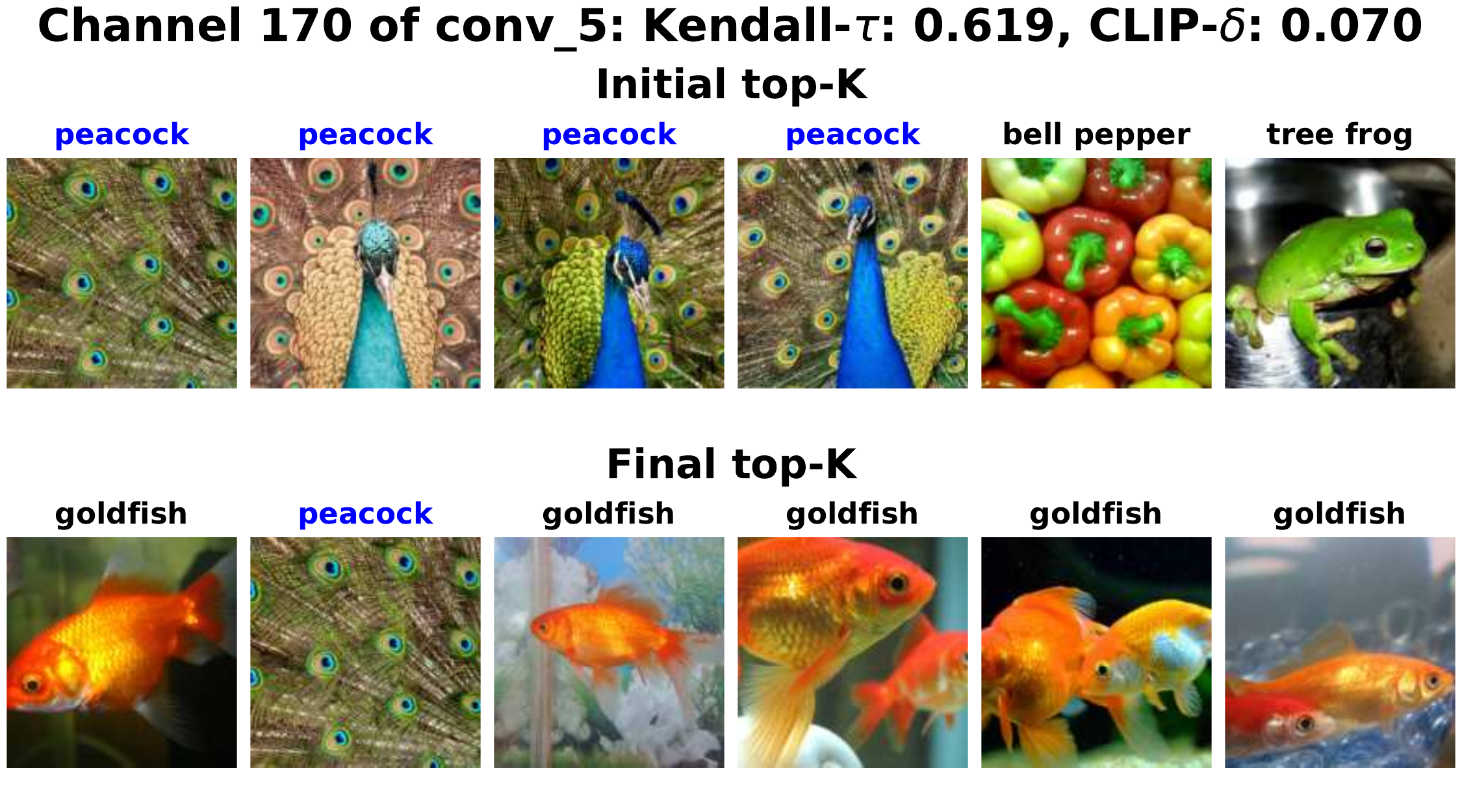}
    \caption{Examples of channels in all-channel push-up attack. The top images were successfully put in top images. The Kendall-$\tau$ remains relatively high ($>0.5$) suggesting much of the channel behavior is preserved while the top activating images completely obfuscate the behavior.}
    \label{fig:push_up}
\end{figure}

Fig. \ref{fig:beforeafter} shows the result of the push-up attack using a collection of images with the Imagenet label ``Goldfish'' as the decoy set.
Further, in Fig. \ref{fig:push_up} we show that for many channels of a layer, we can modify the top-$10$ to contain a few or consist entirely of Goldfish images. The metrics in Table~\ref{tab:main_overall} also demonstrate substantial change and a low likelihood of whack-a-mole behavior.
Studying the figure more closely, we observe that not only Goldfish, but also other images that share certain traits with the Goldfish images are also boosted, suggesting a degree amount of generality of the newly imposed selectivity, further explored in the Appendix. 

\begin{figure}
\centering
\includegraphics[width=0.65\textwidth]{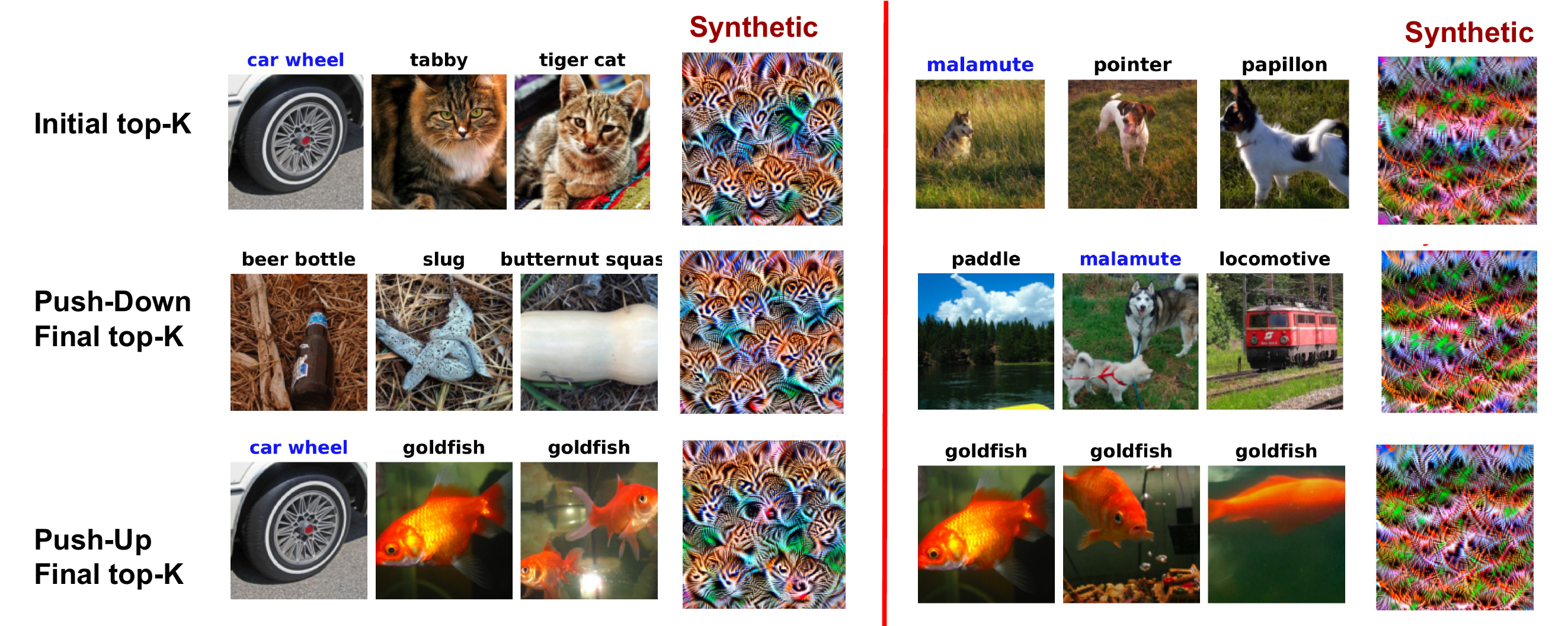}
    \caption{ Synthetic feature visualization after our attack. We observe the visualization is largely decorrelated to top-$k$ natural images.}
    \label{fig:synthetic}
\end{figure}
\subsubsection{Synthetic Feature Visualization}\label{sec:synthetic_feature_visualization}
We study the impact of the Push-Down and Push-Up attacks 
on the synthetic activation-maximizing images of the channels under attack \cite{zeiler2014visualizing}.
%
Synthetic
activation-maximizing images are the result of an optimization problem over input pixels 
solved
by gradient ascent on the channel activation under 
a norm constraint in pixel space.
To avoid adversarial noise samples \cite{goodfellow2014explaining} it is necessary to jitter the input image or parameterize it as a smooth function\cite{olah2017feature}.

In Fig. \ref{fig:synthetic}, we study the synthetic optimal images for several channels before and after the attack.
%
By visual inspection, while the top-$k$ images change drastically, the synthetic optimal image is largely unaffected.
%
The most common observed change (see also Appendix) for $conv5$ is a low-frequency modulation of the pattern. We hypothesize that this is 
because
the top-$k$ attack most significantly modifies the weights of the attacked layer, which is a later layer preceded by several downsamplings.
%

The lack of change in the synthetic optimal image suggests that the synthetic feature visualization and the top-$k$ analysis are, counter-intuitively, highly de-correlatable.
Observe, for instance, that the left-hand synthetic image suggests selectivity for cats even when most of the top-$k$ images are goldfish.
This is a worrying prospect for the top-$k$ interpretability method.
Further, this does not permit the conclusion that the synthetic optimal image is more robust to attack, since we have not explicitly run an attack against it.
Rather, this suggests the space of NN weights and the possible functions they span is quite large, and can possibly accommodate more functionality, and attacks, than one might expect.

\subsection{Fairwashing Feature Visualization}\label{sec:fairwashing_results}
\begin{figure}
\begin{minipage}{.52\textwidth}
    \centering
		\includegraphics[width=0.95\linewidth]{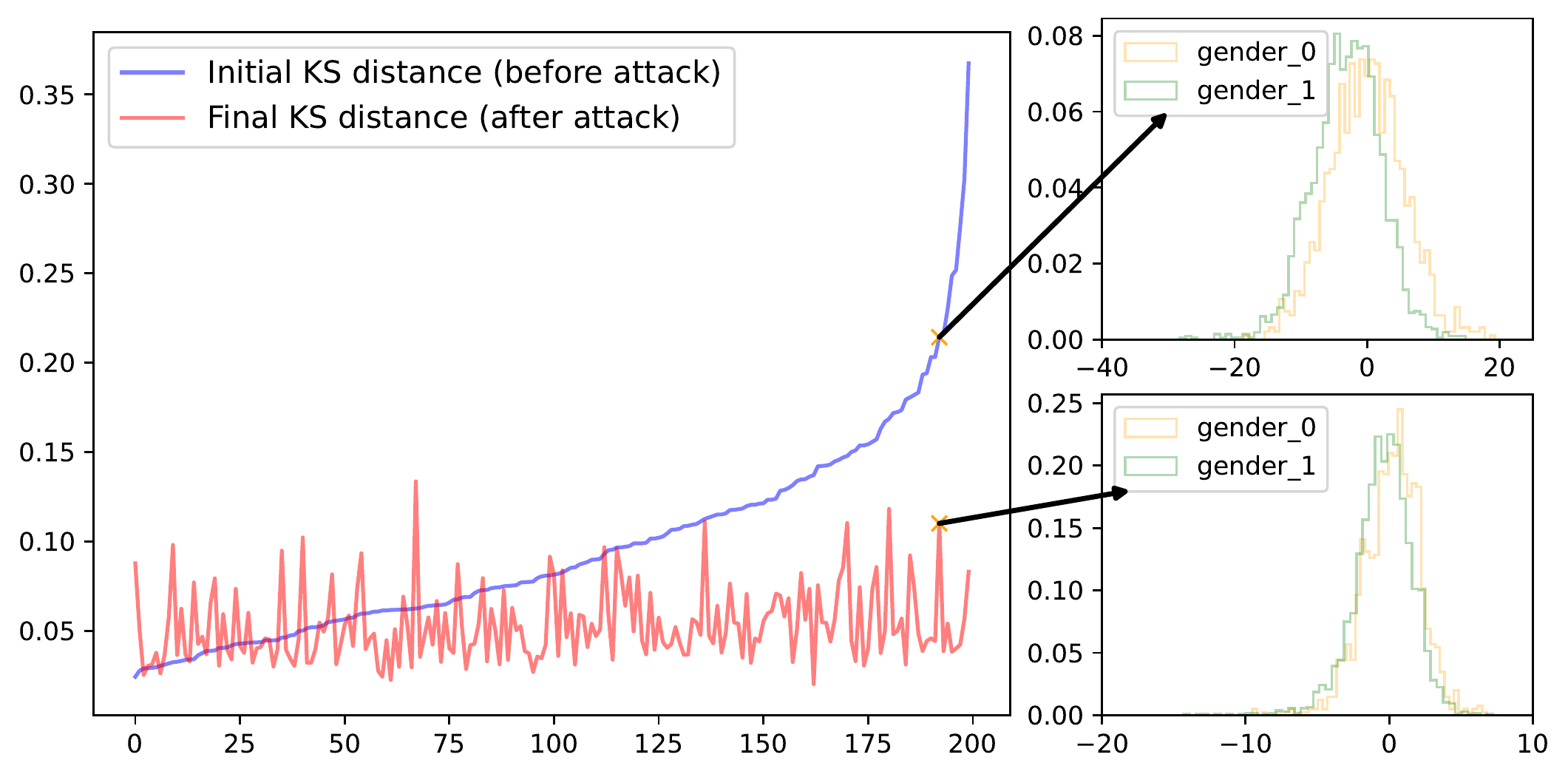}\vspace{-5pt}
		\captionof{figure}{\small Kolmogorov-Smirnov (KS) distance between the conditional distributions of each condition estimated on the annotated testing set. We sort the channels based on the initial KS and observe that after our fairwashing interpretability attack,  each channels KS is drastically reduced.}
		\label{fig:KS_distance}
\end{minipage}%
\hfill
\begin{minipage}{.45\textwidth}
    \centering
		\includegraphics[width=0.65\linewidth]{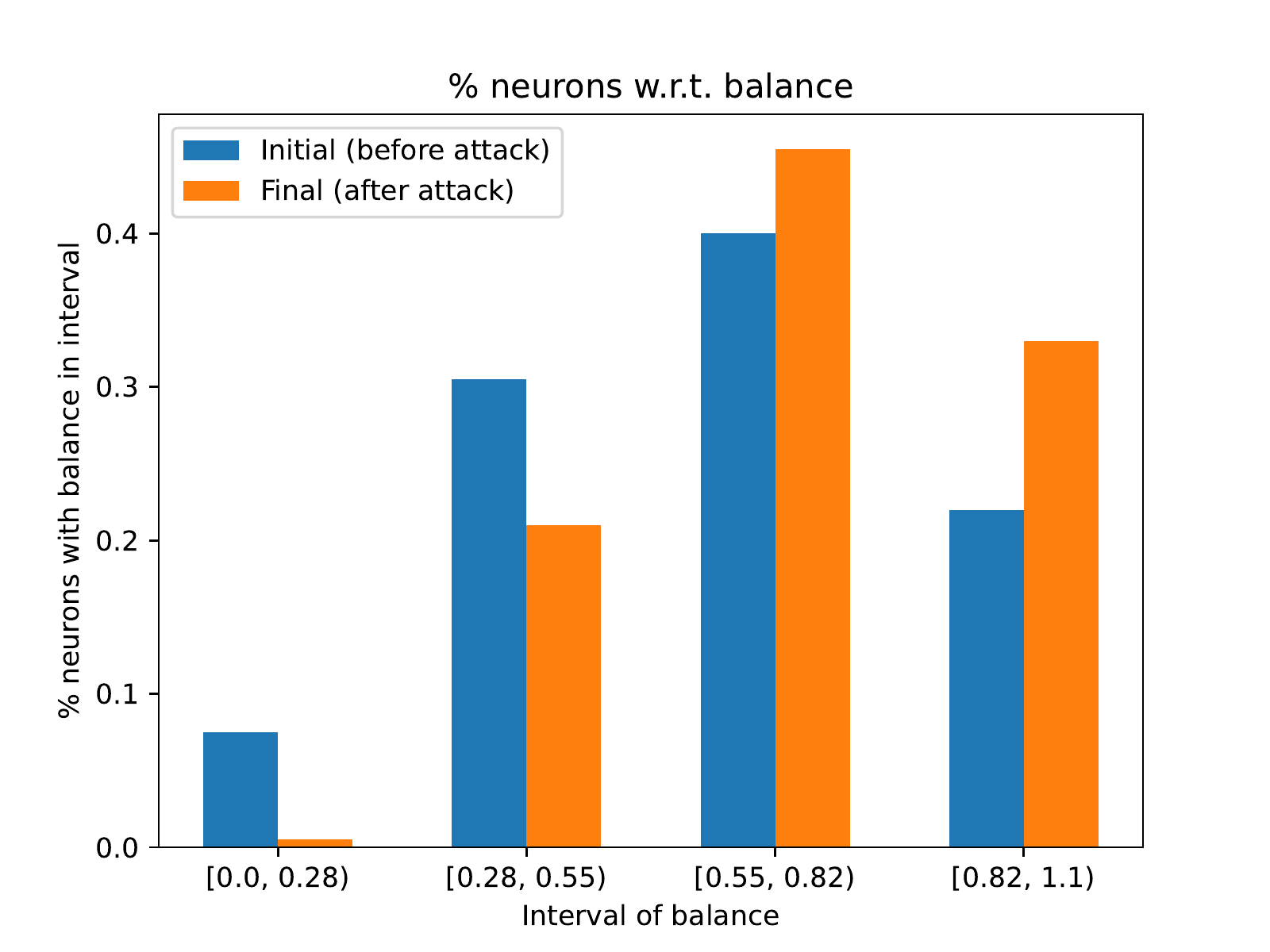}
		\captionof{figure}{\small Percentage of the neurons according to their balance over the annotated testing set. After the attack, the percentage of neurons with low balance has decreased while the percentage of neurons with high balance has increased.}
		\label{fig:test_balance}    
\end{minipage}%

\end{figure}
\begin{figure}[t]
\centering
    \begin{subfigure}[b]{0.67\linewidth}
		\includegraphics[width=\linewidth]{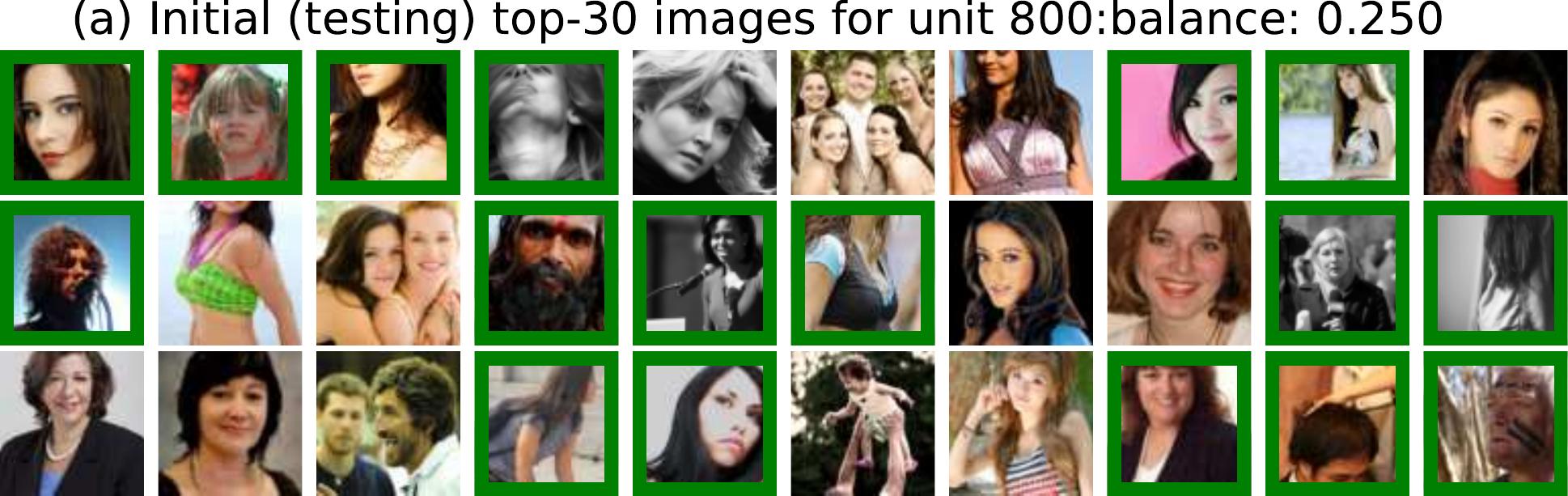}
		\label{fig:initial_images_test}
    \end{subfigure}\\ 
    \begin{subfigure}[b]{0.67\linewidth}
		\includegraphics[width=\linewidth]{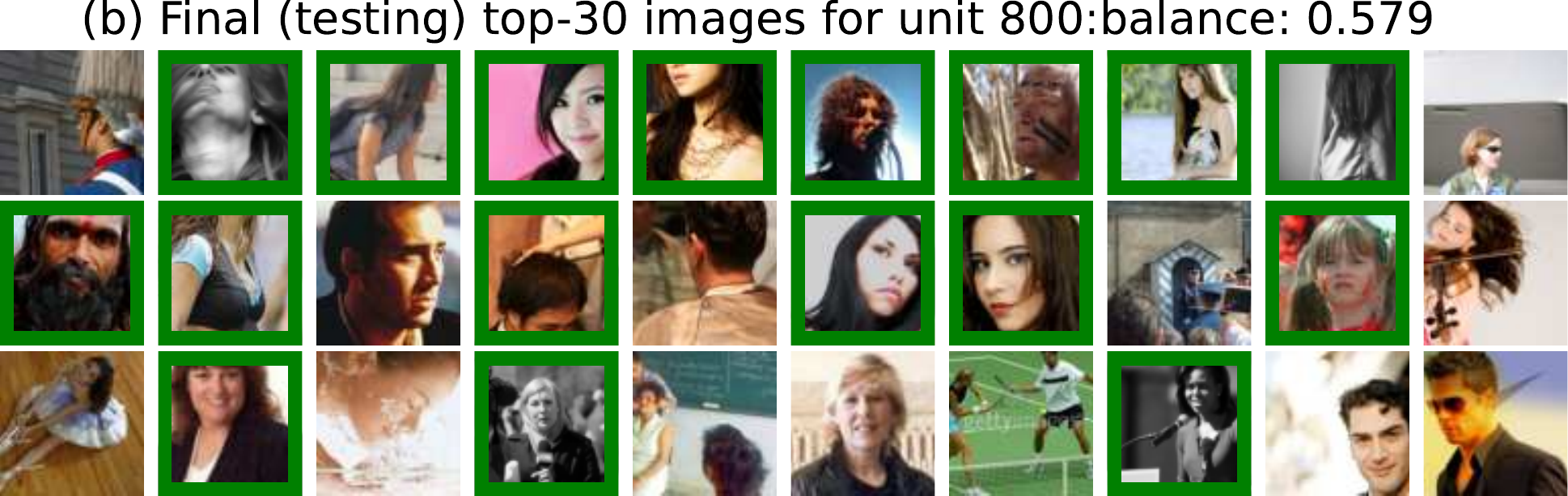}
		\label{fig:final_images_test}    
    \end{subfigure}
    \caption{ Top-30 images obtained for unit 800 of the last but one layer of AlexNet. Green is used for images that stay in top-30 images after attack. Before the fairwahsing attack, (a) the initial top-30 images are gender-biased. After the fairwashing attack, (b) the top-30 are less gender-biased: balance (fairness) measure has almost doubled. On the other hand, the model's unfairness has not changed.}
    \label{fig:fairwashing_unit_800}

\end{figure}
We demonstrate the application of our fairwashing attack for feature visualization as defined Sec.~\ref{section:fairwashing}. 
Given an \textit{unfair} (according to a certain metric of unfairness) model and a set of neurons whose top-activating images look \textit{unfair}, we ask ourselves whether it is possible, by fine-tuning, to make the new set of images for the same neurons appear \textit{fairer} while maintaining the same performance and bias of the initial model.
We instantiate this fairwashing attack on an annotated subset of Imagenet data~\cite{yang2020towards} (as described in Sec.~\ref{sec:experiments}) with gender as the protected attribute.
We first estimate the model unfairness of the pre-trained AlexNet model using DDI and DEO unfairness measures. 
Tab.~\ref{tab:fairwashing_table} reports these measures for the three \text{human} classes of the ImageNet-1k dataset on which AlexNet is trained. According to this table, the initial AlexNet model is not totally fair, with the largest values of unfairness on the \textit{Baseball player} class.
\begin{table}
    \centering
    \begin{tabular}{rrcc|cc|cc}
       &      & \multicolumn{6}{c}{Class}\\
          \cmidrule(lr){3-8}
		&	 & \multicolumn{2}{c|}{Baseball player} & \multicolumn{2}{c|}{Bridegroom} & \multicolumn{2}{c}{Scuba diver} \\
        \cmidrule(lr){3-8}
		 & Acc. & DDI & DEO & DDI & DEO & DDI & DEO \\\hline
 Pre-Attack& 56.45 &  3.38 & 76.92&  2.67 &12.34 & 0.28 & 5.26\\
 Post-Attack & 56.56 &  3.14 &  73.07 & 1.90 &12.34  & 0.24 & 5.26\\
			\bottomrule 
	\end{tabular}
    \caption{Accuracy/fairness measures (DDI/DEO) computed respectively on the ImageNet val. set and on the annotated testing set. Both measures are relatively similar before and after the fairwashing attack while the model has decreased the bias perceived by the interpreter for feature visualizations.}
    \label{tab:fairwashing_table}
\end{table}
We identified 200 neurons of the last but one layer whose MILAN~\cite{hernandez2022natural} descriptions are related to humans (see Appendix for more details). 
We run our attack on all these neurons to prevent missing neurons whose biases may transfer to other ones. Fig.~\ref{fig:KS_distance} shows the results of Kolmogorov-Smirnov distance between the distributions of activations conditioned on the two gender groups. It can be observed that after the attack, this distance has been drastically reduced, especially for highly biased neurons. This suggests the balance of the top-$k$ is also improved.
As can be seen in Fig.~\ref{fig:test_balance}, the percentage of neurons whose top-$k$ images have a low balance (low \textit{fairness}) has decreased, while the percentage of neurons with high balance has increased, thus making feature visualization fairer. Moreover, according to Tab.~\ref{tab:fairwashing_table}, the model has almost the same accuracy and almost the same measures of unfairness (all cases $\le 1\%$ of relative difference for DDI and $\le 4\%$ for DEO). Note that our attack did not enforce any fairness constraint on the output, the maintain loss $\mathcal{L}_\text{M}$ described in Sec.~\ref{sec:attack_framework} was enough to also maintain model unfairness. We also depicted in Fig.~\ref{fig:fairwashing_unit_800} an example of a unit whose top-$k$ images were initially \textit{biased}, but have been fairwashed after running the attack by almost doubling the balance measure. More examples of training and testing sets can be found in the appendix. 

\section{Conclusions, Limitations, and Broader Impact} 
We demonstrated the adversarial model manipulability of feature visualization with top-$k$, proposing three attacks that pose varying threats.
We provide experimental evidence that supports the success of our attacks, with little to no evidence of a \textit{whack-a-mole} issue.  Our metrics to systematically detect the presence of whack-a-mole may be imperfect as validating them requires inspecting all channels to validate correspondence. Future work may consider investigation of synthetic feature maps and how they may be attacked and generalization of the fairwashing attack beyond binary attributes. 
\\
\textbf{Broader Impact.} 
The goal of our study has been to demonstrate a potential vulnerability in current interpretability methods and raise awareness of reliability and ethical risks. By showing the fairwashing attack, an apparent consequence is the possibility that an ill-intentioned individual uses this work to perform these attacks in order to release models that marginalize minority groups. However, we think that raising these risks is an essential first step towards addressing these vulnerabilities, and we hope our contributions provide a springboard for future discussion and protection efforts.  

\section{Acknowledgements}
We acknowledge support from OpenPhilanthropy and  resources provided by Compute Canada and Calcul Quebec. We also thank Kaiyu Yang for the access to annotations of the ImageNet Subtree People dataset.


\printbibliography

\newpage
\appendix

\section*{\centering \LARGE Appendix}

\section{Hyperparameters and Training Details}\label{app:hyperparams}
This section presents the details of the hyperparamters and training settings used to run our attacks.
\subsection{Push-Up and Push-Down Attacks}
We train for 2 epochs over the ImageNet-1k training set with a batch size of 256. We use the \textit{Adam} optimizer with learning rate {1e-5}. 

Regarding $\alpha$, we employ a dynamic updating rule inspired by \textit{Algorithm  1: Dynamical balancing of Distillation and CKA map loss} in appendix A of Davari et al's \cite{davari2022reliability} in order to have better control over loss in accuracy. We initialize $\alpha$ as $0.1$ (except for on the push-down attack for \textit{conv-2} where use $\alpha=0.01$ had more stable results).  If the accuracy loss is greater than 0.5\% we halve the current $\alpha$. If it is less than 0.1\% we double $\alpha$. With this dynamic update, we aim to minimize the loss in accuracy while still ensuring the top images shifts.
\subsection{Fairwashing Attack}
Similarly to push-up and push-down attacks, for the maintain loss, we use the ImageNet-1k training set. For the fairwashing attack, we need annotations for the protected attribute. We consider the set of $14865$ images derived from ImageNet-21k for which annotations of labeled demography (gender, race, and age) are available in the ImageNet People Subtree dataset~\cite{yang2020towards}.  
We use $75\%$ of these images (annotated training set) in the maintain loss and use the rest of $25\%$ images (annotated testing set) for fairness assessment.
We perform the attack with gender as the protected attribute and we binarize this attribute using the majority group defined as “male in the image''.

We also use the Adam optimizer with a learning rate of $1e$-$5$ and we use a batch size of $256$ for losses. No dynamic update for $\alpha$ was needed, and we keep it to $\alpha=0.1$, corresponding to the initial value of $\alpha$ for push-up and push-down attacks.

Finally, for the attack loss, we attack the last but one layer of AlexNet by considering the neurons (200 in total) whose MILAN~\cite{hernandez2022natural} descriptions likely relate to humans. We accomplish this by inspecting the neurons' MILAN descriptions to get neurons whose descriptions contain one of the following words ``faces'', ``skin'', ``person'', ``human'' and ``people''.

\begin{figure}[!ht]
	\includegraphics[width=\linewidth]{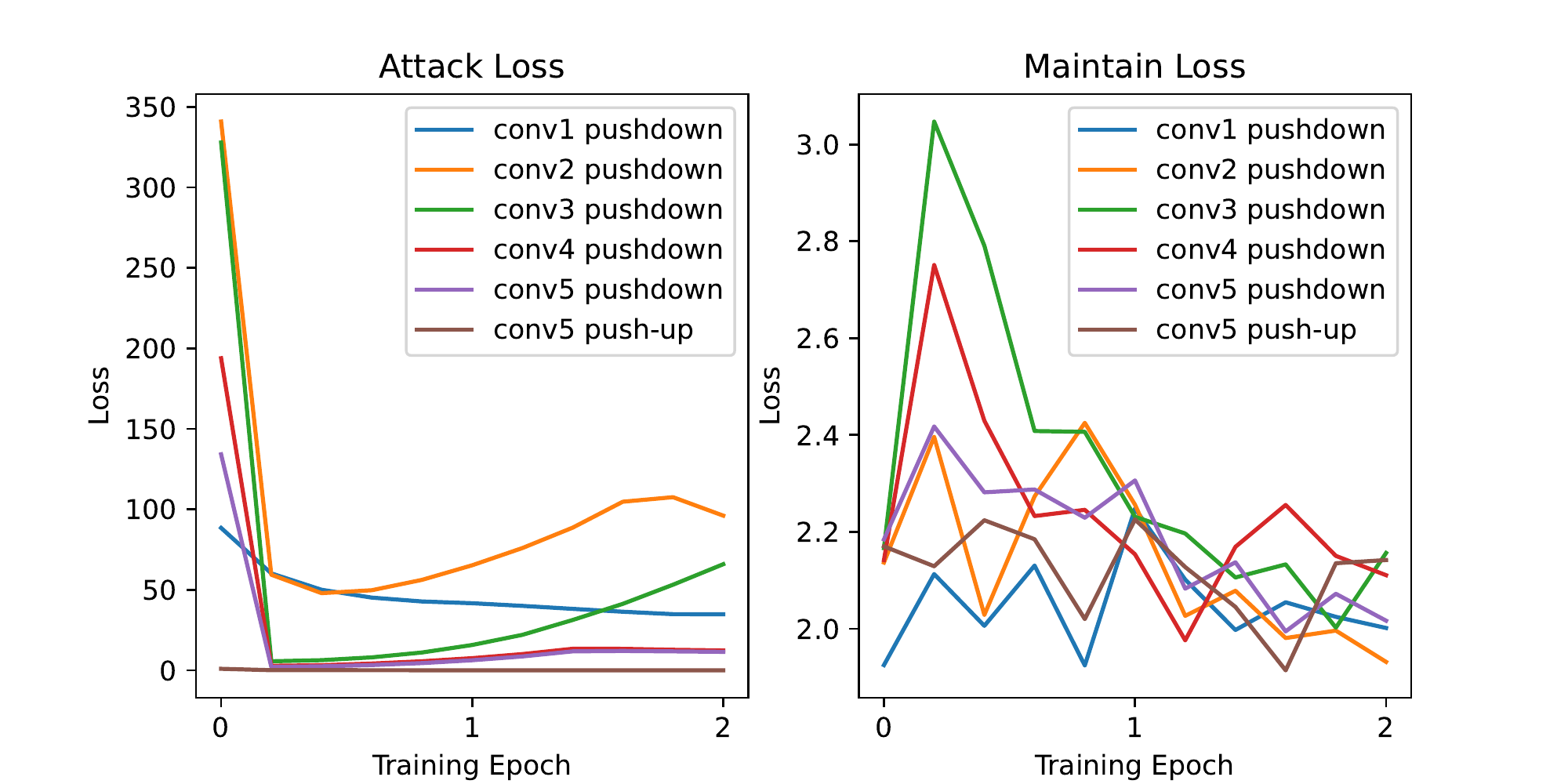}
	\caption{Sample training curves for the maintain and attack objectives. Late layers (conv5, conv4) are easier to attack compared to early ones (conv1, conv2 and conv3). The maintain loss is very close to its initial value after two epochs.}
		\label{fig:training_curves}

\end{figure}
\subsection{Optimization Curves}
We show in Figure~\ref{fig:training_curves} the evolution of attack and maintain losses across two epochs. It can be observed that the attack loss of late layers (conv 4, conv 5) decreases very quickly, and almost monotonically, showing the easiness to attack late layers. In contrast, early layers do not have the same behavior. We can also observe from the training curves that the maintain loss is almost close to its initial value after 2 epochs. This corroborates the observed accuracy preservation as shown in Table~\ref{tab:main_overall}.
\section{Additional Results}
This section shows additional illustrations and results for all the attacks.
\subsection{Additional Results for Push-down Attack on a Single Channel and on all Channels}
We show additional results for the push-down attacks on a single channel and on all channels simultaneously.
\subsubsection{Push-up Attack on Single Channel}
\begin{figure}
\begin{subfigure}[]{0.49\linewidth}
    \includegraphics[width=\textwidth]{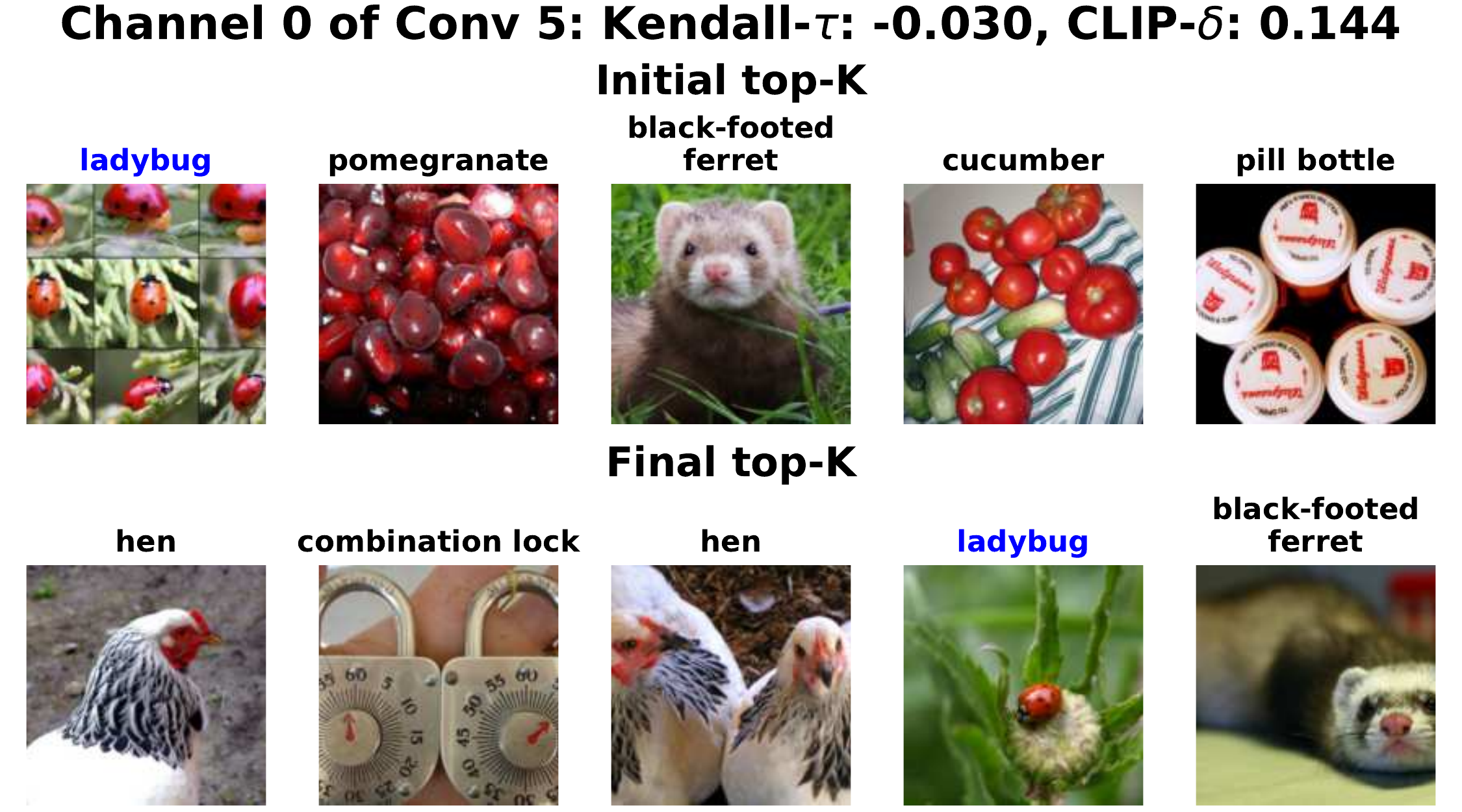}
\end{subfigure}\hfill
\begin{subfigure}[]{0.49\linewidth}
    \includegraphics[width=\textwidth]{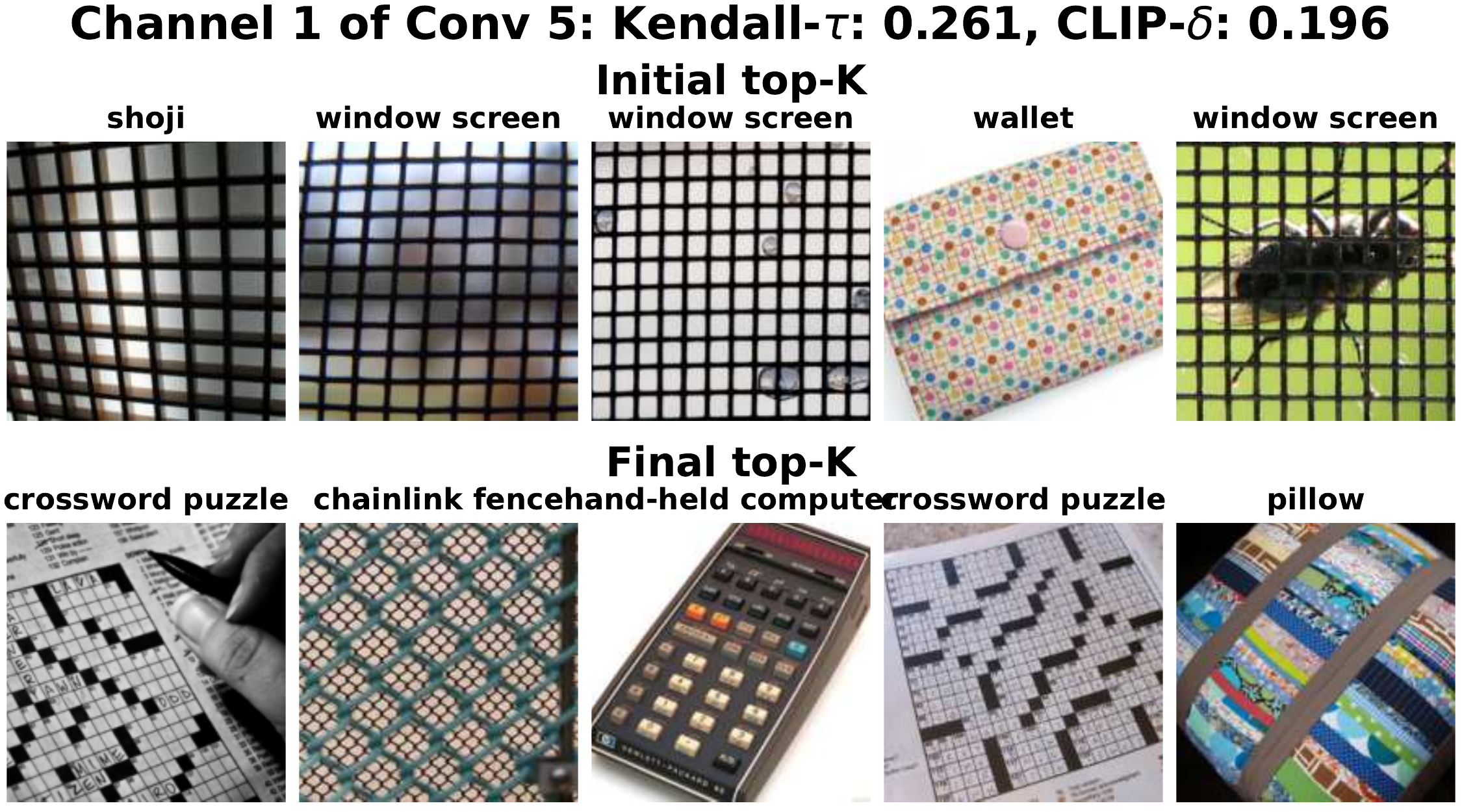}
\end{subfigure}\\
\vspace{.8cm}
\begin{subfigure}[]{0.49\linewidth}
    \includegraphics[width=\textwidth]{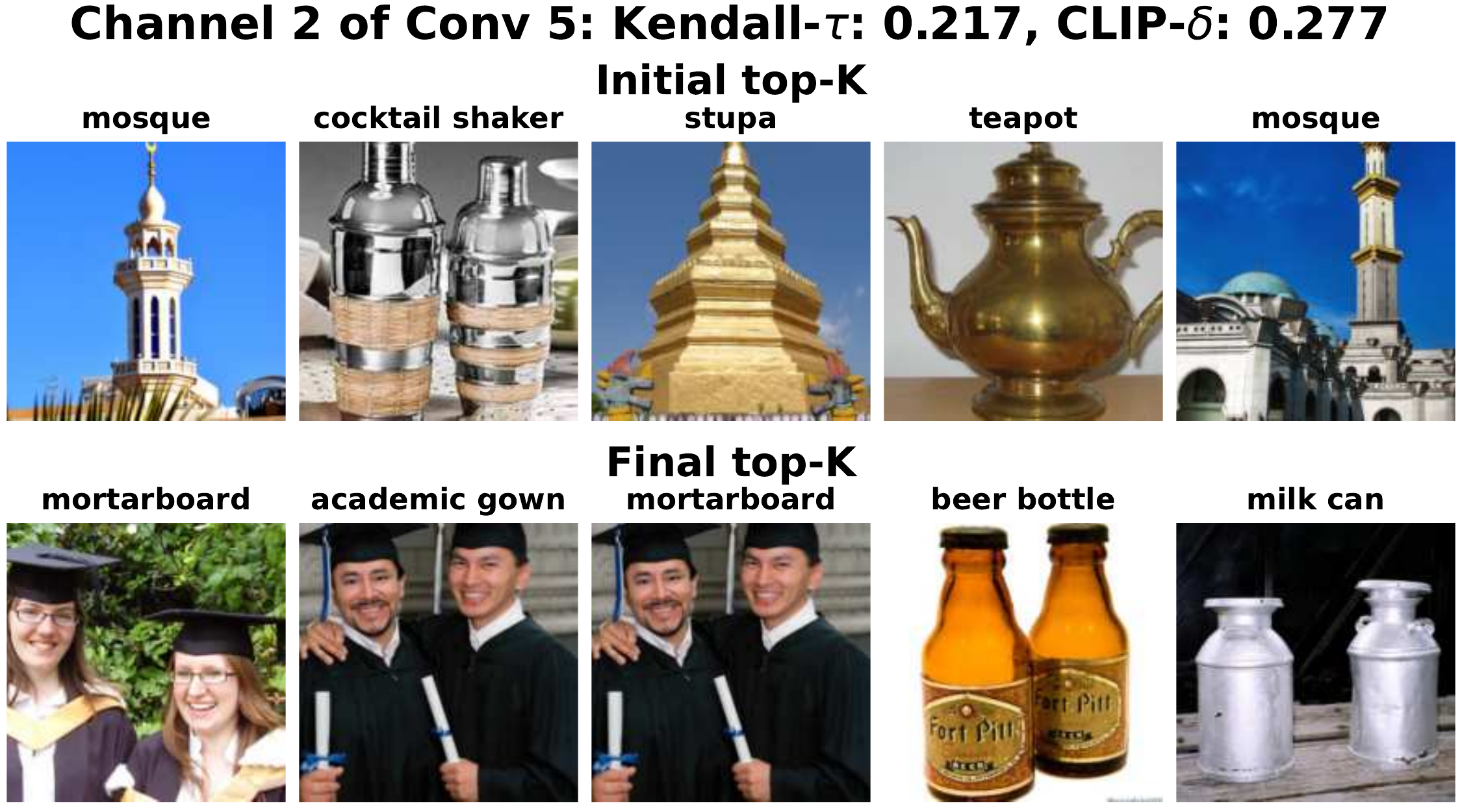}
\end{subfigure}\hfill
\begin{subfigure}[]{0.49\linewidth}
    \includegraphics[width=\textwidth]{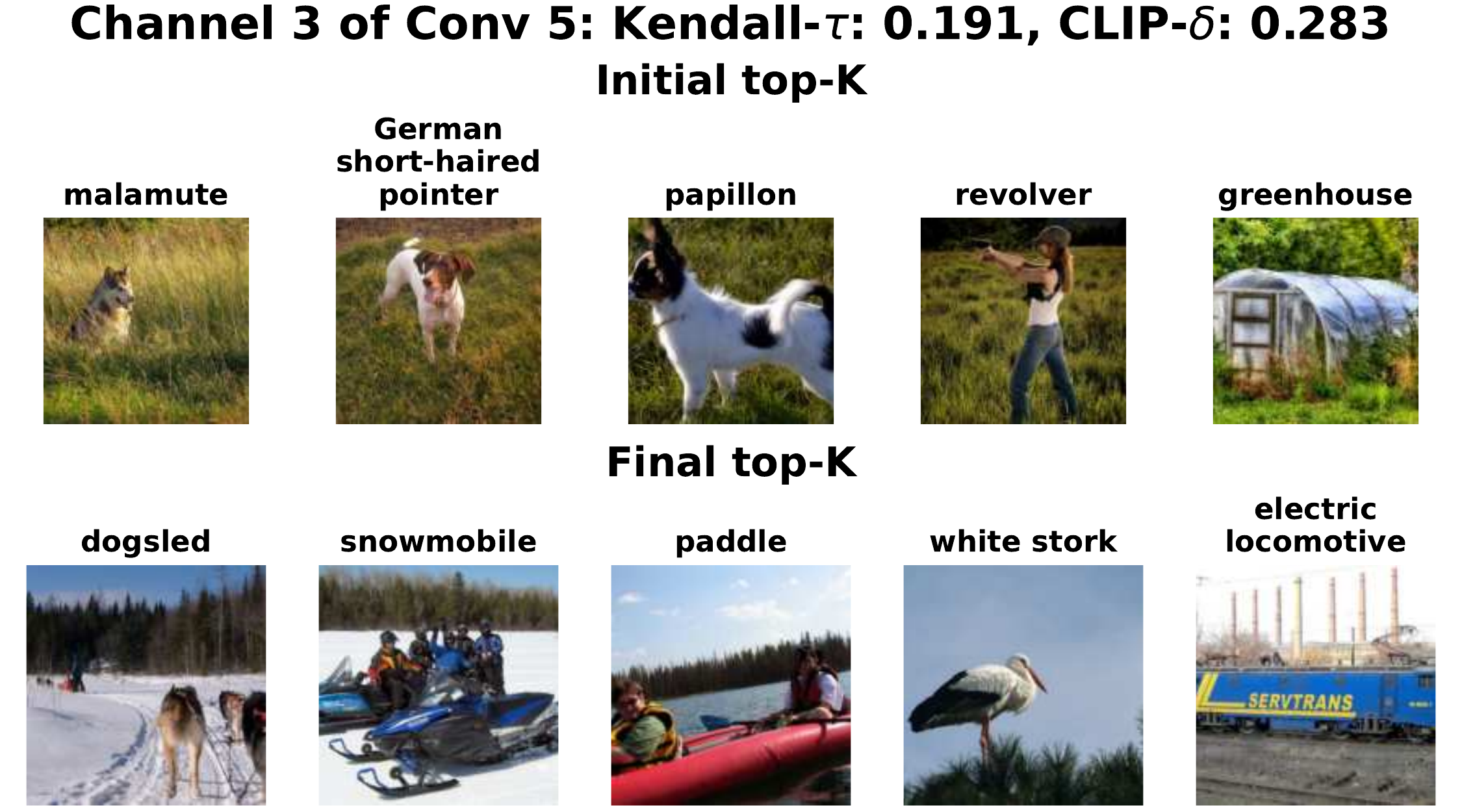}
\end{subfigure}\\
\vspace{.8cm}
\begin{subfigure}[]{0.49\linewidth}
    \includegraphics[width=\textwidth]{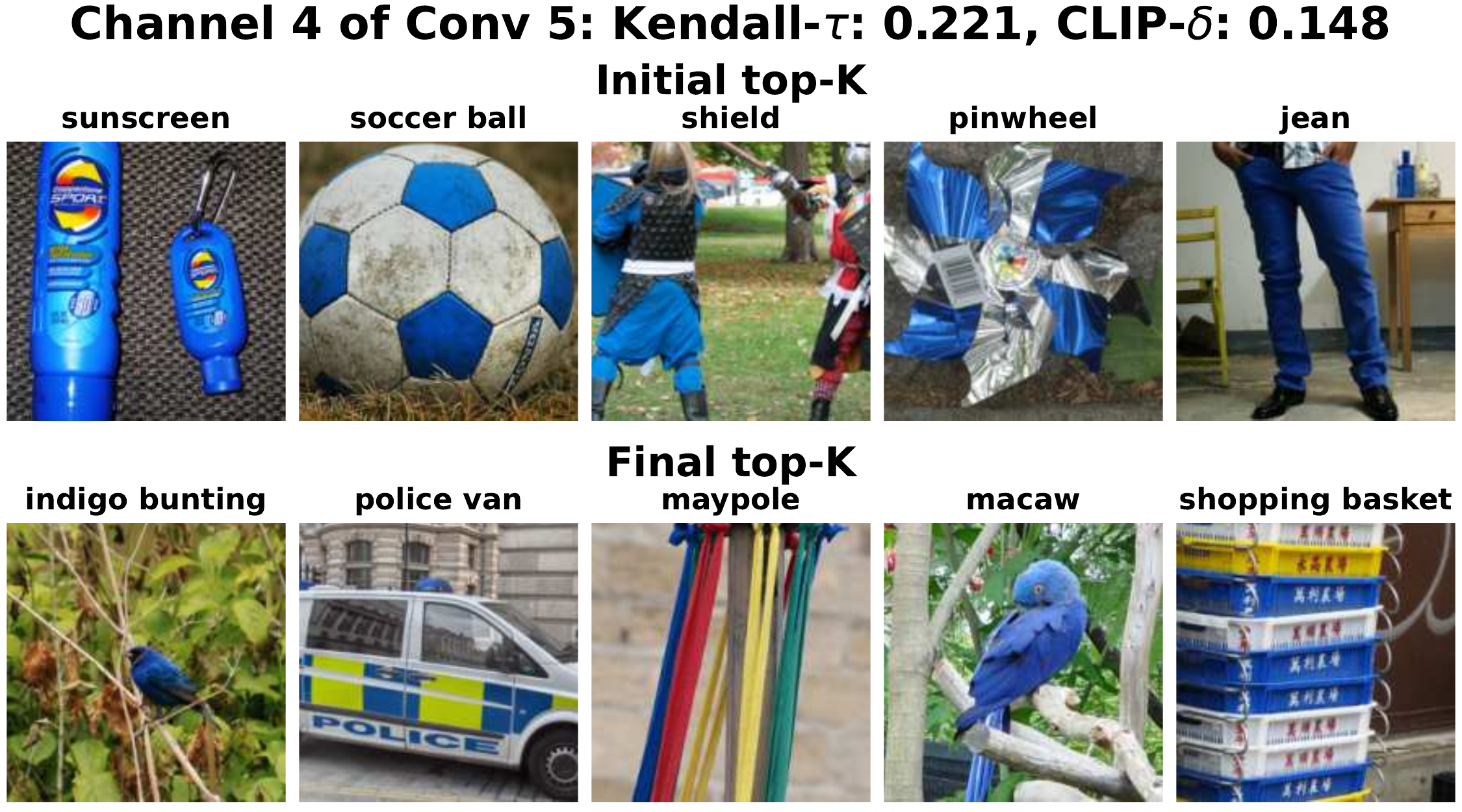}
\end{subfigure}\hfill
\begin{subfigure}[]{0.49\linewidth}
    \includegraphics[width=\textwidth]{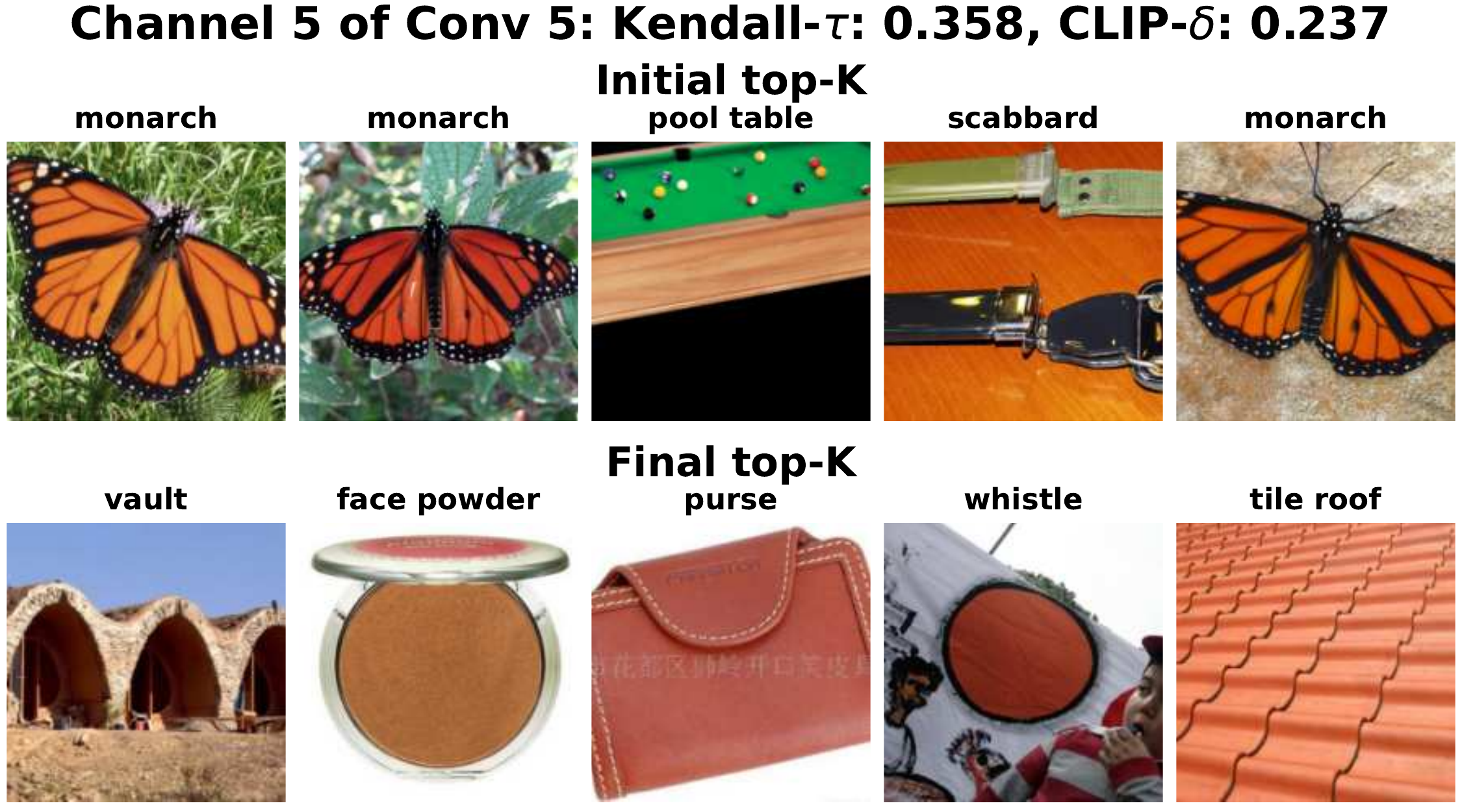}
\end{subfigure}\\
\vspace{.8cm}
\begin{subfigure}[]{0.49\linewidth}
    \includegraphics[width=\textwidth]{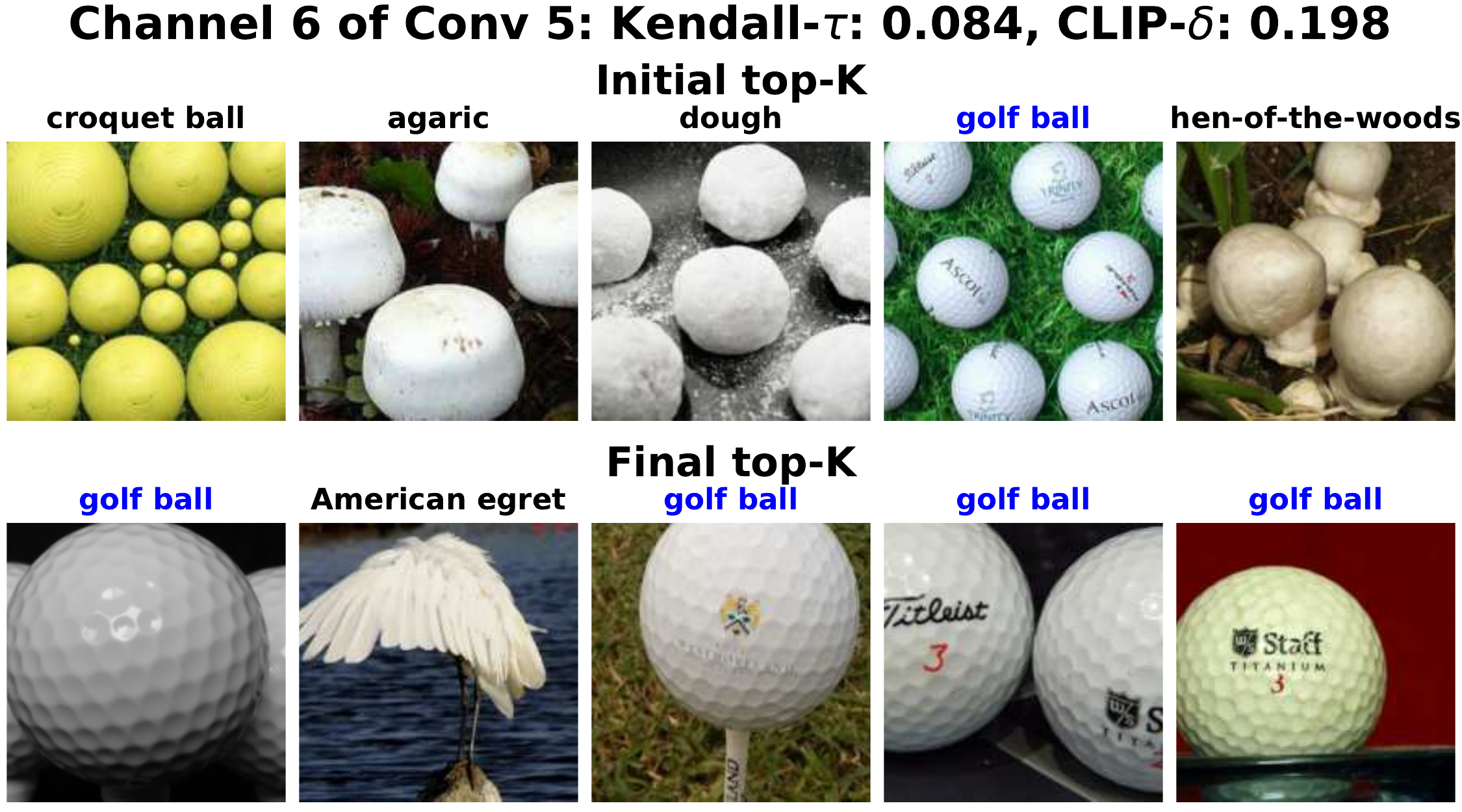}
\end{subfigure}\hfill
\begin{subfigure}[]{0.49\linewidth}
    \includegraphics[width=\textwidth]{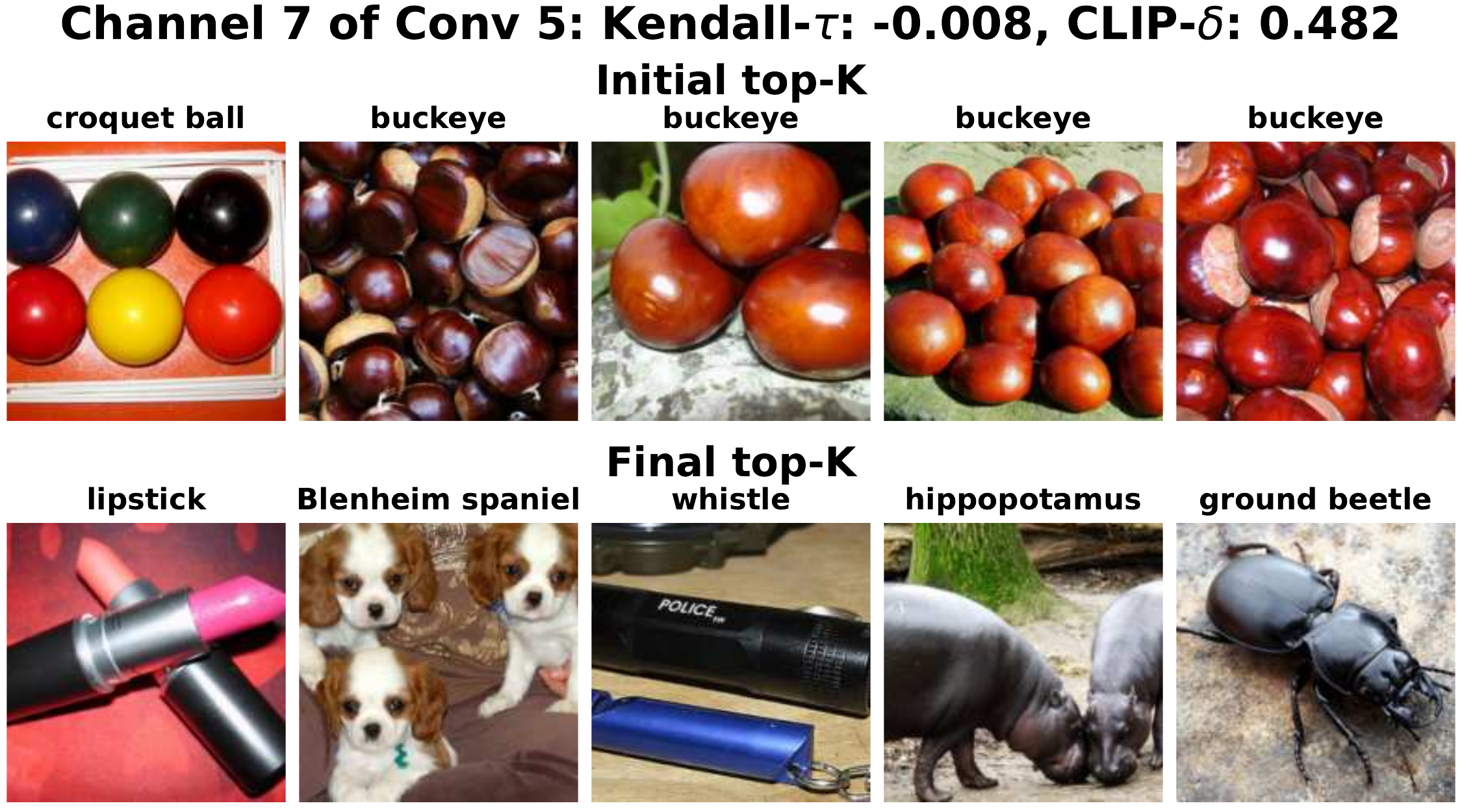}
\end{subfigure}\\
\vspace{.8cm}
\begin{subfigure}[]{0.49\linewidth}
    \includegraphics[width=\textwidth]{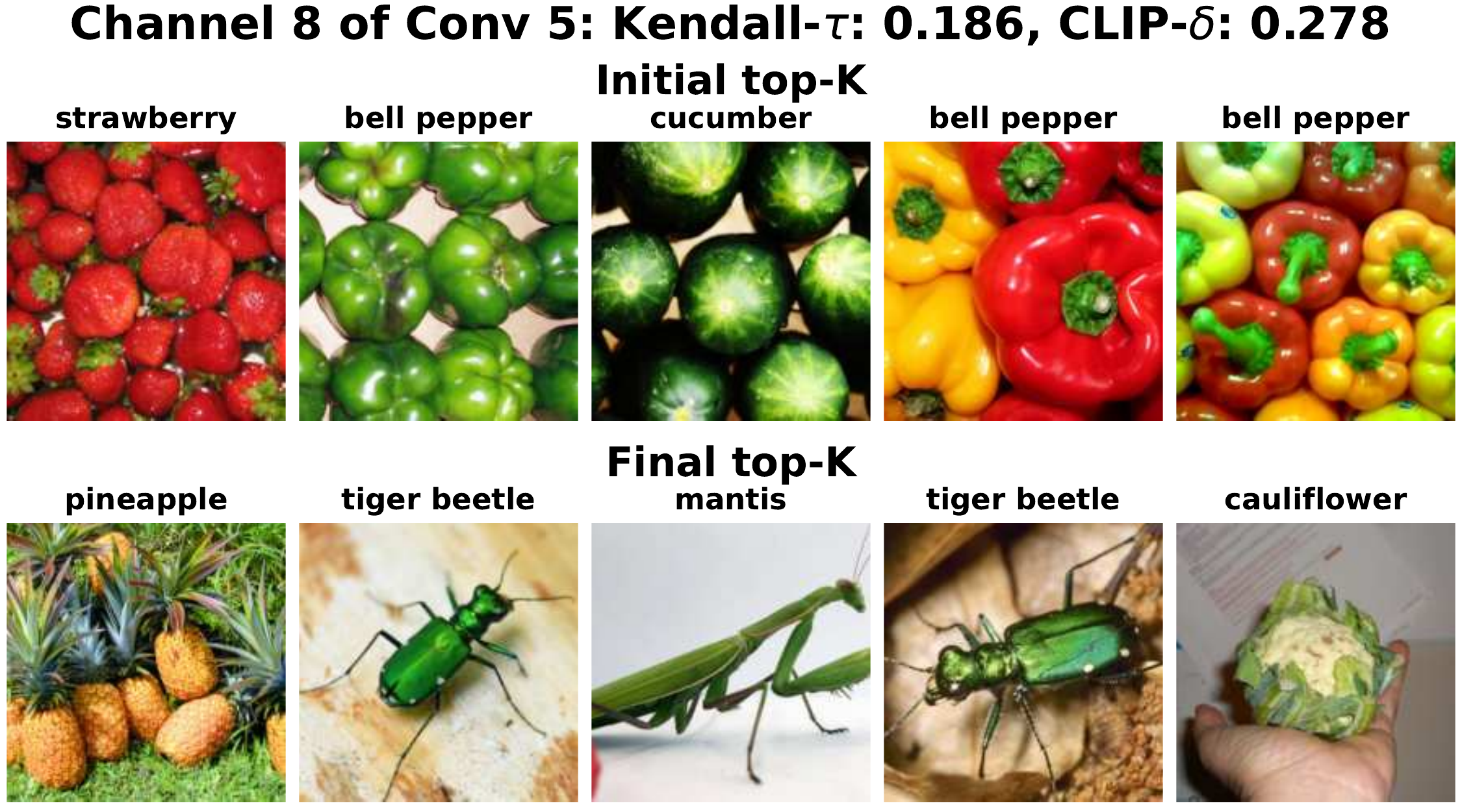}
\end{subfigure}\hfill
\begin{subfigure}[]{0.49\linewidth}
    \includegraphics[width=\textwidth]{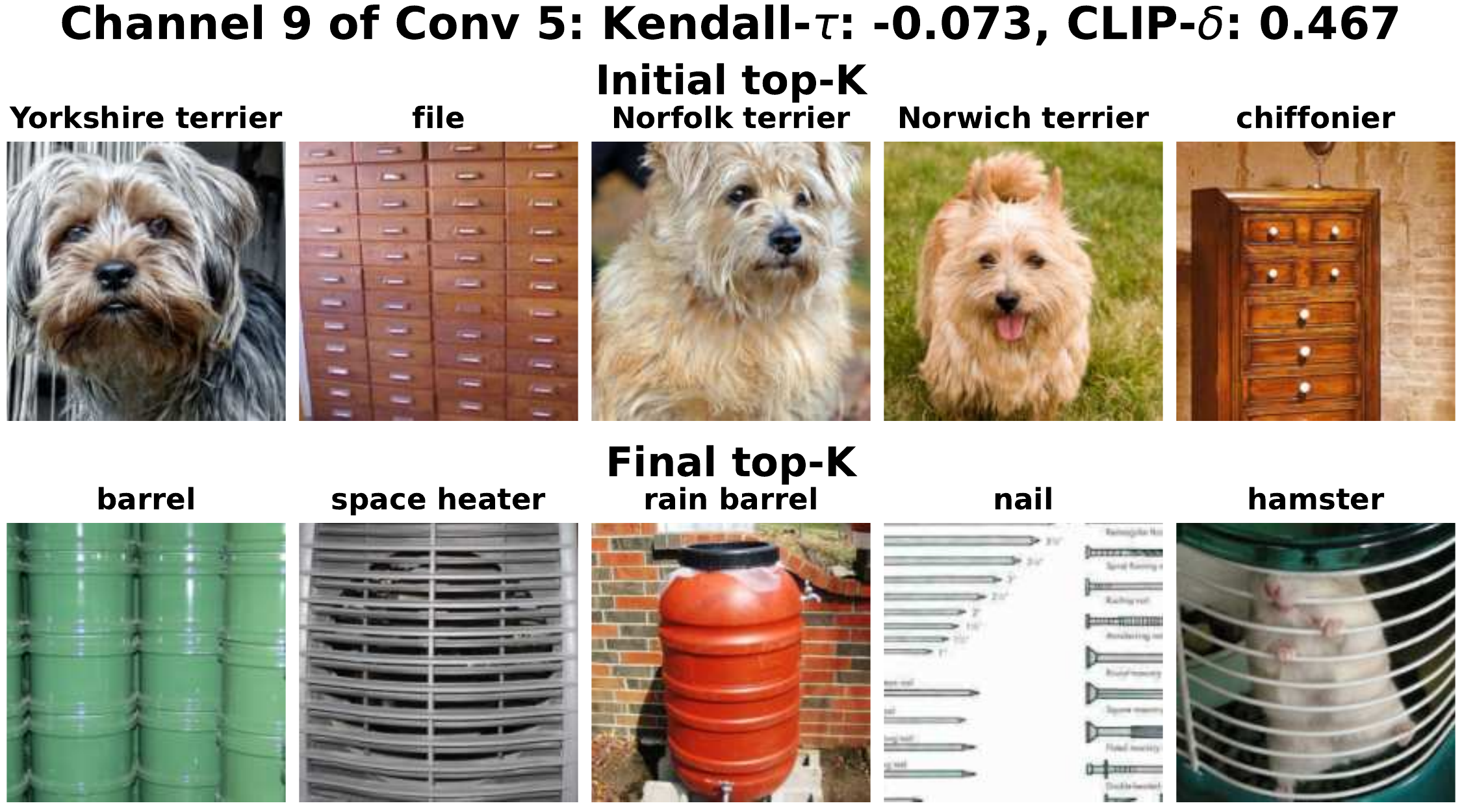}
\end{subfigure}
    \caption{\small Push-down attack on a single-channel of \textit{Conv5} of AlexNet. All initial images have been replaced by other images.} 
        \label{fig_add:single_channel}

\end{figure}

Figure~\ref{fig_add:single_channel} shows the results of initial top-$k$ images and final ones after running the push-down attack on every single channel. Except for channels 6 and 4 with relatively low CLIP-$\delta$ scores, it can be observed that all other channels have semantically different final top-$k$ images compared to the initial ones. This can be also seen by higher values of CLIP-$\delta$ scores.

\clearpage

\subsubsection{Push-down All-Channel Attack}
\begin{figure}[!ht]
\centering
\begin{subfigure}[]{0.49\linewidth}
    \includegraphics[width=\textwidth]{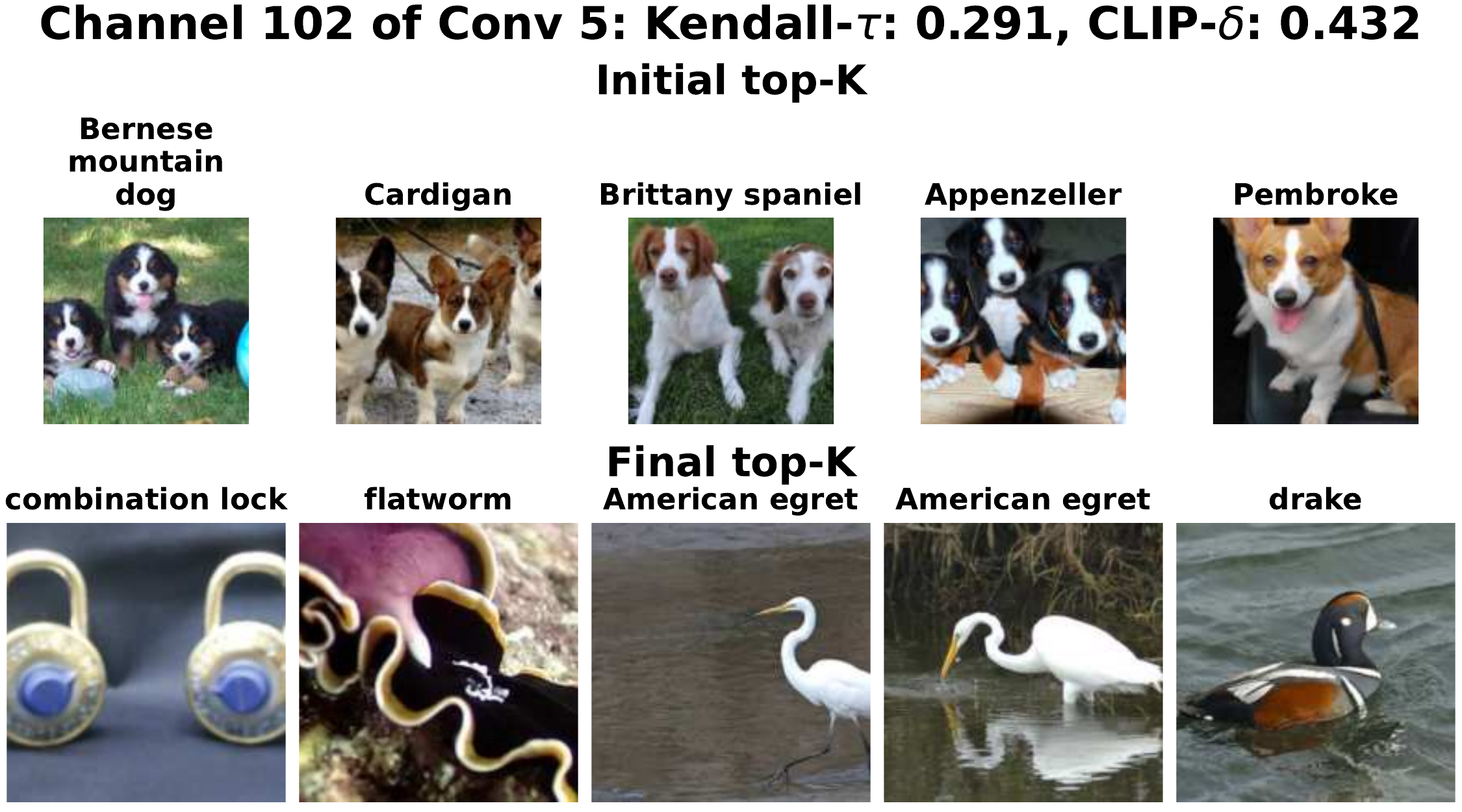}
\end{subfigure}\hfill
\begin{subfigure}[]{0.49\linewidth}
    \includegraphics[width=\textwidth]{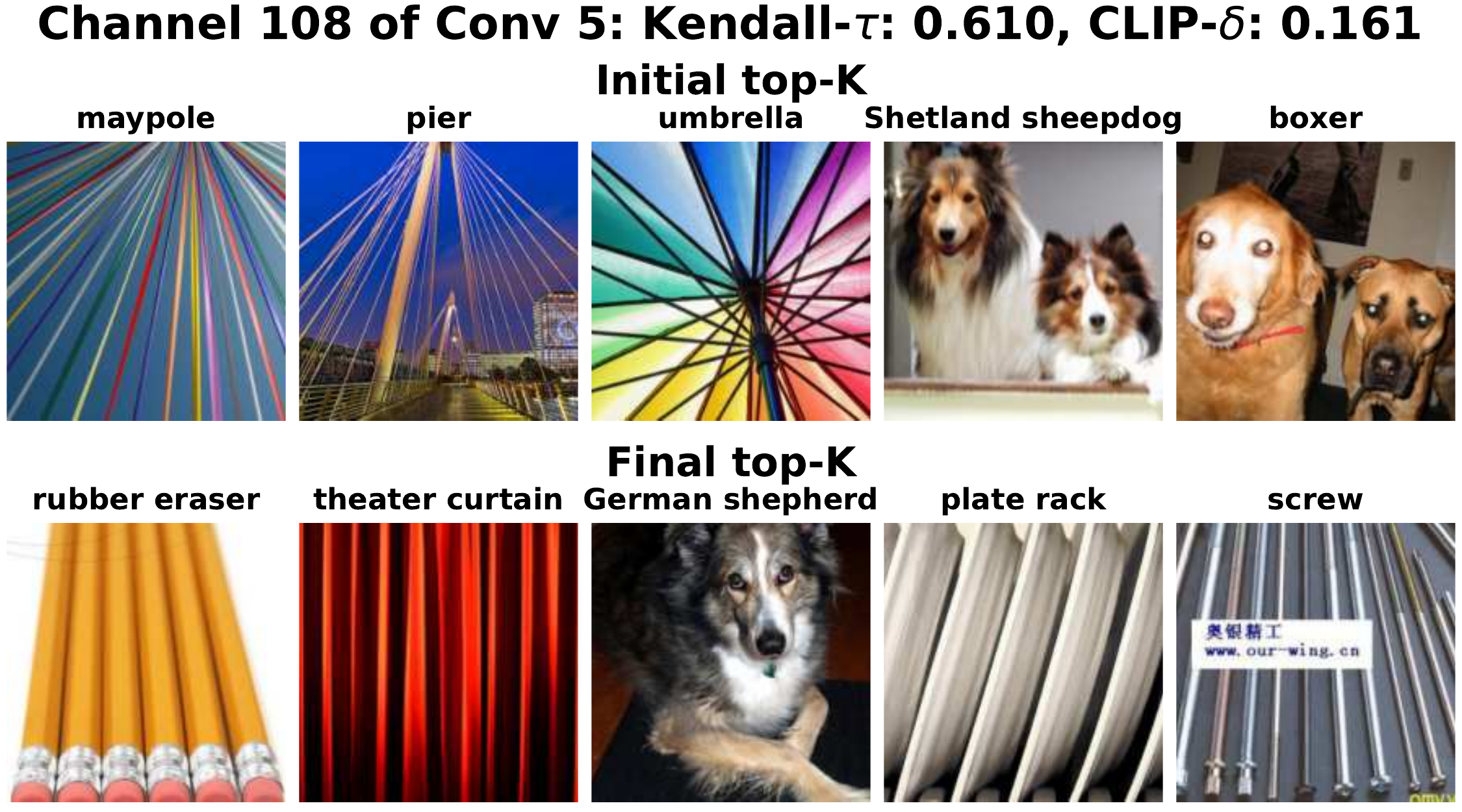}
\end{subfigure}\\
\vspace{.8cm}
\begin{subfigure}[]{0.49\linewidth}
    \includegraphics[width=\textwidth]{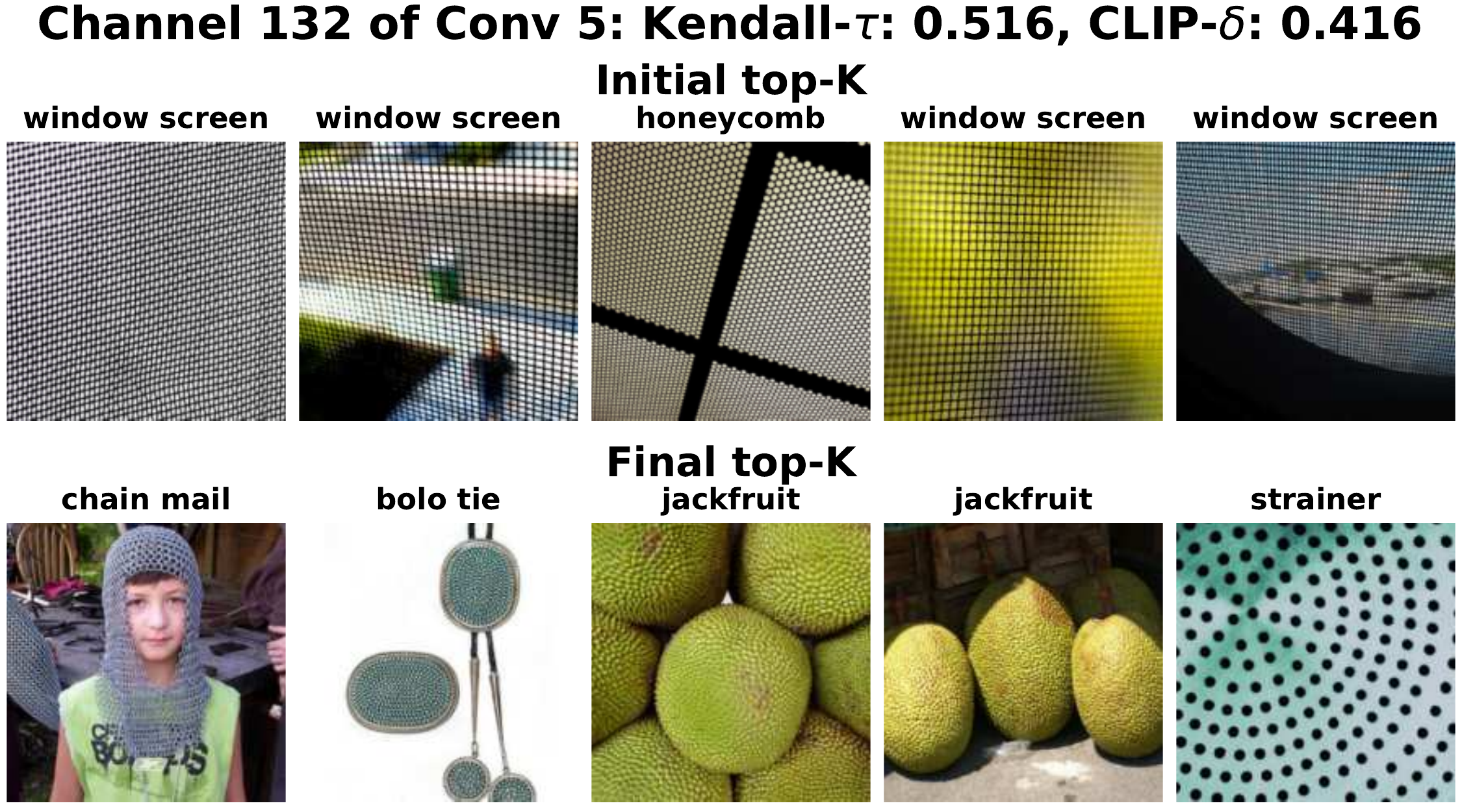}
\end{subfigure}\hfill
\begin{subfigure}[]{0.49\linewidth}
    \includegraphics[width=\textwidth]{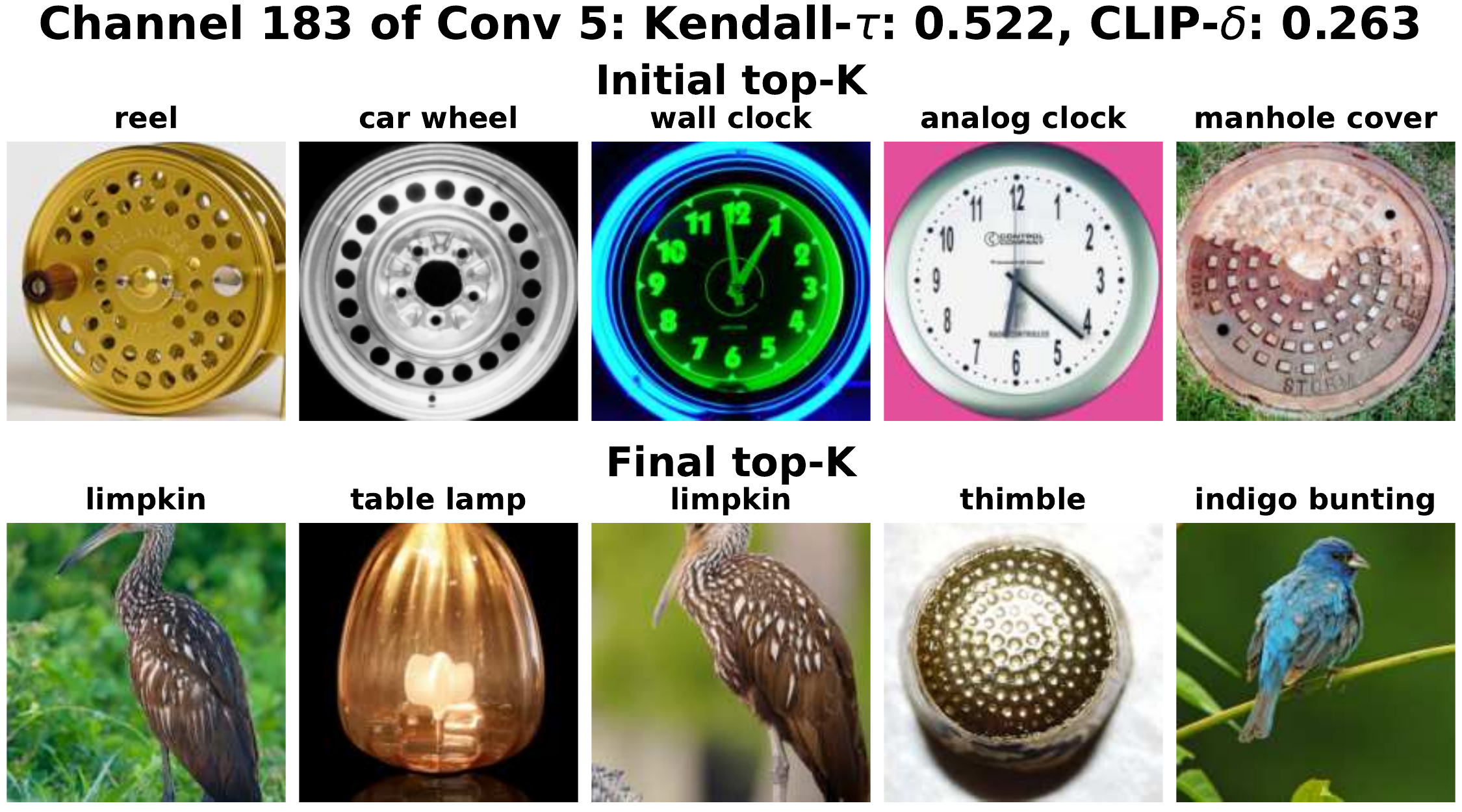}
\end{subfigure}\\
\vspace{.8cm}
\begin{subfigure}[]{0.49\linewidth}
    \includegraphics[width=\textwidth]{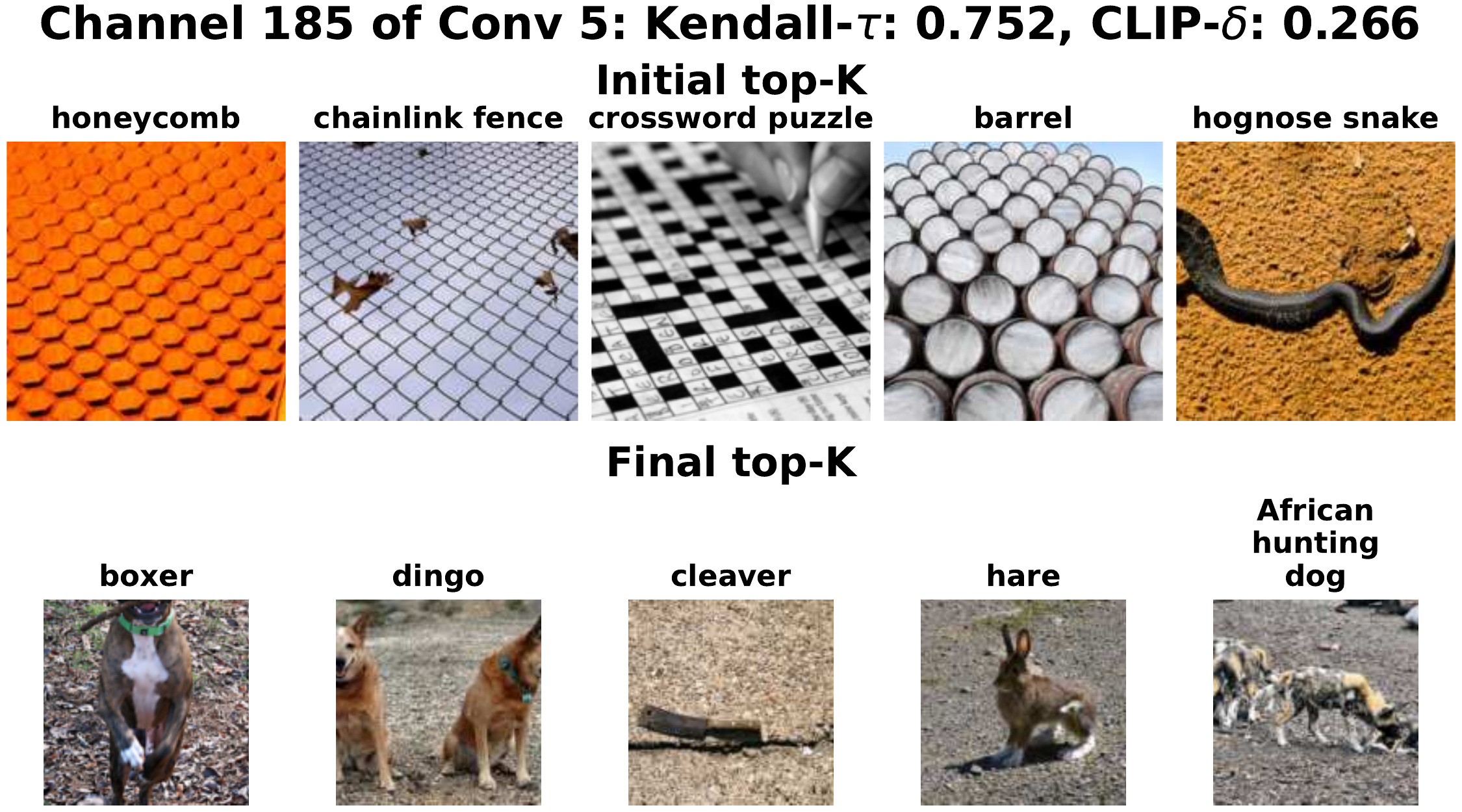}
\end{subfigure}\hfill
\begin{subfigure}[]{0.49\linewidth}
    \includegraphics[width=\textwidth]{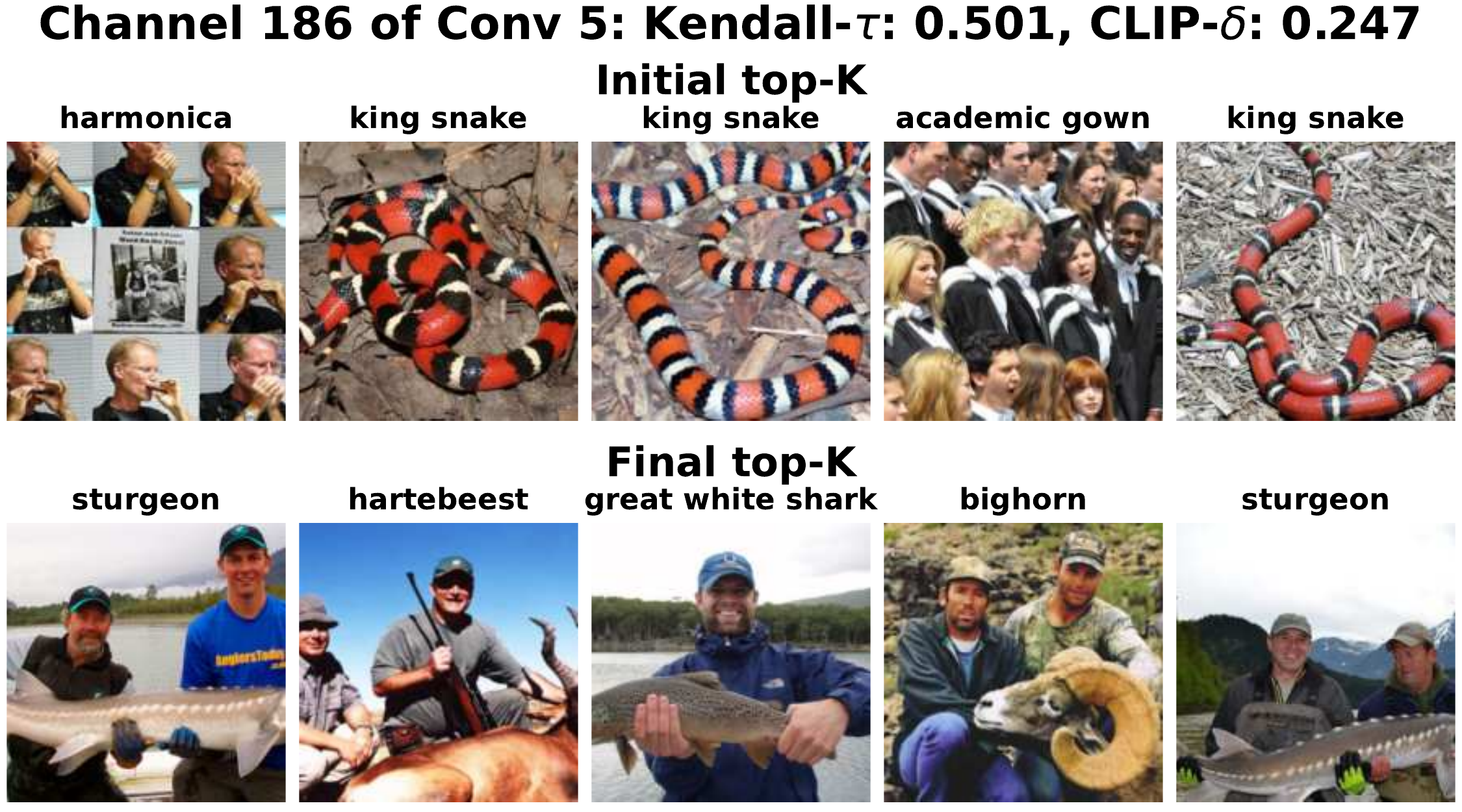}
\end{subfigure}\\
\vspace{.8cm}
\begin{subfigure}[]{0.49\linewidth}
    \includegraphics[width=\textwidth]{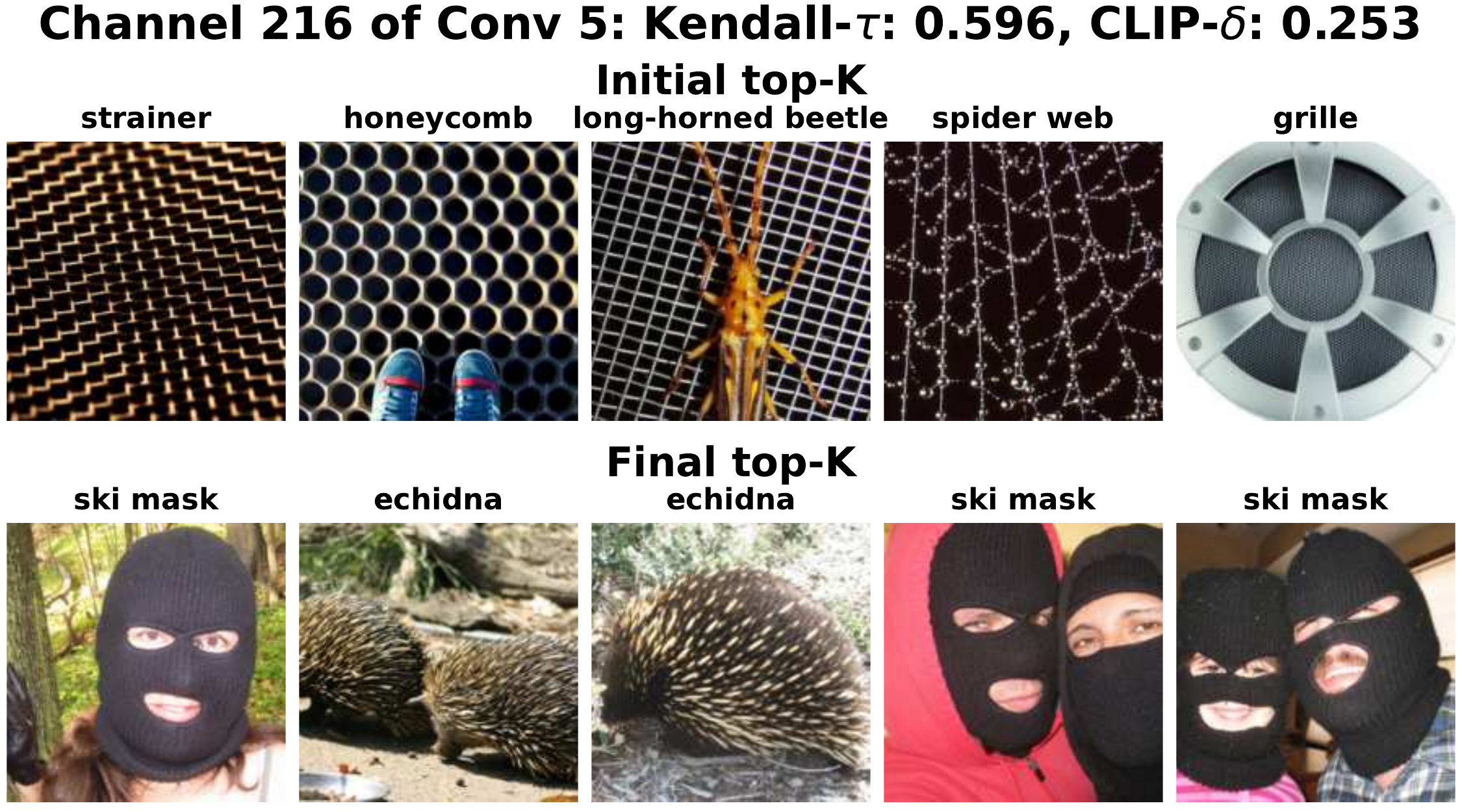}
\end{subfigure}\hfill
\begin{subfigure}[]{0.49\linewidth}
    \includegraphics[width=\textwidth]{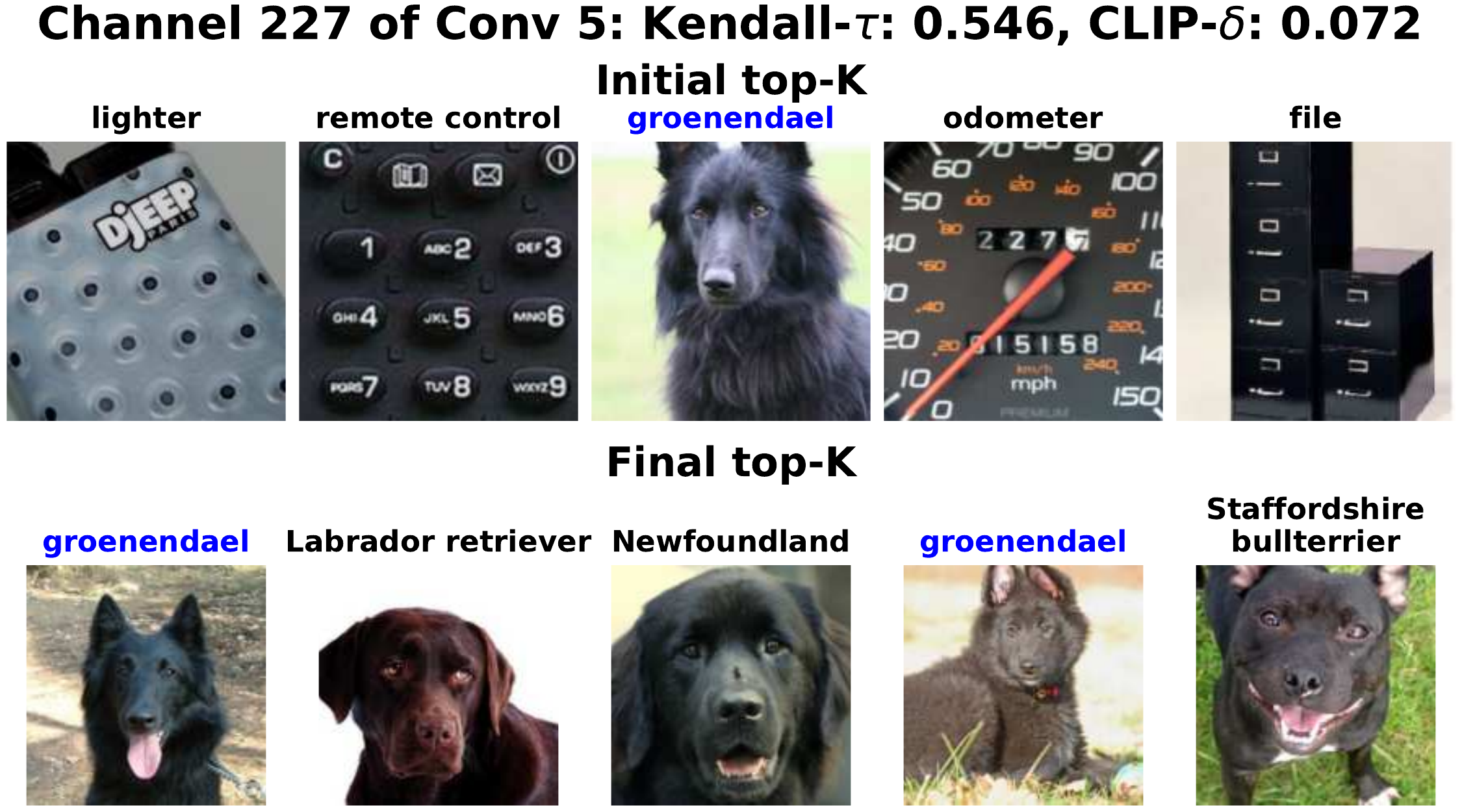}
\end{subfigure}\\
\vspace{.8cm}
\begin{subfigure}[]{0.49\linewidth}
    \includegraphics[width=\textwidth]{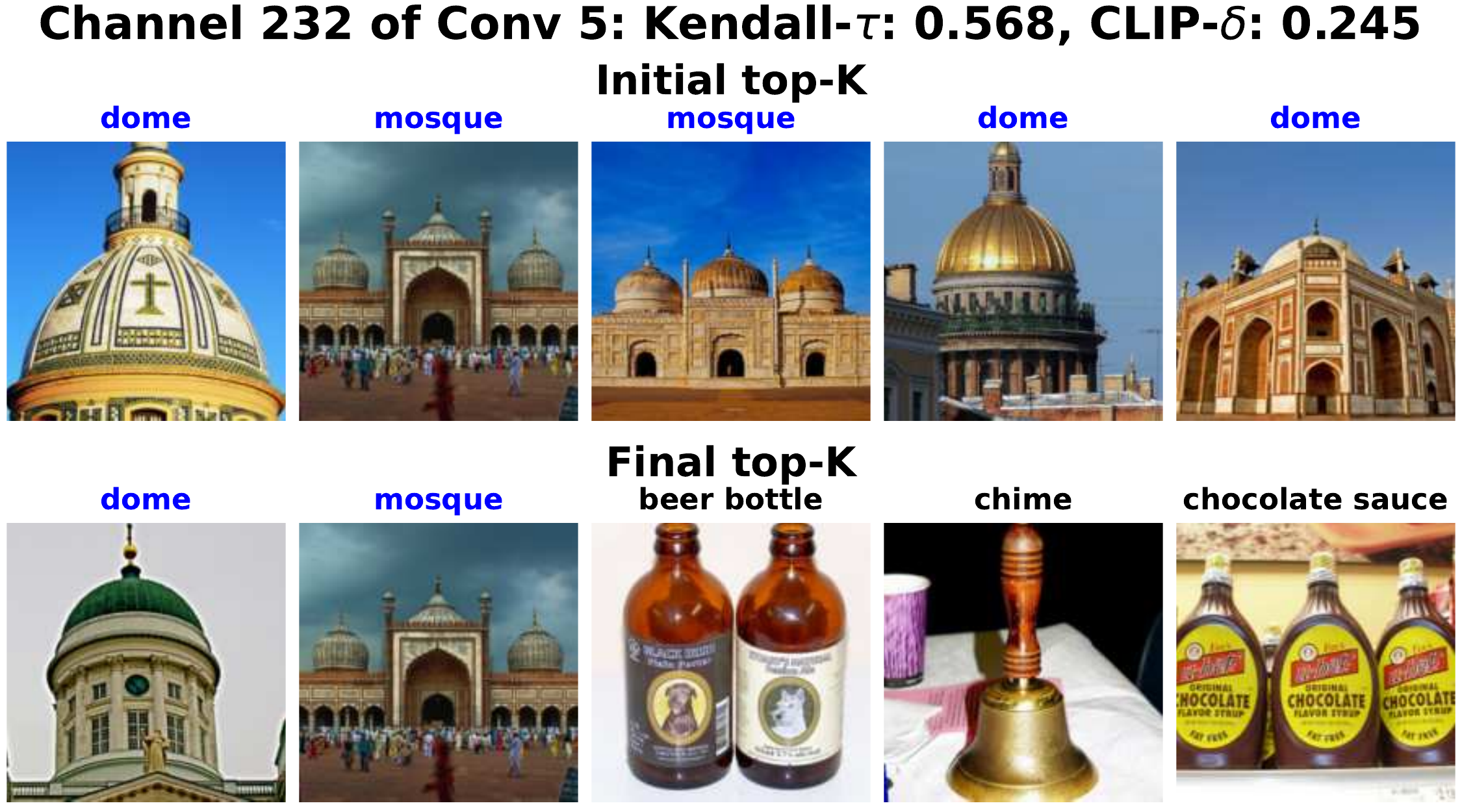}
\end{subfigure}\hfill
\begin{subfigure}[]{0.49\linewidth}
    \includegraphics[width=\textwidth]{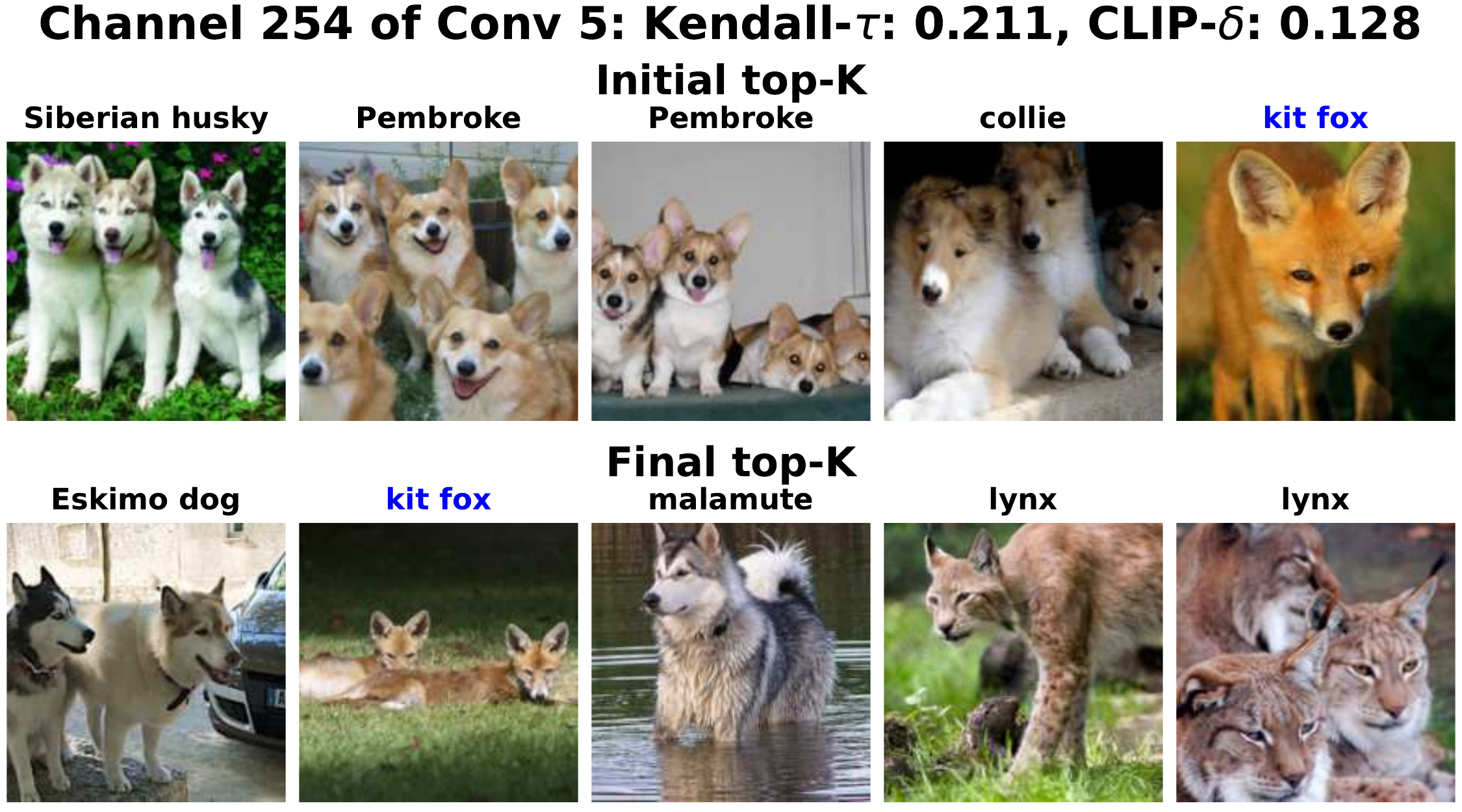}
\end{subfigure}
    \caption{\small Push-down all-channel attack of \textit{Conv5} of AlexNet. All initial top-5 images were completely removed from the new set of top-5 images, demonstrating the success of the attack. Channel indexes were chosen randomly.} 
        \label{fig_add:all_channel}
\end{figure}

This section presents additional results for the push-down attack on all channels at once. The results are obtained by attacking all the channels of the conv5 layer of AlexNet. We first show visual examples of results obtained from the training set of ImageNet and show its generalization to the validation set.
\paragraph{Visual Examples.}
Figure~\ref{fig_add:all_channel} shows results obtained on 10 randomly chosen channels. It can be observed that all initial top-5 images were completely removed from the set of top-activating images. Additionally, channels with high CLIP-$\delta$ scores such as channels 102 and 132, present semantically different images (initial vs final) with no overlap classes. In contrast, we observe that channels with low CLIP-$\delta$ scores such as channels 254 and 227 usually share similar classes in top-activating images. Finally, from Kendall-$\tau$ scores, we observe that channels that have high Kendall-$\tau$ (e.g., channel 108 and 185) do not often have high values of CLIP-$\delta$ scores, indicating that the weak change in channel behavior assessed by the Kendall-$\tau$ is often related to low semantic change.

\begin{figure}[!ht]
\centering
\begin{subfigure}[]{0.49\linewidth}
    \includegraphics[width=\textwidth]{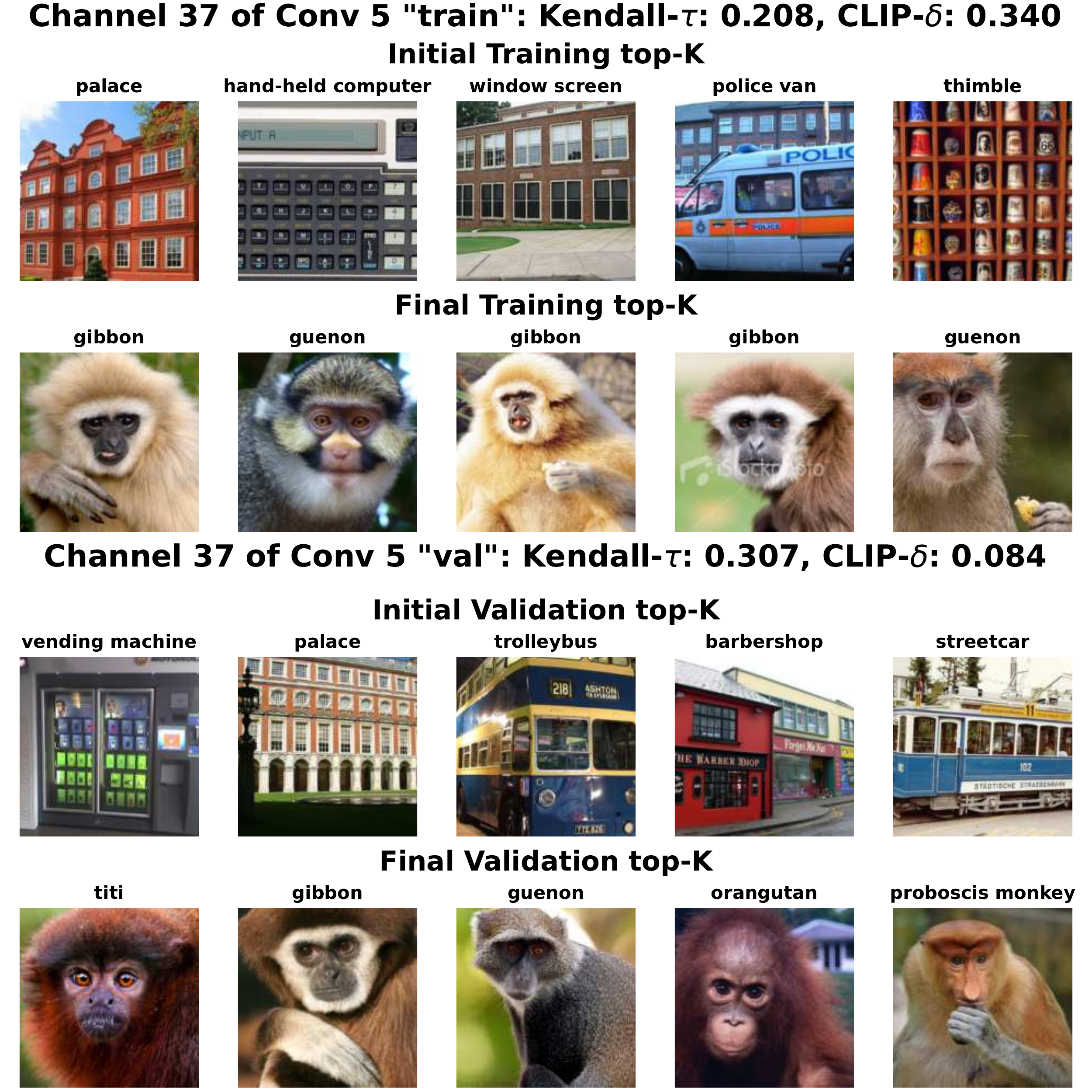}
\end{subfigure}\hfill
\begin{subfigure}[]{0.49\linewidth}
    \includegraphics[width=\textwidth]{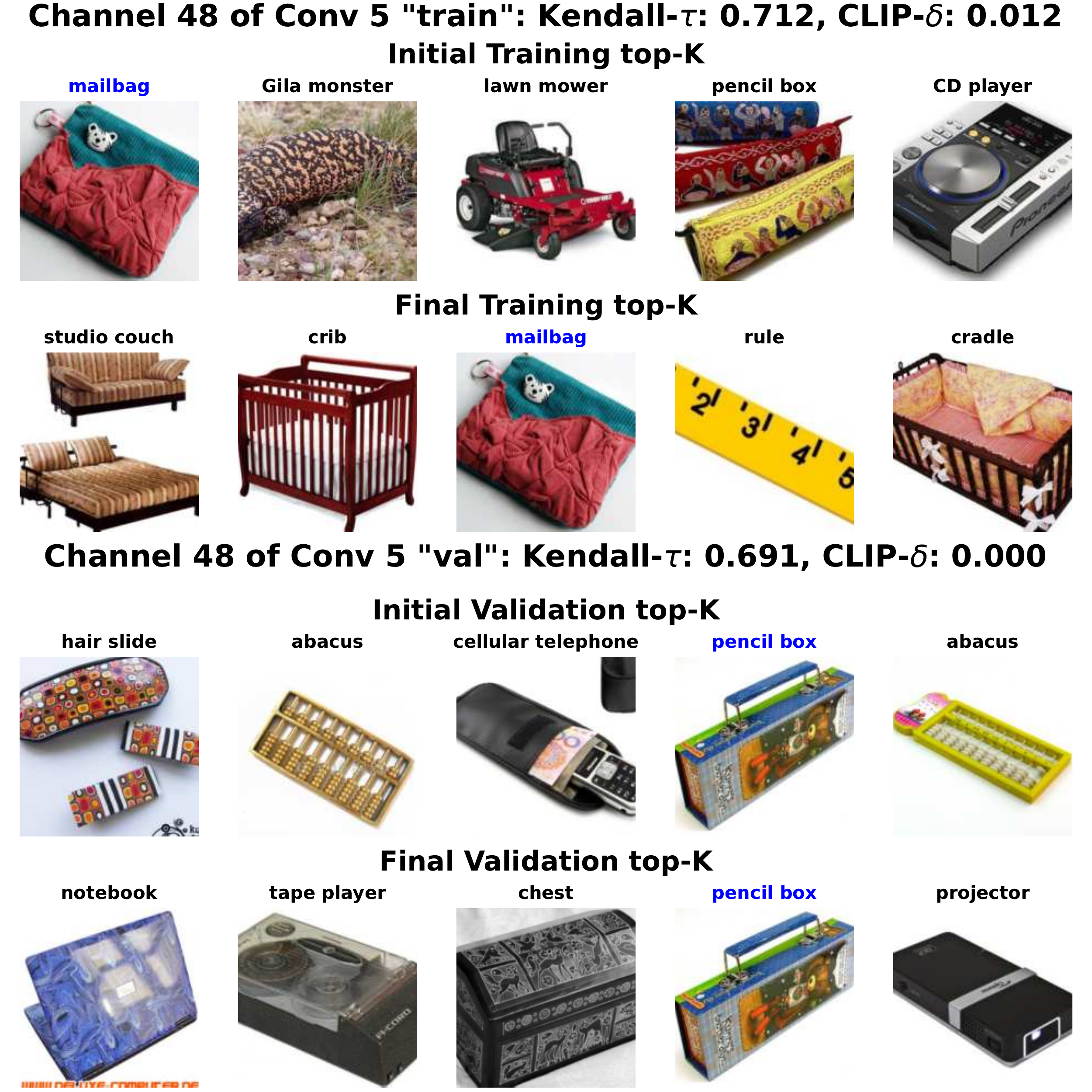}
\end{subfigure}\\
\vspace{.8cm}
\begin{subfigure}[]{0.49\linewidth}
    \includegraphics[width=\textwidth]{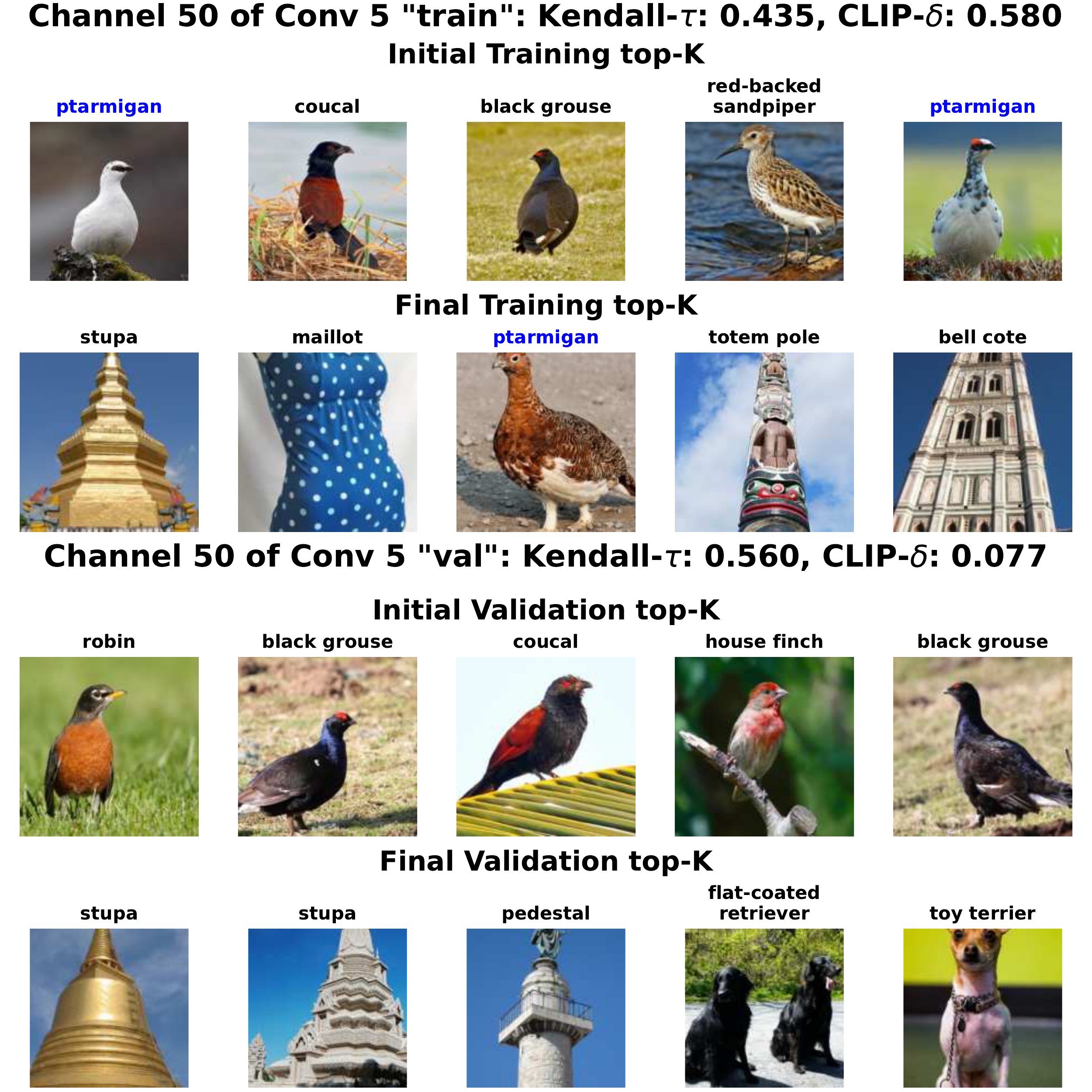}
\end{subfigure}\hfill
\begin{subfigure}[]{0.49\linewidth}
    \includegraphics[width=\textwidth]{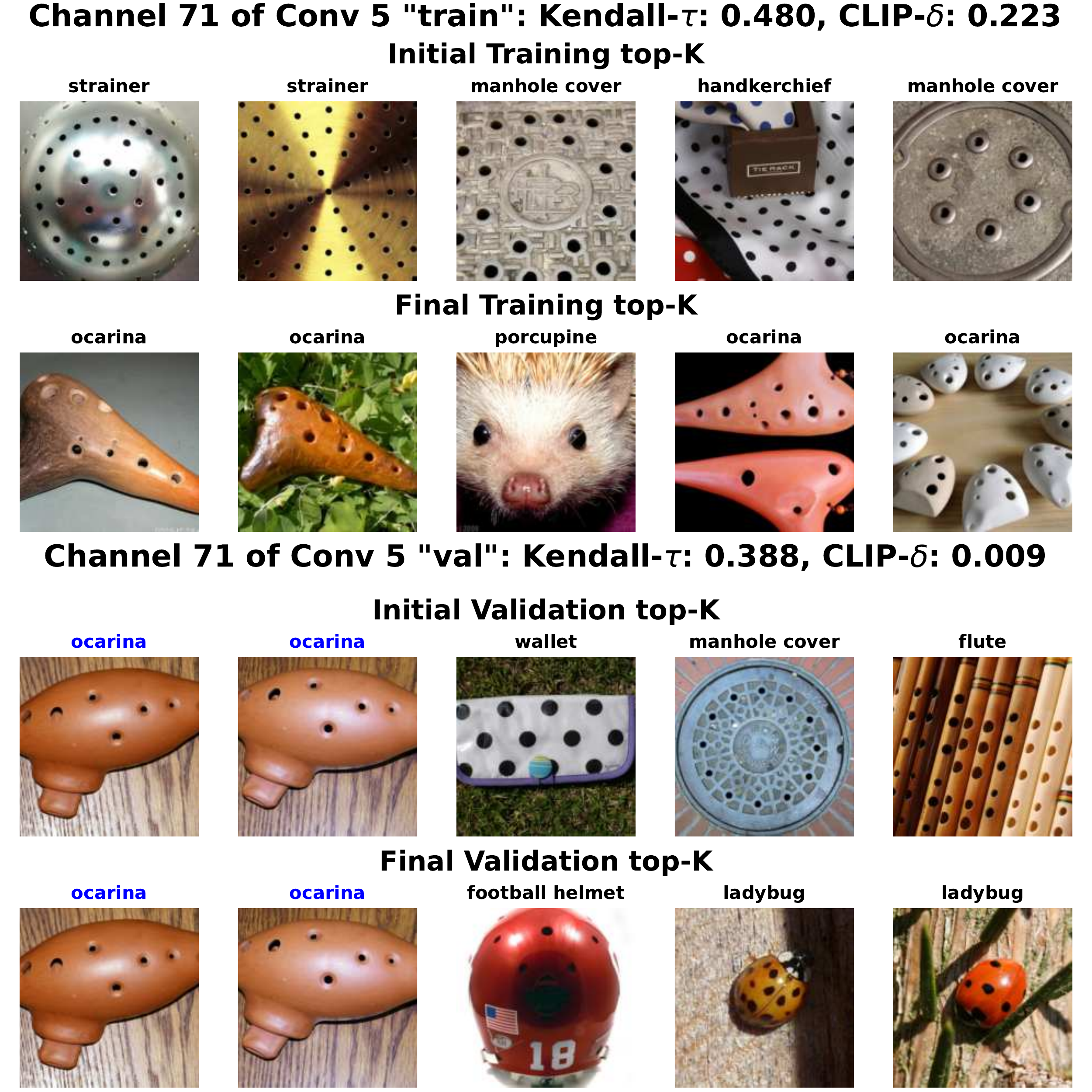}
\end{subfigure}\\
\vspace{.8cm}
\begin{subfigure}[]{0.49\linewidth}
    \includegraphics[width=\textwidth]{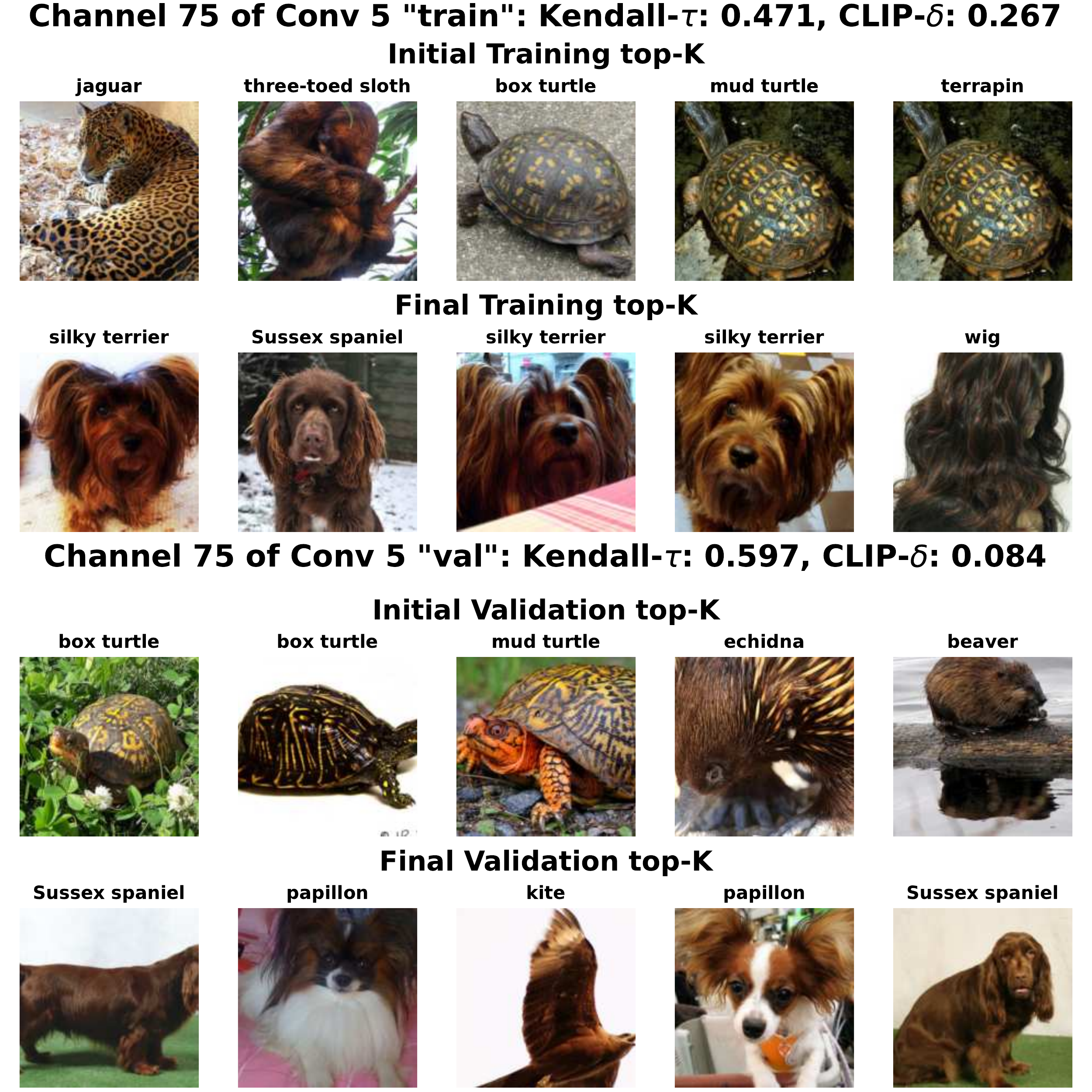}
\end{subfigure}\hfill
    \caption{\small Push-down all-channel attack of \textit{Conv5} of AlexNet. For each channel, the first two rows are top-$k$ images derived from the training set while the last two are derived from the validation set.} 
        \label{fig_add:all_channel_validation}
\end{figure}

\begin{figure}
\centering
\begin{subfigure}[]{0.49\linewidth}
    \includegraphics[width=\textwidth]{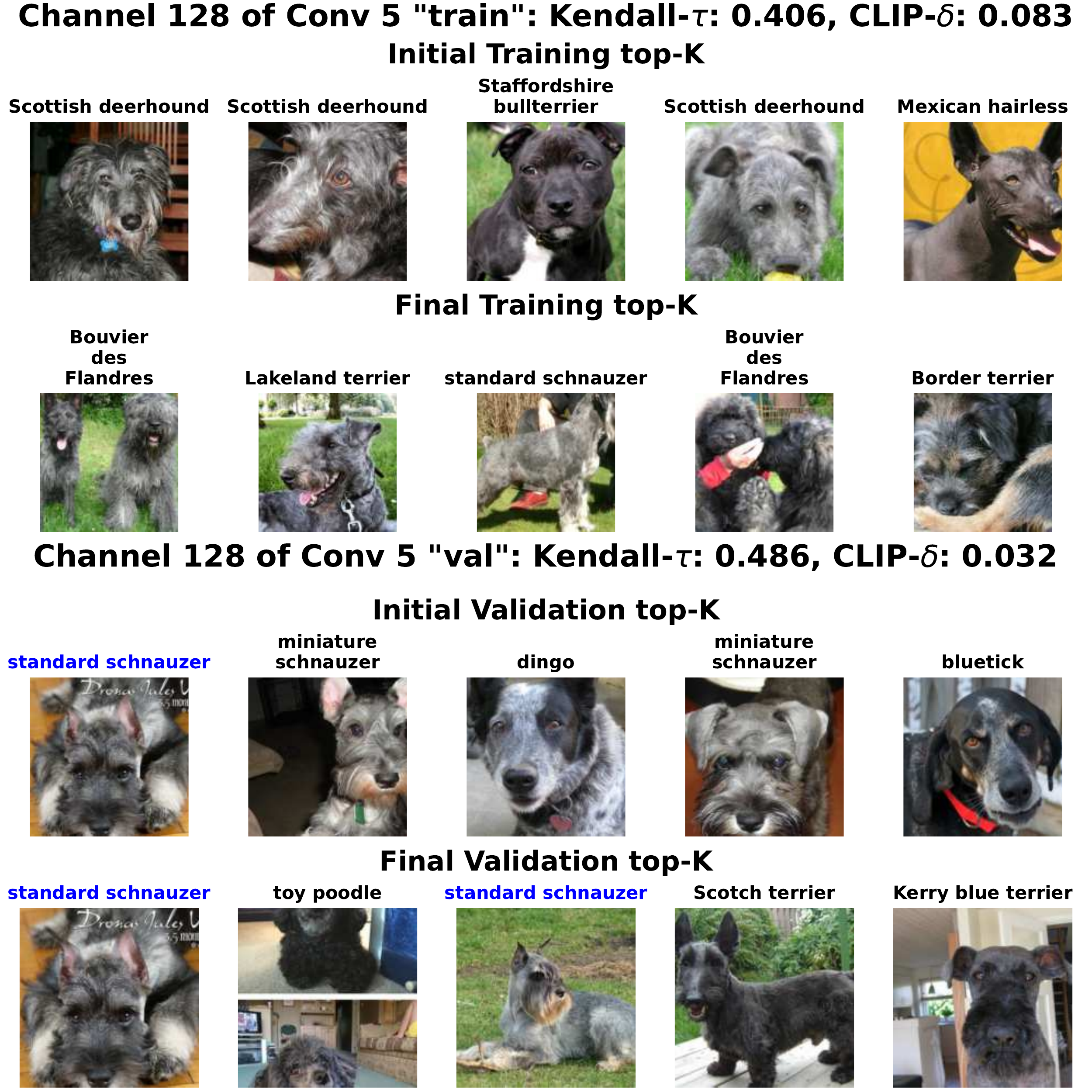}
\end{subfigure}\hfill
\begin{subfigure}[]{0.49\linewidth}
    \includegraphics[width=\textwidth]{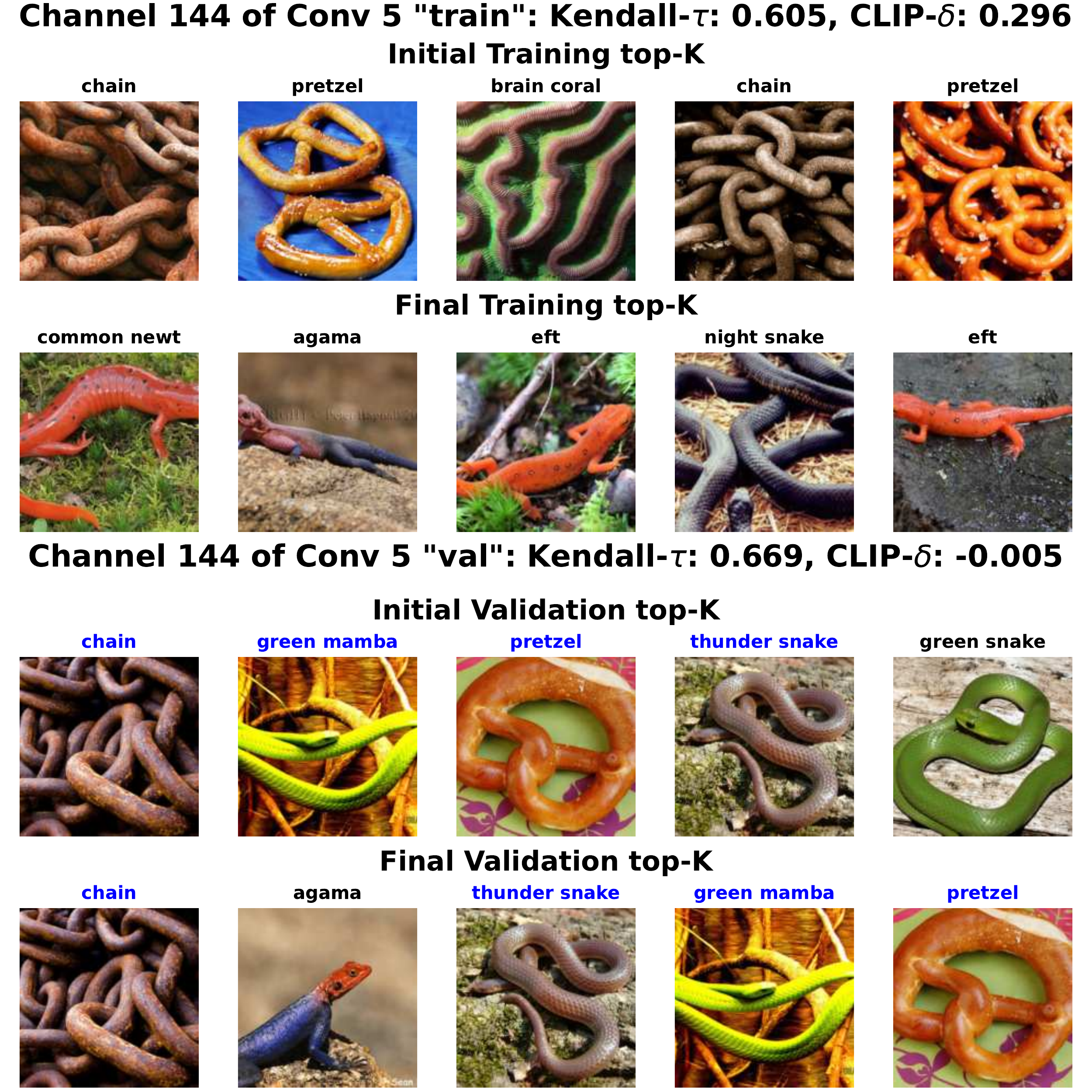}
\end{subfigure}\\
\vspace{.8cm}
\begin{subfigure}[]{0.49\linewidth}
    \includegraphics[width=\textwidth]{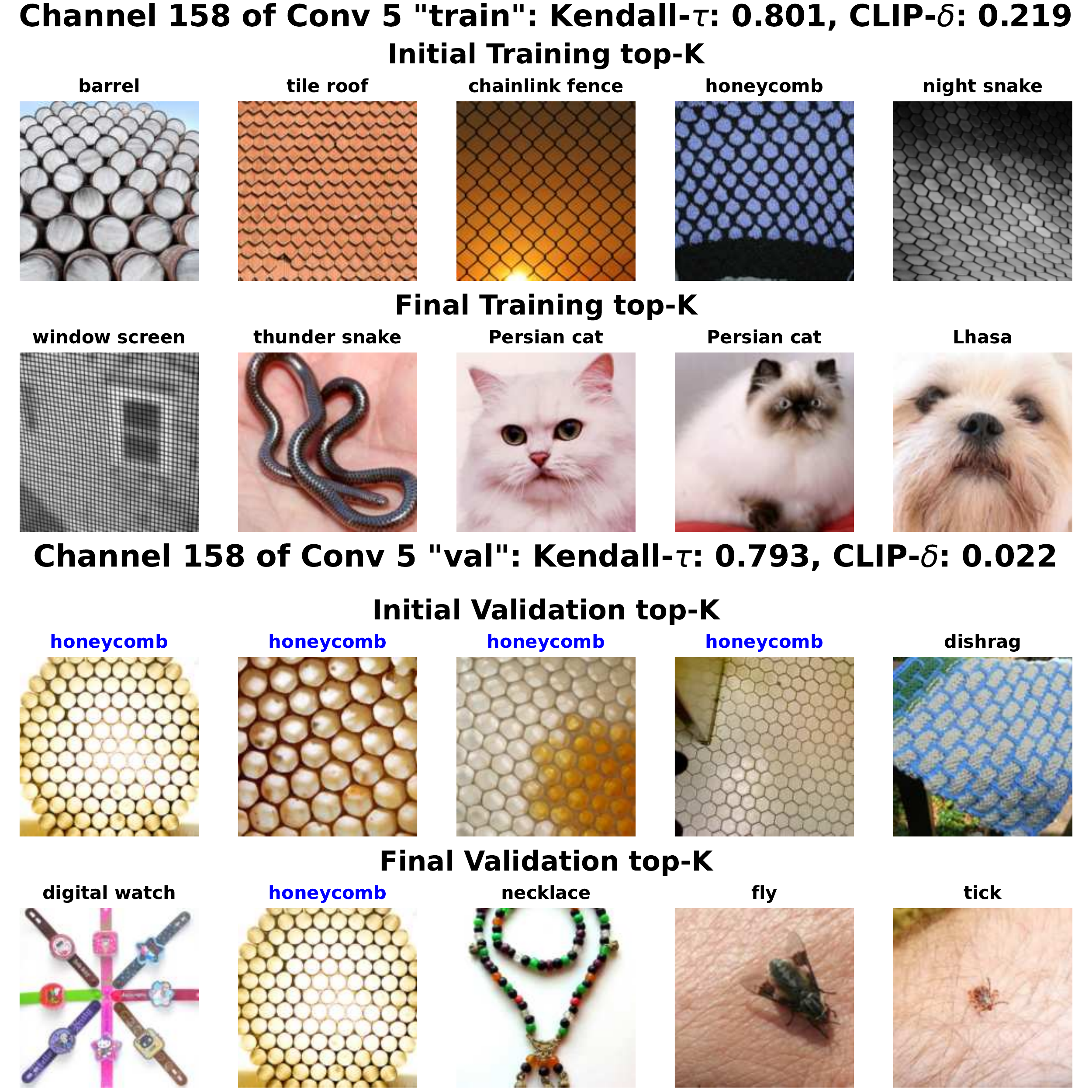}
\end{subfigure}\hfill
\begin{subfigure}[]{0.49\linewidth}
    \includegraphics[width=\textwidth]{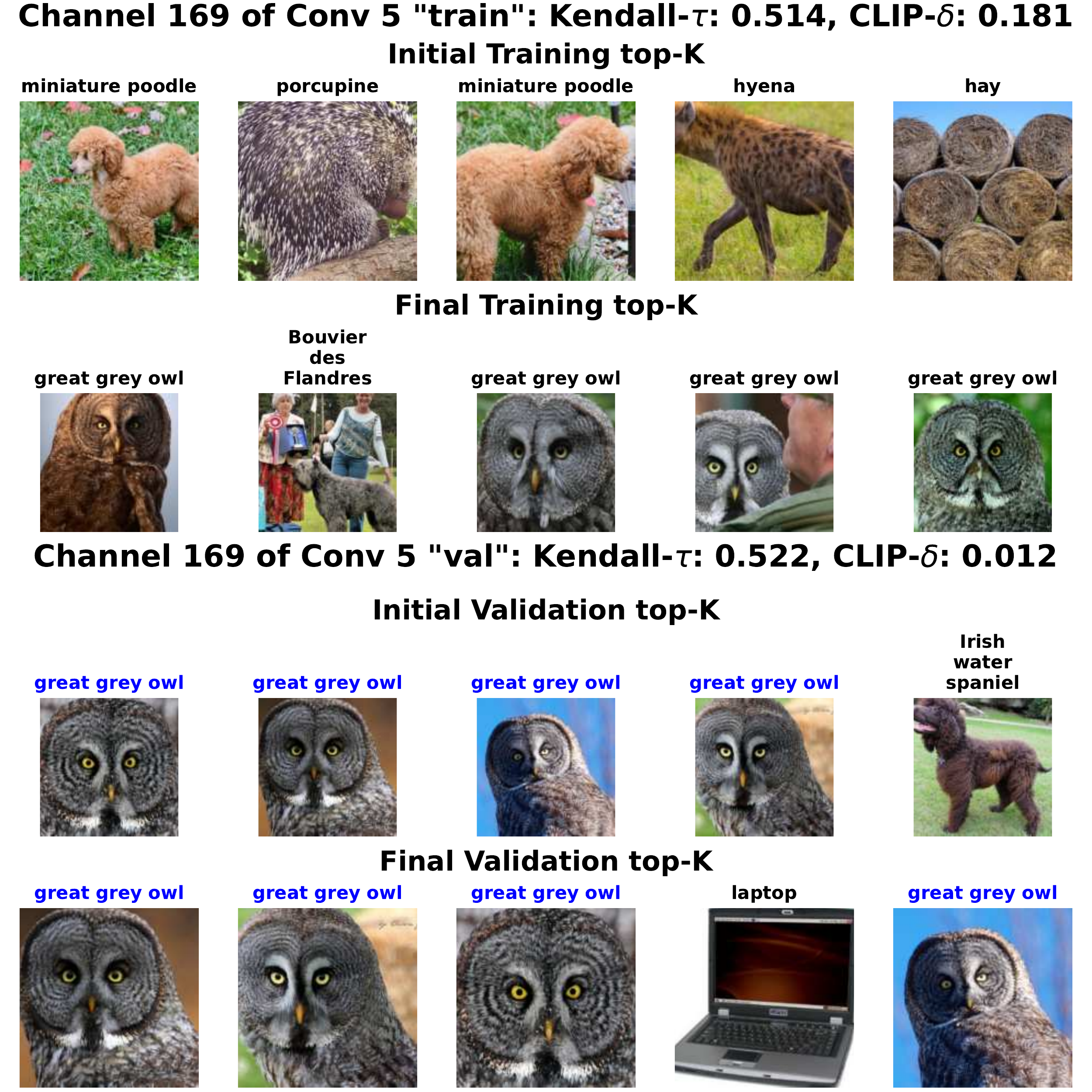}
\end{subfigure}\\
\vspace{.8cm}
\begin{subfigure}[]{0.49\linewidth}
    \includegraphics[width=\textwidth]{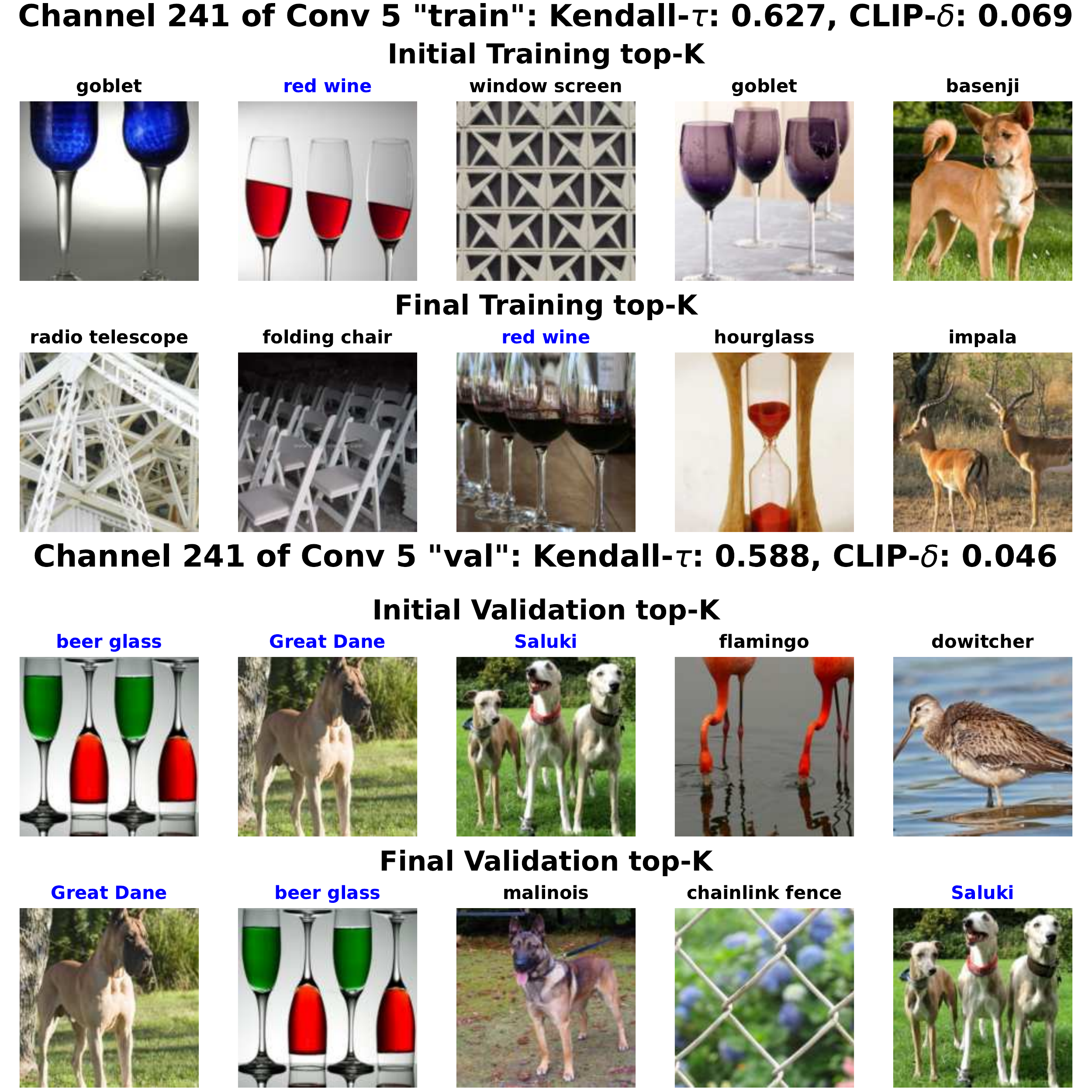}
\end{subfigure}
\caption{\small Push-down all-channel attack of \textit{Conv5} of AlexNet. For each channel, the first two rows are top-$k$ images derived from the training set while the last two are derived from the validation set.} 
        \label{fig_add:all_channel_validation_next}
\end{figure}

\clearpage
\paragraph{Generalization on Validation Set.}
We evaluate the generalization of our attack on the validation set of ImageNet. This gives more insights to the change of feature visualization. Figures~\ref{fig_add:all_channel_validation} and ~\ref{fig_add:all_channel_validation_next} show the initial top-$k$ images and final ones from training and validation sets for 10 randomly chosen channels.  

It can be observed that on every channel, from the validation set, at least one image from the initial top-$5$ images is no longer present in final top-$5$ images (for the majority of these channels, the first top-activating is no longer the top one). We also observe a complete replacement of top-$5$ images on the validation set when Kenall-$\tau$ scores and CLIP-$\delta$ are respectively low and high simultaneously (e.g., channels 37 and 50 of Figure~\ref{fig_add:all_channel_validation}). Moreover, the general trends in training and validation are similar suggesting the attack is not just memorizing specific images but leading to a generalized change.

\clearpage

\subsection{Ablation Study on EfficientNet}
\label{app:other-architectures}
It is important to show that the proposed attack methodology is not limited to AlexNet.   
In order to show that the attack can work on newer, more sophisticated neural nets, we have also run an ablation study on EfficientNet \cite{tan2019efficientnet}. We select the third convolutional block in the Feature 7 layer and perform a push-down attack similar way to AlexNet. The visual results are shown in Appendix A and the metrics for the layer are given in Table~\ref{tab:main_overall}. 
We observe similar effects to AlexNet; the top images are changed in terms of the exact images and the semantic concepts.  We also observe relatively strong CLIP-$\delta$ and Kendall-$\tau$ changes.
Having confirmed the generality of our approach in this way, we leave a survey study over all relevant architectures to future work, computation power permitting.

\begin{figure}[!ht]
\centering
\begin{subfigure}[]{0.49\linewidth}
    \includegraphics[width=\textwidth]{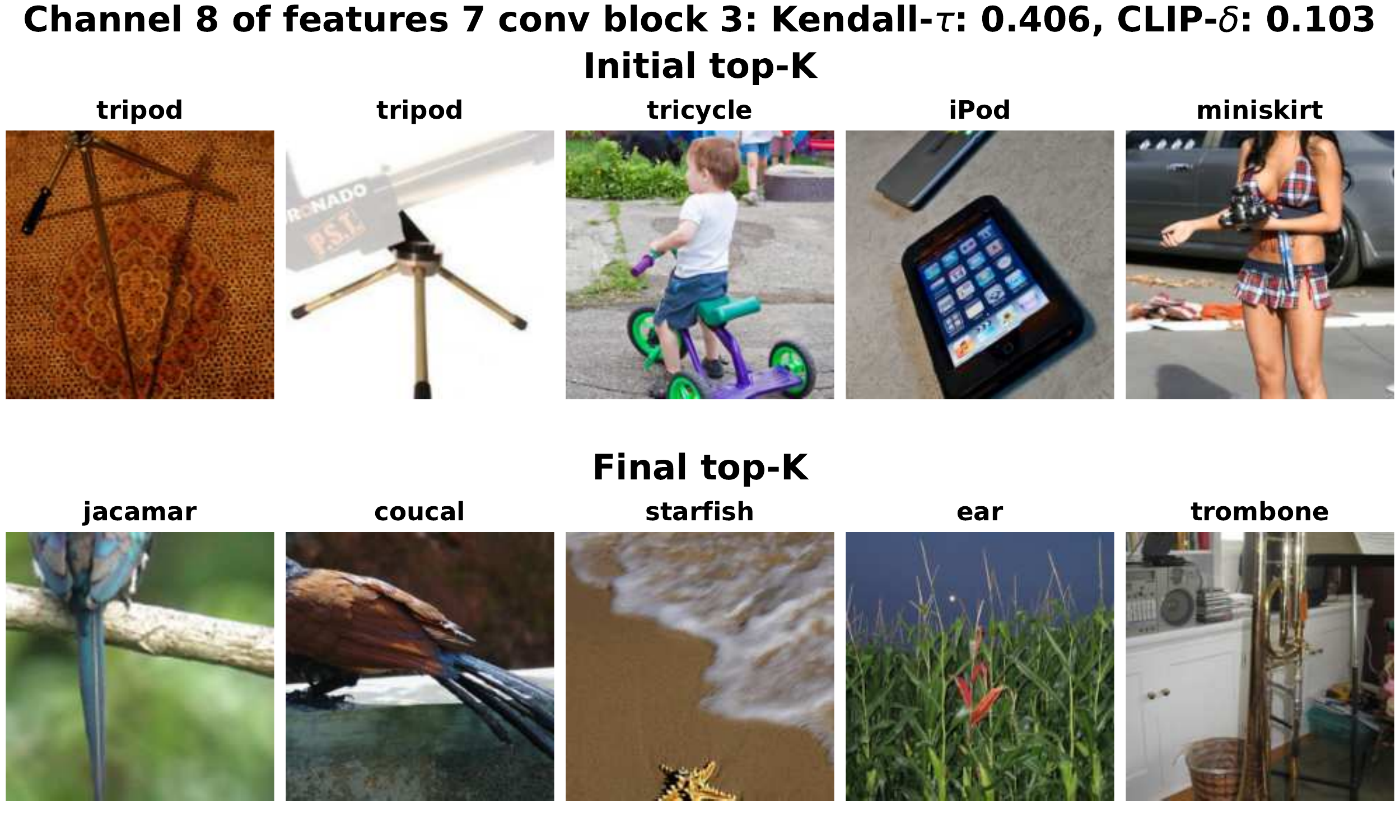}
\end{subfigure}\hfill
\begin{subfigure}[]{0.49\linewidth}
    \includegraphics[width=\textwidth]{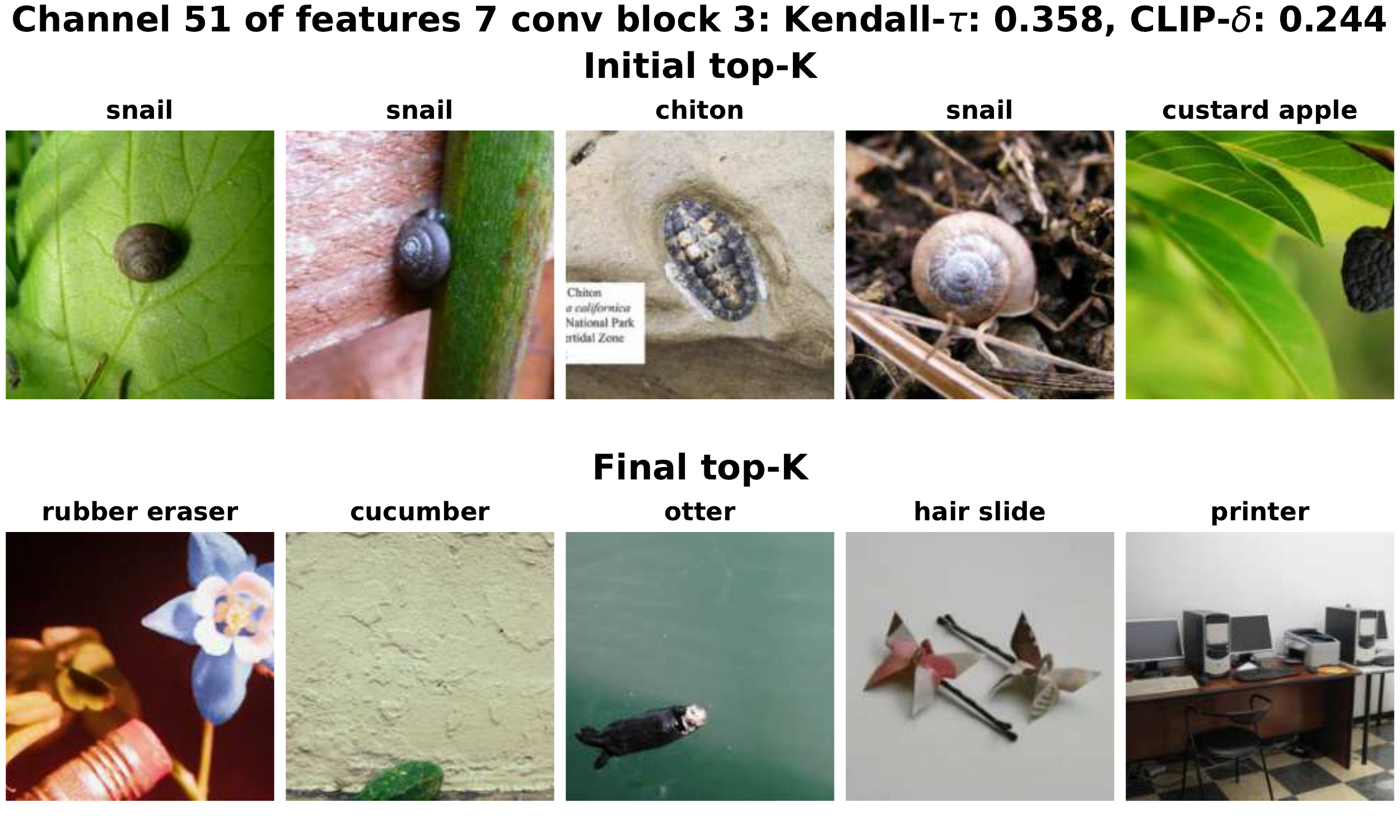}
\end{subfigure}\\
\vspace{.2cm}
\begin{subfigure}[]{0.49\linewidth}
    \includegraphics[width=\textwidth]{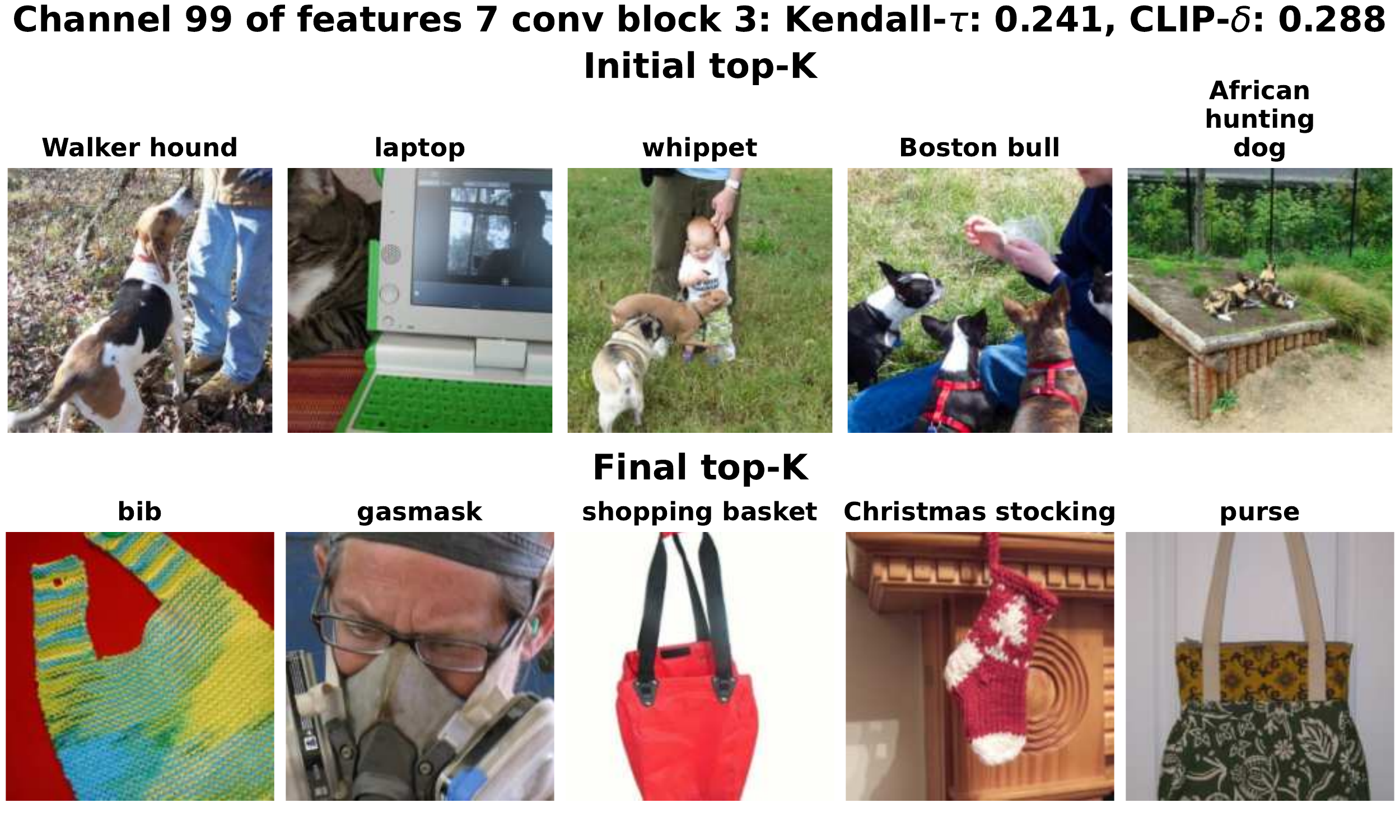}
\end{subfigure}\hfill
\begin{subfigure}[]{0.49\linewidth}
    \includegraphics[width=\textwidth]{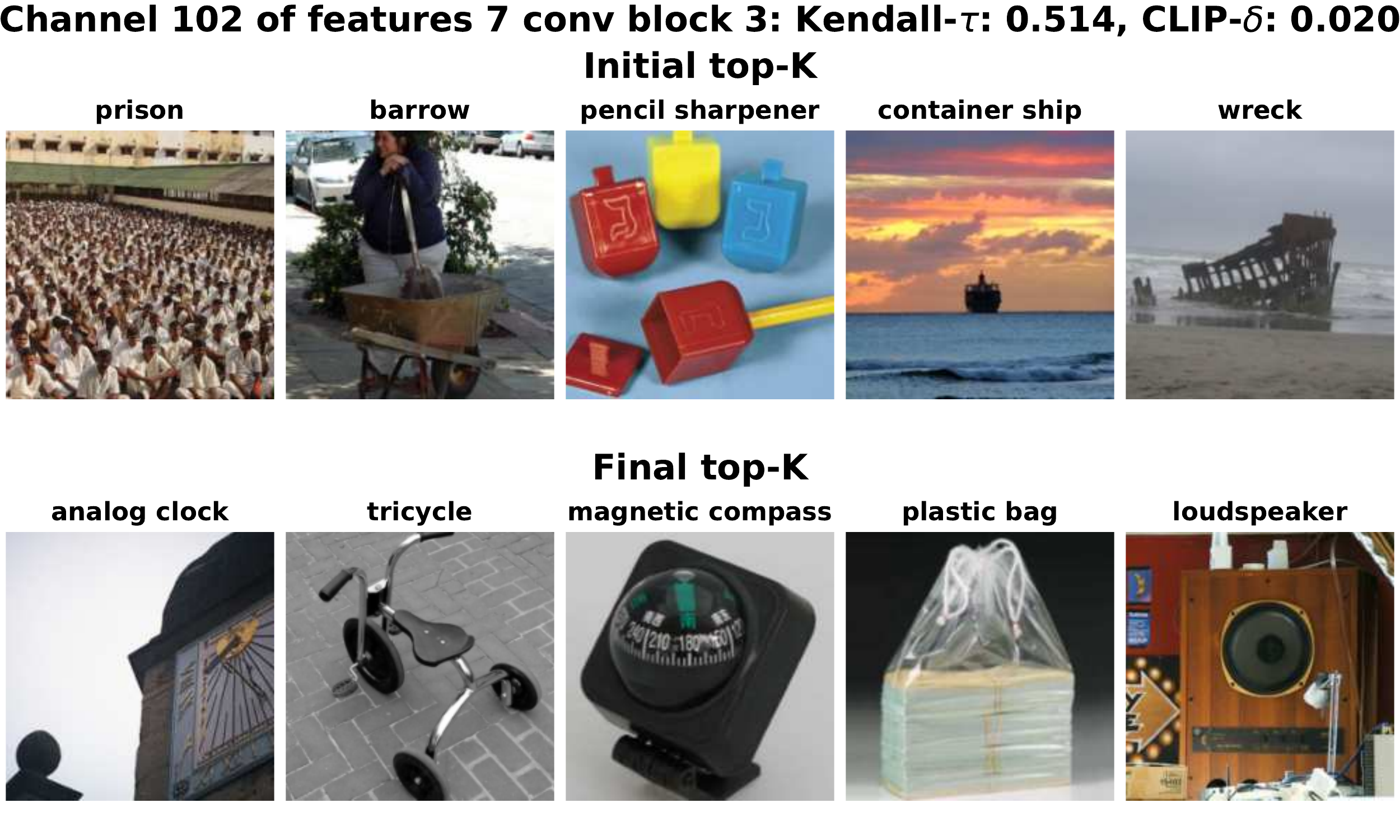}
\end{subfigure}\\
\vspace{.2cm}
\begin{subfigure}[]{0.49\linewidth}
    \includegraphics[width=\textwidth]{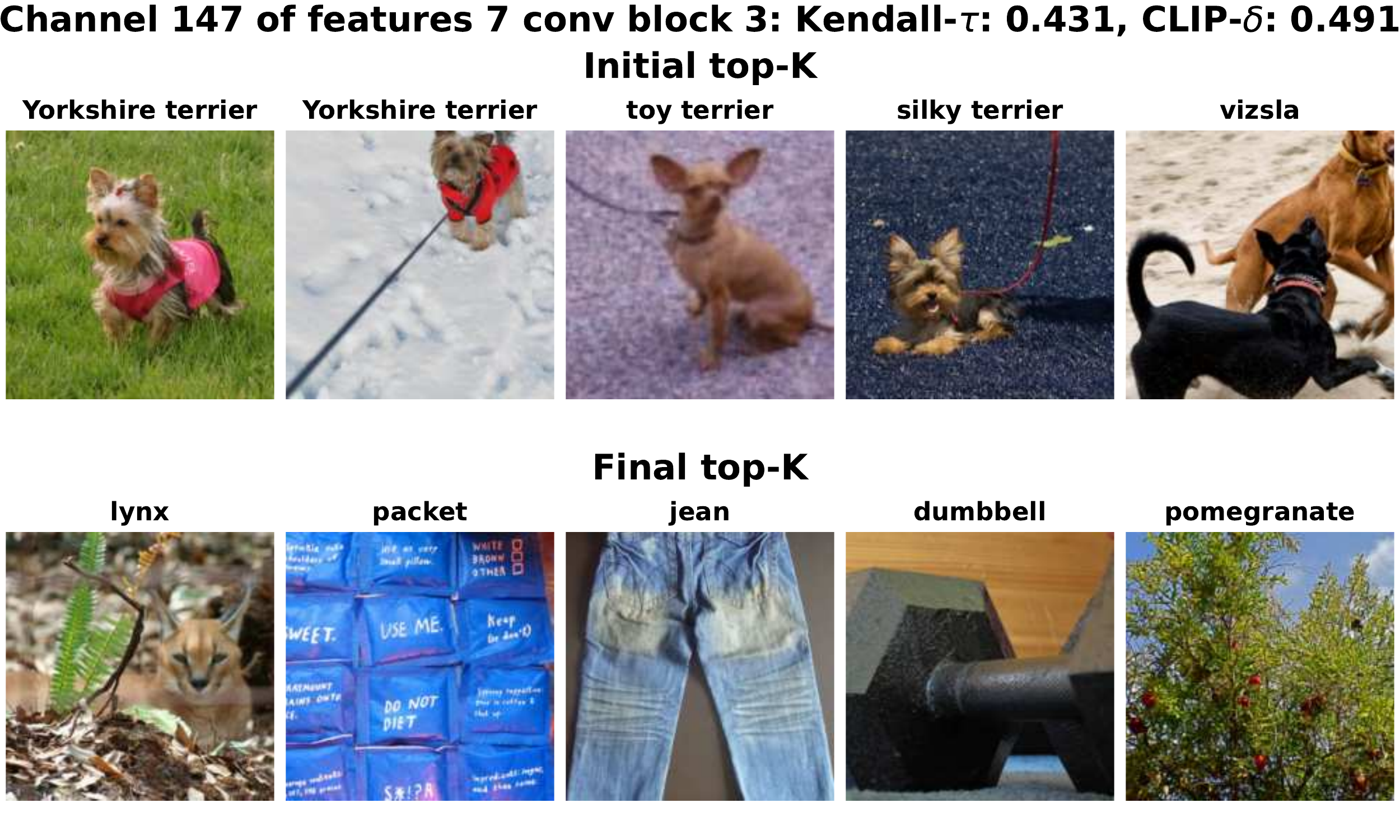}
\end{subfigure}\hfill
\begin{subfigure}[]{0.49\linewidth}
    \includegraphics[width=\textwidth]{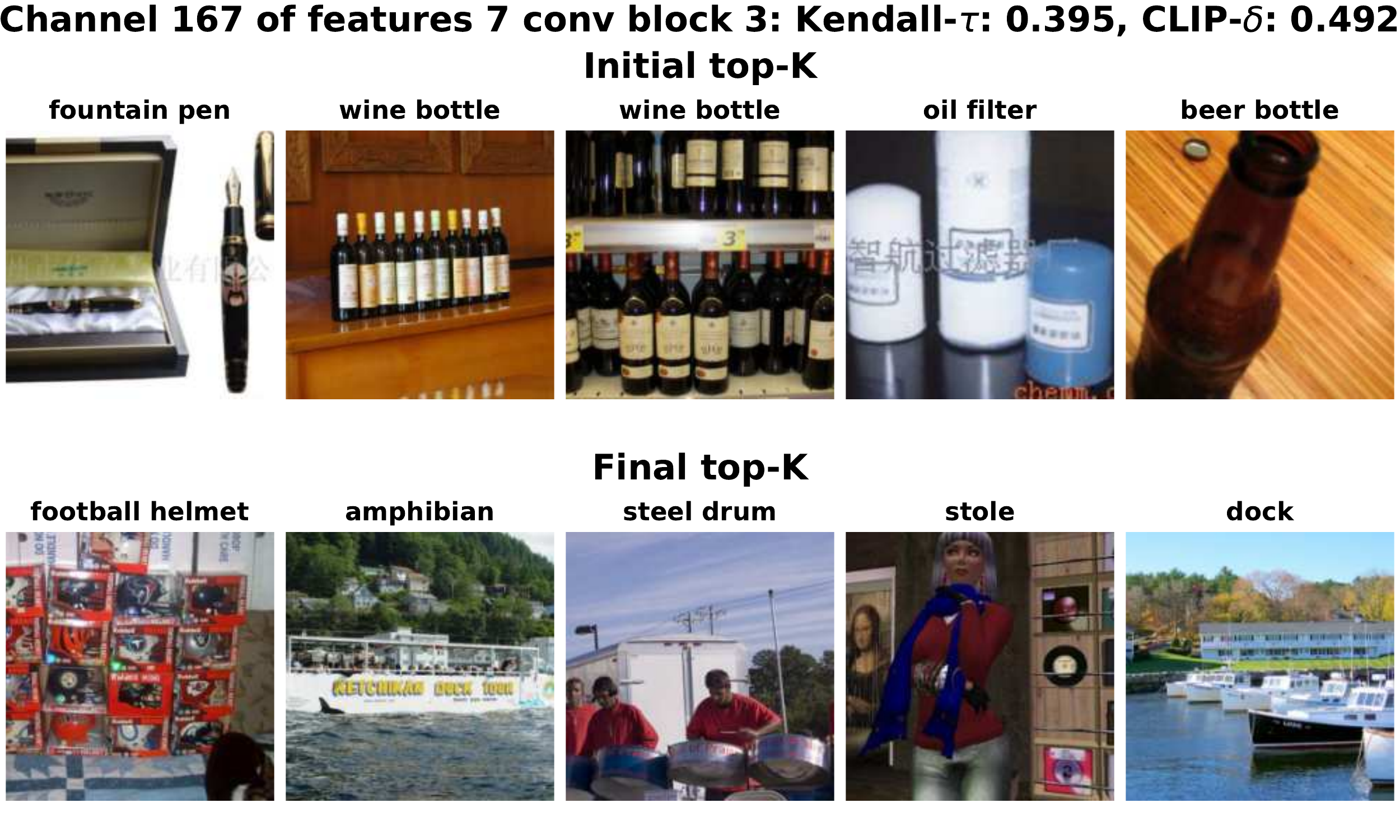}
\end{subfigure}\\
\vspace{.2cm}
\begin{subfigure}[]{0.49\linewidth}
    \includegraphics[width=\textwidth]{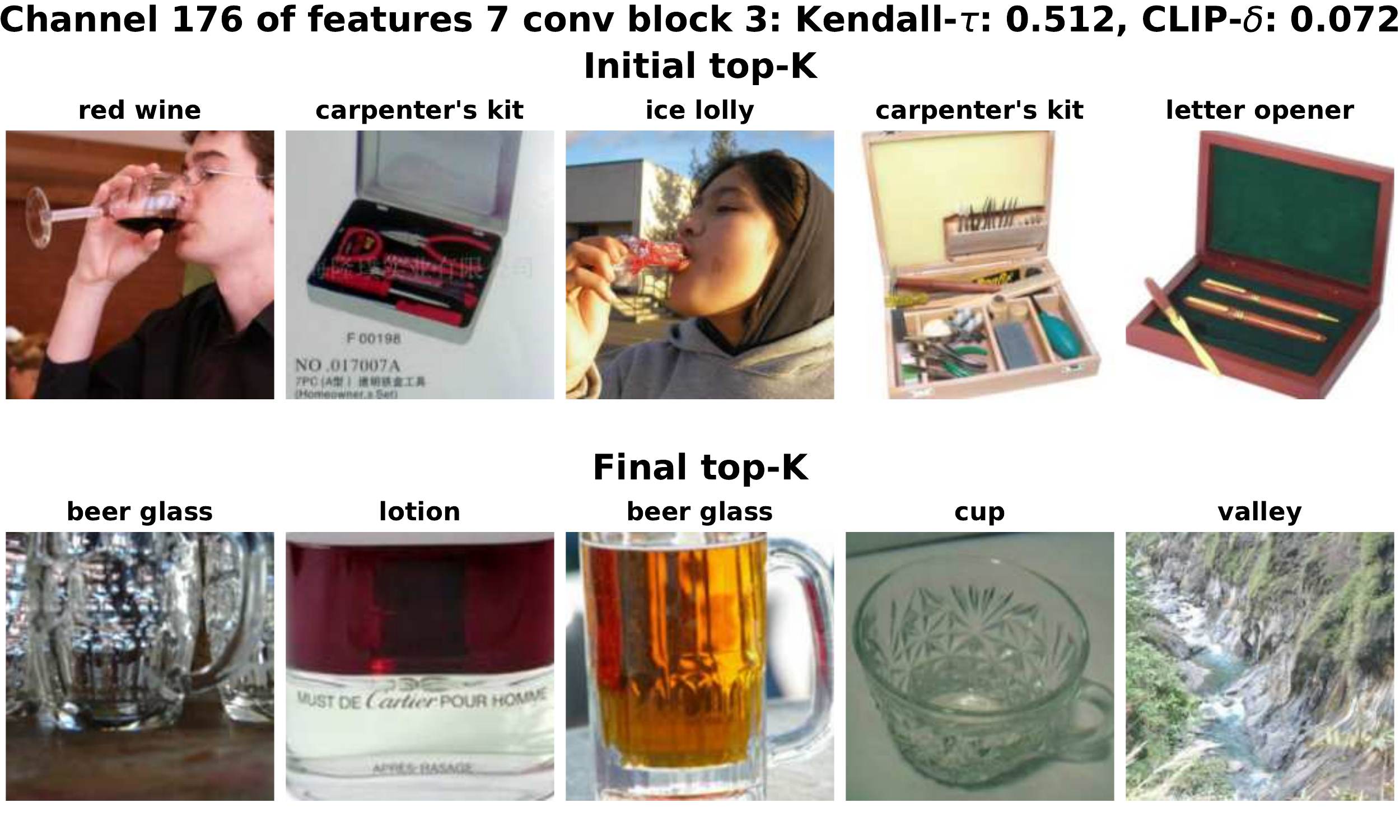}
\end{subfigure}\hfill
\begin{subfigure}[]{0.49\linewidth}
    \includegraphics[width=\textwidth]{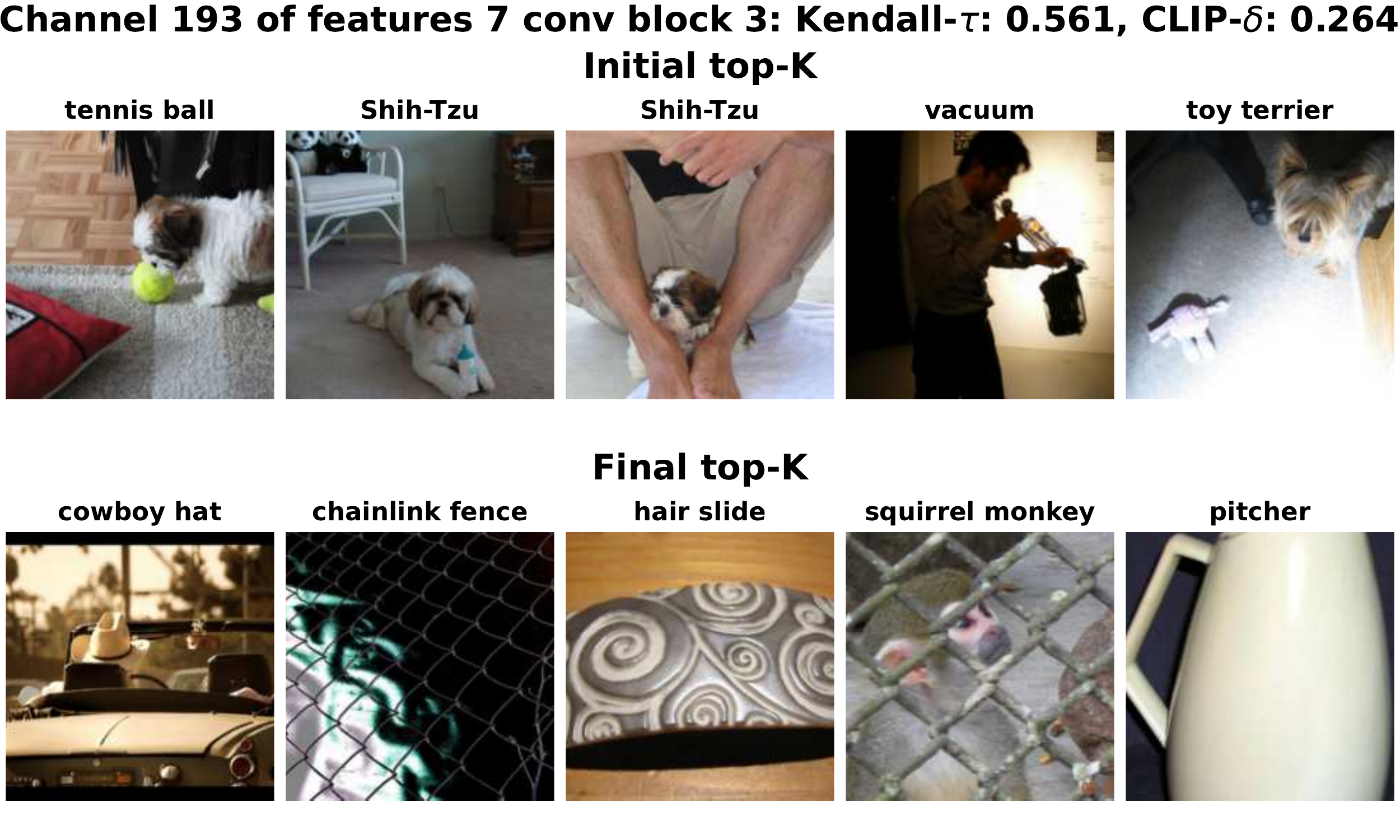}
\end{subfigure}\\
\vspace{.2cm}
\begin{subfigure}[]{0.49\linewidth}
    \includegraphics[width=\textwidth]{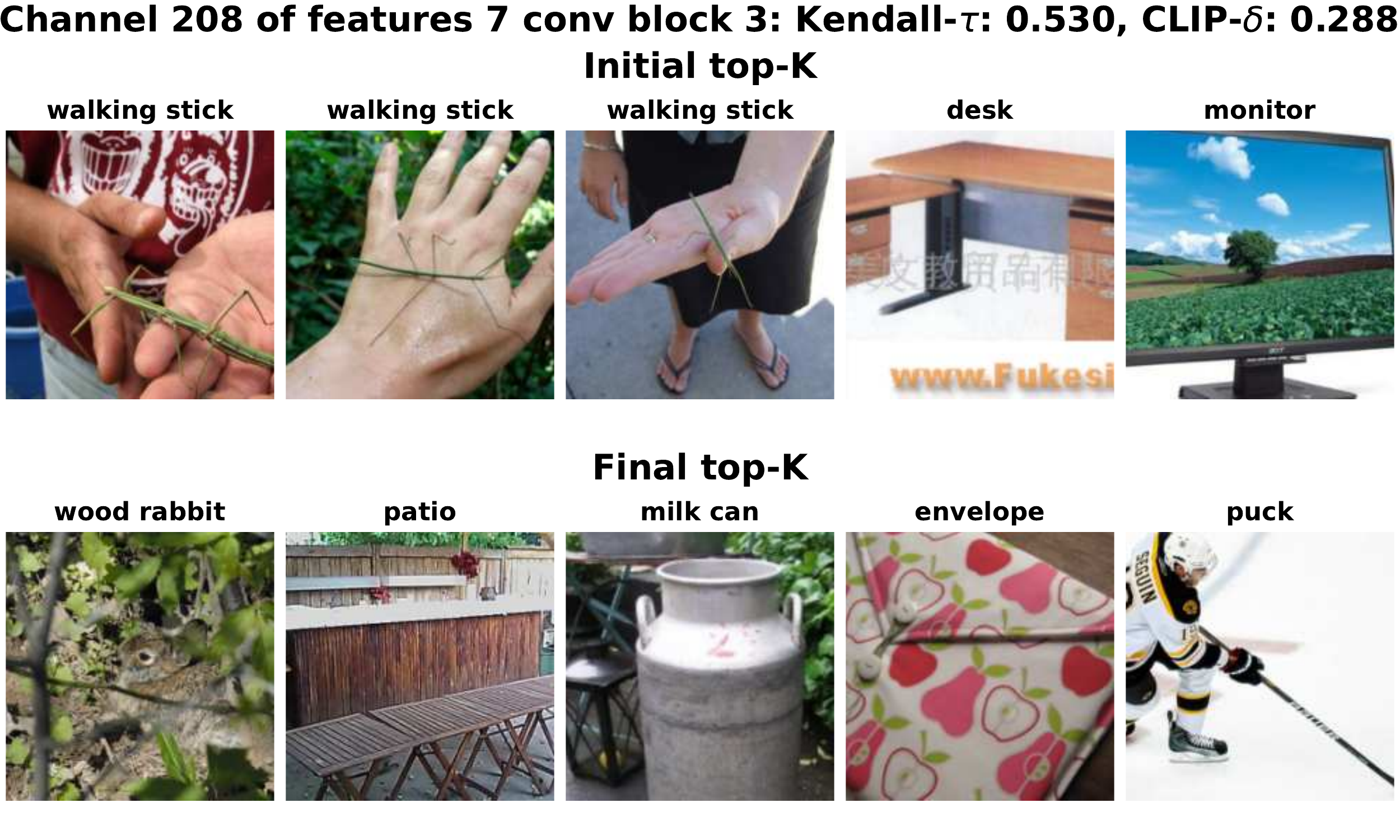}
\end{subfigure}\hfill
\begin{subfigure}[]{0.49\linewidth}
    \includegraphics[width=\textwidth]{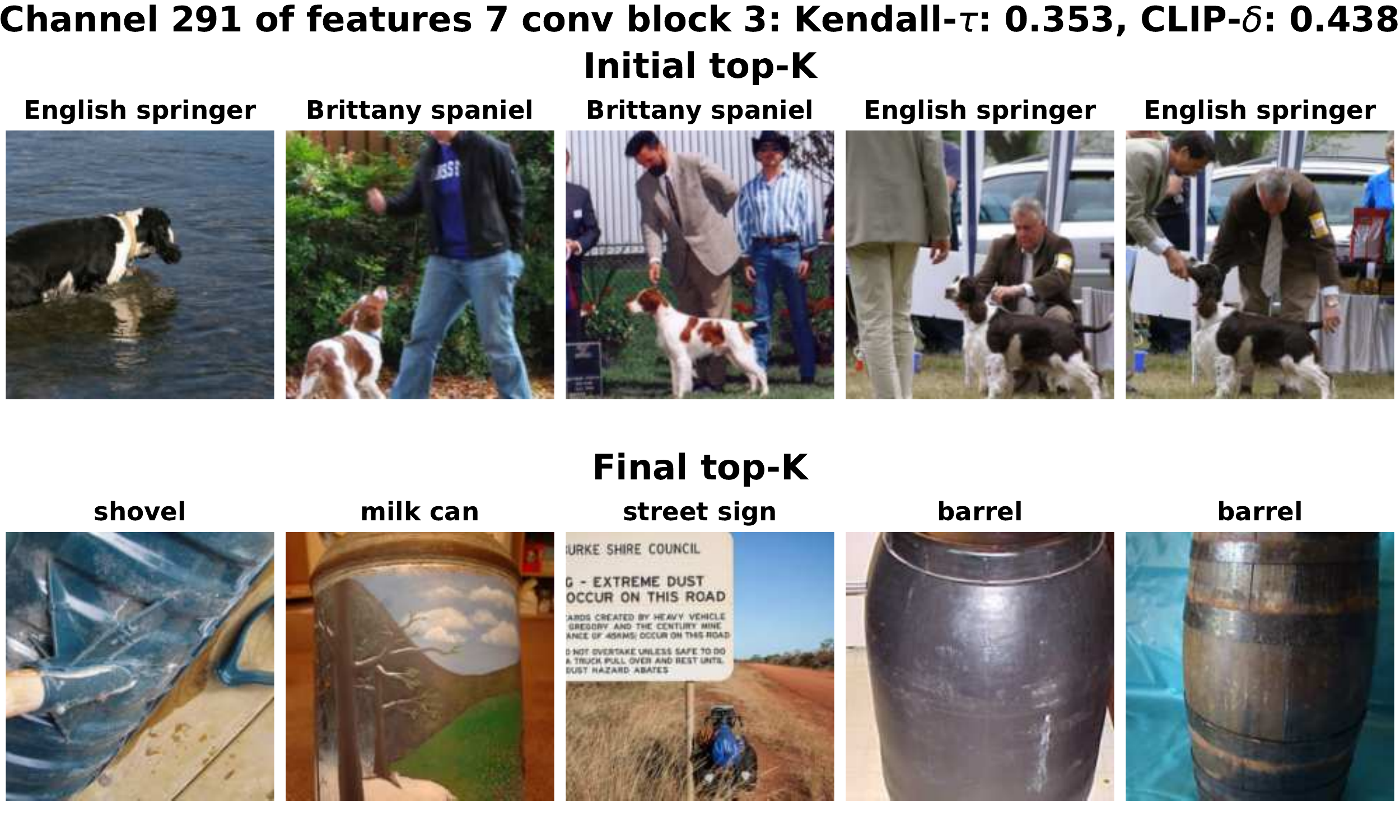}
\end{subfigure}
    \caption{\small Push-down all-channel attack on Feature 7 block 3 of EfficientNet. All initial top-5 images were completely removed from the new set of top-5 images, demonstrating the success of the attack. Channel indexes were randomly chosen.} 
        \label{fig:efficient}
\end{figure}

\subsection{Effect of Depth}
\begin{figure}[!ht]
\centering
\begin{subfigure}[]{0.49\linewidth}
    \includegraphics[width=\textwidth]{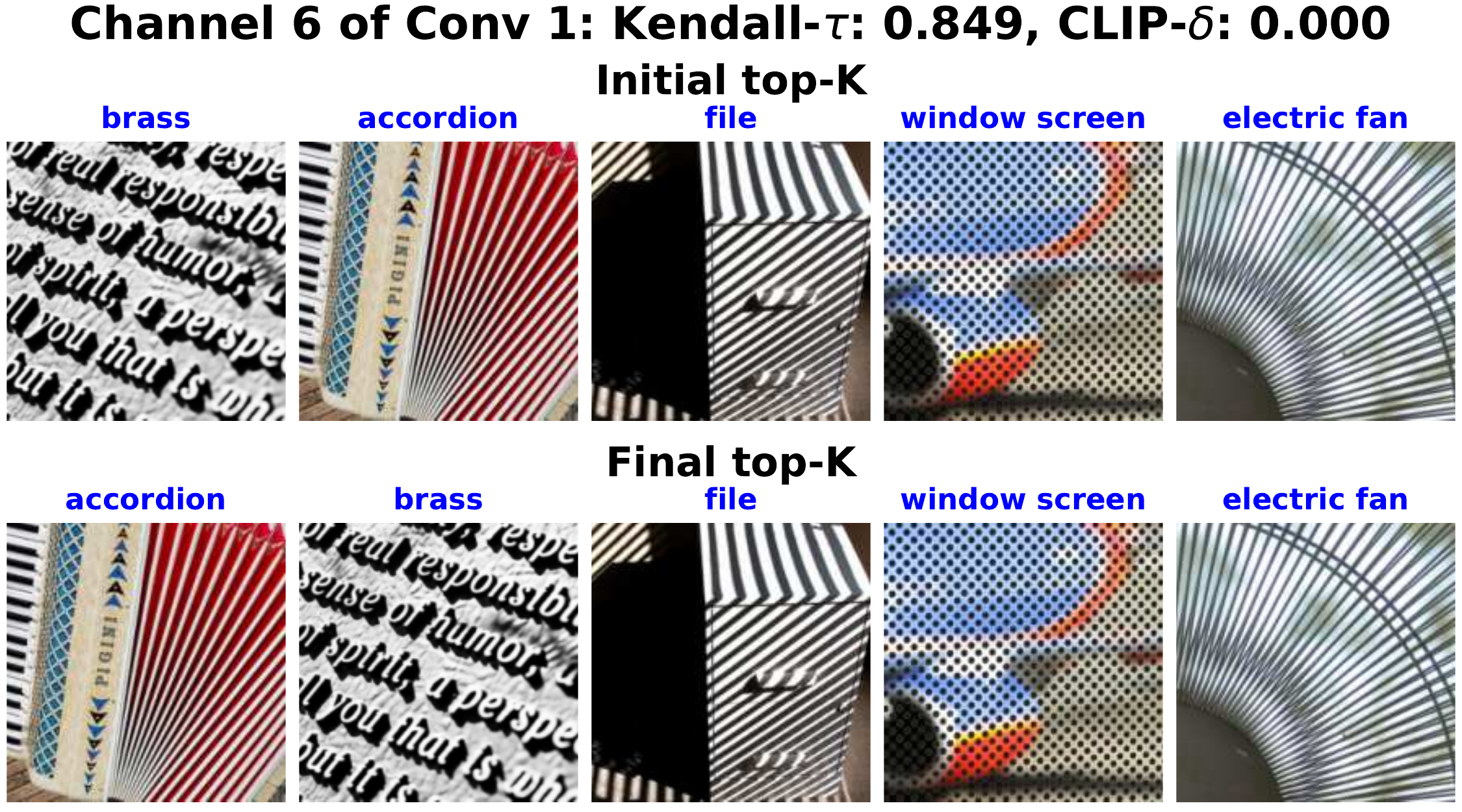}
    \caption{Layer: Conv1.}
\end{subfigure}\hfill
\begin{subfigure}[]{0.49\linewidth}
    \includegraphics[width=\textwidth]{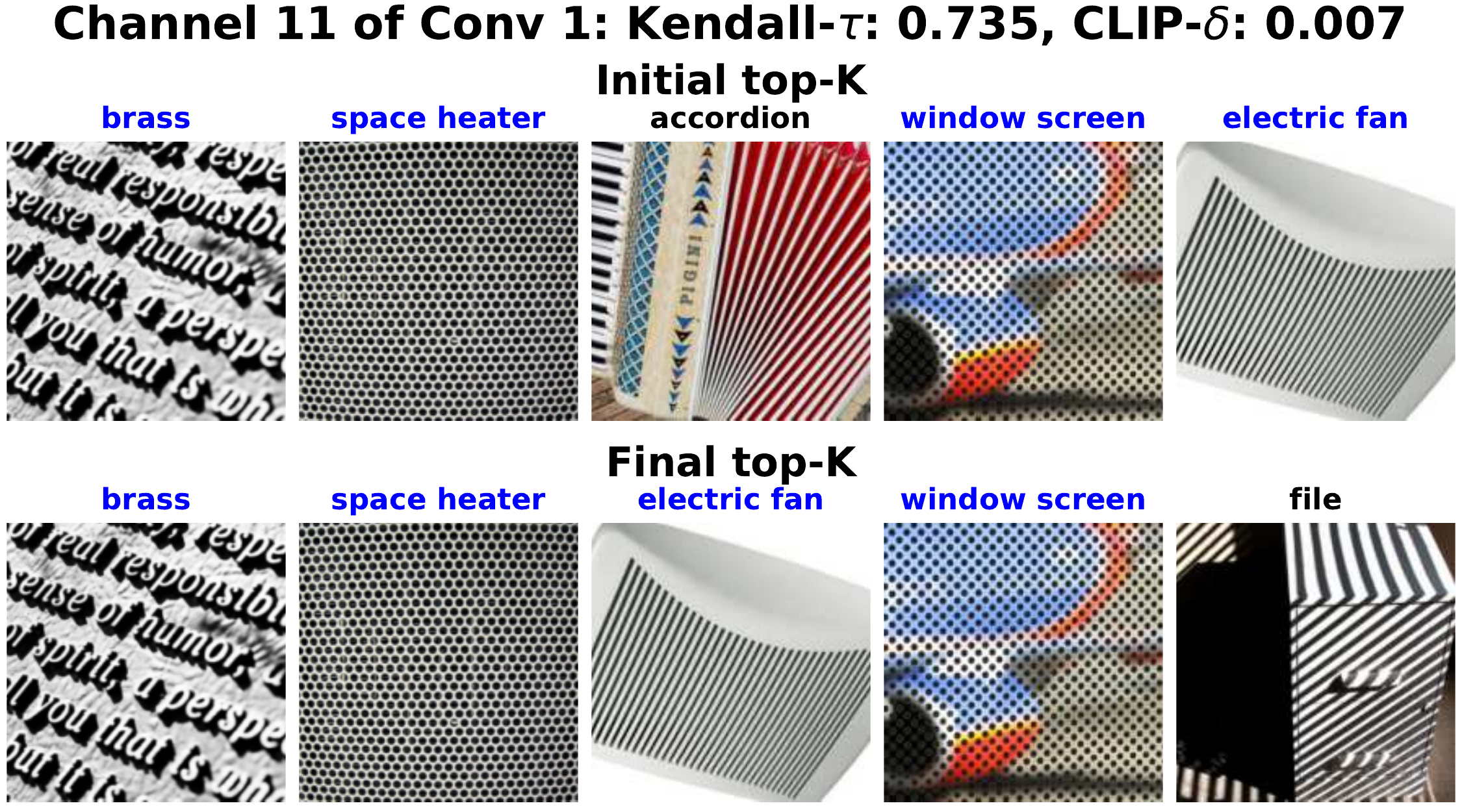}
    \caption{Layer: Conv1.}
\end{subfigure}\\
\vspace{.8cm}
\begin{subfigure}[]{0.49\linewidth}
    \includegraphics[width=\textwidth]{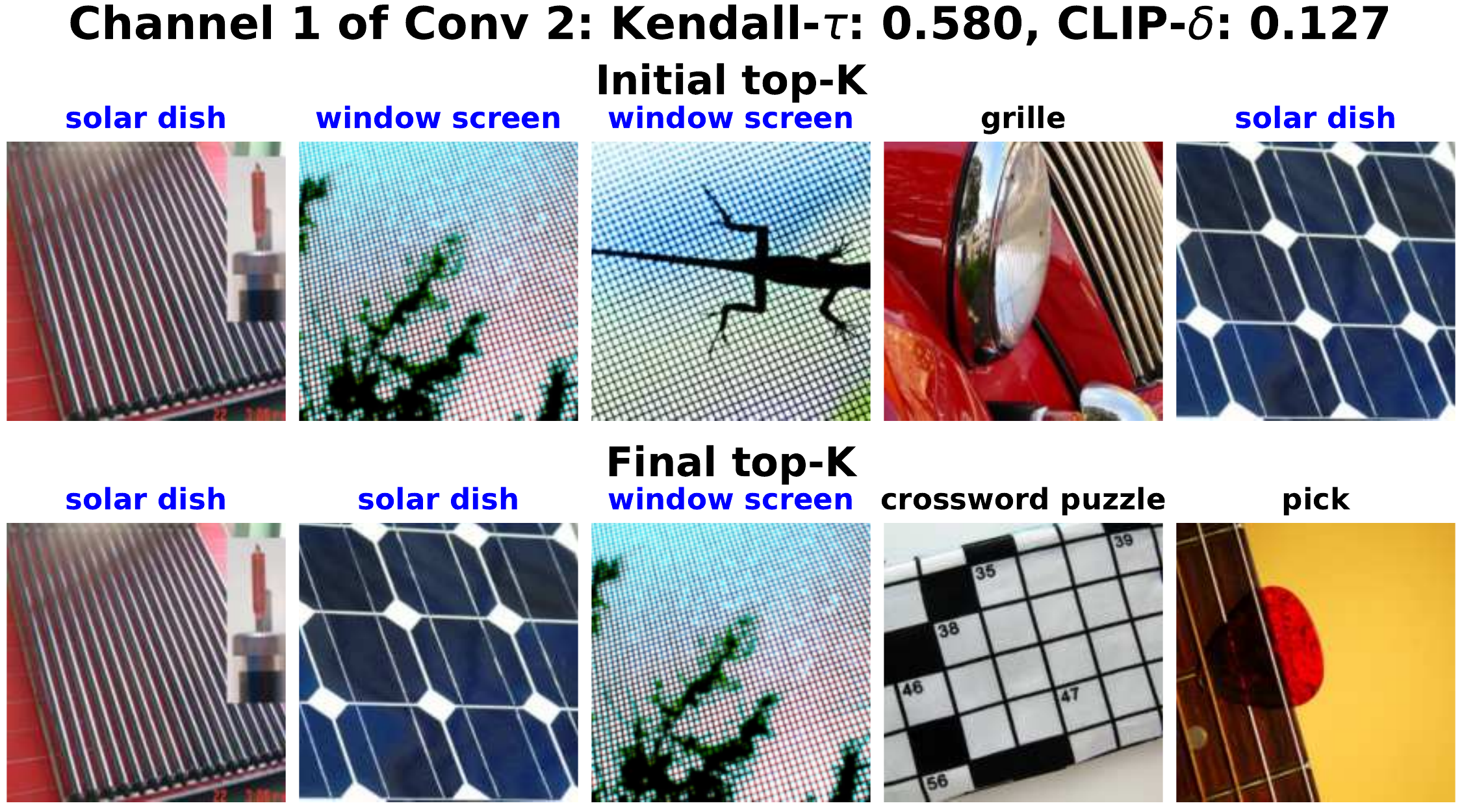}
    \caption{Layer: Conv2.}
\end{subfigure}\hfill
\begin{subfigure}[]{0.49\linewidth}
    \includegraphics[width=\textwidth]{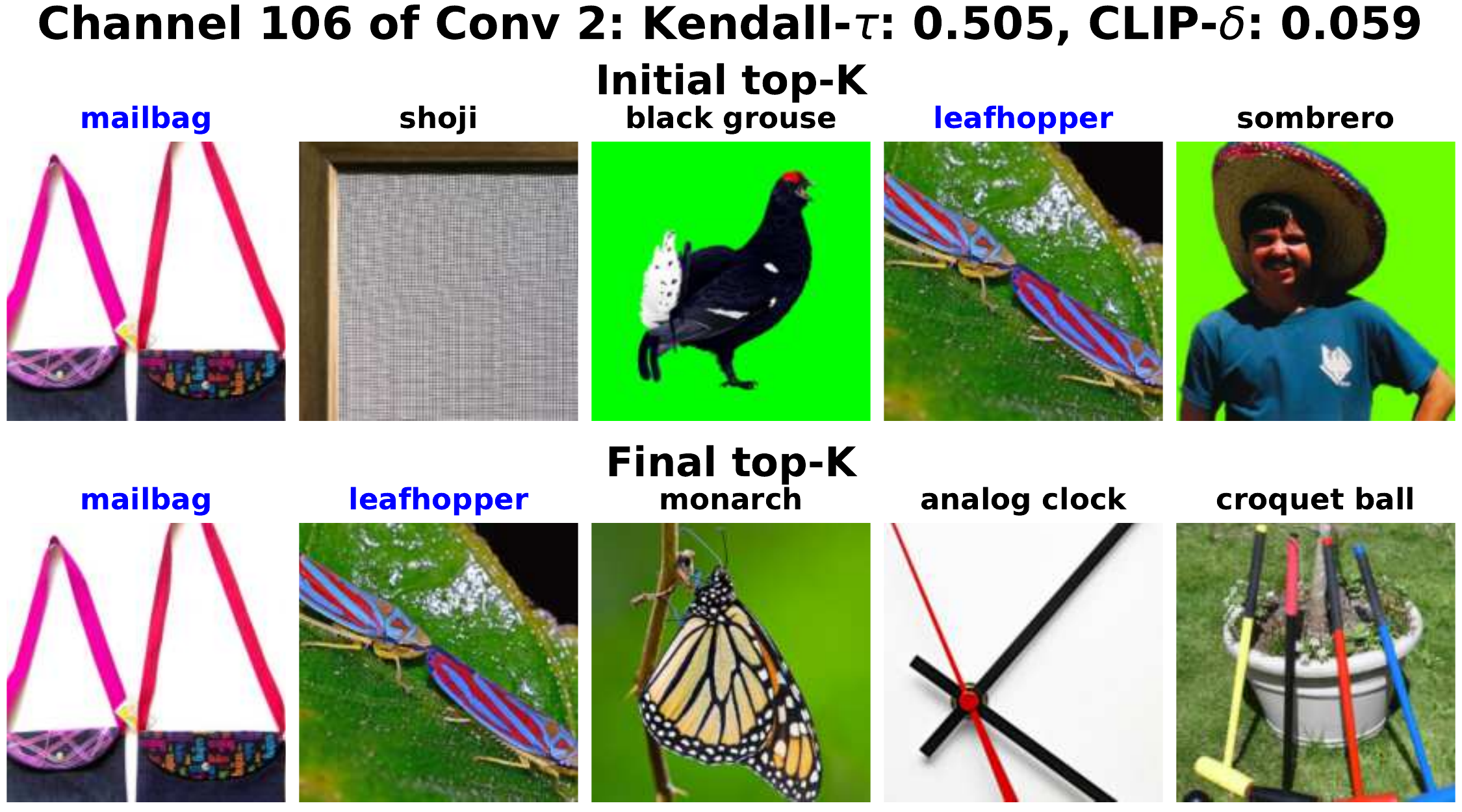}
    \caption{Layer: Conv2.}
\end{subfigure}\\
\vspace{.8cm}
\begin{subfigure}[]{0.49\linewidth}
    \includegraphics[width=\textwidth]{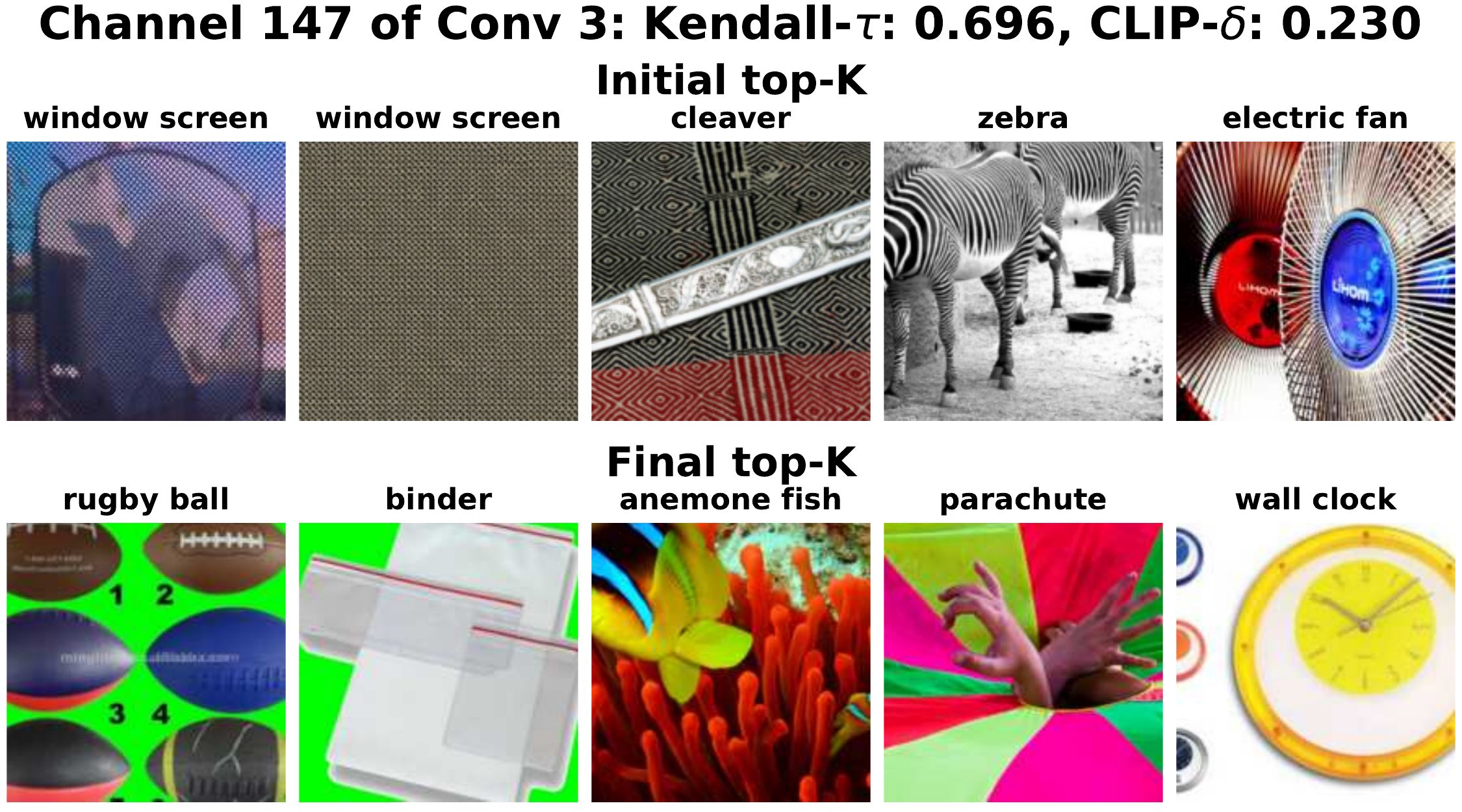}
    \caption{Layer: Conv3.}
\end{subfigure}\hfill
\begin{subfigure}[]{0.49\linewidth}
    \includegraphics[width=\textwidth]{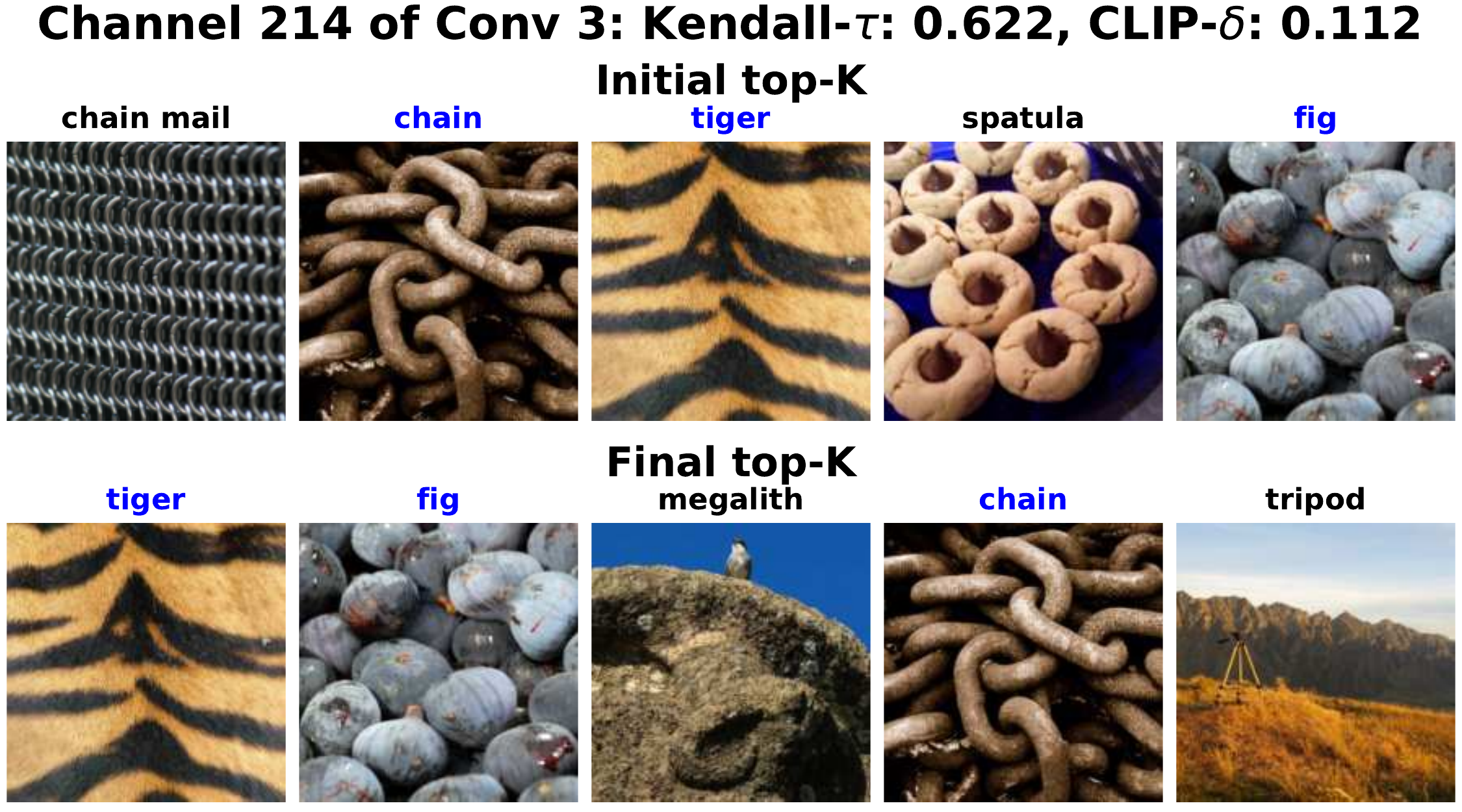}
    \caption{Layer: Conv3.}
\end{subfigure}\\
\vspace{.8cm}
\begin{subfigure}[]{0.49\linewidth}
    \includegraphics[width=\textwidth]{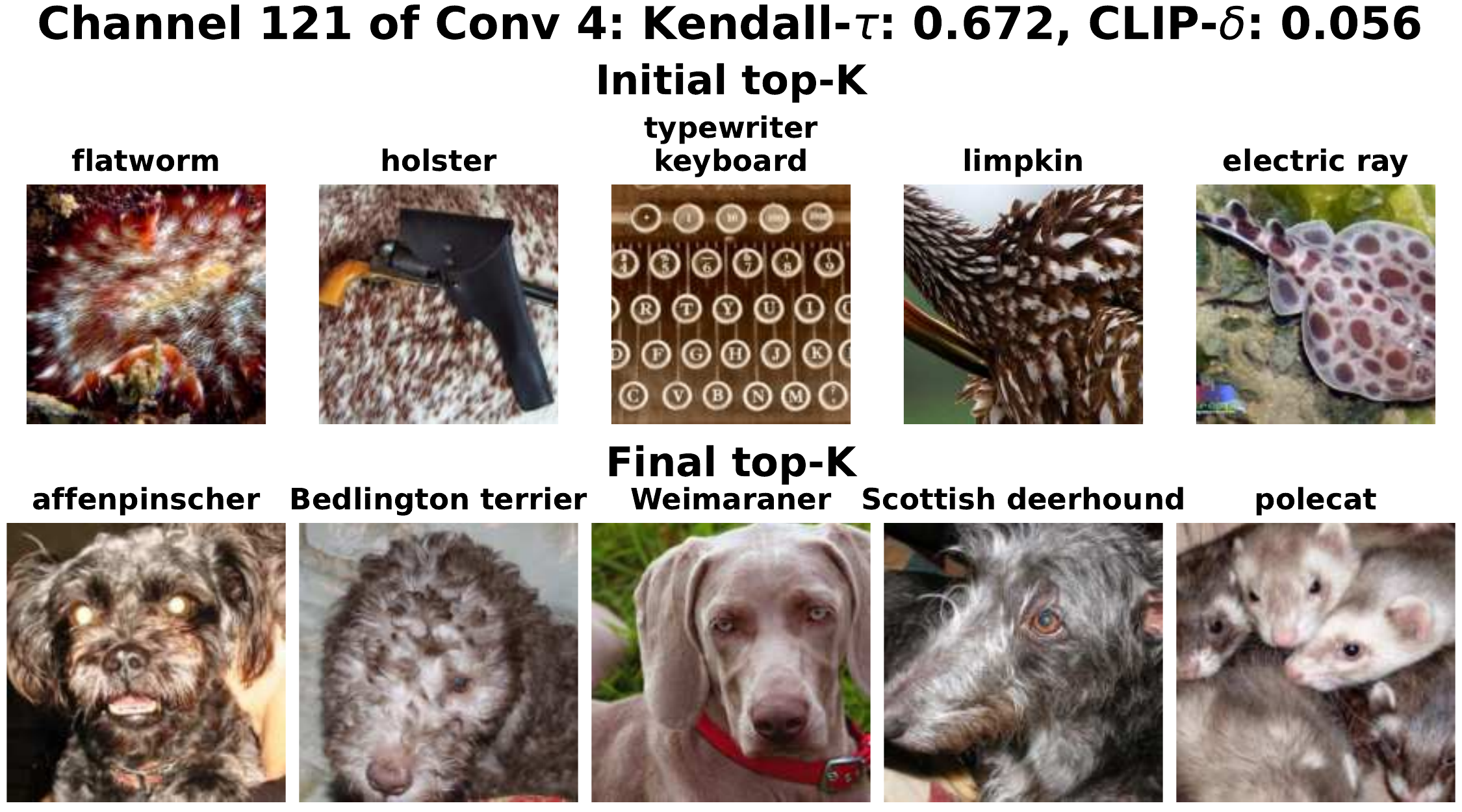}
    \caption{Layer: Conv4.}
\end{subfigure}\hfill
\begin{subfigure}[]{0.49\linewidth}
    \includegraphics[width=\textwidth]{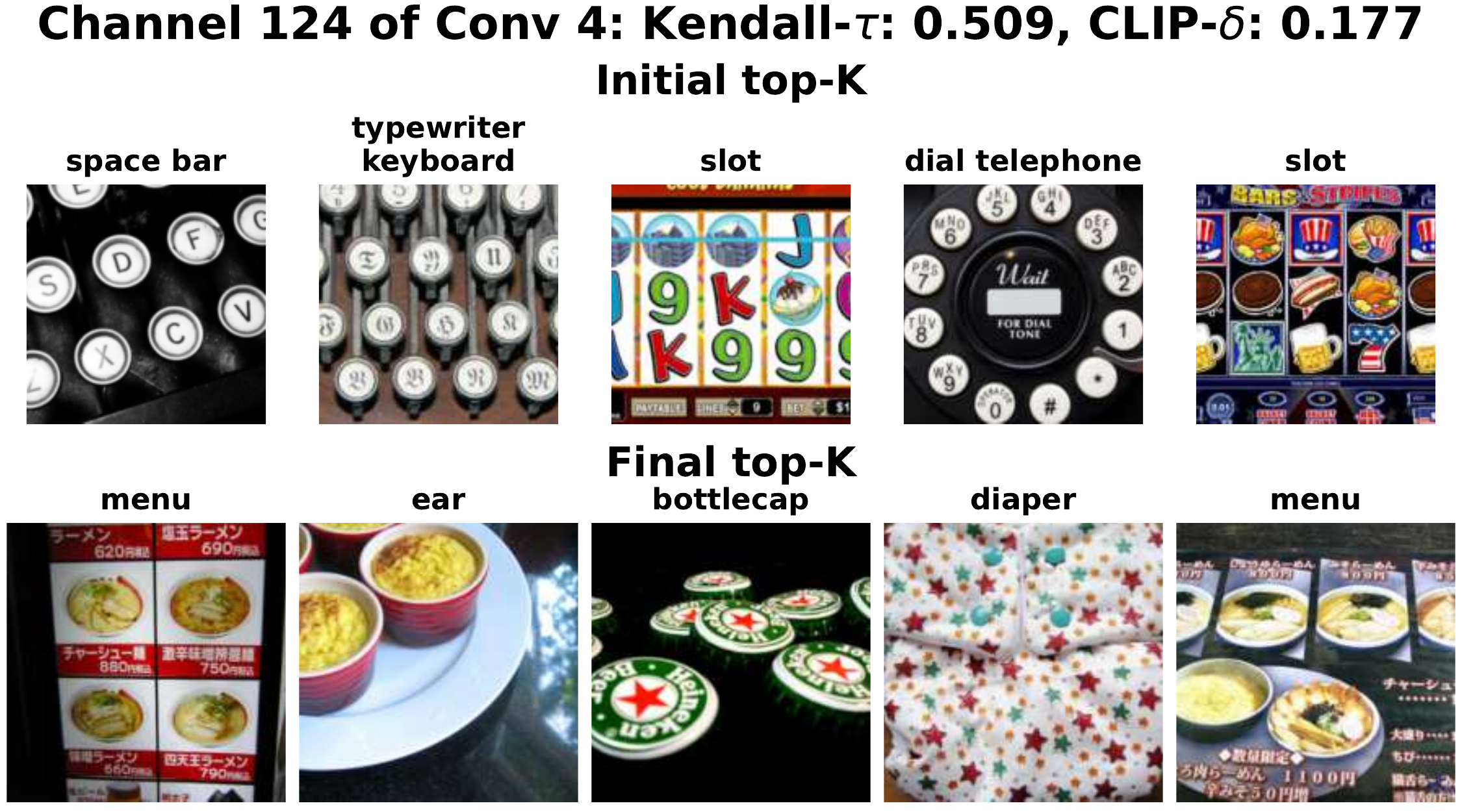}
    \caption{Layer: Conv4.}
\end{subfigure}
    \caption{\small Push-down all-channel attack of on several layers of AlexNet.  Channels indexes were selected randomly. While there are some changes in top-activating images of early layers (conv1 and conv2), they are not significant as materialized by low values of CLIP-$\delta$ and high values of Kendall-$\tau$. For conv3 and conv4, we see a complete replacement of top-5 images on channels 147 (conv3), 121 (conv4), and 124 (conv4).} 
        \label{fig_add:all_channel_depth}
\end{figure}

We vary different layers of AlexNet and evaluate how the attack is affected by depth. 
Figure~\ref{fig_add:all_channel_depth} shows results obtained on randomly chosen channels from conv1, conv2, conv3, and conv4 of AlexNet. It can be observed that the earliest layers conv1 and conv2 are harder to attack. This is materialized by high values of Kendal-$\tau$ and  low values of CLIP-$\delta$ scores. When increasing the depth (conv3 and conv4) we observe a complete replacement in top-$5$ images in channels 147 (conv3), 121 (conv4) and 124 (conv4), although some of these channels have low values of CLIP-$\delta$ scores. 

\clearpage

\subsection{Additional Illustrations for Whack-a-mole}
This section provides further investigations into the existence of the whack-a-mole problem for the push-down attack on AlexNet. 

\begin{figure}[!ht]
\centering
    \includegraphics[width=0.5\textwidth]{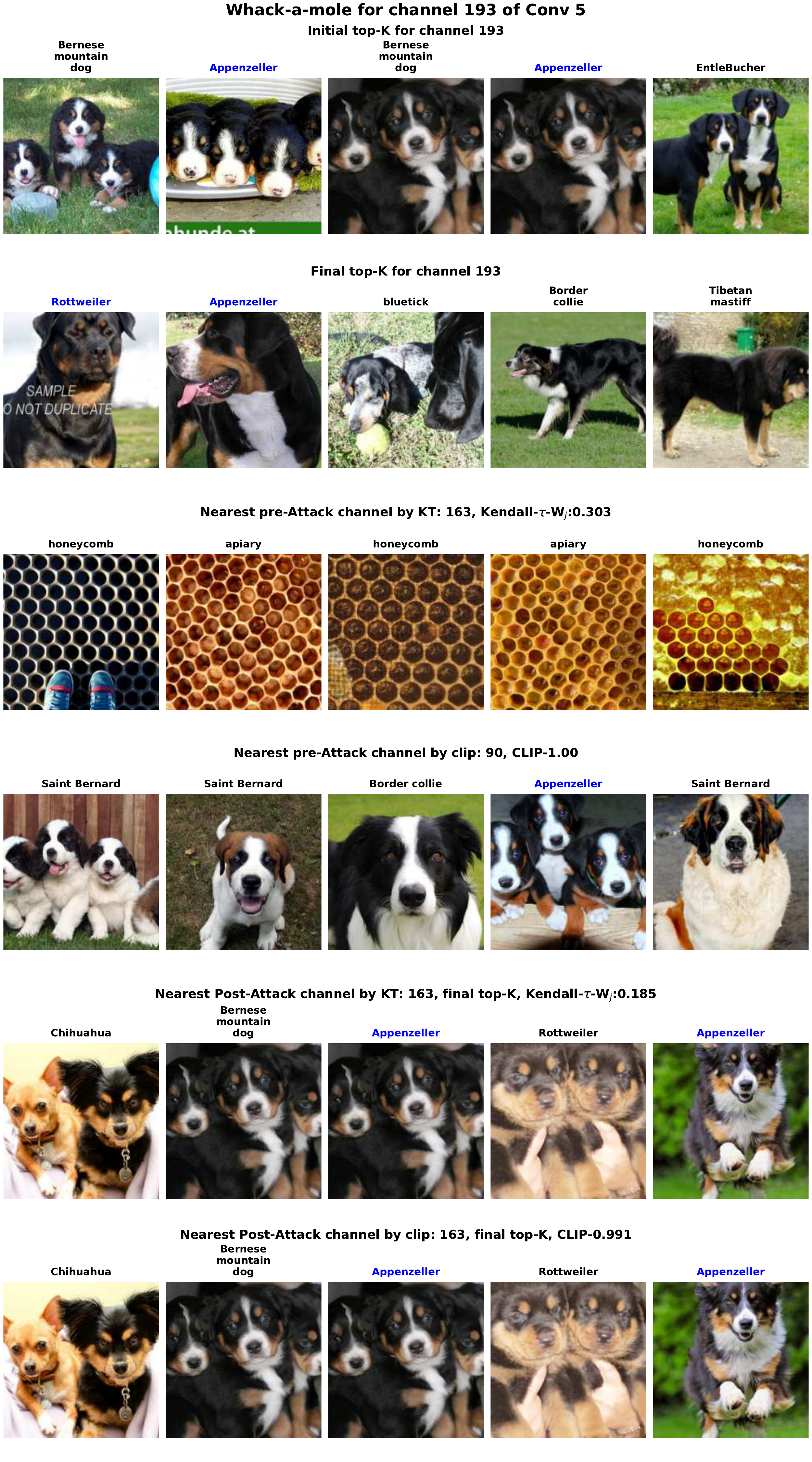}
    \caption{\small Illustrations for the existence of whack-a-mole on the channel 193, found as one of the "hard" case (as presented in Section~\ref{section:single_chanel}, Figure~\ref{fig:whack-a-mole}). The first two rows show the initial and final top-$k$ images for the targeted channel. The third and fourth rows show the initial nearest channels w.r.t. Kendall-$\tau$-$W_j$ and CLIP-$\delta$-$W_j$, respectively. The fifth and sixth rows show the nearest post-attack channel according to Kendall-$\tau$-$W_j$  and CLIP-$\delta$-$W_j$ respectively.} 
        \label{fig_add:whak-a-mole_193}
\end{figure}

\paragraph{Zoom onto Channel 193 for Whak-a-mole.}
We begin by showing the full overview of the behavior of channel 193, selected as one  "hard" case where similar initial images are found in final (post-attack) top-$k$ images of another channel. 
As discussed in Section~\ref{section:single_chanel}, although similar initial images for channel 193 were found in channel 163 after the attack, it appears from the second row of Figure~\ref{fig_add:whak-a-mole_193} that channel 193 was initially highly correlated with the channel 90 according to CLIP-$\delta$ score. Moreover, the fact that the CLIP-$\delta$-$W_j$ is $0.991<1$ shows that the nearest post-attack channel (channel 163) is not more correlated than the nearest pre-attack channel (channel 90) according to CLIP scores. This, therefore, limits the existence of the whack-a-mole problem on this channel.

\begin{figure}[!ht]
\centering
\begin{subfigure}[]{0.49\linewidth}
    \includegraphics[width=\textwidth]{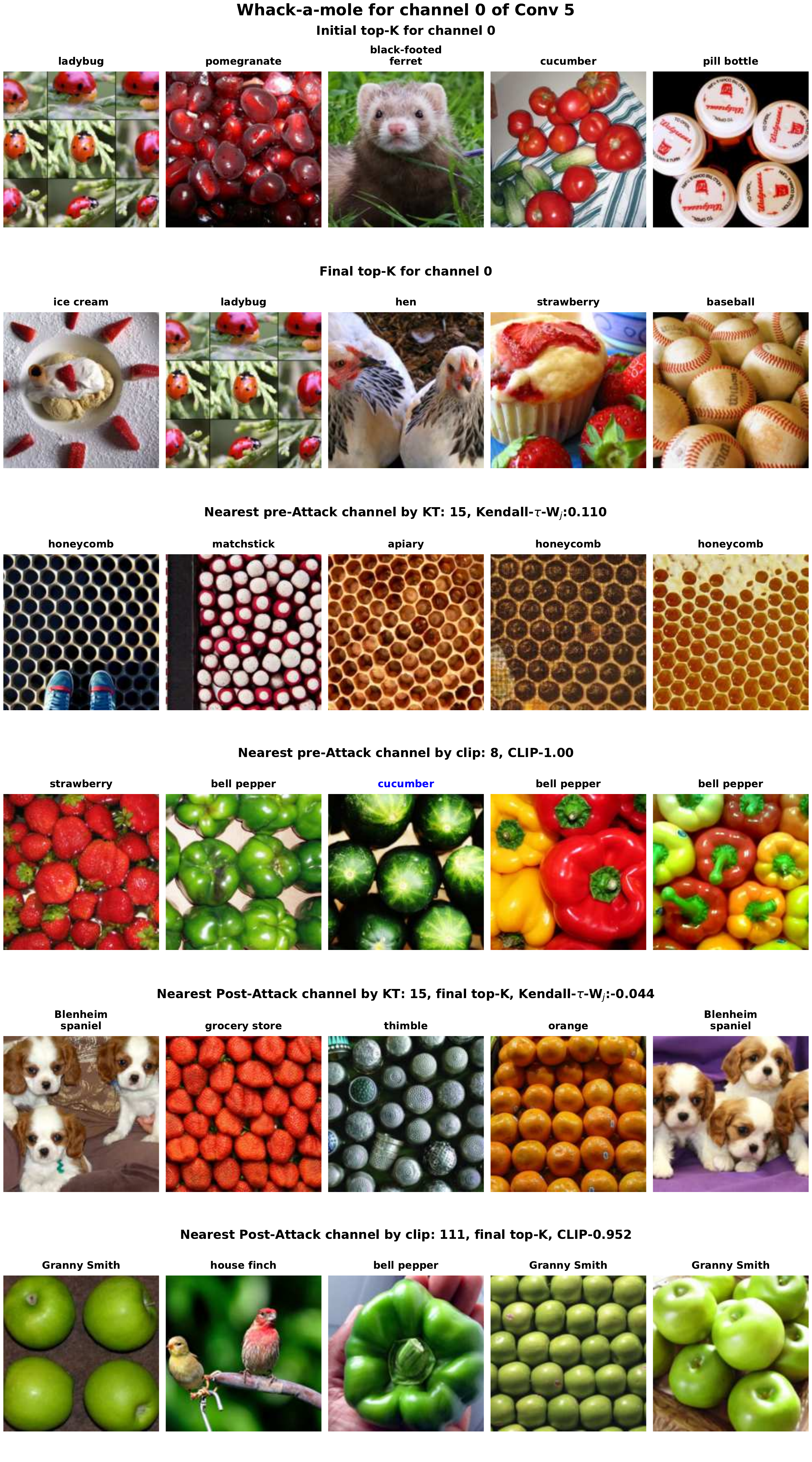}
    \caption{Targeted channel: 0.}
\end{subfigure}\hfill
\begin{subfigure}[]{0.49\linewidth}
    \includegraphics[width=\textwidth]{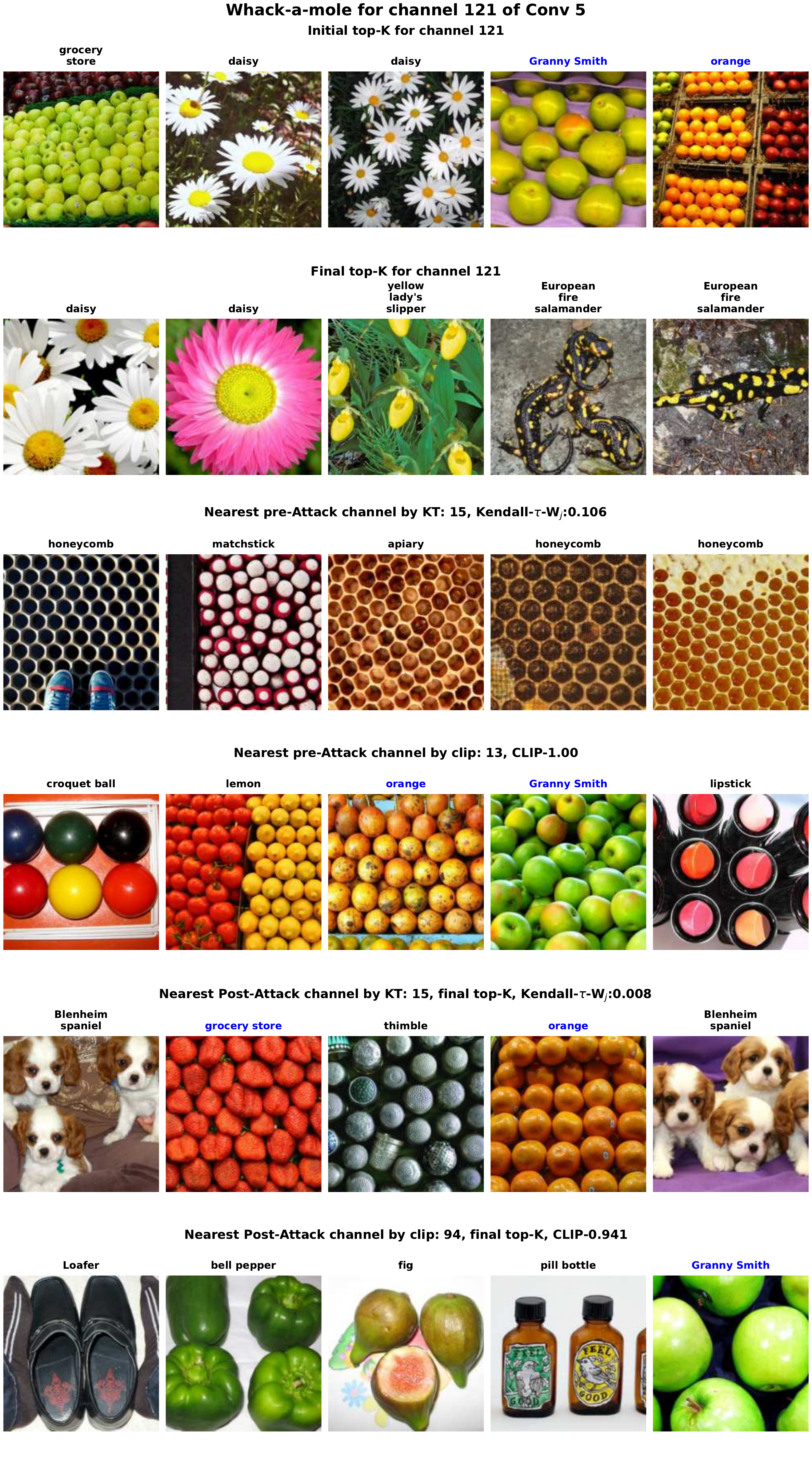}
    \caption{Targeted channel: 121.}
\end{subfigure}
    \caption{\small Illustrations for the existence of whack-a-mole on two randomly chosen channels. The first two rows show the initial and final top-$k$ images for the targeted channel. The third and fourth rows show the initial nearest channels w.r.t. Kendall-$\tau$-$W_j$ and CLIP-$W_j$, respectively. The fifth and sixth rows show the nearest post-attack channel according to Kendall-$\tau$-$W_j$  and CLIP-$W_j$, respectively.} 
        \label{fig_add:whak-a-mole}
\end{figure}
\paragraph{Additional Investigation of Potential Existence of Whack-a-mole.}
These randomly selected examples support the general findings reported in figure-\ref{fig:geo_means_clip_sims}. While certain channels may have similar top images to specific post-attack channels, it is generally the case that even the most similar channels are distinct. In figure-\ref{fig_add:whak-a-mole}, the two bottom rows denote the top 5 images of the most similar channels to the pre-attack channel measured by the Kendall-$\tau$ and CLIP-$W_j$ respectively. 
\clearpage

\subsection{Additional Illustrations for the Push-up Attack}
\begin{figure}[!ht]
\centering
\begin{subfigure}[]{0.49\linewidth}
    \includegraphics[width=\textwidth]{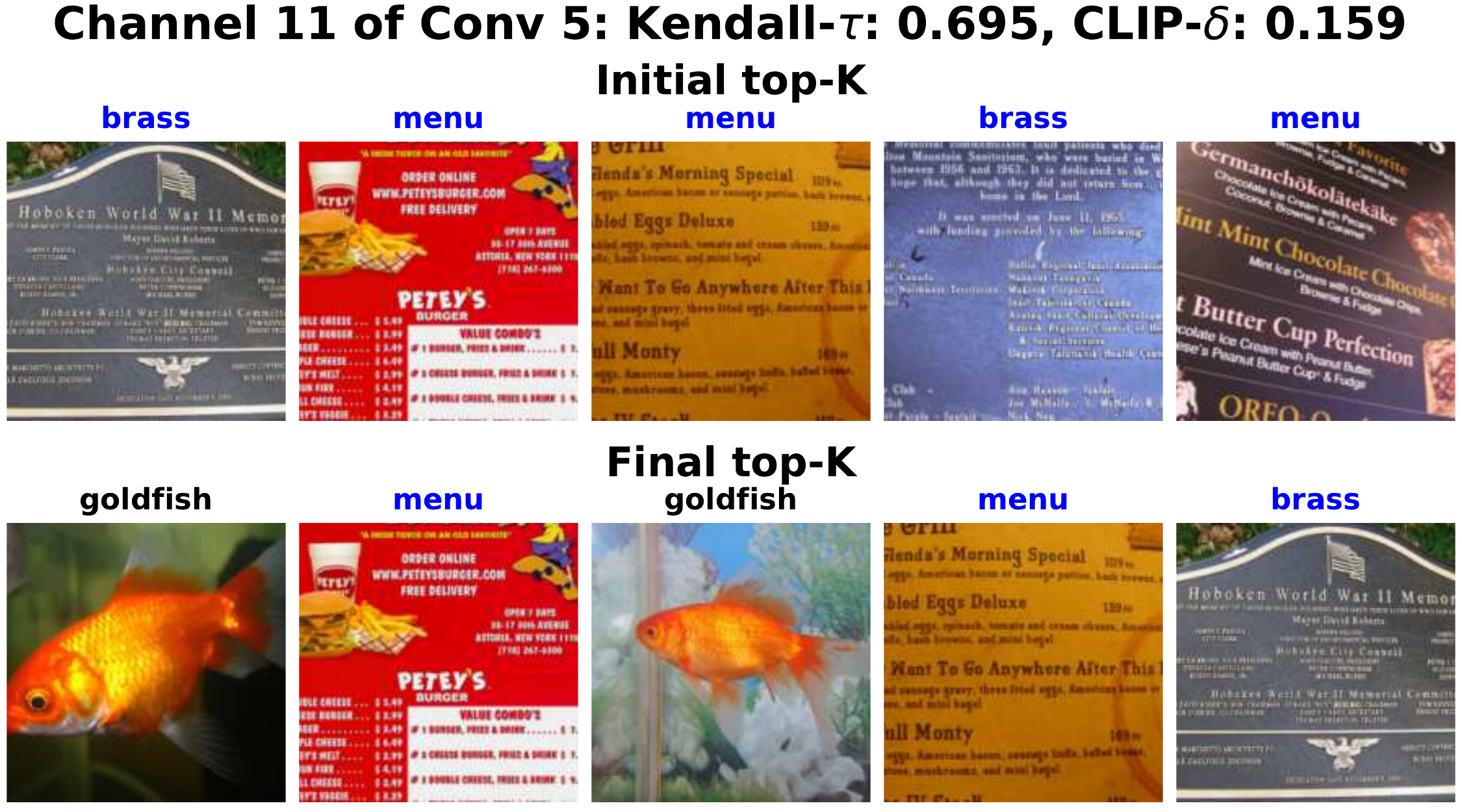}
\end{subfigure}\hfill
\begin{subfigure}[]{0.49\linewidth}
    \includegraphics[width=\textwidth]{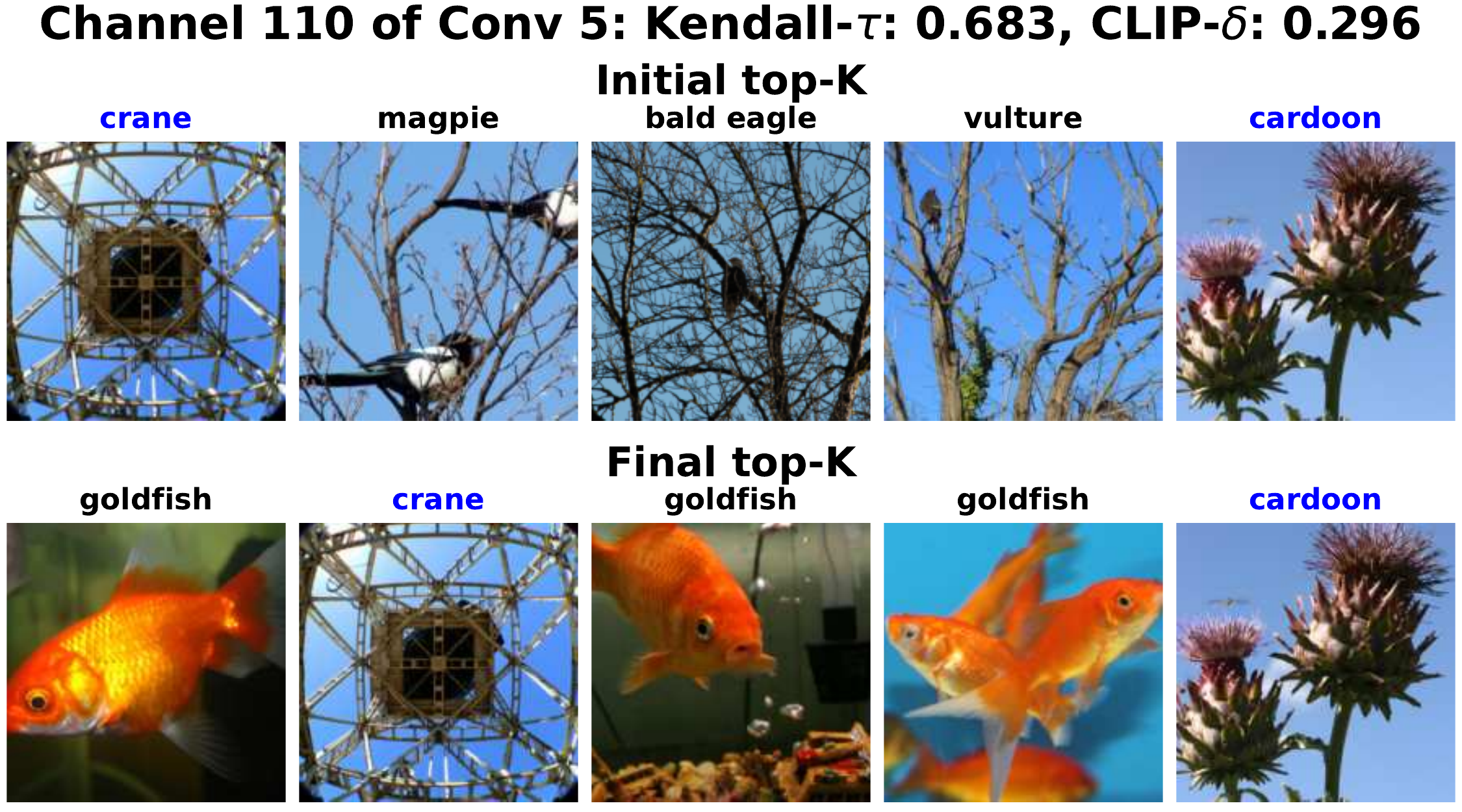}
\end{subfigure}\\
\vspace{.8cm}
\begin{subfigure}[]{0.49\linewidth}
    \includegraphics[width=\textwidth]{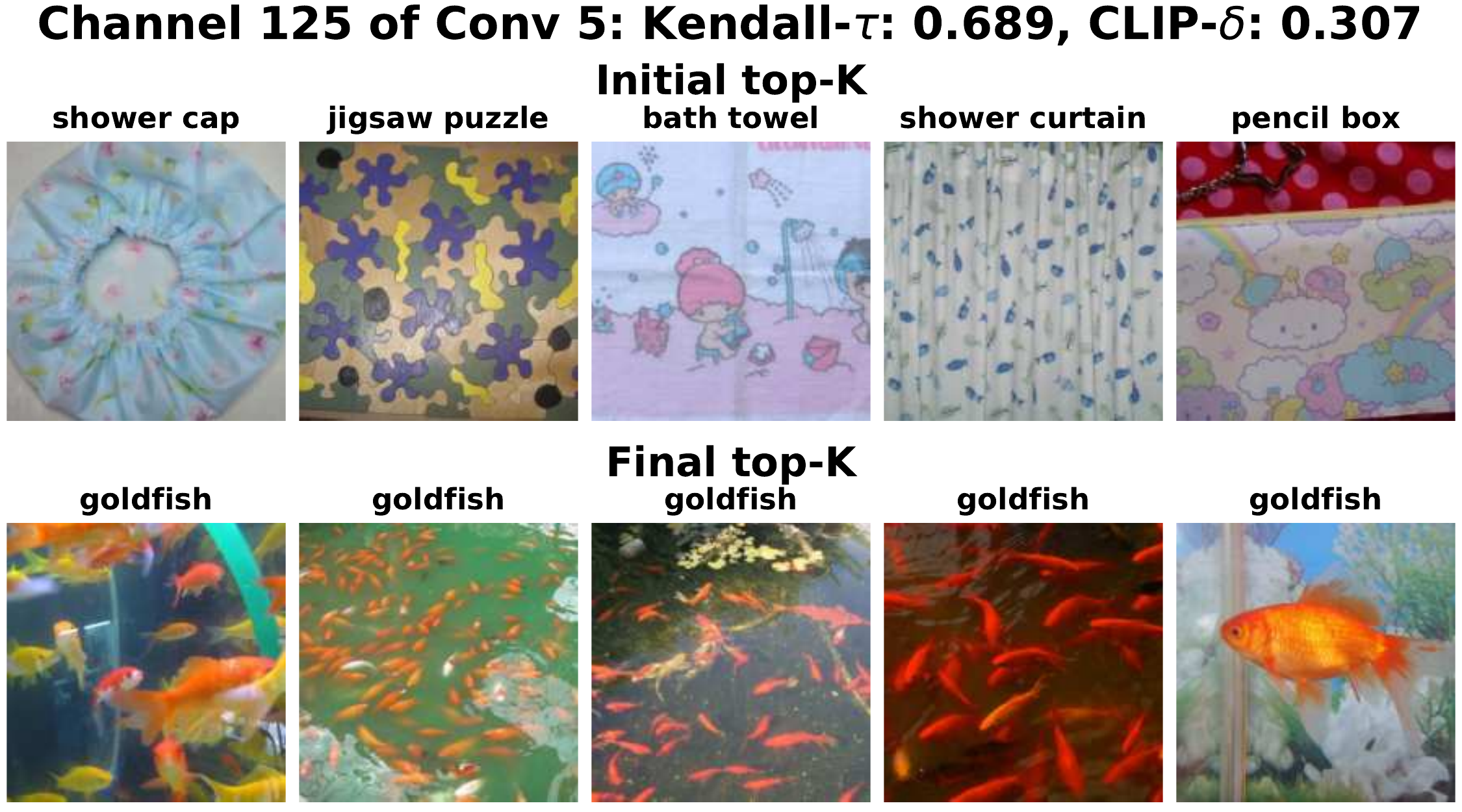}
\end{subfigure}\hfill
\begin{subfigure}[]{0.49\linewidth}
    \includegraphics[width=\textwidth]{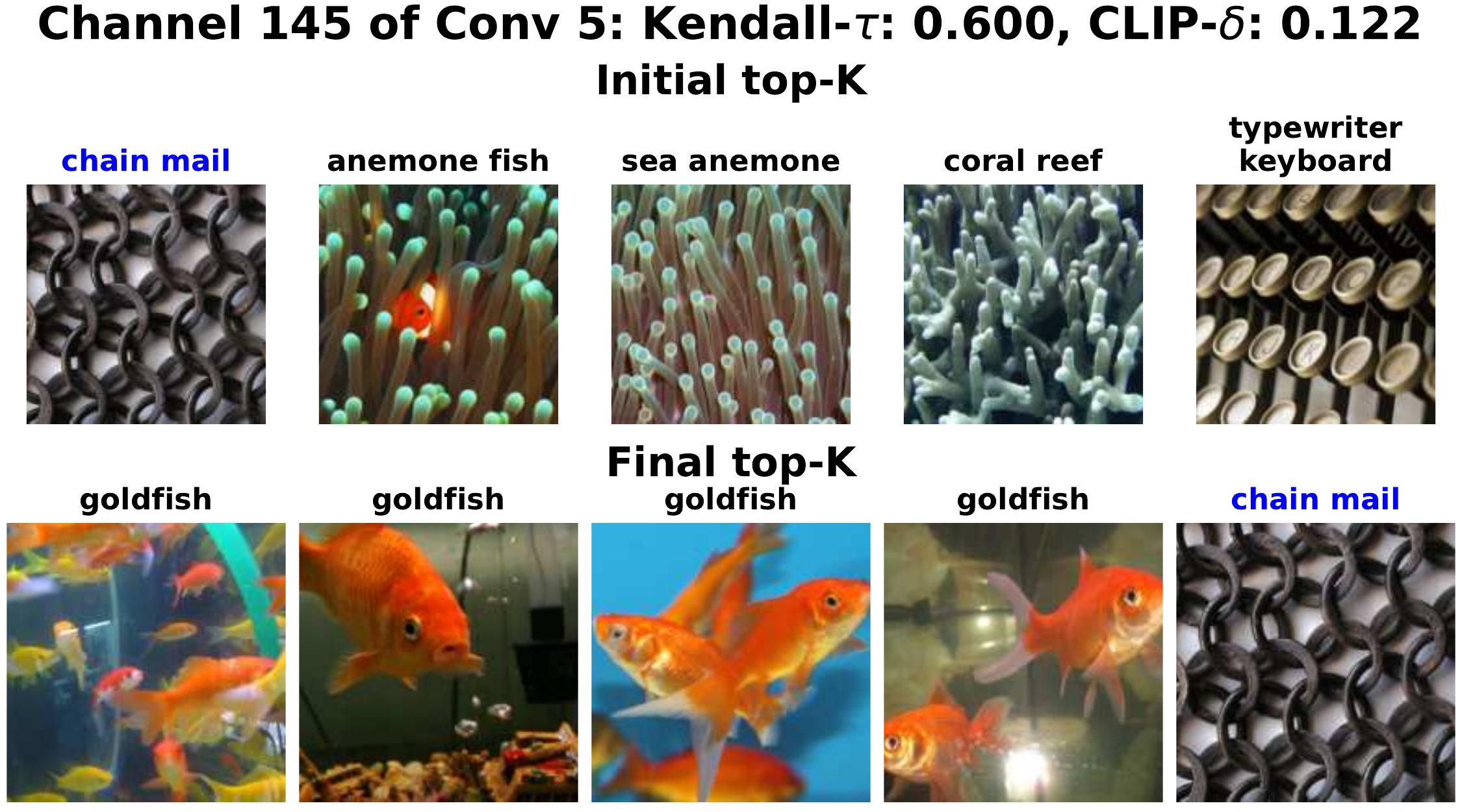}
\end{subfigure}\\
\vspace{.8cm}
\begin{subfigure}[]{0.49\linewidth}
    \includegraphics[width=\textwidth]{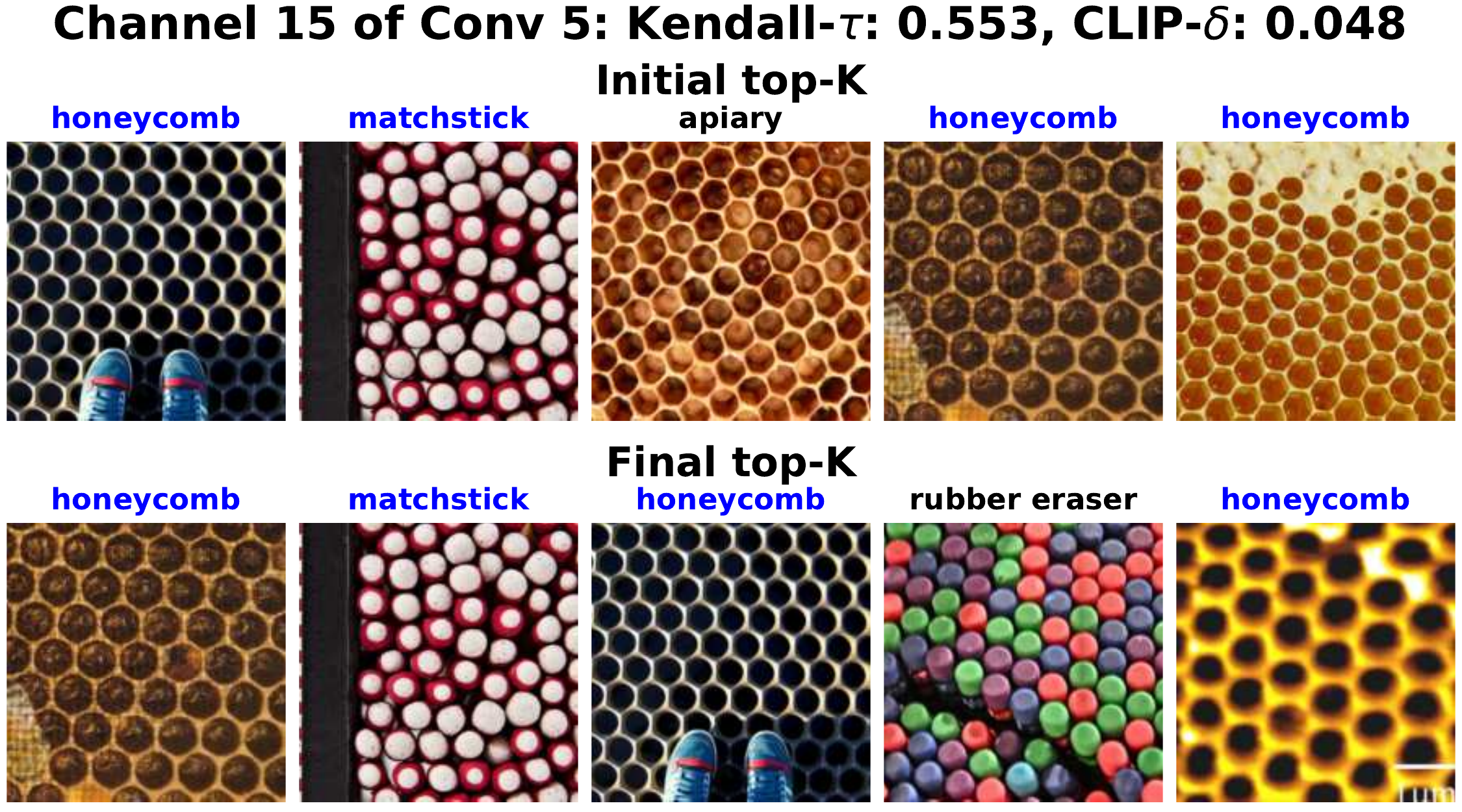}
\end{subfigure}\hfill
\begin{subfigure}[]{0.49\linewidth}
    \includegraphics[width=\textwidth]{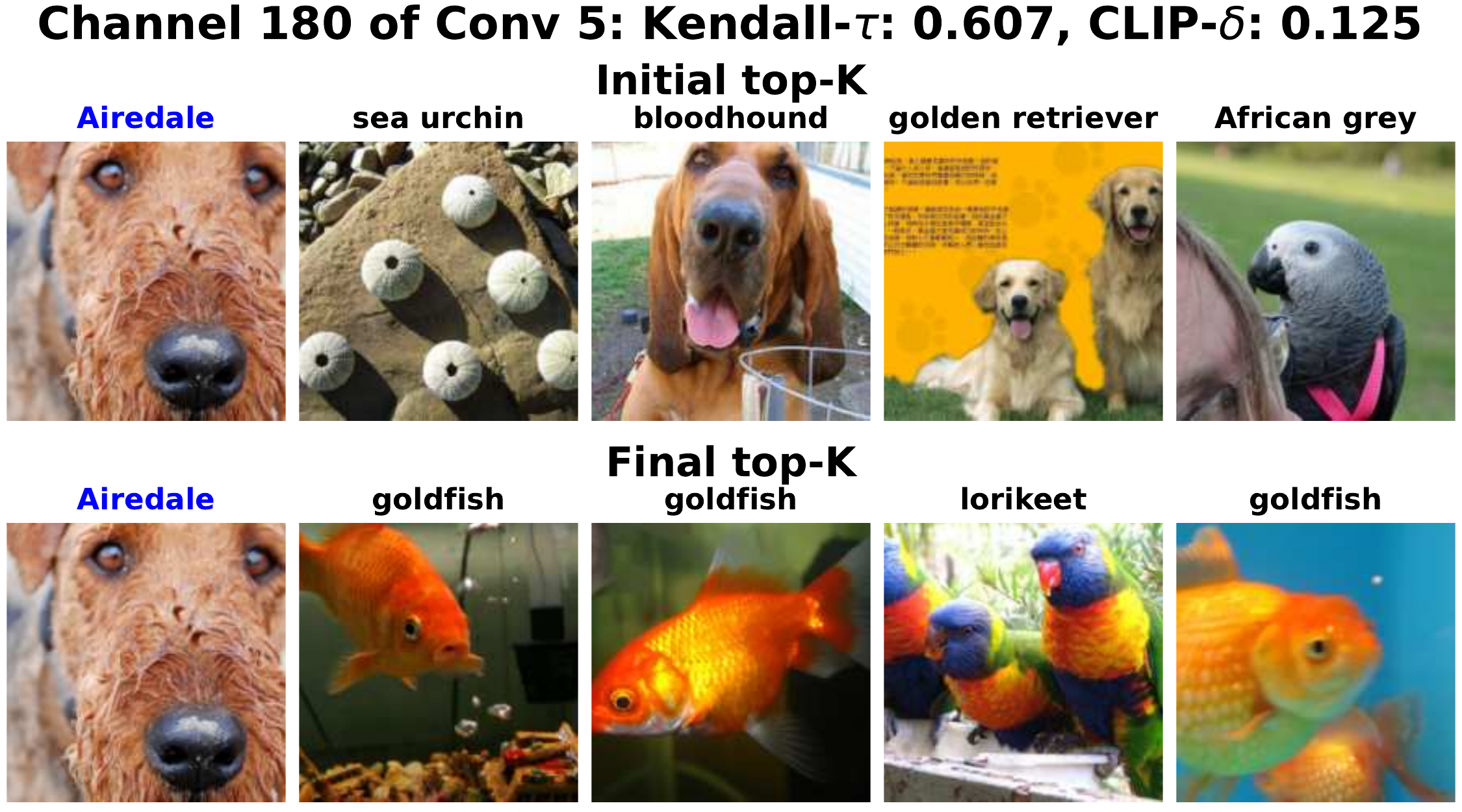}
\end{subfigure}\\
\vspace{.8cm}
\begin{subfigure}[]{0.49\linewidth}
    \includegraphics[width=\textwidth]{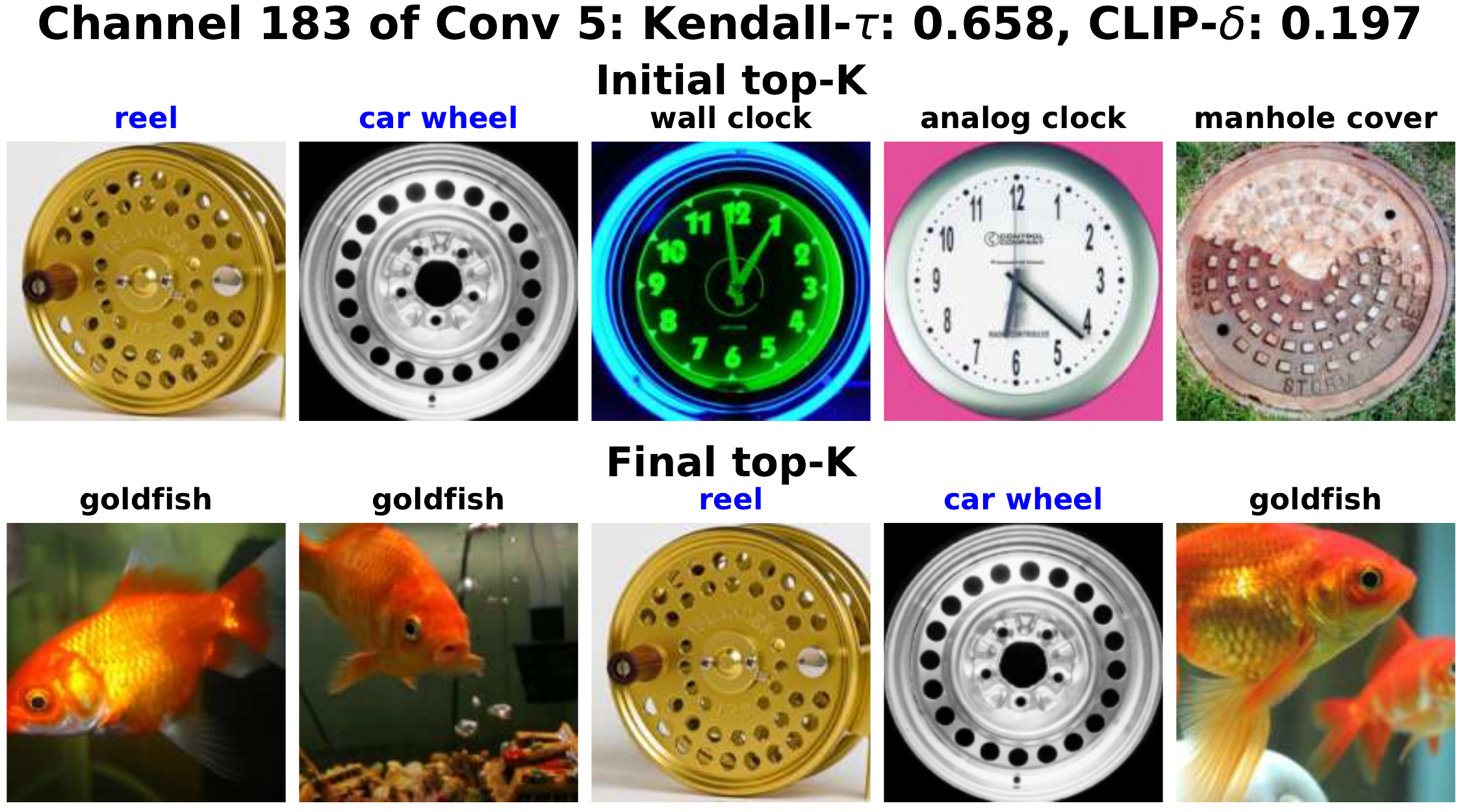}
\end{subfigure}\hfill
\begin{subfigure}[]{0.49\linewidth}
    \includegraphics[width=\textwidth]{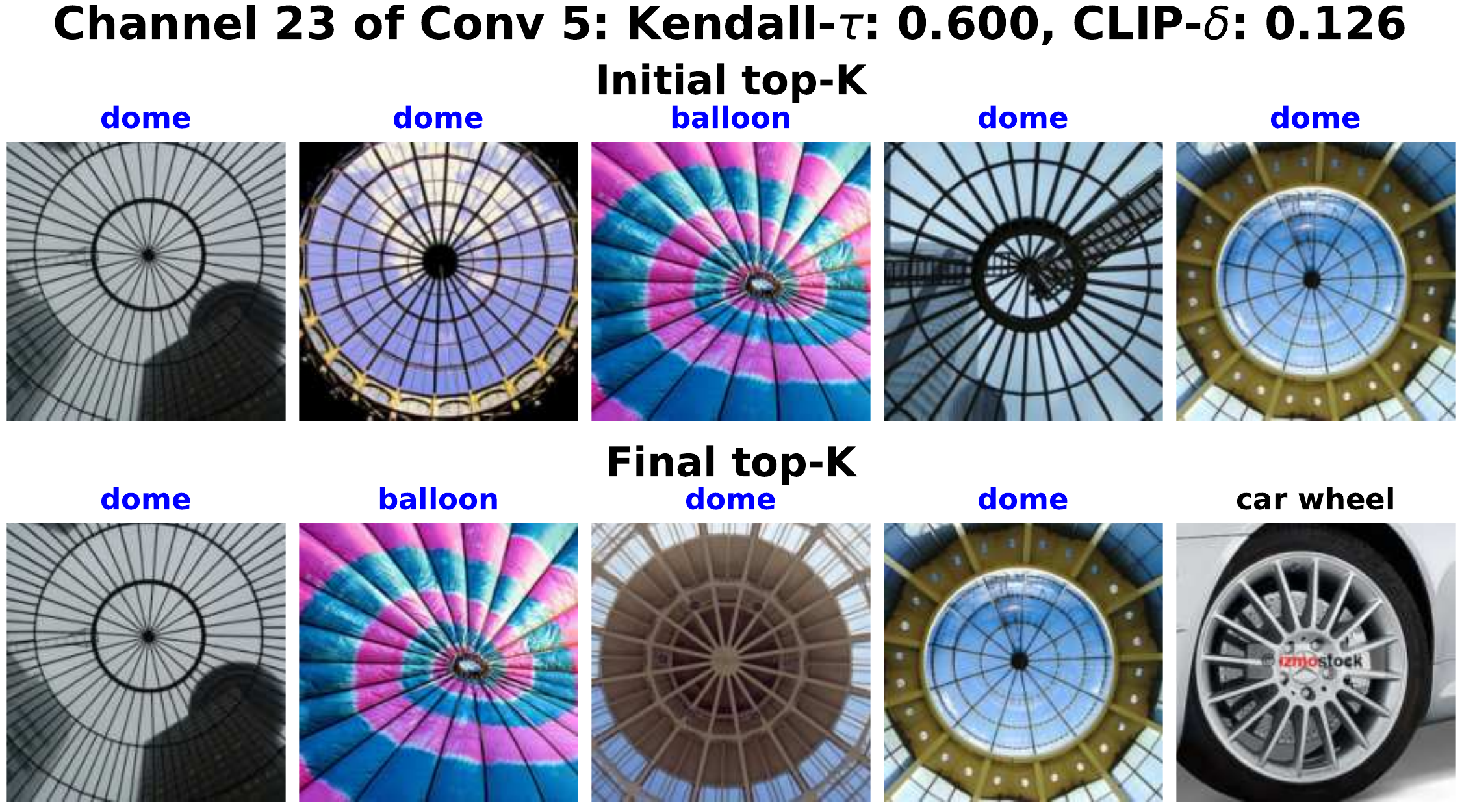}
\end{subfigure}\\
\vspace{.8cm}
\begin{subfigure}[]{0.49\linewidth}
    \includegraphics[width=\textwidth]{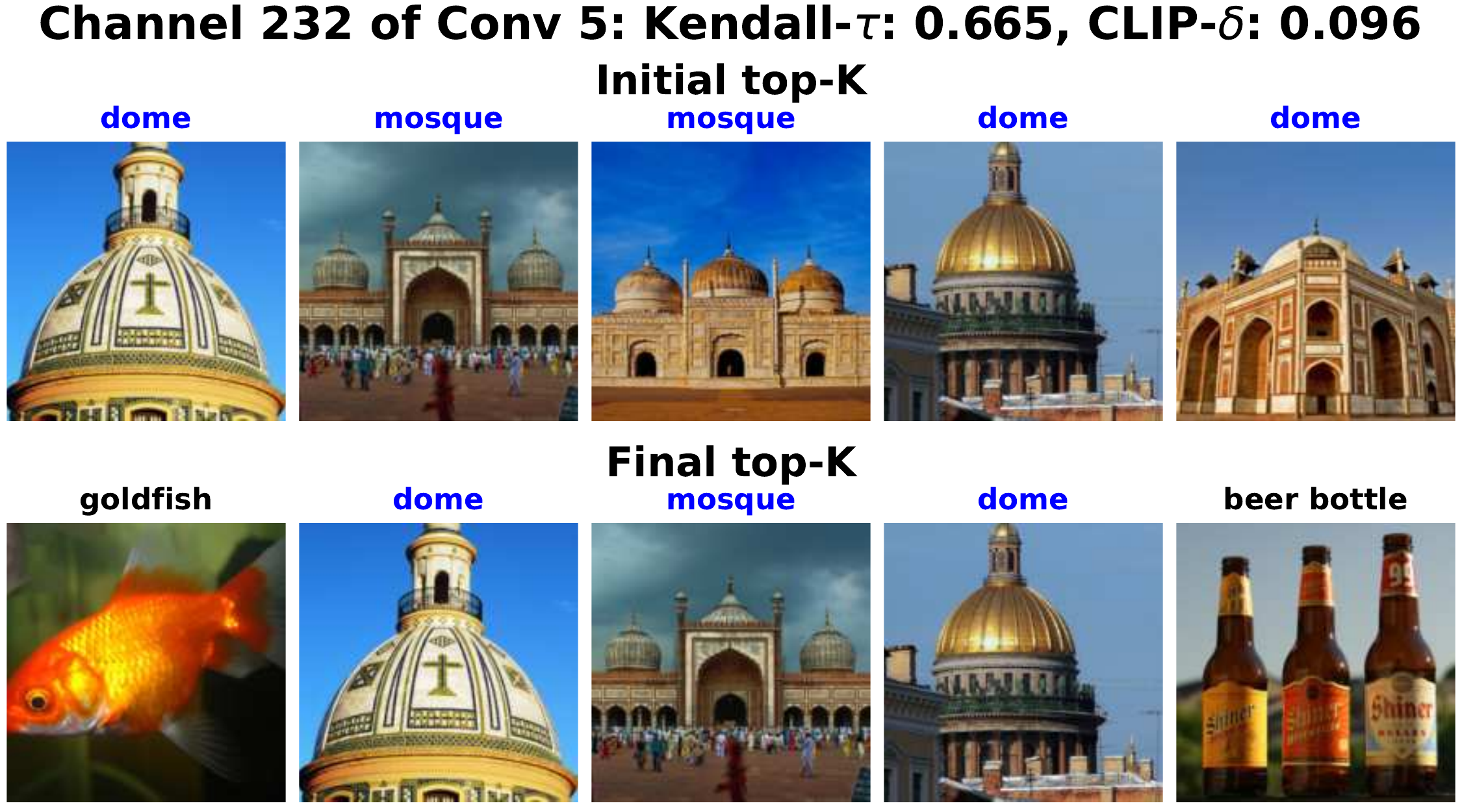}
\end{subfigure}\hfill
\begin{subfigure}[]{0.49\linewidth}
    \includegraphics[width=\textwidth]{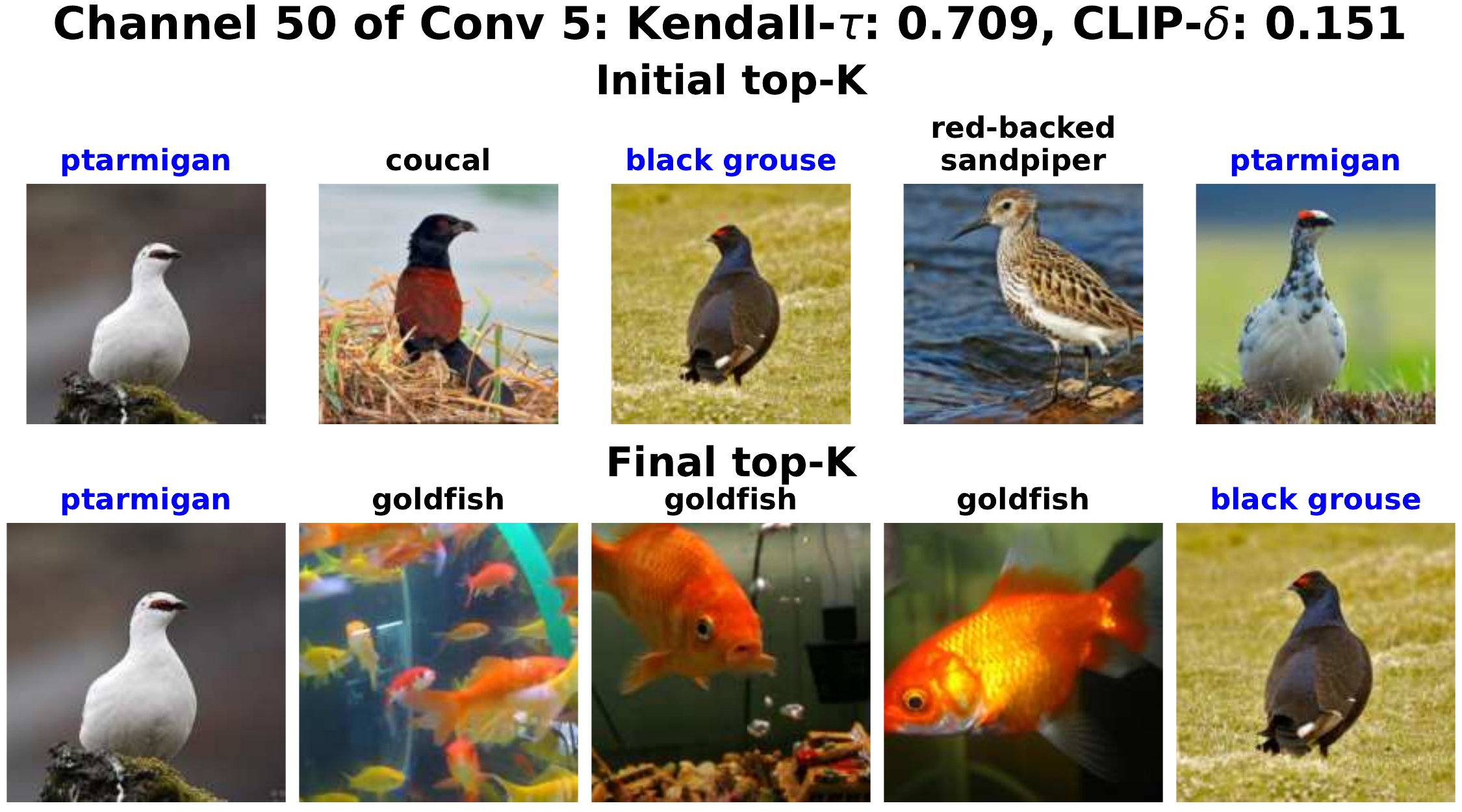}
\end{subfigure}
    \caption{\small Push-up all-channel attack of \textit{Conv5} of AlexNet. Channel indexes were taken randomly.} 
        \label{fig_add:all_channel_push_up}
\end{figure}

This section provides additional visual illustrations of the push-up all-channel attack on the layer conv5 of AlexNet.
\paragraph{Visual Examples.}
We first provide additional visual illustrations in Figure~\ref{fig_add:all_channel_push_up} of the attack on 10 randomly chosen channels. As a reminder, this push-up attack aims to make images of the Goldfish class appear in the top-$k$ images of every channel on the targeted layer. From Figure~\ref{fig_add:all_channel_push_up}, a first observation is the fact that out of these 10 randomly chosen channels, only two channels (channel 15 and channel 23) do not show an image with the Goldfish class. On the rest of the channels, an image with Goldfish was successfully inserted in the final top images. Furthermore, in several cases (channels 110, 125, 145, 180, 183, and 50) is the majority class of final top-5 images, demonstrating the success of this attack. It is also important to note the complete replacement of images with the Goldfish class in some channels (e.g., channel 125).
\clearpage

\begin{figure}[!ht]
\centering
\begin{subfigure}[]{0.49\linewidth}
    \includegraphics[width=\textwidth]{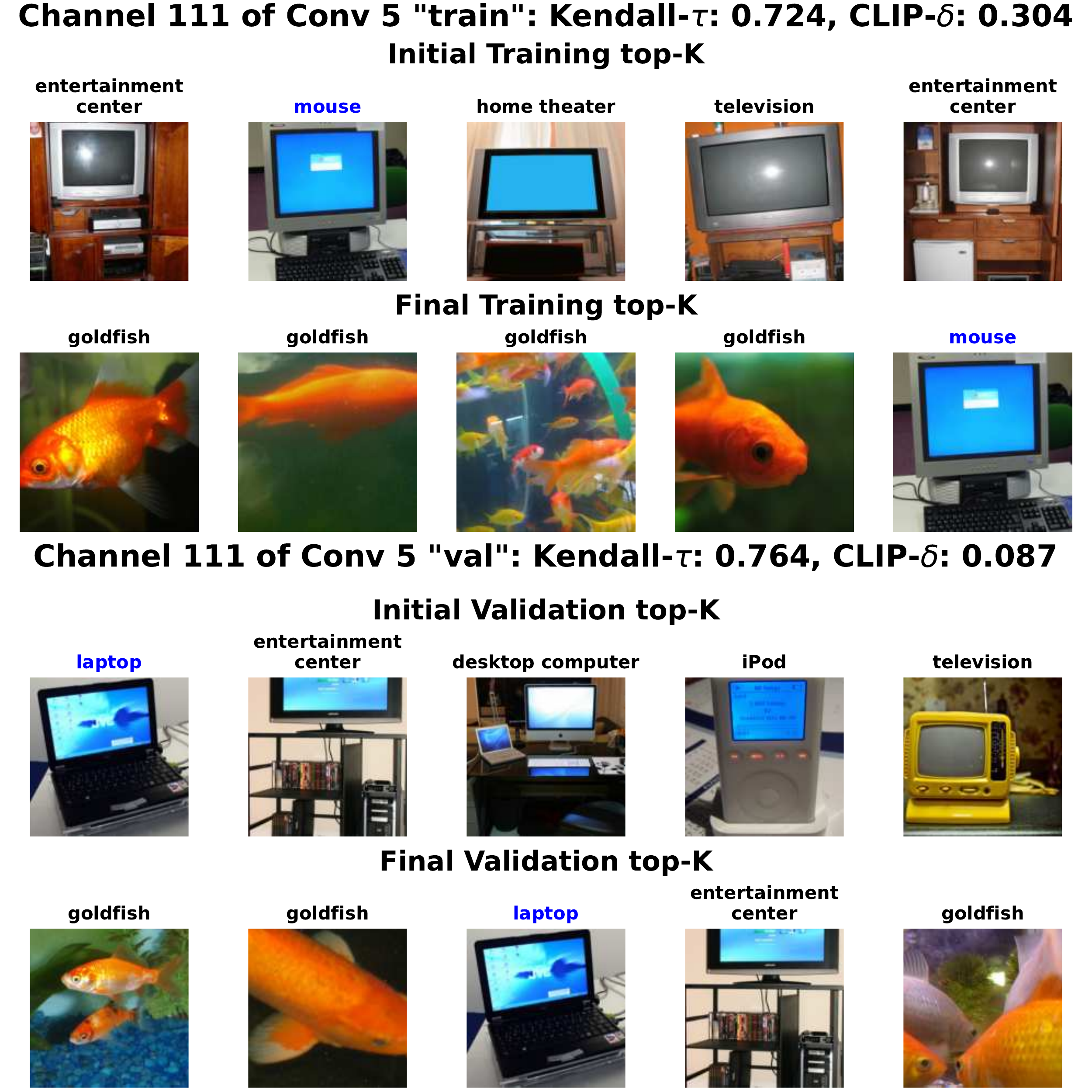}
\end{subfigure}\hfill
\begin{subfigure}[]{0.49\linewidth}
    \includegraphics[width=\textwidth]{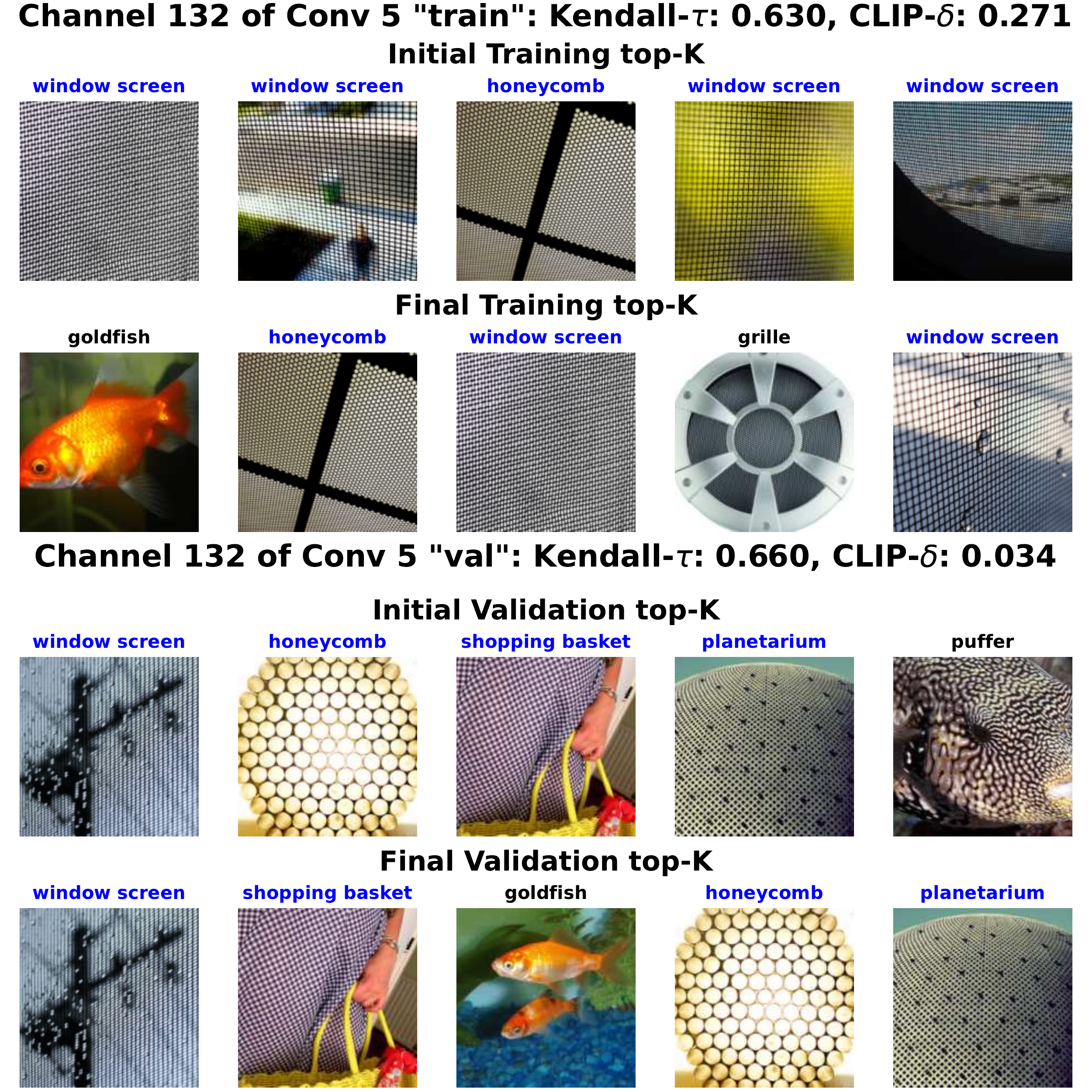}
\end{subfigure}\\
\vspace{.8cm}
\begin{subfigure}[]{0.49\linewidth}
    \includegraphics[width=\textwidth]{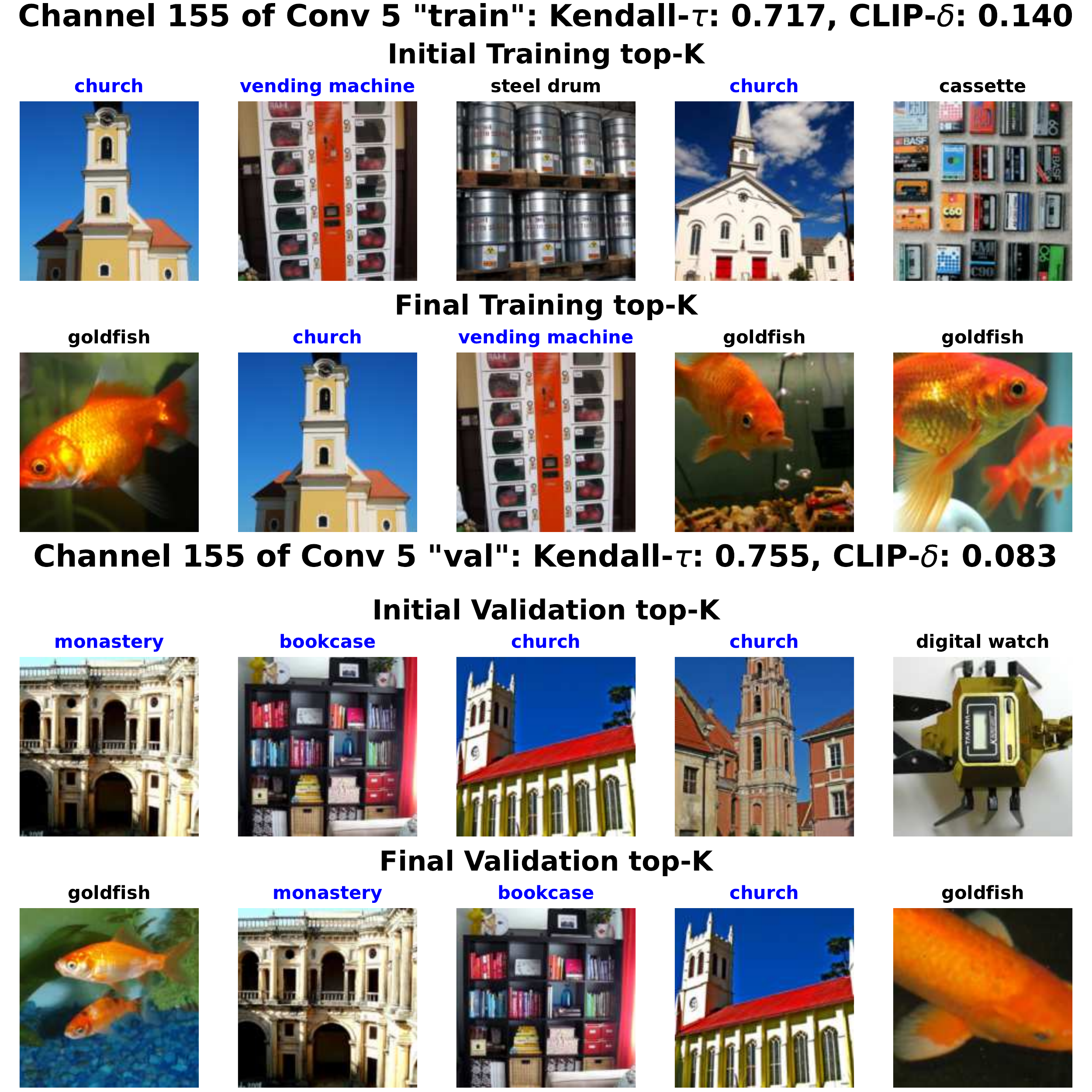}
\end{subfigure}\hfill
\begin{subfigure}[]{0.49\linewidth}
    \includegraphics[width=\textwidth]{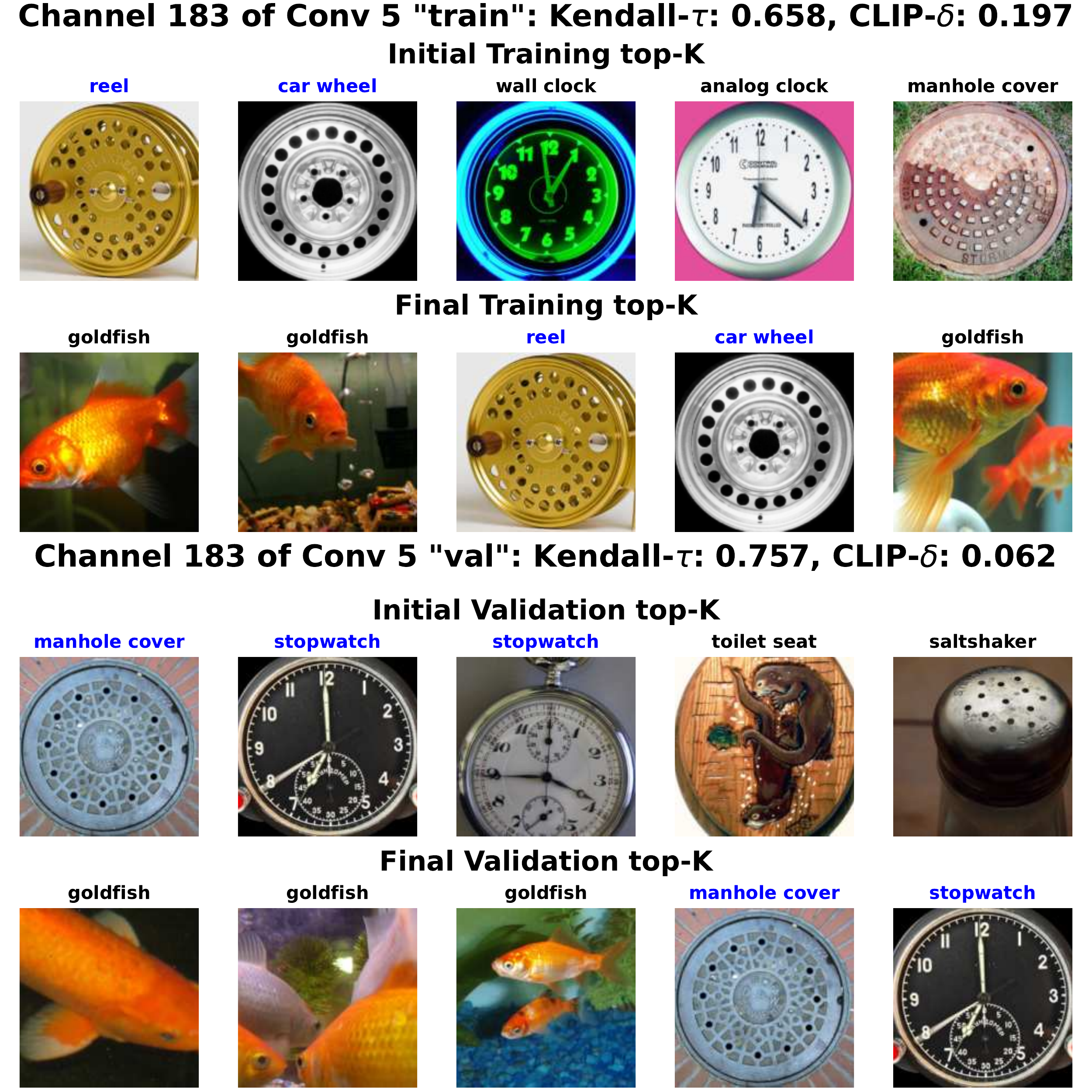}
\end{subfigure}\\
\vspace{.8cm}
\begin{subfigure}[]{0.49\linewidth}
    \includegraphics[width=\textwidth]{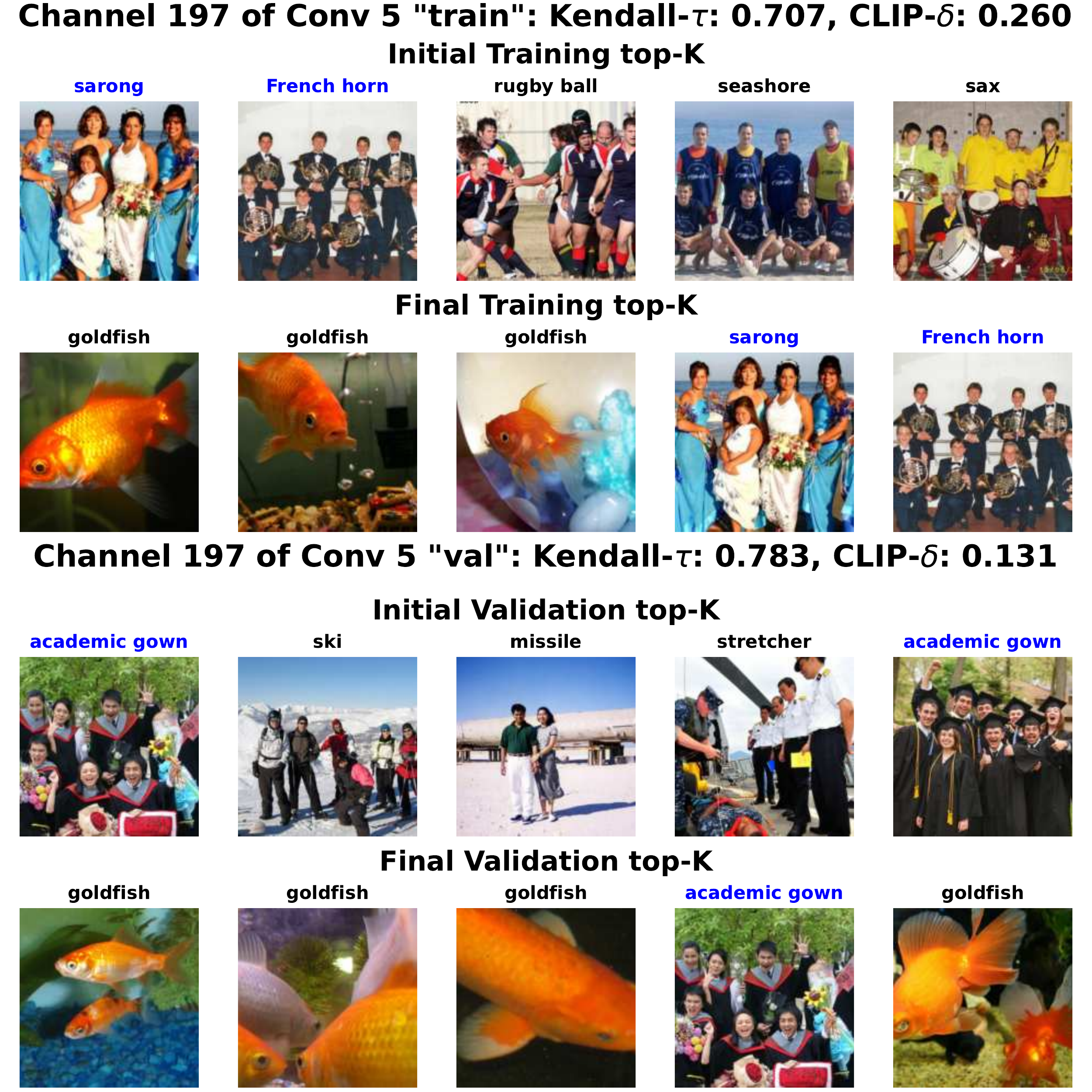}
\end{subfigure}\hfill
    \caption{\small Push-up all-channel attack of Conv5 of AlexNet. For each channel, the first two rows are top-$k$ images derived from the training set while the last two are derived from the validation set.} 
        \label{fig_add:all_channel_validation_pushup}
\end{figure}

\paragraph{Generalization for the Push-Up attack.}
After demonstrating the success of achieving target manipulability of top-$k$ feature visualization through the push-up attack on training images, it is also important to evaluate whether this success generalizes to unseen data. Figure~\ref{fig_add:all_channel_validation_pushup} shows not only top-$k$ images from the training but also from the validation set of ImageNet. We can observe that on all the 10 randomly chosen channels not only at least one image of the Goldfish class is present in the final top-5 images of the training but also at least one image of the Goldfish class is in the final top-5 images from the validation set. Moreover, we also observe a similar number of images of the Goldfish class present in top-5 images from both training and validation sets. This indicates the ability of the push-up attack to generalize on the same distribution from where training examples were drawn.

\begin{figure}
\centering
\begin{subfigure}[]{0.49\linewidth}
    \includegraphics[width=\textwidth]{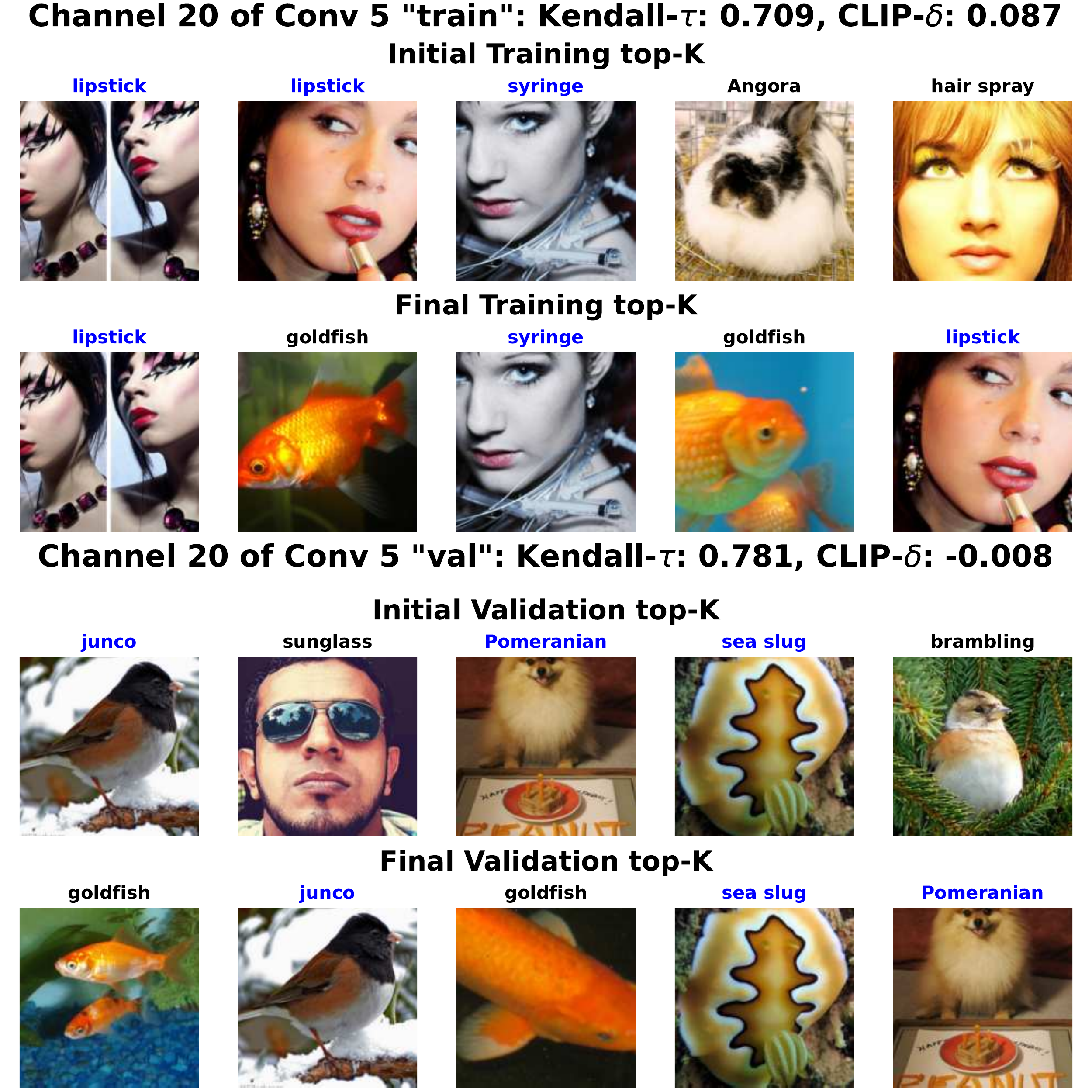}
\end{subfigure}\hfill
\begin{subfigure}[]{0.49\linewidth}
    \includegraphics[width=\textwidth]{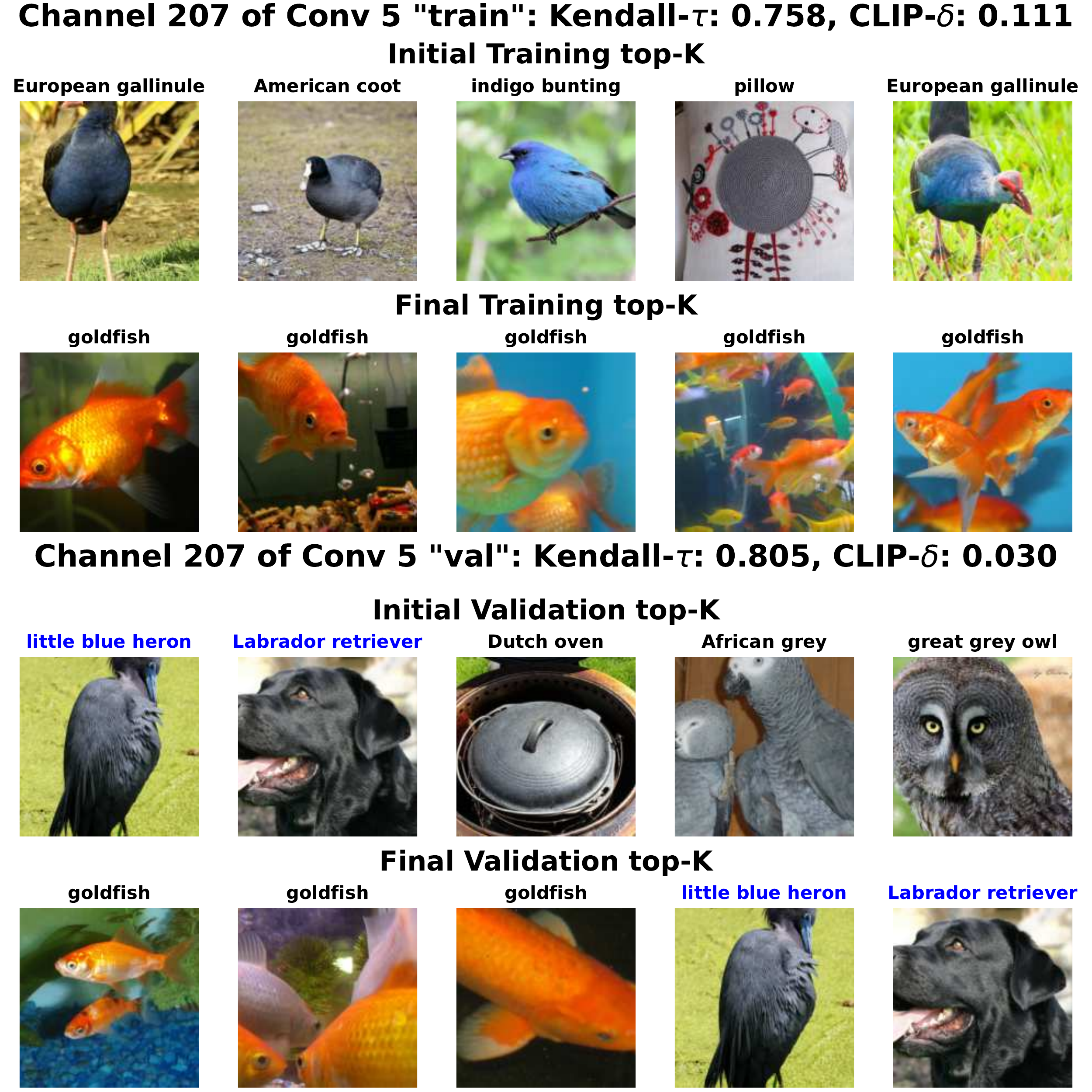}
\end{subfigure}\\
\vspace{.8cm}
\begin{subfigure}[]{0.49\linewidth}
    \includegraphics[width=\textwidth]{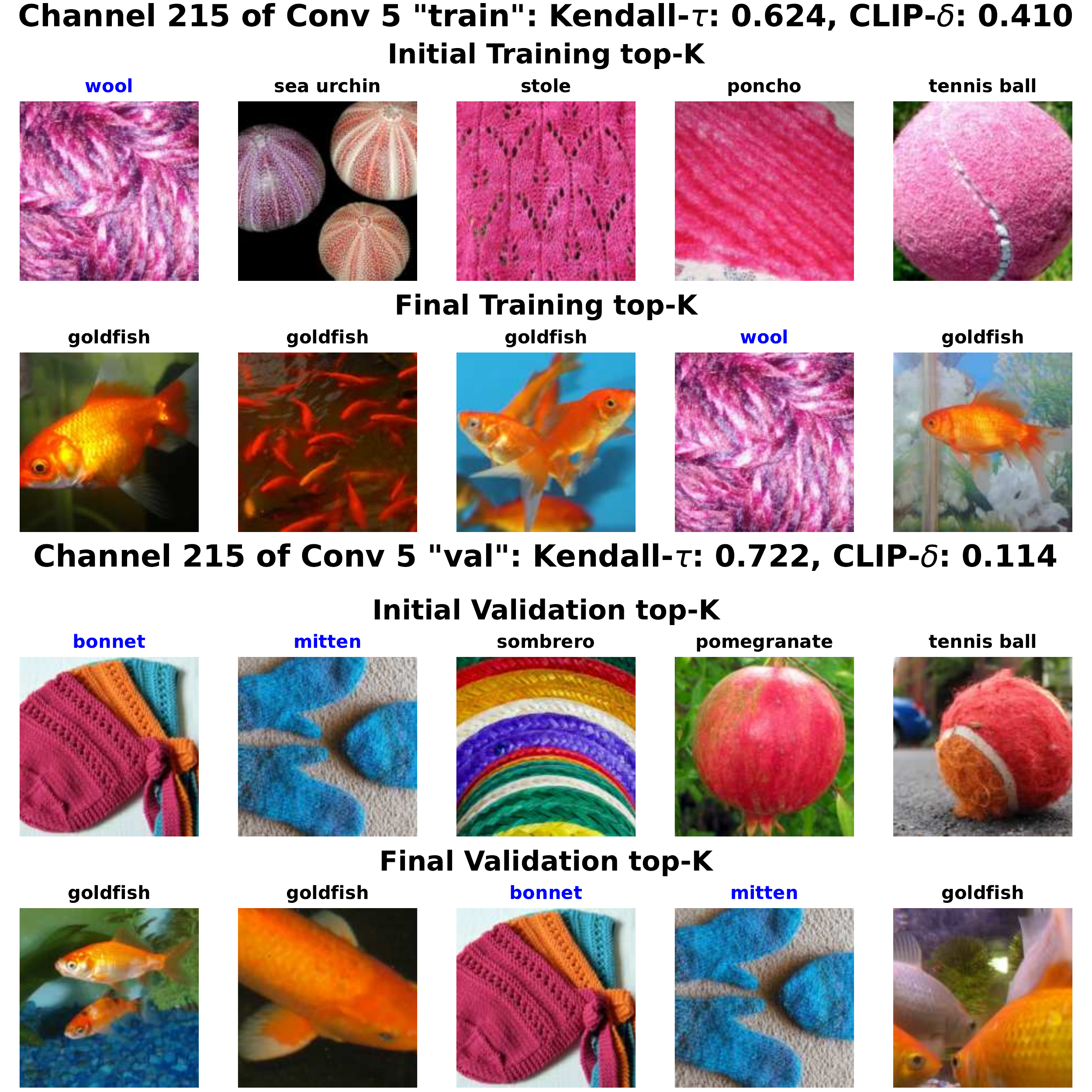}
\end{subfigure}\hfill
\begin{subfigure}[]{0.49\linewidth}
    \includegraphics[width=\textwidth]{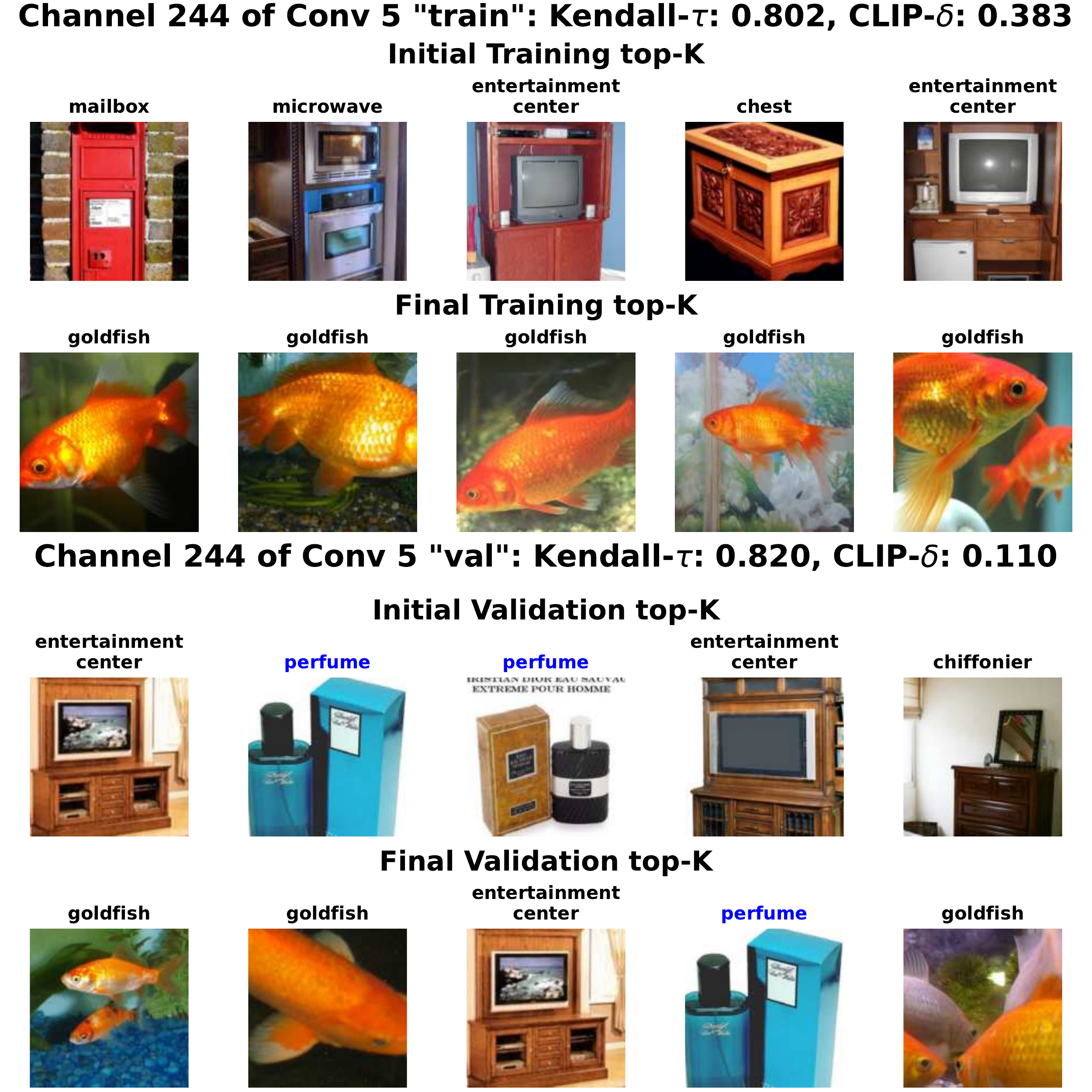}
\end{subfigure}\\
\vspace{.8cm}
\begin{subfigure}[]{0.49\linewidth}
    \includegraphics[width=\textwidth]{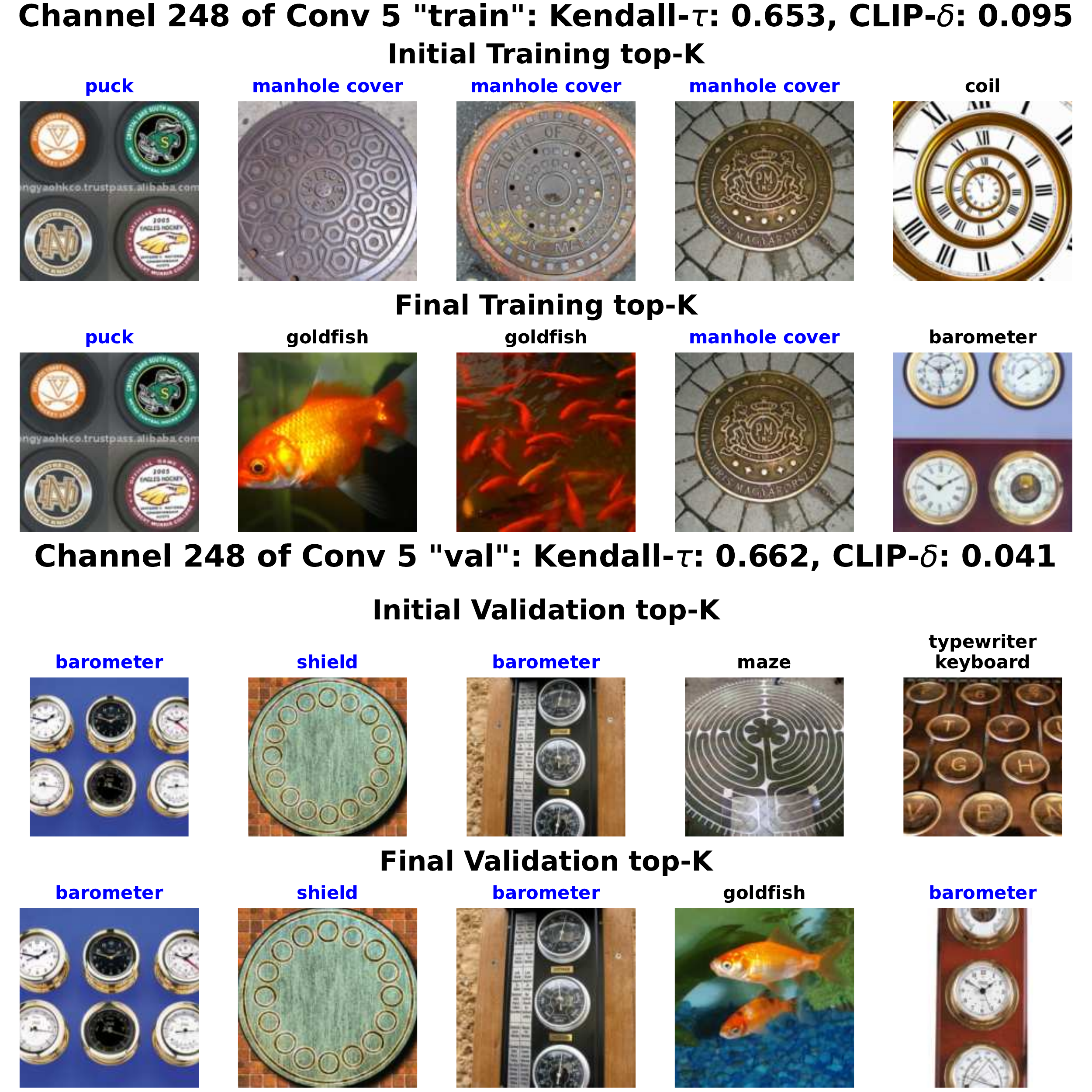}
\end{subfigure}
\caption{\small Push-up all-channel attack of Conv5 of AlexNet. For each channel, the first two rows are top-$k$ images derived from the training set while the last two are derived from the validation set.} 
        \label{fig_add:all_channel_validation_next_pushup}
\end{figure}
\clearpage

\subsection{Additional Illustrations for Synthetic Feature Visualization}
This section provides additional illustrations of the decorrelation between synthetic and natural (through top-$k$ images) feature visualization.  

\begin{figure}[!ht]
\vspace{-.5cm}
\centering
\begin{subfigure}[]{0.49\linewidth}
    \includegraphics[width=\textwidth]{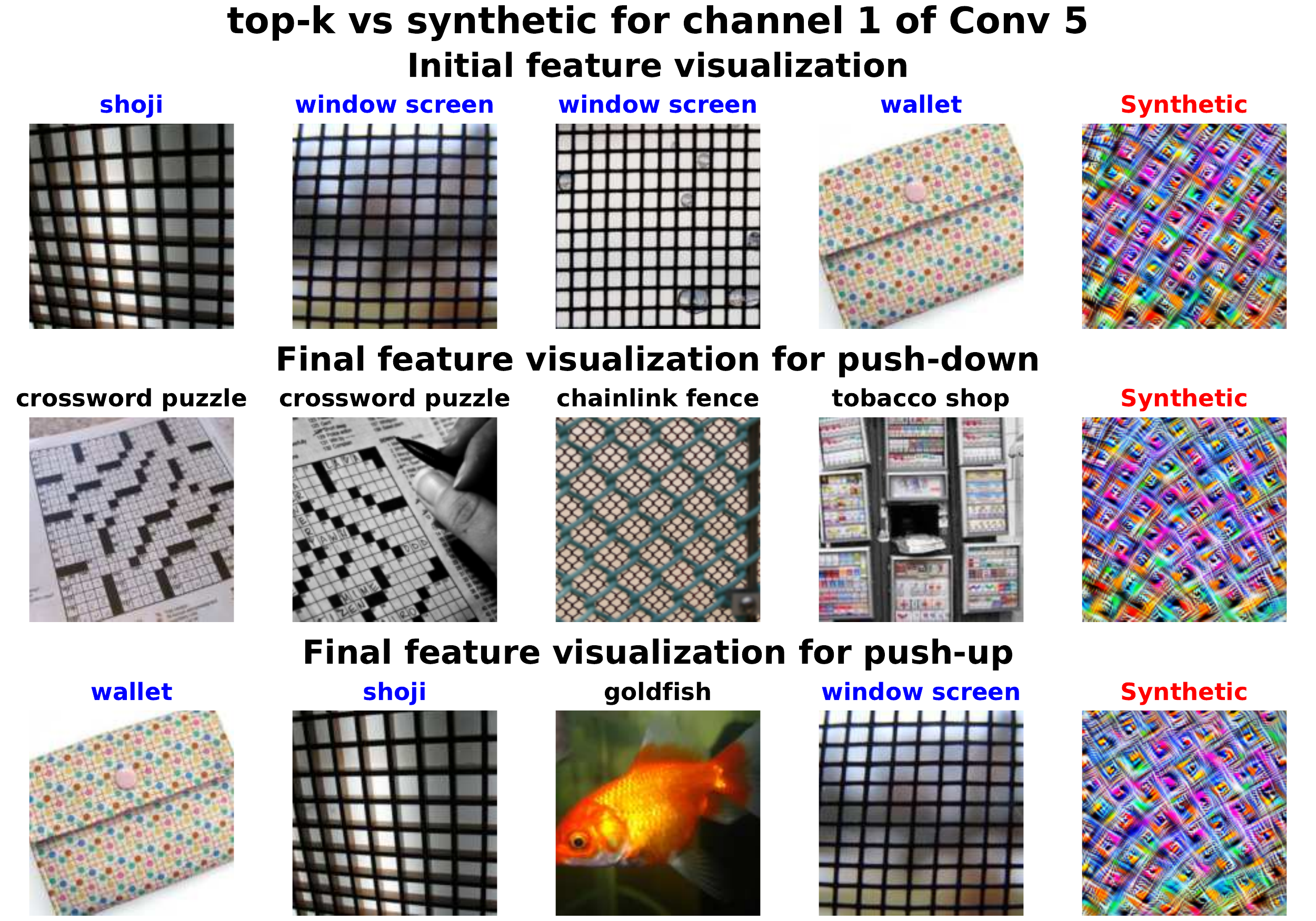}
\end{subfigure}\hfill
\begin{subfigure}[]{0.49\linewidth}
    \includegraphics[width=\textwidth]{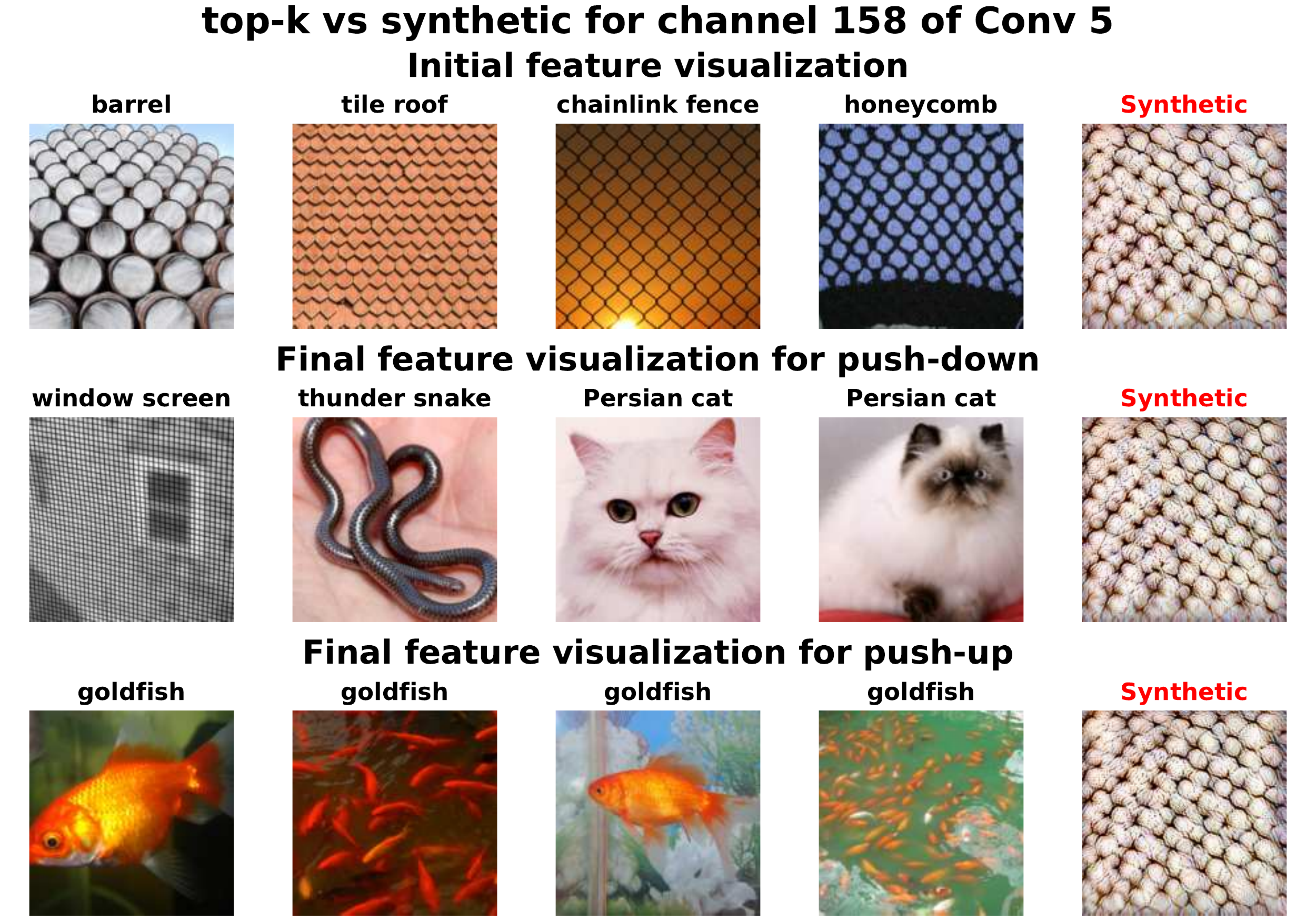}
\end{subfigure}\\
\vspace{.8cm}
\begin{subfigure}[]{0.49\linewidth}
    \includegraphics[width=\textwidth]{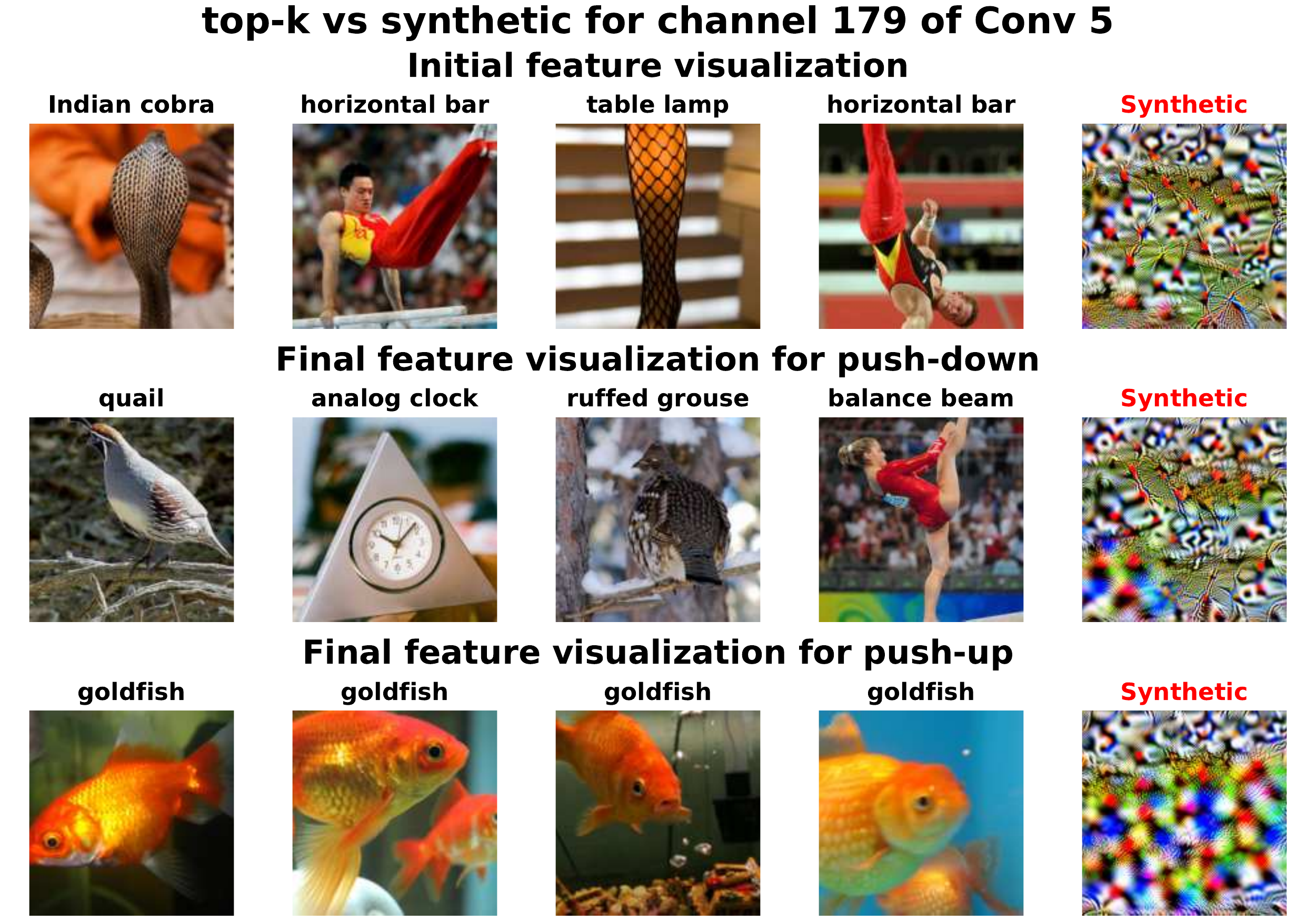}
\end{subfigure}\hfill
\begin{subfigure}[]{0.49\linewidth}
    \includegraphics[width=\textwidth]{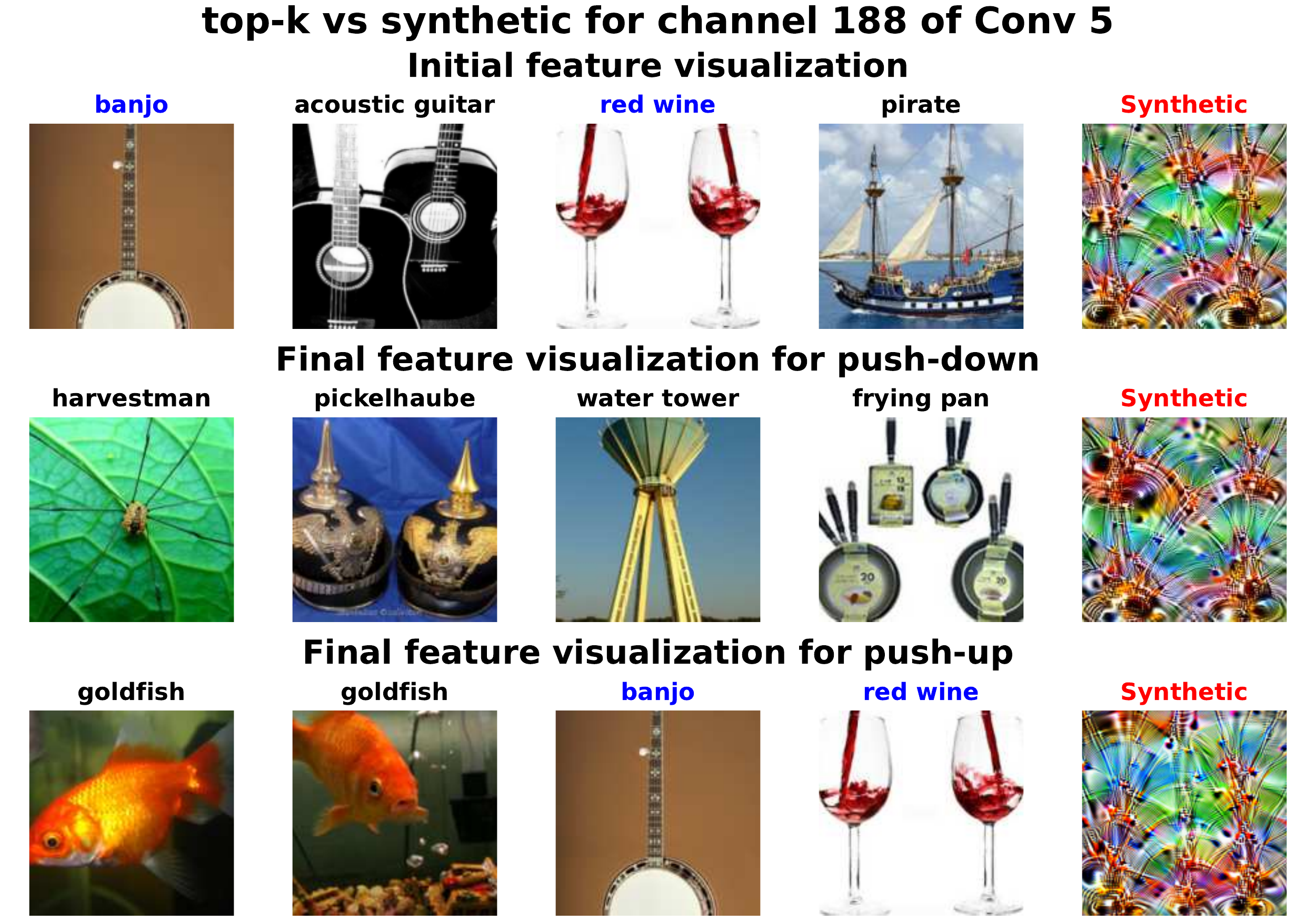}
\end{subfigure}\\
\vspace{.8cm}
\begin{subfigure}[]{0.49\linewidth}
    \includegraphics[width=\textwidth]{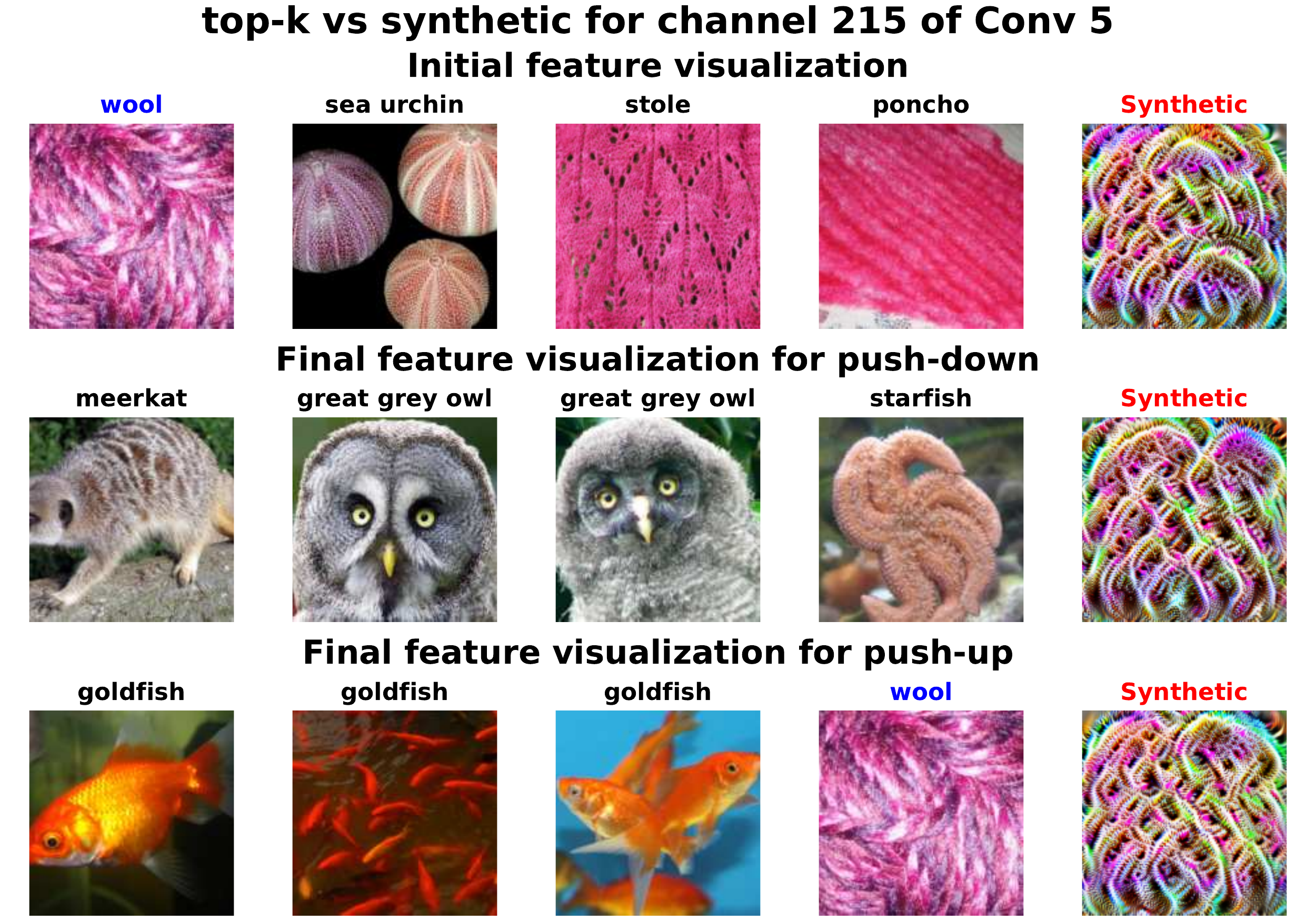}
\end{subfigure}\hfill
\begin{subfigure}[]{0.49\linewidth}
    \includegraphics[width=\textwidth]{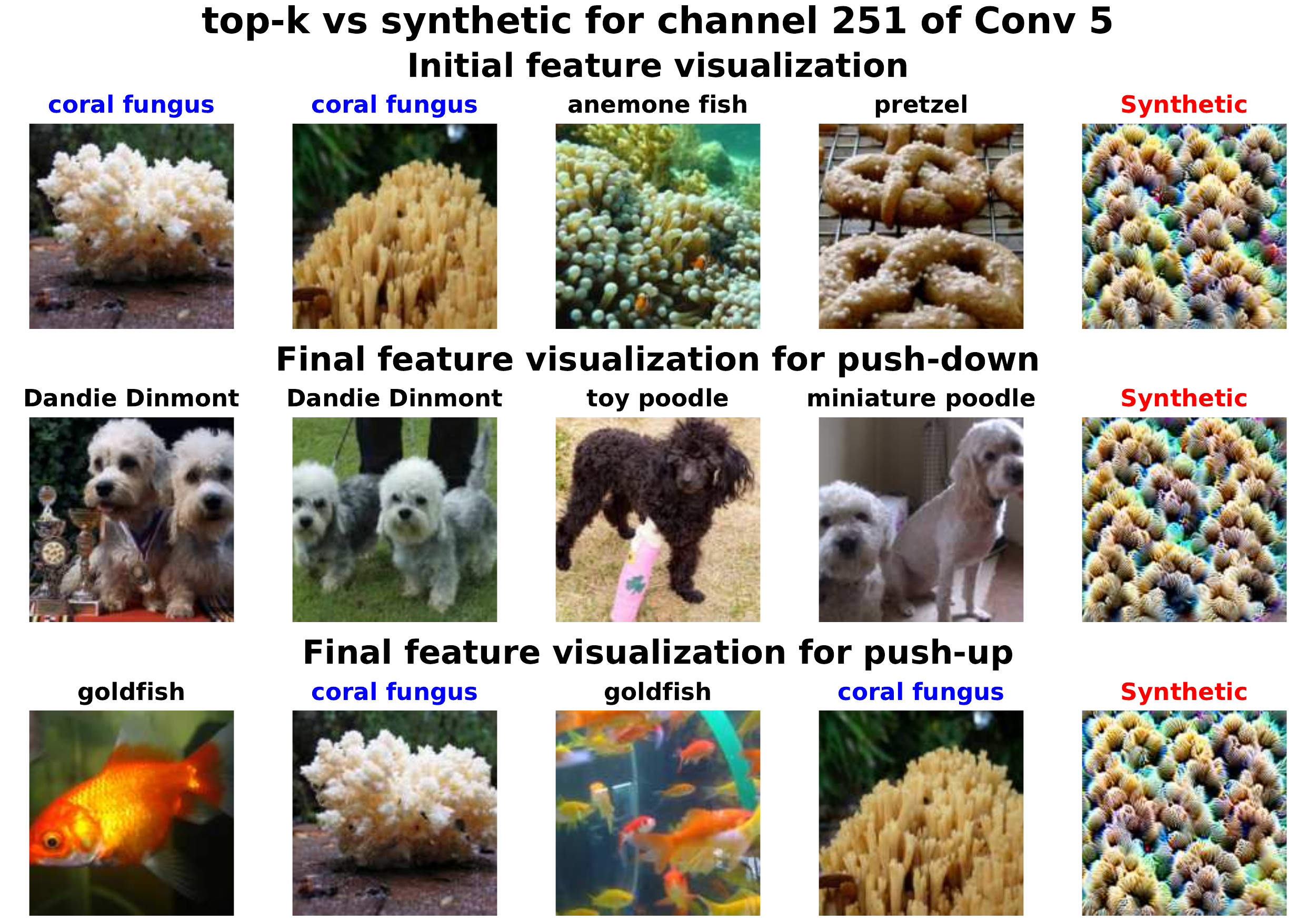}
\end{subfigure}\\
\vspace{.8cm}
\begin{subfigure}[]{0.49\linewidth}
    \includegraphics[width=\textwidth]{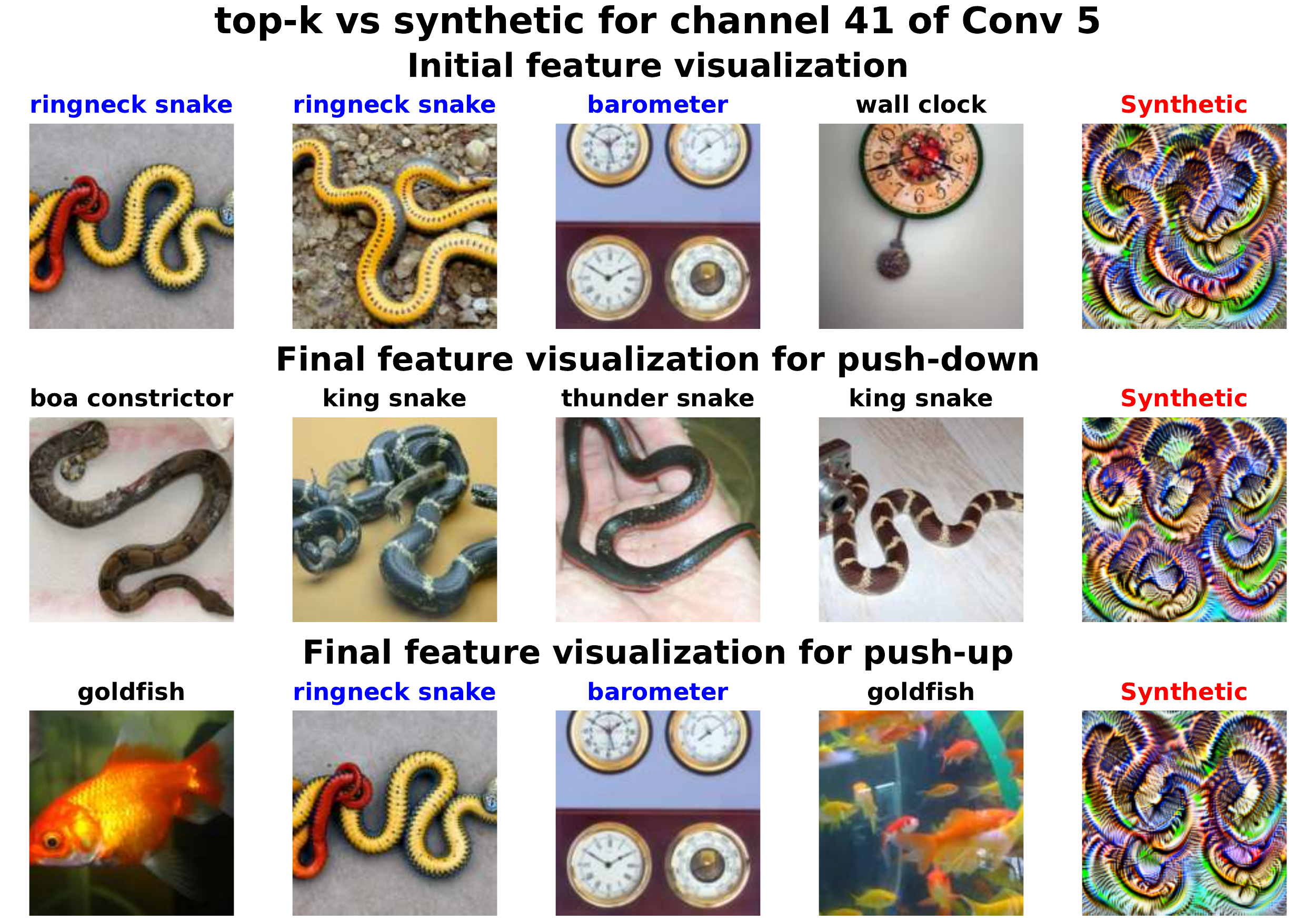}
\end{subfigure}\hfill
\begin{subfigure}[]{0.49\linewidth}
    \includegraphics[width=\textwidth]{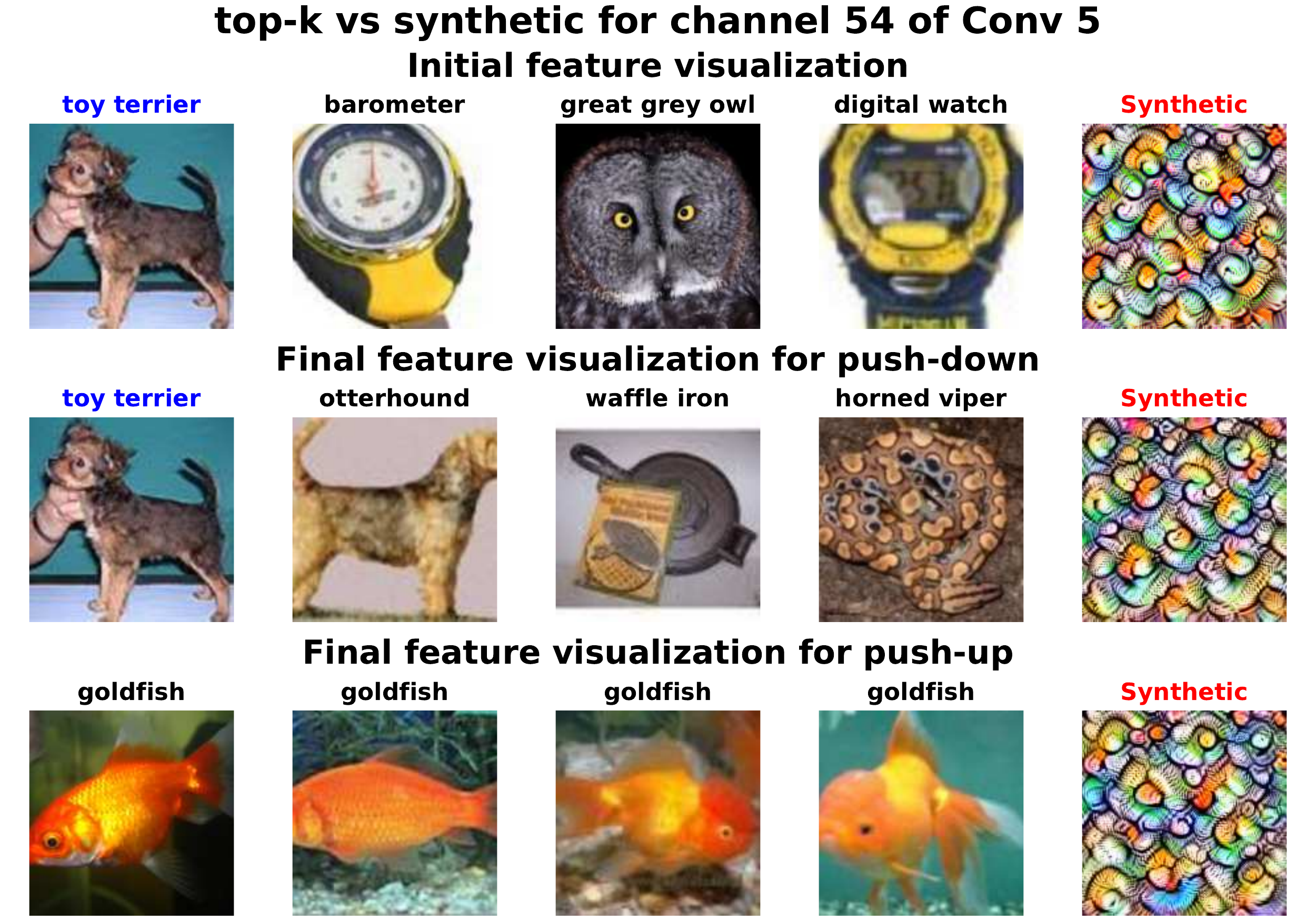}
\end{subfigure}
\caption{\small Synthetic Feature Visualization attack after push-down and push-up attacks on Conv5 of AlexNet. Channels indexes were taken randomly. We observe a decorrelation between natural top-activating images and synthetic optimal images.} 
        \label{fig_add:all_channel_synthetic}

\end{figure}

Figure~\ref{fig_add:all_channel_synthetic} shows the natural and synthetic feature visualization before and after the attack on 4 randomly chosen channels of conv5 of AlexNet. As stated in Section~\ref{sec:synthetic_feature_visualization}, from this figure,  we observe a lack of change in the synthetic optimal image (even when top images have been completely replaced by images of the Goldfish class, e.g., in channel 54). We, therefore, reemphasize that attacking the natural feature visualization does not transpose to attacking the synthetic feature visualization. This indicates a decorrelation between the synthetic feature visualization and the top-k images.

\clearpage


\subsection{Additional Fairwashing Results.}
\begin{figure}[!hb]
\centering
\begin{subfigure}[]{0.49\linewidth}
    \includegraphics[width=\textwidth]{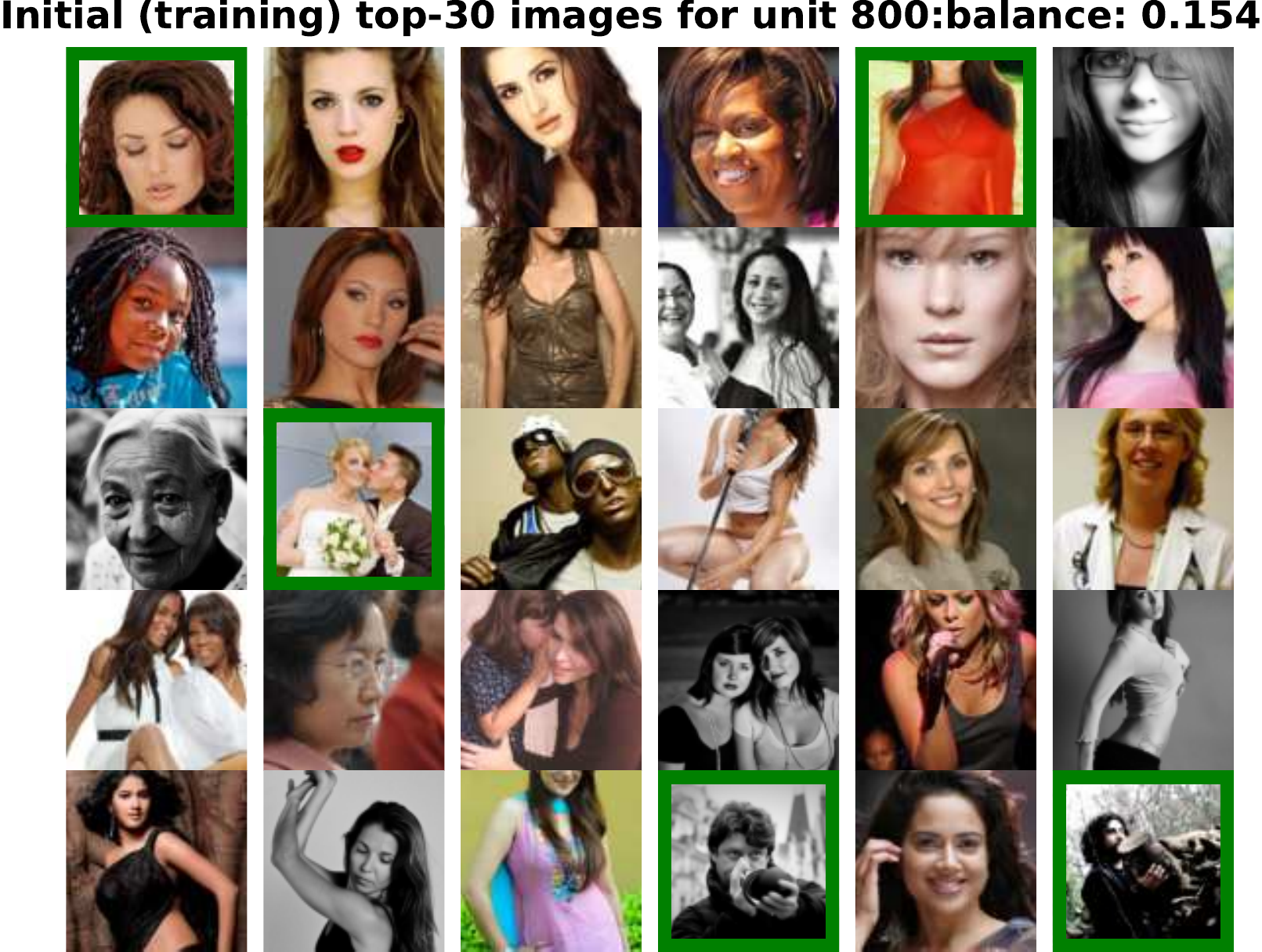}
\end{subfigure}\hfill
\begin{subfigure}[]{0.49\linewidth}
    \includegraphics[width=\textwidth]{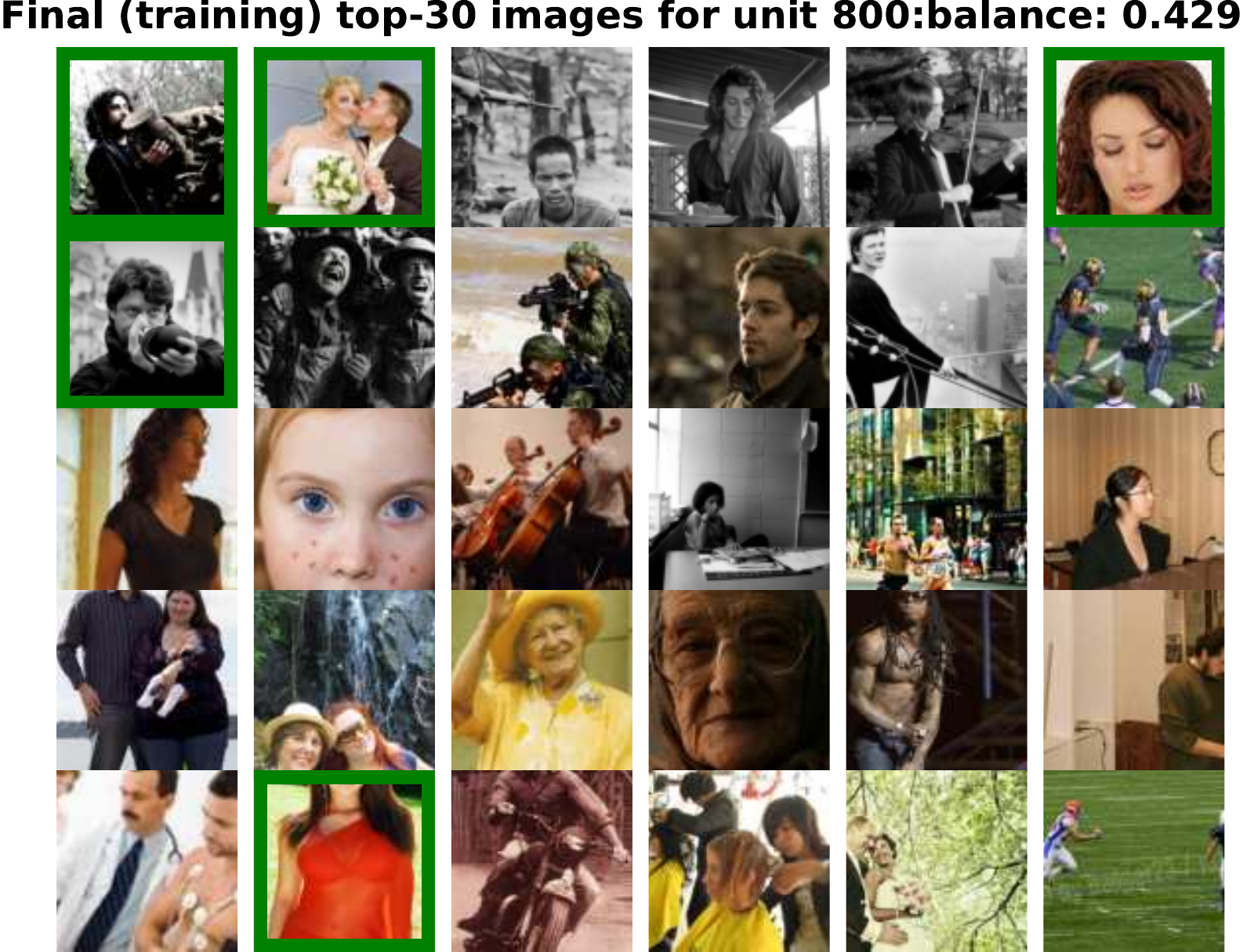}
\end{subfigure}\\
\vspace{.3cm}
\begin{subfigure}[]{0.49\linewidth}
    \includegraphics[width=\textwidth]{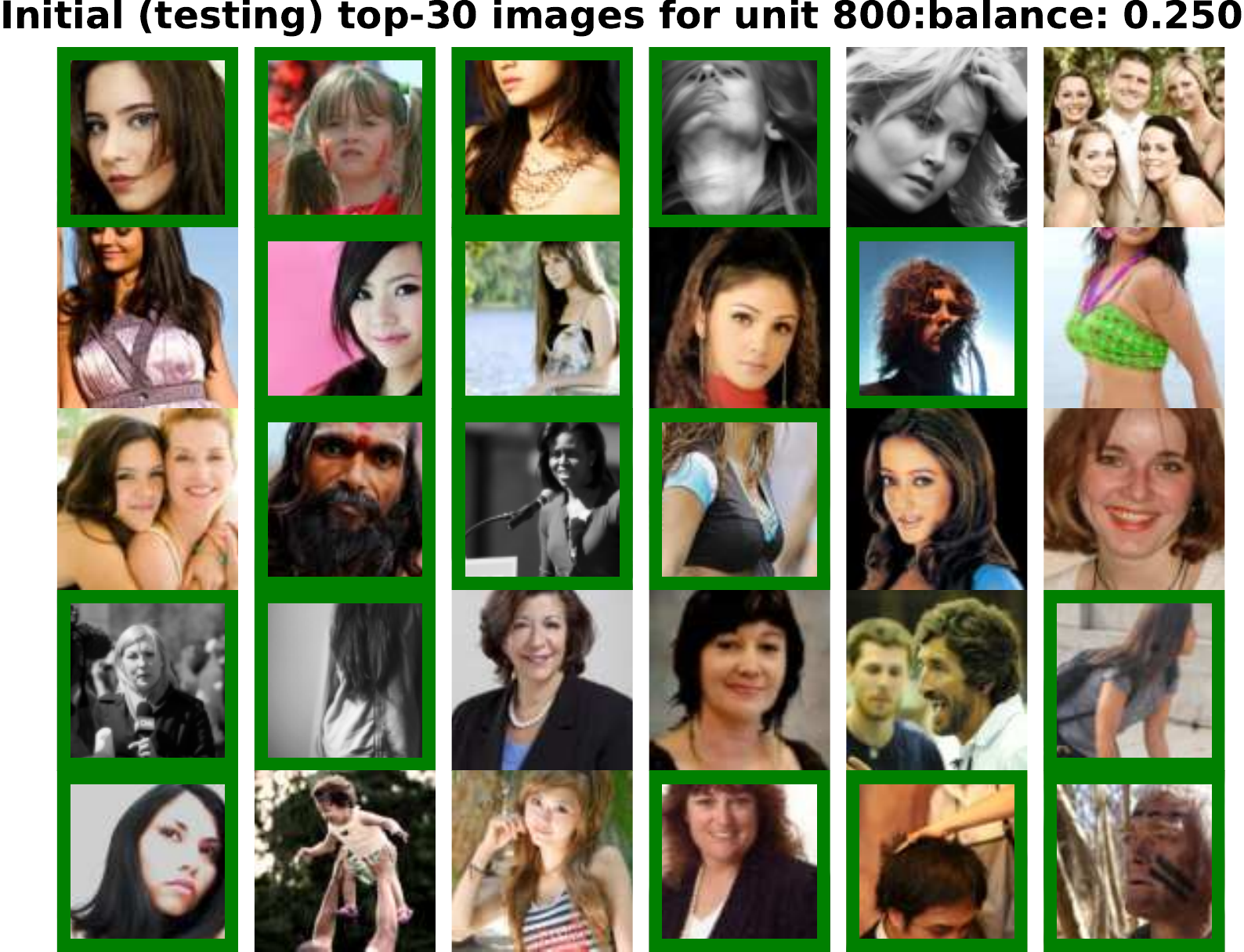}
\end{subfigure}\hfill
\begin{subfigure}[]{0.49\linewidth}
    \includegraphics[width=\textwidth]{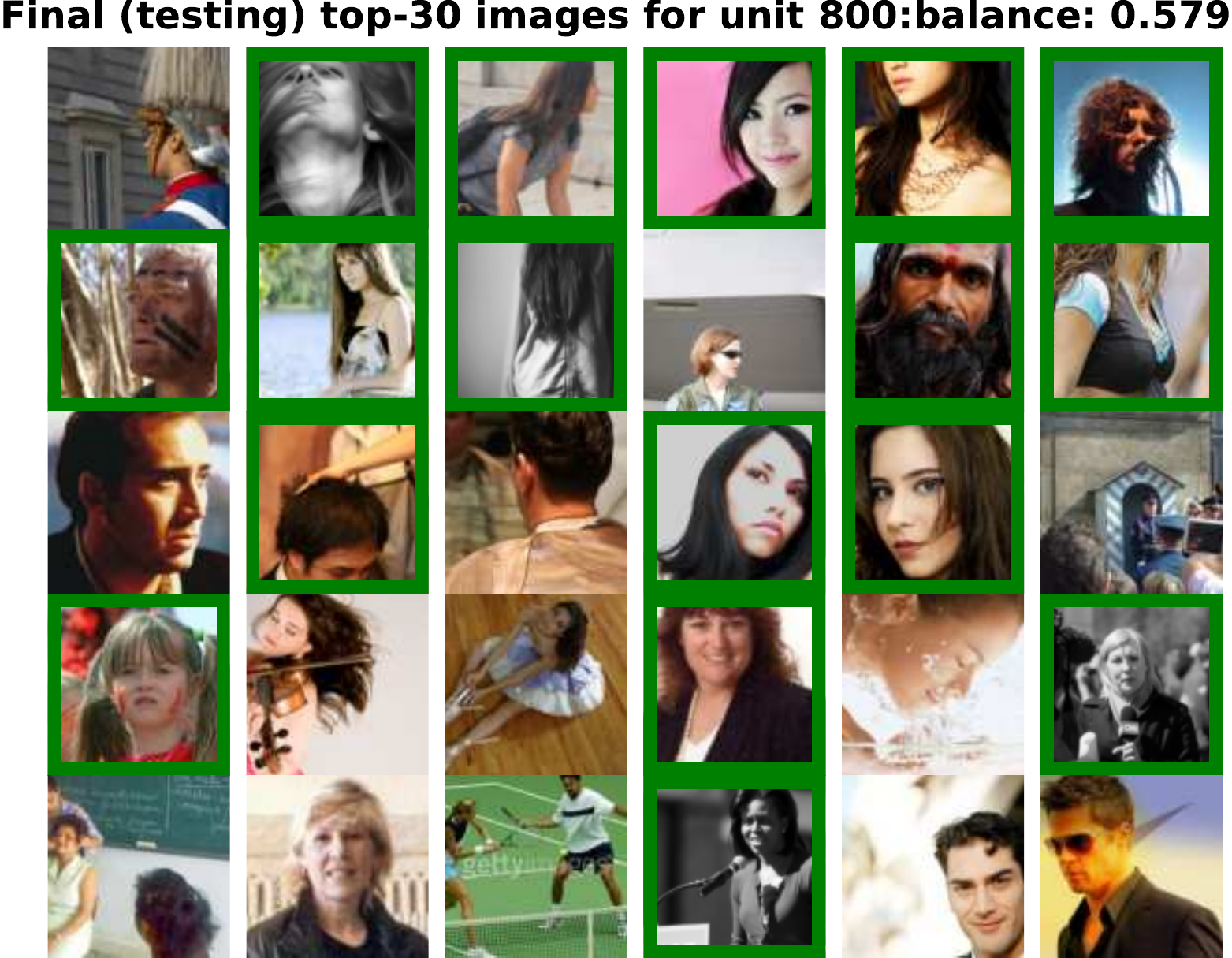}
\end{subfigure}
\caption{\small Results on channel 800 for training and testing annotated data. Note that the initial top images are shown here only for illustration as these images are used in the fairwashing loss. Only annotated testing data simulate the visualization seen by the interpreter or the regulator.} 
        \label{fig_add:fairwashing_attack_channel_800}
\end{figure}

This section presents the results obtained after the fairwashing attack on the last but one layer of AlexNet. 
\paragraph{Example of the Paper.}
We begin by showing in Figure~\ref{fig_add:fairwashing_attack_channel_800}, the top-$30$ images before and after the attack from both training and testing annotated data.
As a reminder, we assume that the interpreter has access to (testing) non-annotated data with a protected attribute (here the gender) and the attacker uses annotated training data to fairwash (making the top-$k$ look fairer) feature visualization.

Training annotated data (first row of Figure~\ref{fig_add:fairwashing_attack_channel_800}) is shown only for illustration.

\begin{figure}[!ht]
\centering
\begin{subfigure}[]{0.49\linewidth}
    \includegraphics[width=\textwidth]{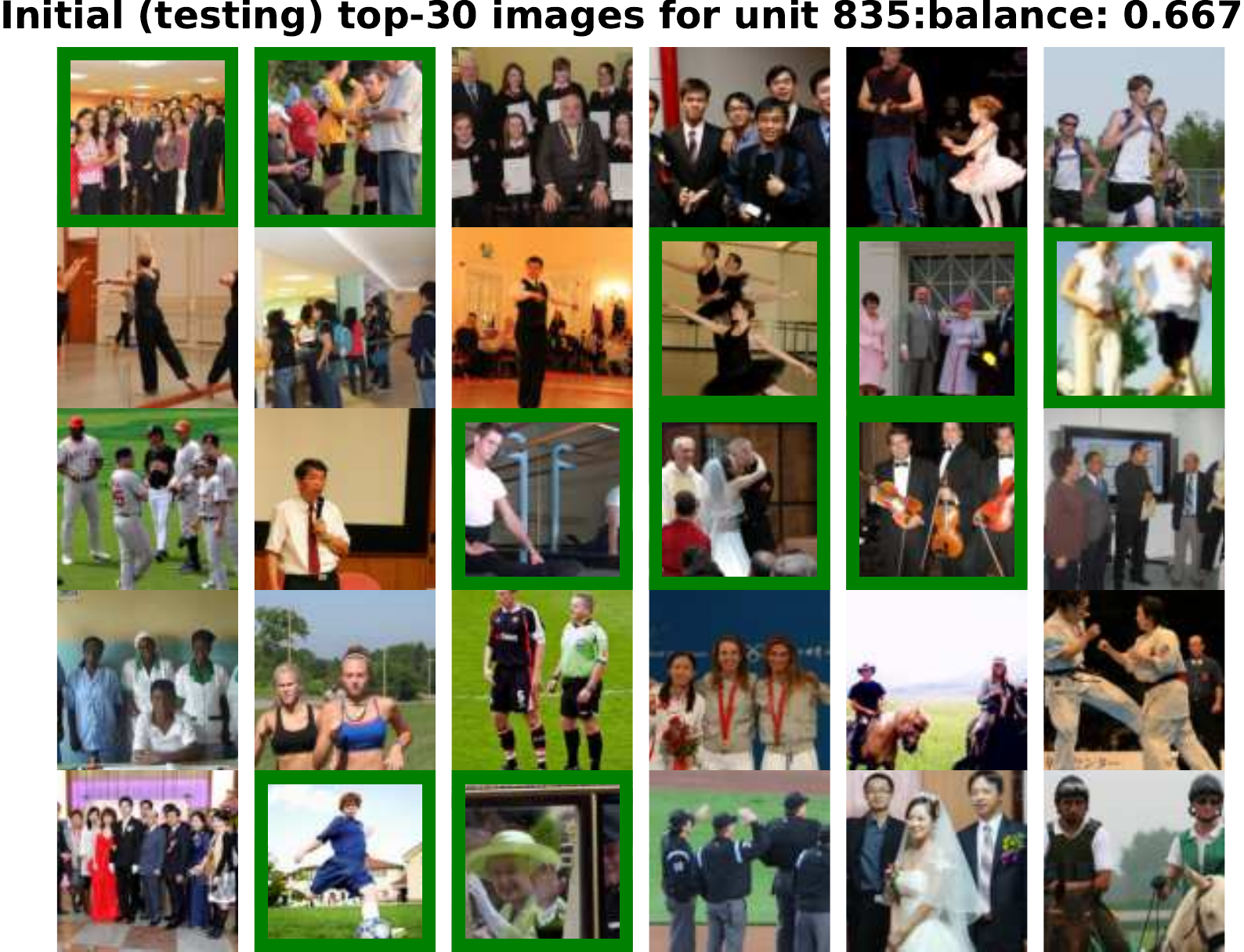}
\end{subfigure}\hfill
\begin{subfigure}[]{0.49\linewidth}
    \includegraphics[width=\textwidth]{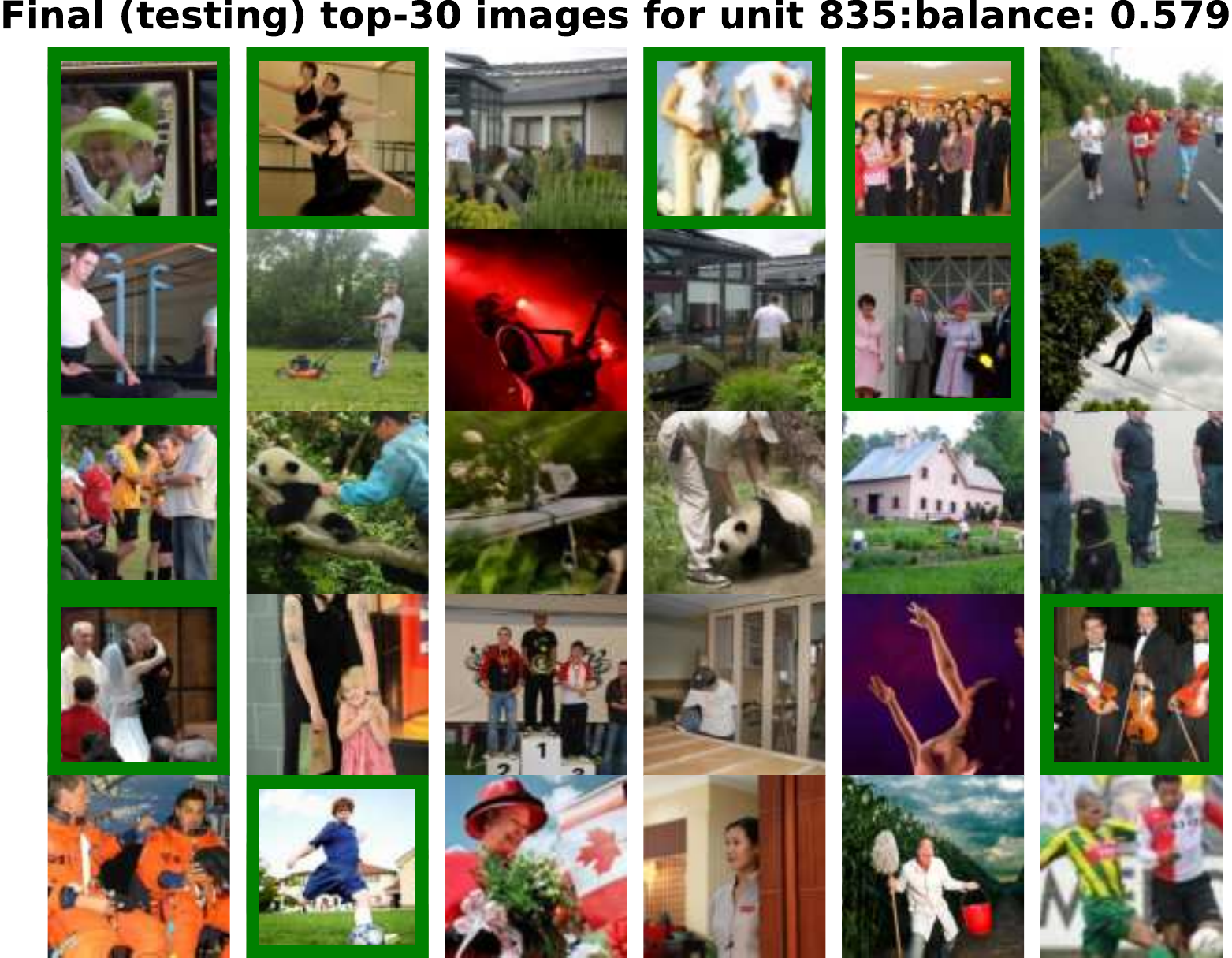}
\end{subfigure}\\
\vspace{.3cm}
\begin{subfigure}[]{0.49\linewidth}
    \includegraphics[width=\textwidth]{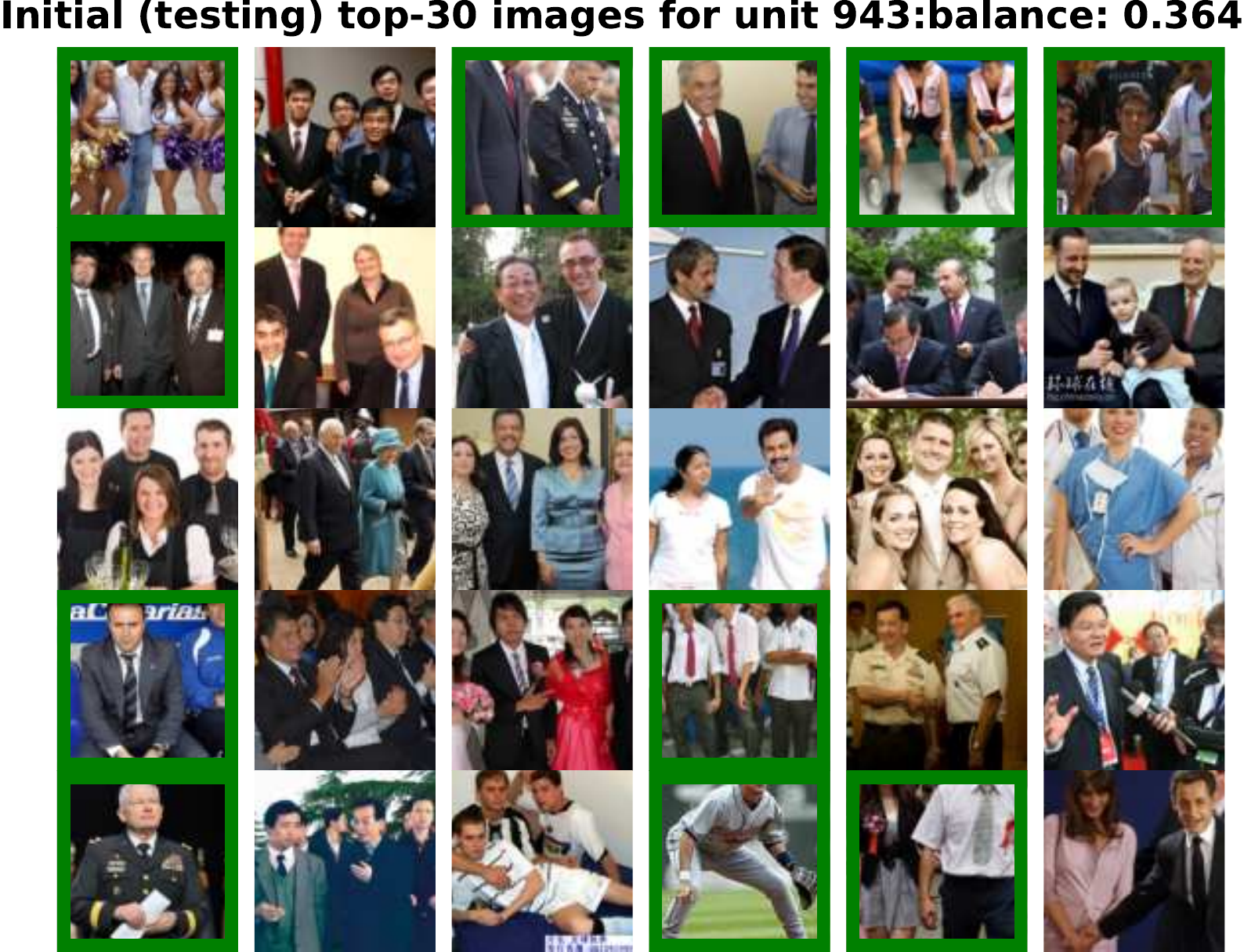}
\end{subfigure}\hfill
\begin{subfigure}[]{0.49\linewidth}
    \includegraphics[width=\textwidth]{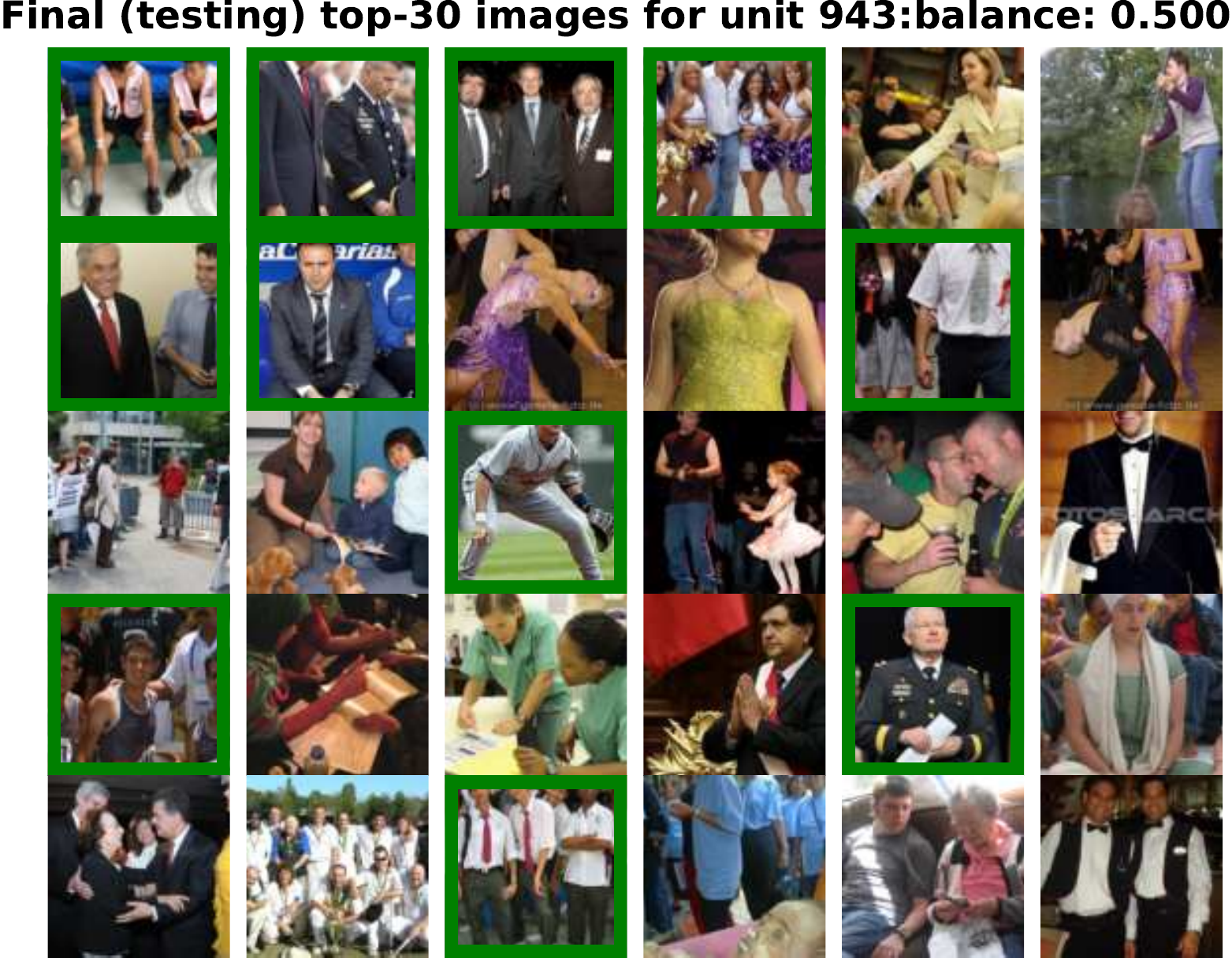}
\end{subfigure}\\
\vspace{.3cm}
\begin{subfigure}[]{0.49\linewidth}
    \includegraphics[width=\textwidth]{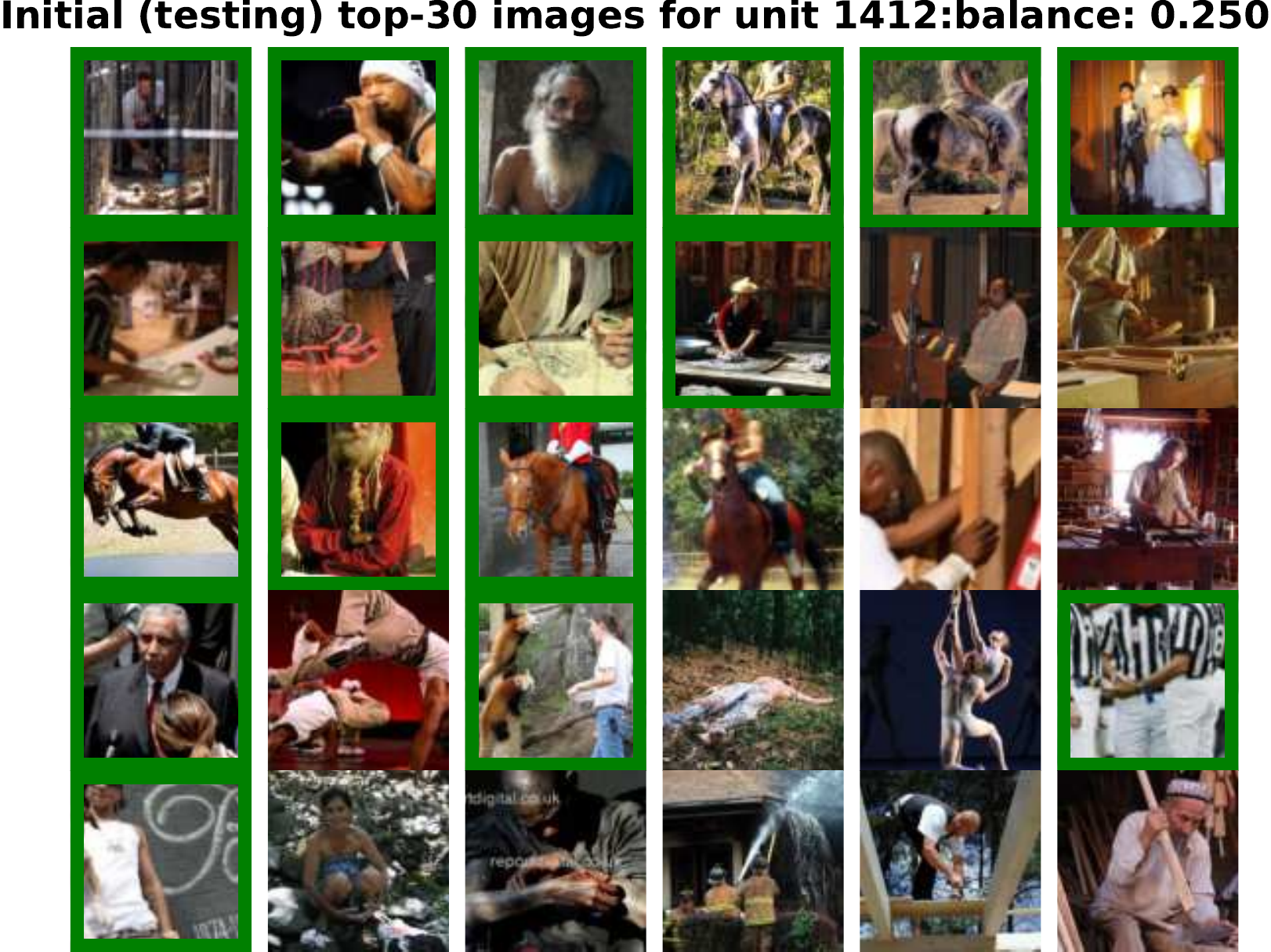}
\end{subfigure}\hfill
\begin{subfigure}[]{0.49\linewidth}
    \includegraphics[width=\textwidth]{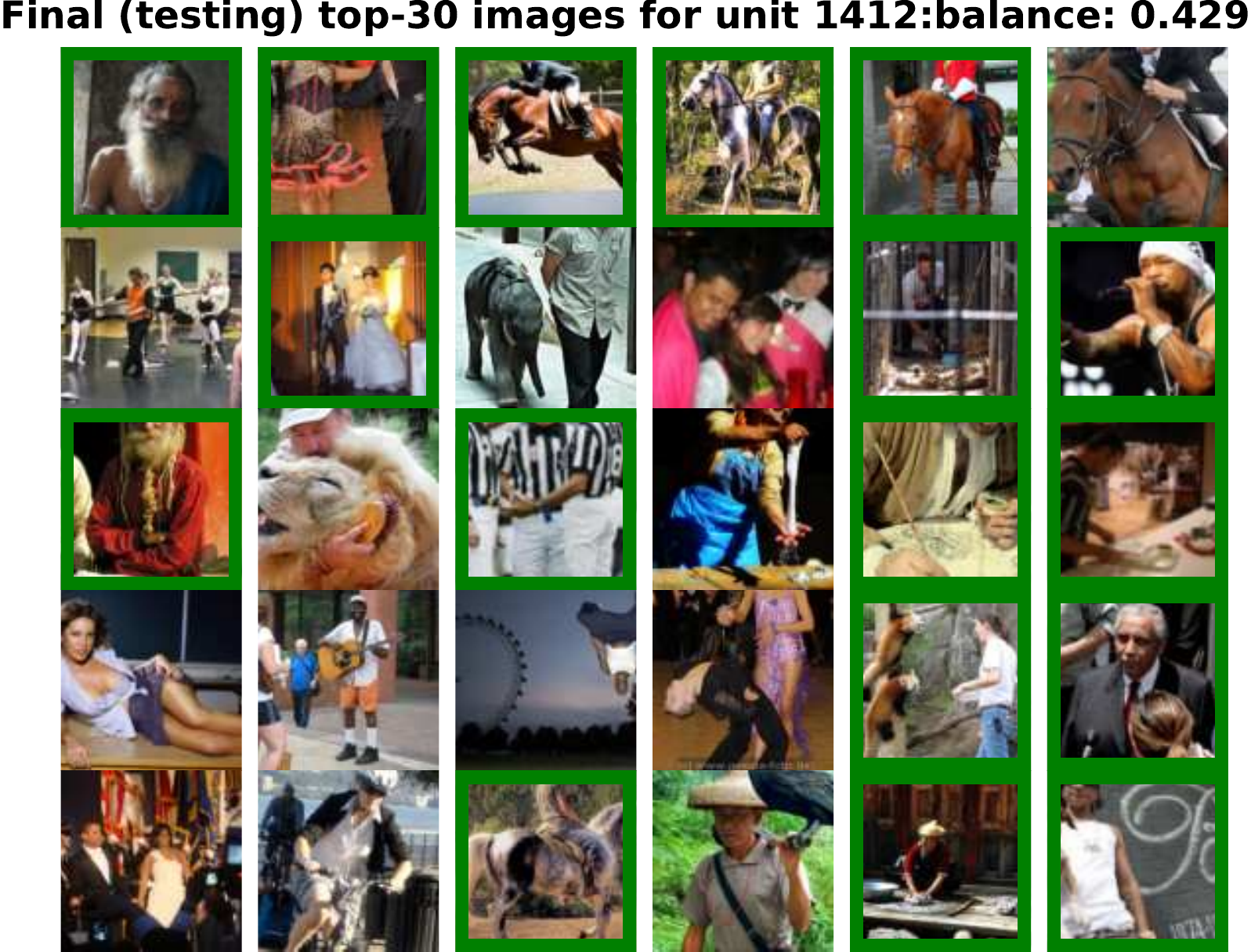}
\end{subfigure}\\
\vspace{.3cm}
\begin{subfigure}[]{0.49\linewidth}
    \includegraphics[width=\textwidth]{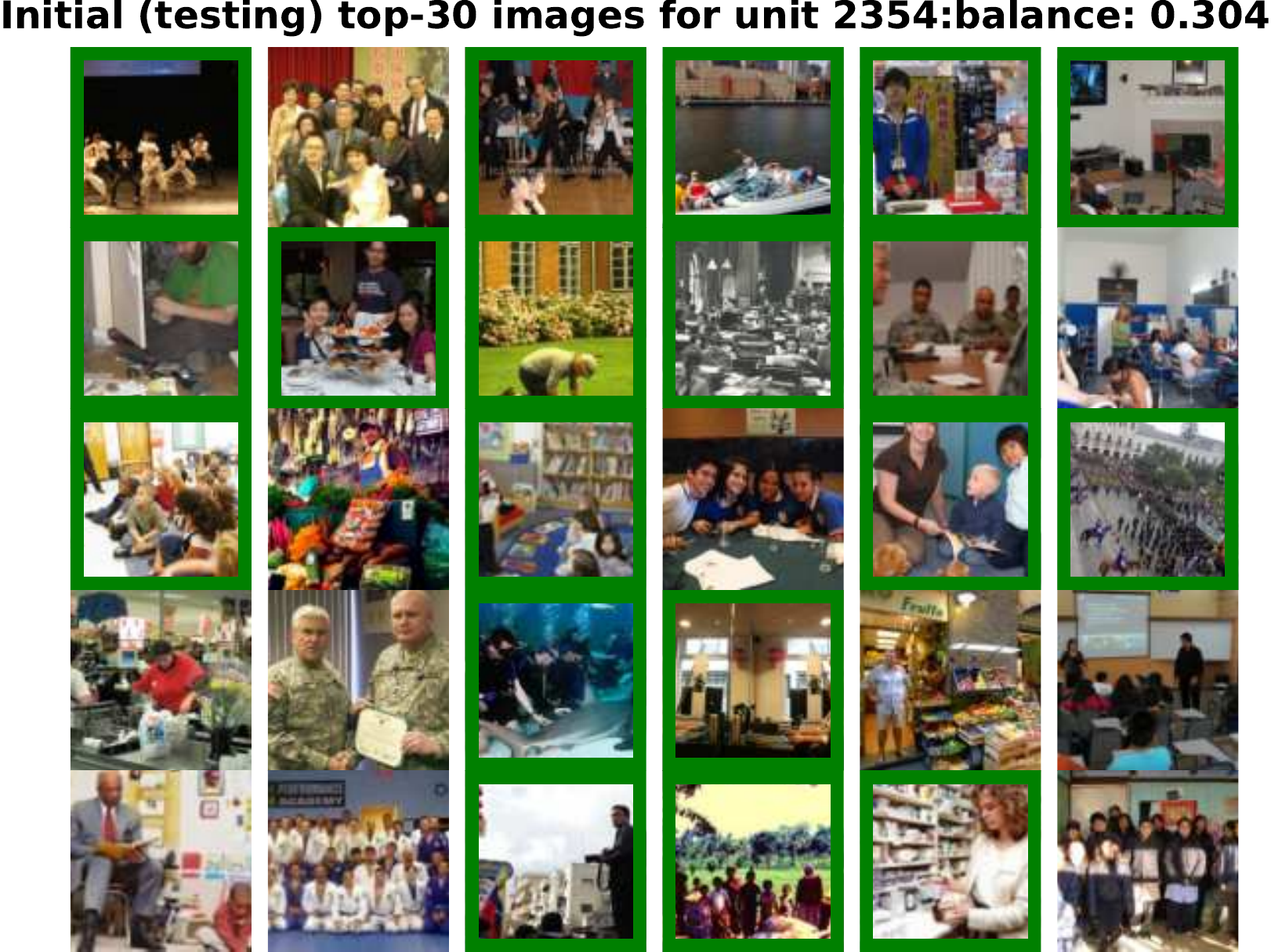}
\end{subfigure}\hfill
\begin{subfigure}[]{0.49\linewidth}
    \includegraphics[width=\textwidth]{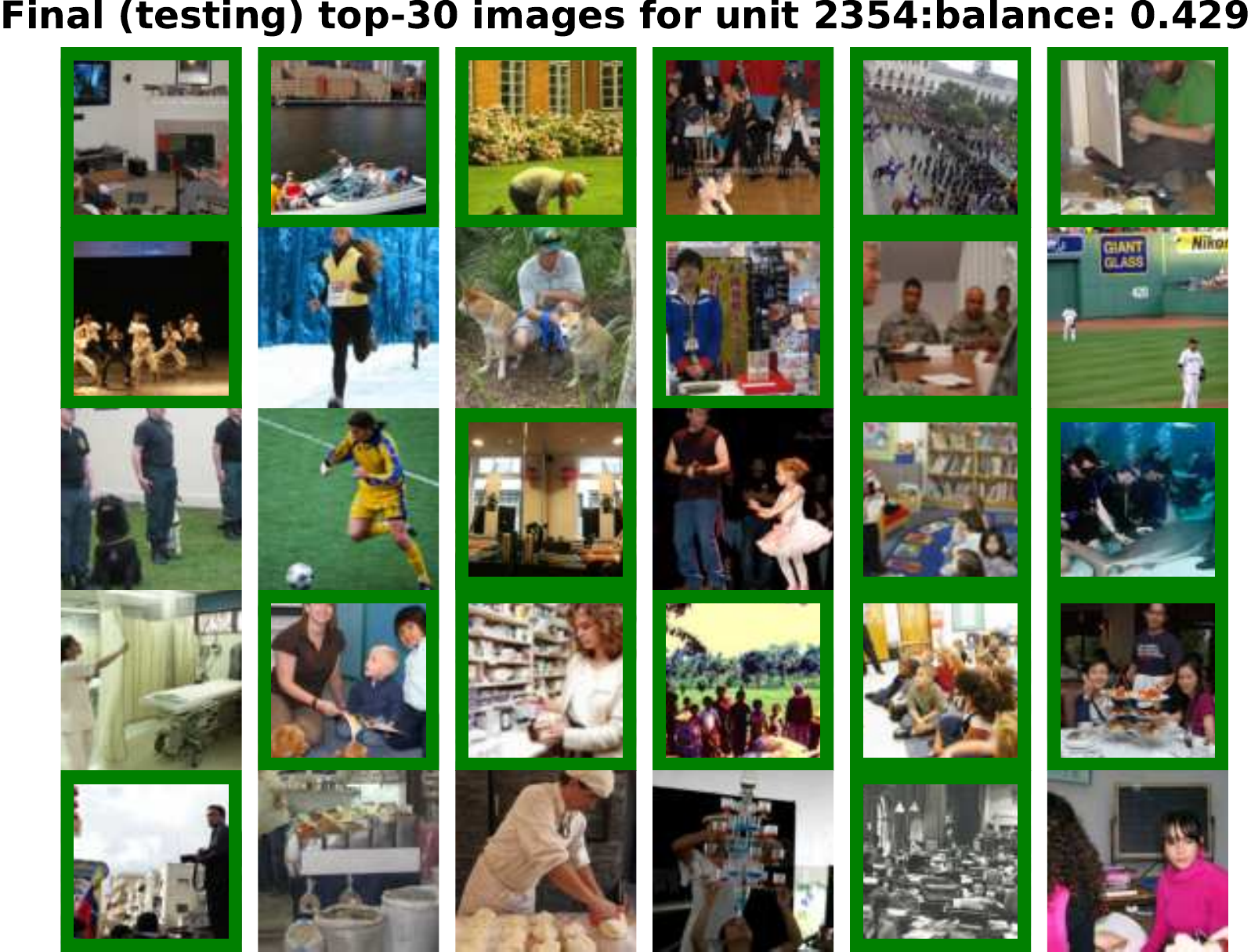}
\end{subfigure}

\caption{\small Results on several channels for the fairwashing attack. Units were randomly chosen. Balance (\textit{fairness}) is usually improved in cases of severe bias, in particular when there are not many people in images.} 
        \label{fig_add:fairwashing_attack_all}

\end{figure}

\begin{figure}[!ht]
\centering
\begin{subfigure}[]{0.49\linewidth}
    \includegraphics[width=\textwidth]{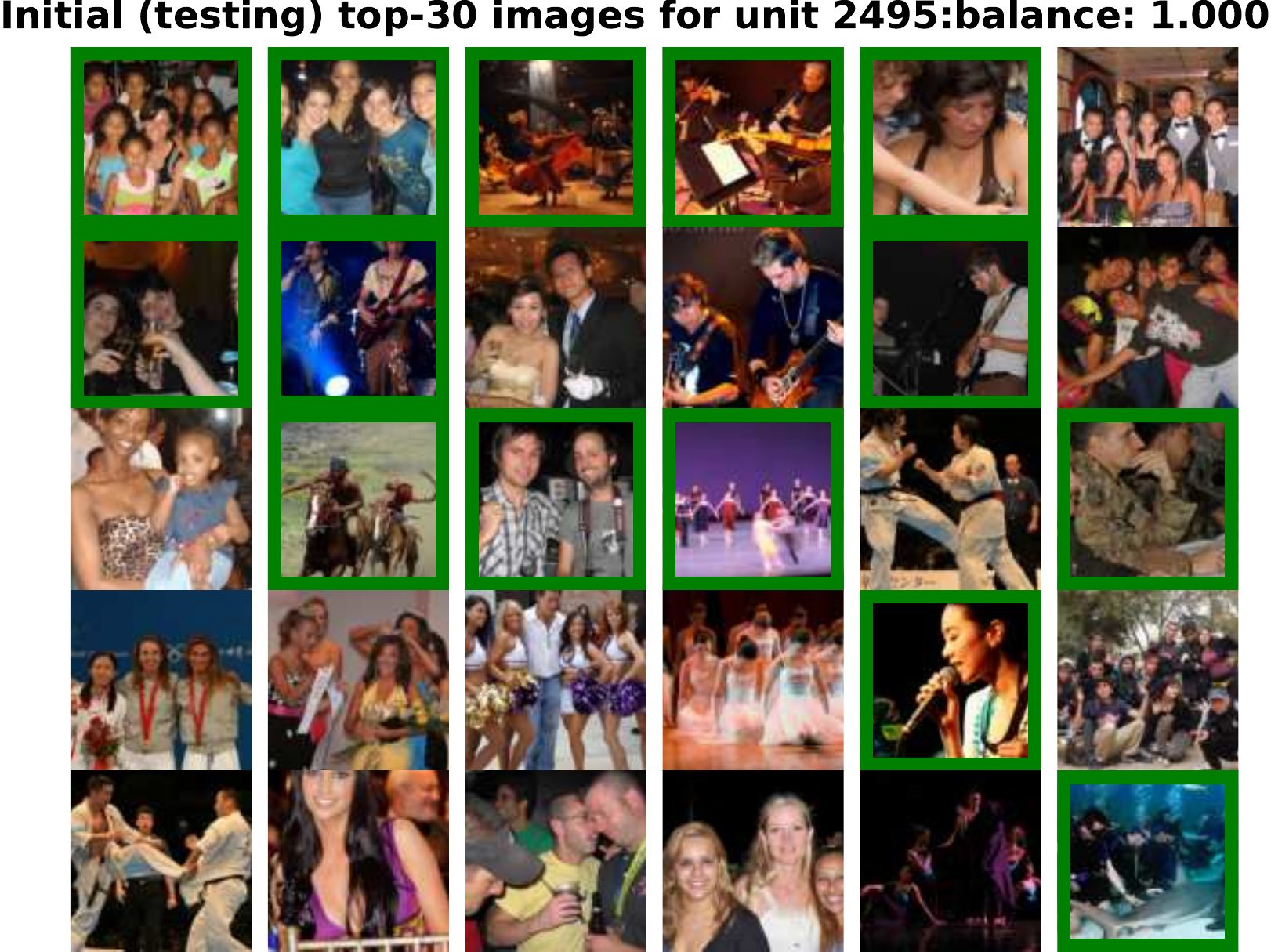}
\end{subfigure}\hfill
\begin{subfigure}[]{0.49\linewidth}
    \includegraphics[width=\textwidth]{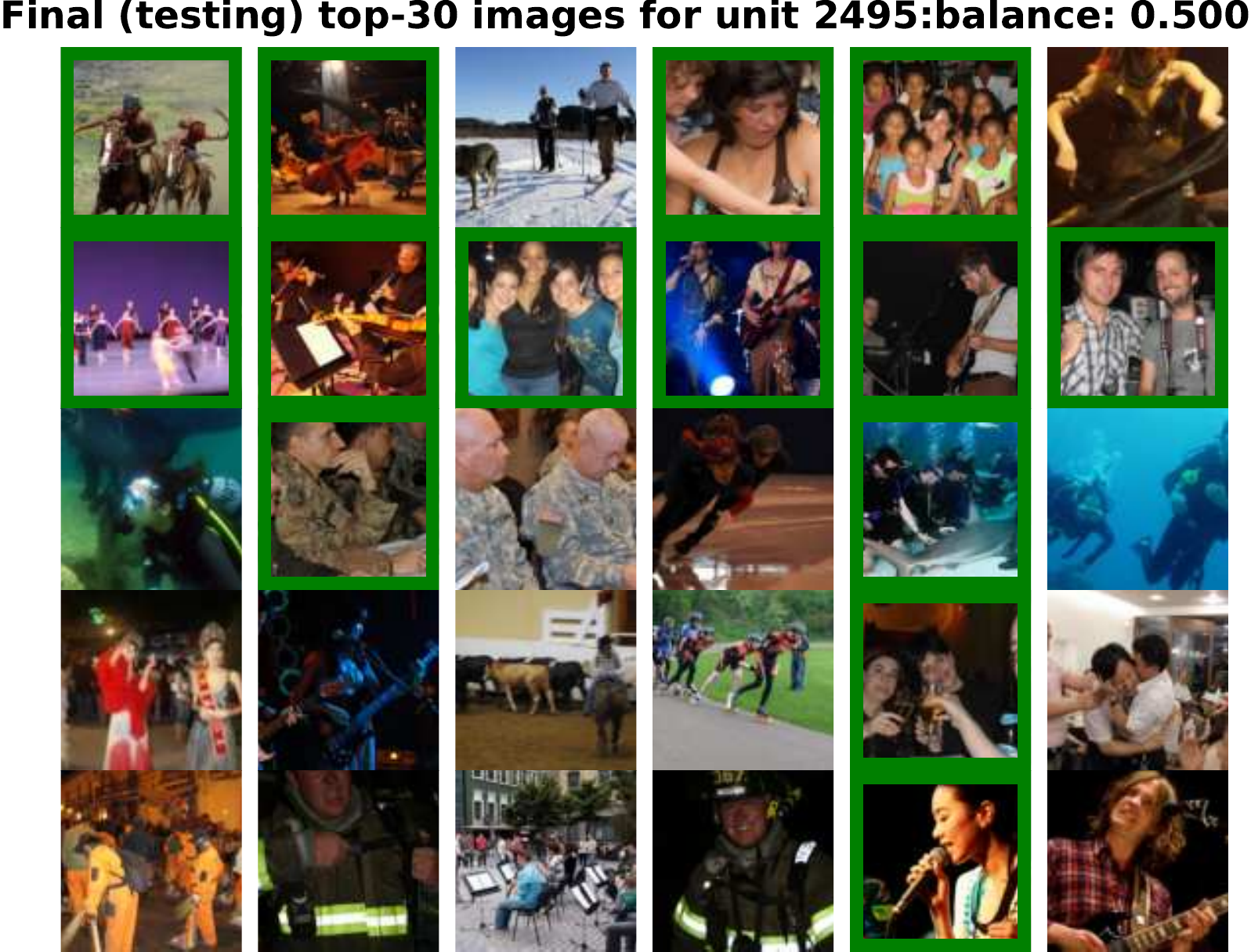}
\end{subfigure}\\
\vspace{.3cm}
\begin{subfigure}[]{0.49\linewidth}
    \includegraphics[width=\textwidth]{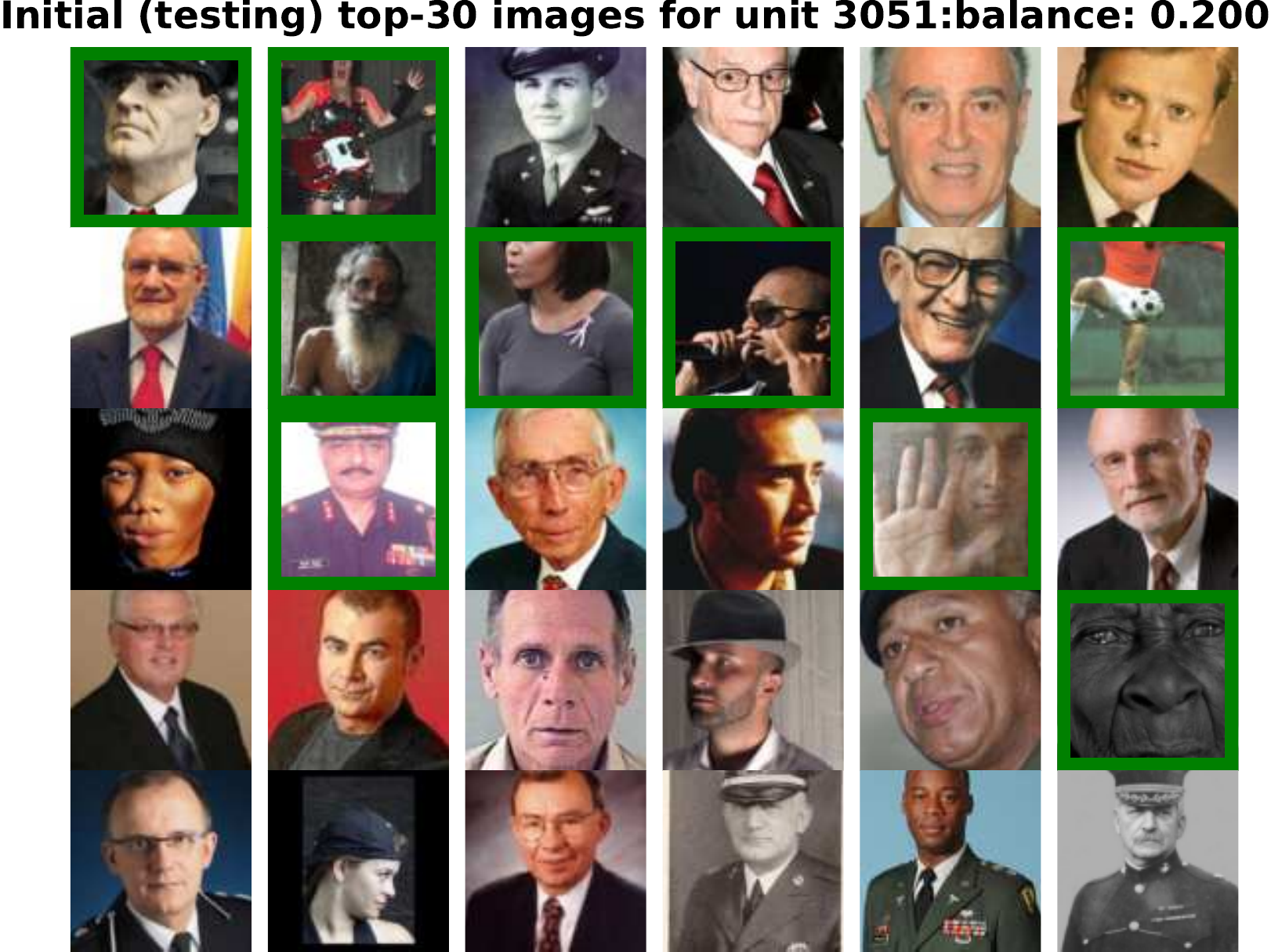}
\end{subfigure}\hfill
\begin{subfigure}[]{0.49\linewidth}
    \includegraphics[width=\textwidth]{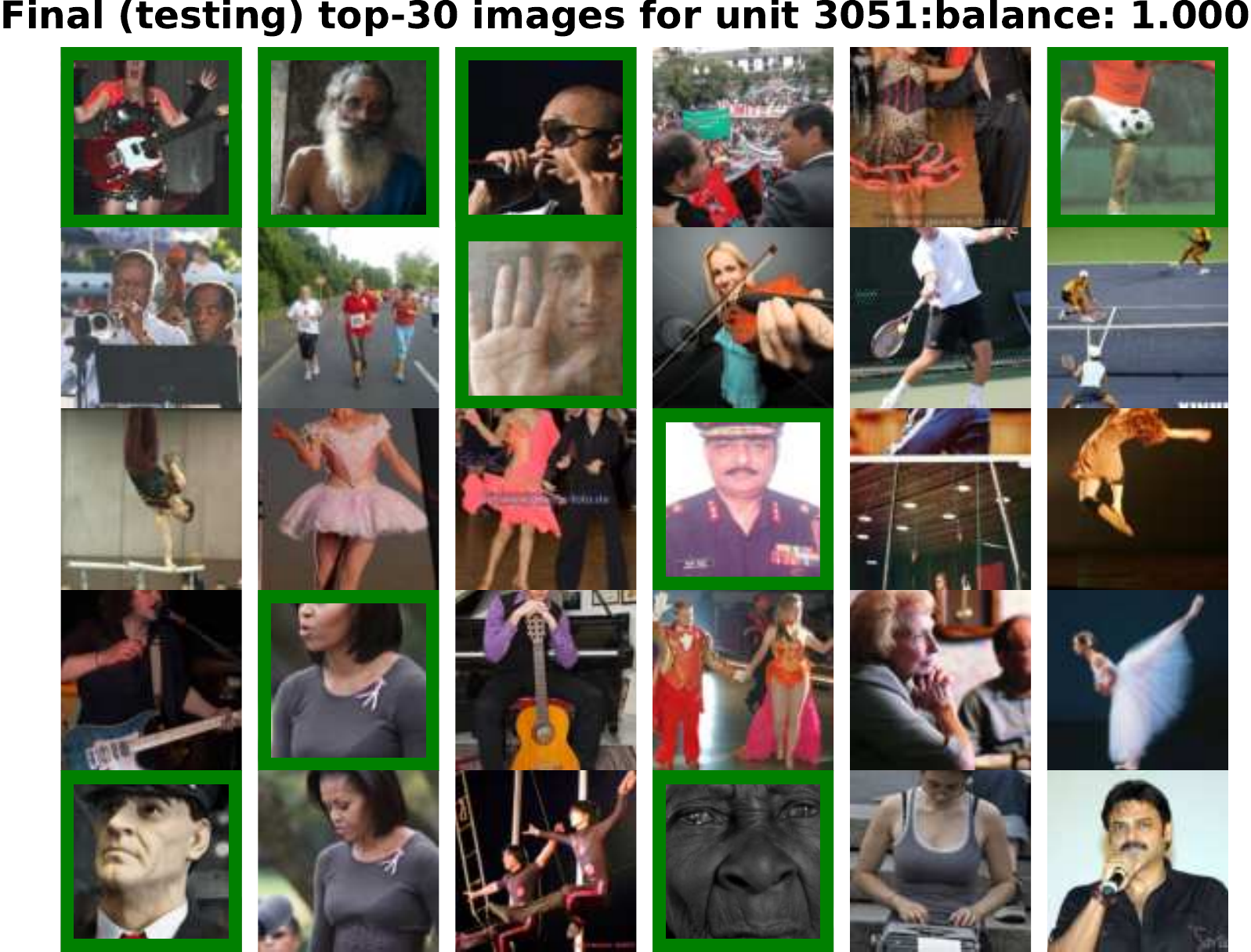}
\end{subfigure}\\
\vspace{.3cm}
\begin{subfigure}[]{0.49\linewidth}
    \includegraphics[width=\textwidth]{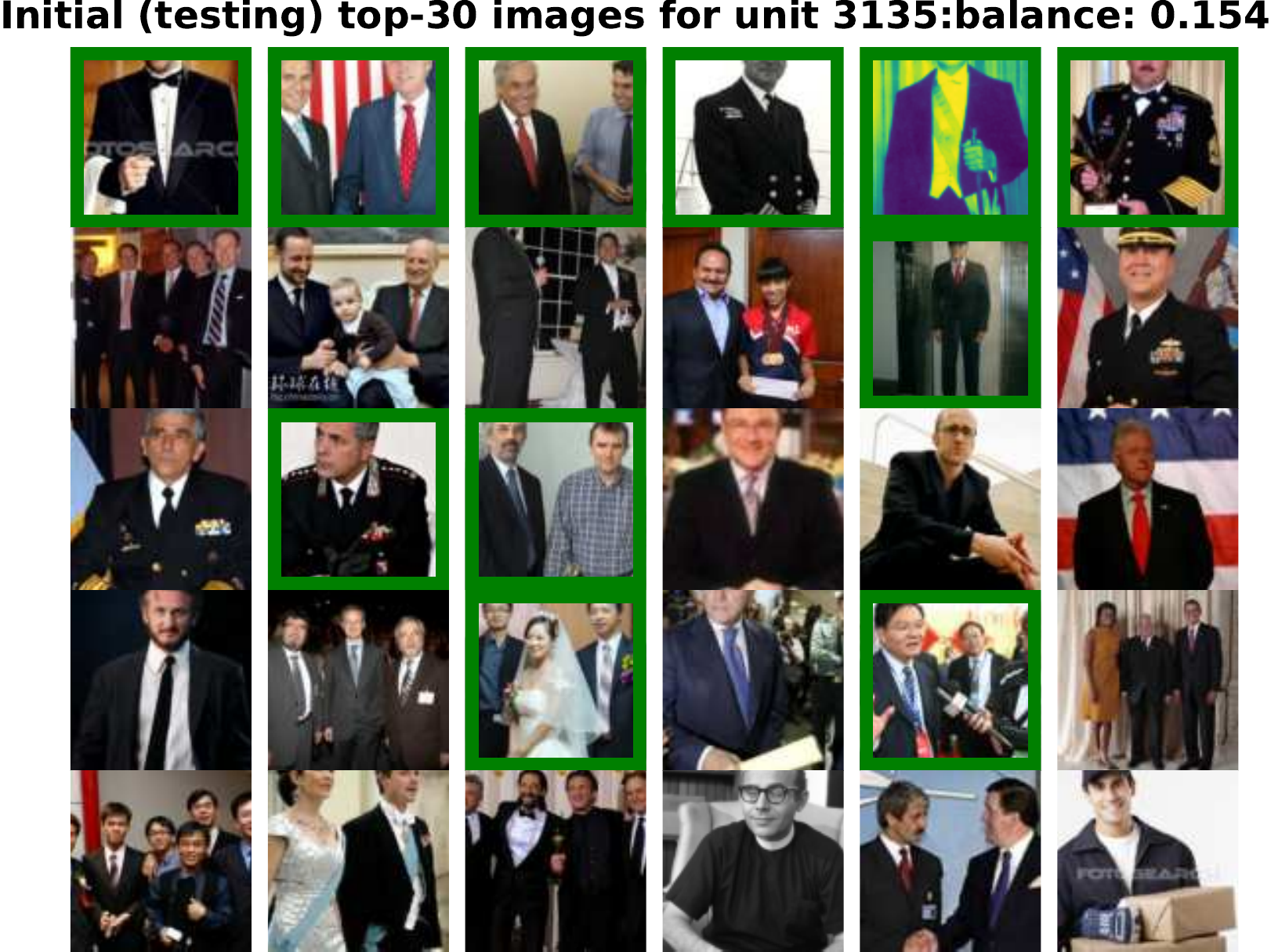}
\end{subfigure}\hfill
\begin{subfigure}[]{0.49\linewidth}
    \includegraphics[width=\textwidth]{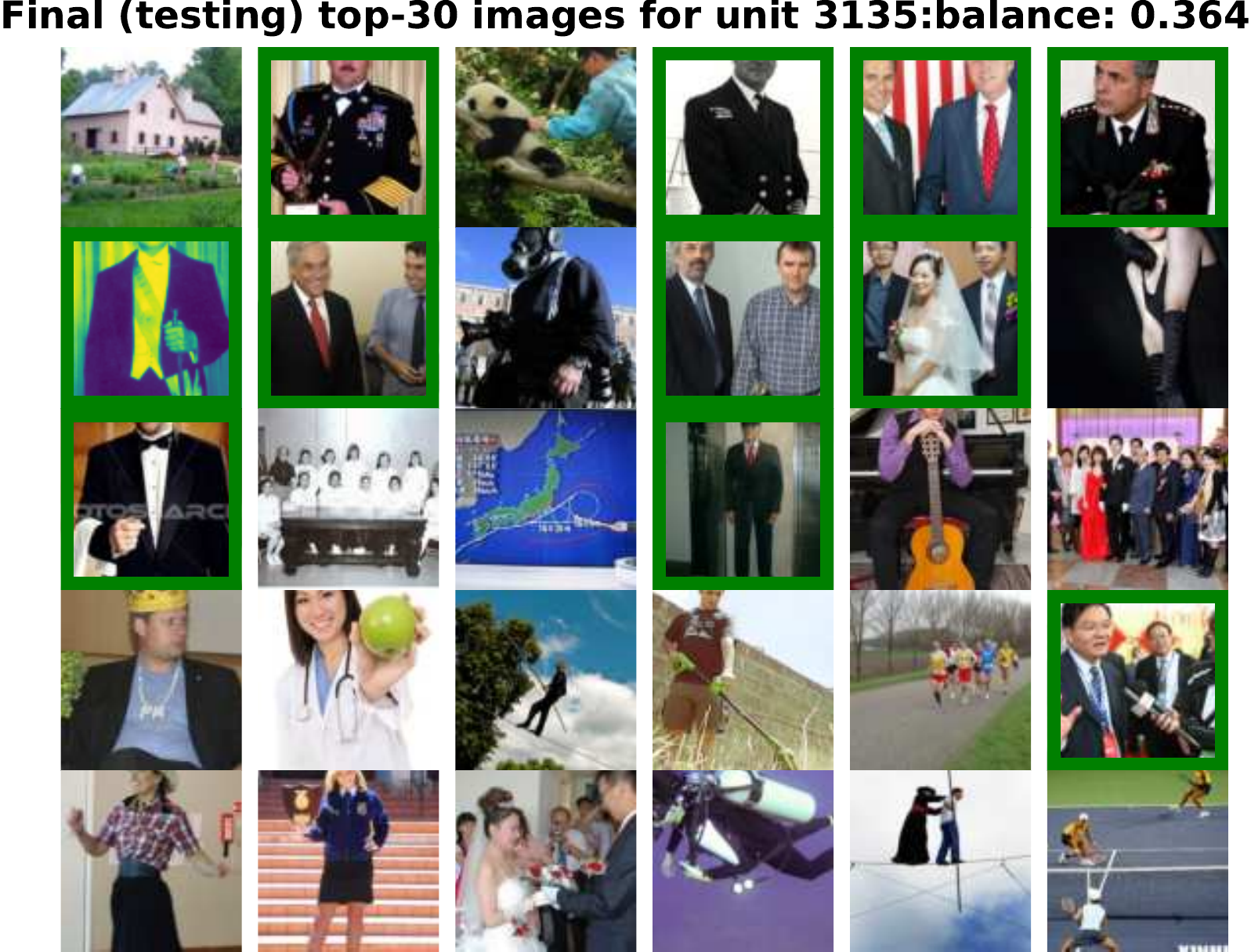}
\end{subfigure}\\
\vspace{.3cm}
\begin{subfigure}[]{0.49\linewidth}
    \includegraphics[width=\textwidth]{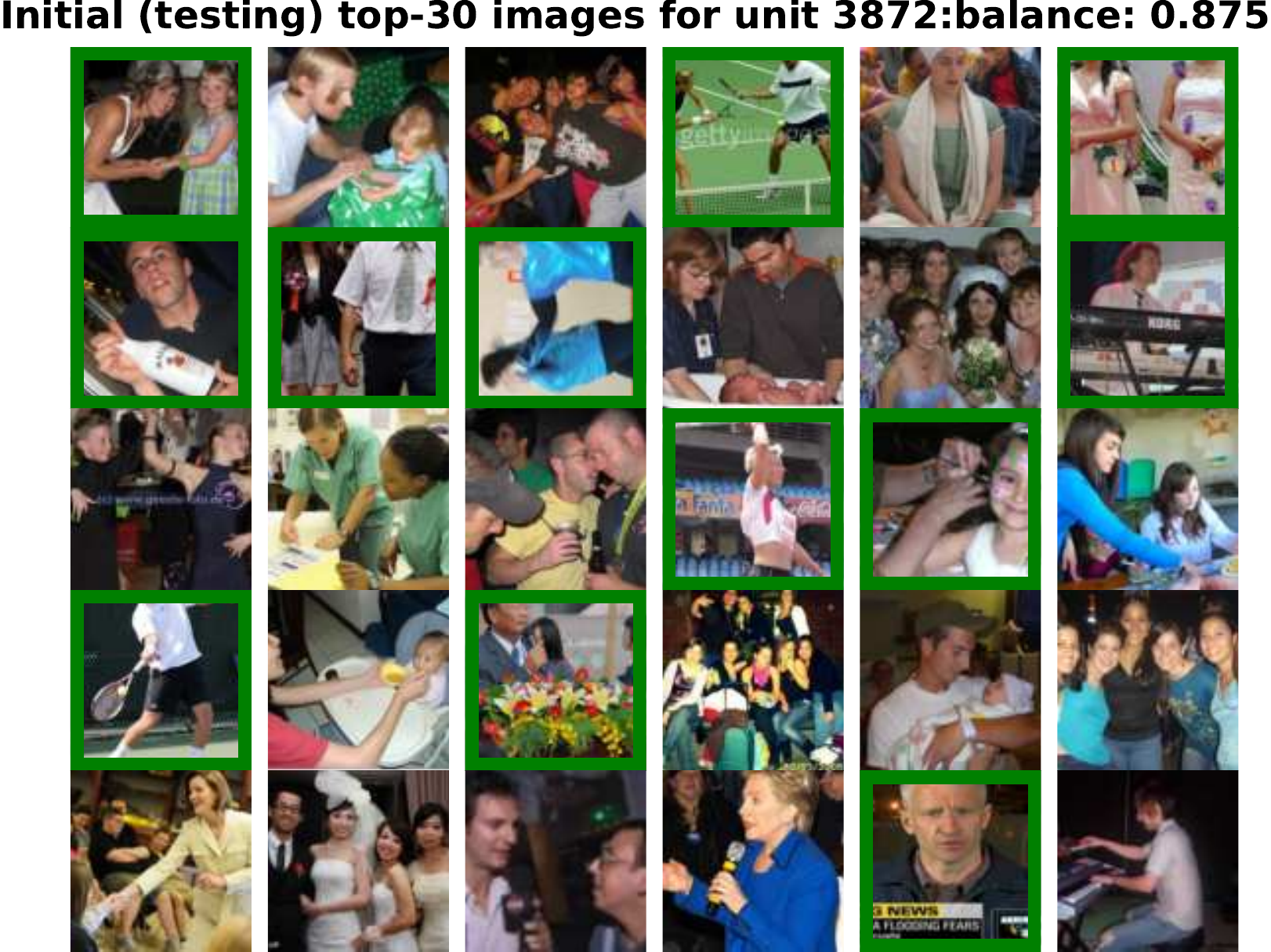}
\end{subfigure}\hfill
\begin{subfigure}[]{0.49\linewidth}
    \includegraphics[width=\textwidth]{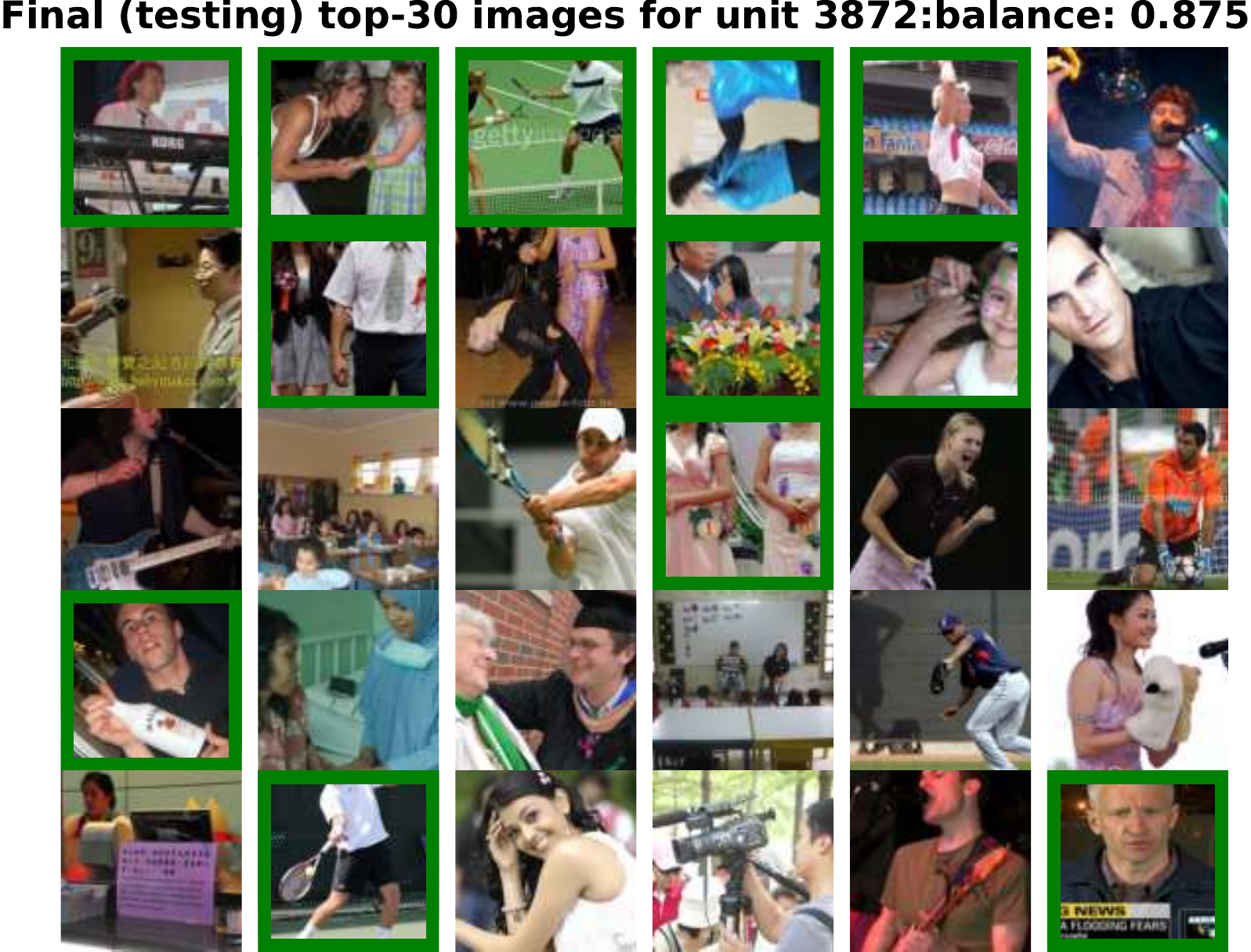}
\end{subfigure}\\

\caption{\small  Results on several channels for the fairwashing attack. Units were randomly chosen. The balance (\textit{fairness}) is usually improved in cases of severe bias, in particular when there are not many people in images.} 
        \label{fig_add:fairwashing_attack_next}
\end{figure}

\paragraph{More examples.} Figure~\ref{fig_add:fairwashing_attack_all} and ~\ref{fig_add:fairwashing_attack_next} simulate what the interpreter or regulator may see on testing annotated data before and after the fairwashing attack on 4 randomly chosen units. We can observe from this figure that when the balance (\textit{fairness} measure on top-$30$ images) is relatively low the fairwashing attack makes the top images look fairer (e.g., units 943, 1412, 3051, 3135). In particular, we observe (e.g., unit 3051) that the fairwashing attack is usually very effective in cases of severe bias in top-$k$ images with not many people in each image.

\clearpage

\end{document}